\documentclass[11pt]{article}
\usepackage[numbers]{natbib}
\usepackage{packages}
\usepackage{statistics}
\usepackage{lscape}
\usepackage{statistics-macros}
\usepackage{manyfoot}
\usepackage{setspace}
\usepackage{stackengine}
\usepackage{siunitx}
\usepackage{graphicx}
\usepackage{changepage}
\usepackage[most]{tcolorbox} 
\definecolor{innerboxcolor}{rgb}{.9,.95,1}
\definecolor{outerlinecolor}{rgb}{.6,0,.2}
\definecolor{outerlinecoloreb}{rgb}{0,1,0}
\definecolor{innerboxcoloreb}{rgb}{1,1,1}

\newcommand{\dottedline}[1]{%
  \multicolumn{1}{c}{\dotfill} & \multicolumn{1}{c}{\dotfill}\\
}

\DeclareNewFootnote{A}
\DeclareNewFootnote{B}

\let\footnoteR\footnoteB
\let\footnote\footnoteA

\def\tyw#1{\textcolor{black}{#1}}
\def\twy#1{\textcolor{black}{#1}}

\def\wty#1{\textcolor{black}{#1}}
\def\ljs#1{\textcolor{black}{#1}}

\begin{document}

\begin{center}
  {\huge Rethinking Distribution Shifts: Empirical Analysis and Modeling for Tabular Data}\\
  \vspace{.5cm}
 {\Large
   Tianyu Wang\footnoteR{Equal contribution, listed in reverse alphabetical order}$^{,1}$ \quad Jiashuo Liu\textsuperscript{*}$^{,2}$ \quad Peng Cui$^2$ \quad Hongseok Namkoong$^3$} \\
 \vspace{.2cm}
   $^1$Department of Industrial Engineering and Operations Research,
   $^2$Department of Computer Science and Technology,
   $^3$Decision, Risk, and Operations Division \\
 \vspace{.2cm}
 {\large  Columbia University$^{1,3}$ \hspace{3em} Tsinghua University$^{2}$}  \\
 \vspace{.1cm}
 \texttt{tw2837@columbia.edu, liujiashuo77@gmail.com}\\
 \texttt{cuip@tsinghua.edu.cn, namkoong@gsb.columbia.edu} \\
\end{center}

\begin{abstract}
  Different distribution shifts require different interventions, and algorithms must be grounded in the specific shifts they address. However,  methodological development for robust algorithms typically relies on structural assumptions that lack empirical validation. Advocating for an empirically grounded \emph{data-driven} approach to algorithm development, we build an empirical testbed comprising natural shifts across {8 tabular datasets, 172 distribution pairs over 45 methods and 90,000 method configurations} encompassing empirical risk minimization and distributionally robust optimization (DRO) methods. We find $Y|X$-shifts are most prevalent in our testbed, in stark contrast to the heavy focus on $X$ (covariate)-shifts in the ML literature, and that the performance of robust algorithms is no better than that of vanilla methods. To understand why, we conduct an in-depth empirical analysis of DRO methods and find that underlooked implementation details---such as the choice of underlying model class (e.g., LightGBM) and hyperparameter selection---have a bigger impact on performance than the ambiguity set or its radius. We illustrate via case studies how a data-driven understanding of distribution shifts can provide a new approach to algorithm development.
  \footnote{More information on the data, codes and python
    packages about \textsc{WhyShift} are available at 
    \url{https://github.com/namkoong-lab/whyshift}.
    \tyw{A conference version of this paper is here \citep{liu2024need}.}}
\end{abstract}

\section{Introduction}\label{sec:intro}

The misalignment between the data distribution in the training and post-deployment phase, commonly referred to as \emph{distribution shift}, is a critical challenge for robust predictions. As \citet{knight1921risk} articulates in his classical work, in contrast to statistical uncertainty where the likelihood of an event can be determined using a statistical model, epistemic uncertainty arises when there is insufficient information to even construct a probabilistic model. 
 Frameworks like distributionally robust optimization (DRO)~\citep{rahimian2019distributionally} provide an approach to decision-making under epistemic uncertainty by requiring modelers to (i) articulate the set of distribution shifts of interest through ambiguity sets and (ii) explicitly optimize worst-case performance within these sets.

To model epistemic uncertainty, past works typically develop algorithms for specific distribution shifts posited by researchers~\cite{kuhn2025distributionally}. 
We refer to these distribution shifts as \twy{\emph{controlled}} since \twy{they are typically constructed by specifying a priori assumptions about how distributions may change} rather than being informed by empirical data. When modeling assumptions are not validated, the resulting predictions can perform poorly~\citep{hu2018does}.

Motivated by empirical benchmarking practices that have driven progress in machine learning~\citep{recht2019imagenet}, we advocate for a fresh perspective in this work. 
Rather than starting with mathematical \twy{assumptions}, we study distribution shifts as they manifest in the data, 
and then evaluate every available method across diverse model classes to empirically determine what works.
\twy{Unlike previous benchmarks that construct \twy{\emph{controlled}} shifts where datasets are perturbed to match specific ambiguity set assumptions, the distribution shifts we examine are larger in scale compared to those typically considered in the literature (172 distribution pairs) and quantitatively more complex. The shifts we consider often involve intricate mixtures of covariate changes and evolving relationships that resist simple mathematical characterization.}


We refer to the distribution shifts we study as \twy{\emph{semi-synthetic}}. It is important to note that such shifts are neither more nor less realistic than shifts constructed through controlled perturbations. Rather, they shed light on shifts different from those typically considered in the literature.  Throughout this work, we focus on the data structured in a table format (``tabular data''), where each row corresponds to a data instance (e.g., a person) and each column corresponds to a specific feature describing the instance (e.g., age, salary, product category). 
 Such data modalities dominate electronic healthcare records~\cite{fatima2017survey}, financial
data~\cite{dastile2020statistical}, natural
sciences~\cite{shwartz2022tabular}, online advertising~\citep{guo2017deepfm,mcmahan2013ad} and many other domains~\citep{johnson2016mimic,ulmer2020trust}.

\paragraph{A new distribution shift benchmark.}
To substantiate our proposal, we construct 172 distribution shift pairs from 8 real-world tabular datasets spanning \wty{education, healthcare}, commute times, road accidents, income, and insurance coverage. \wty{Based on these, we evaluate 45 methods across 90,000 configurations--including DRO, imbalanced learning, and fairness algorithms. Our benchmark, \textsc{WhyShift}, provides the \emph{first systematic empirical evaluation of commonly used DRO methods} under \twy{semi-synthetic} distribution shifts that cannot be easily characterized by existing ambiguity sets,} capturing socioeconomic, geographic, demographic, and temporal diversity. We release our implementation and evaluation protocols as a Python 
package (\url{https://pypi.org/project/whyshift/}).

\twy{Our benchmark aims to bridge fundamental differences in research culture between OR and ML communities; these two modes are complementary, each with its unique set of advantages and limitations. OR research emphasizes rich theoretical insights based on mathematically tractable model classes that admit closed-form solutions or convex reformulations. In contrast, ML practice prioritizes performance on benchmarks, readily adopting complex model architectures like tree-based ensembles that have demonstrated superior performance without analytical treatment. }\twy{We advocate for bridging the gap between the DRO community's focus on modeling distribution shifts and the ML community's emphasis on empirical validation. In particular, the scale and diversity of our benchmark enable three key contributions that advance our understanding of both robust algorithms and distribution shifts that resist simple mathematical representations:
}



\vspace{-10pt}
\twy{\paragraph{Understanding when blind benchmarking fails: the role of shift type.}
To illustrate the need for careful modeling of distribution shifts, we first show that even widely cited empirical findings in the ML literature fail to generalize 
without a grounded understanding of distribution shifts
(Section \ref{subsec:acc-on-the-line}). 
In particular, \citet{miller2020effect} observed that across vision and language tasks, out-of-distribution performance strongly correlates with in-distribution performance (``accuracy-on-the-line''), suggesting practitioners can reliably use validation performance to predict deployment performance.}

\twy{Our analysis reveals that the ``accuracy-on-the-line'' phenomenon fundamentally breaks down on many of the tabular shifts we consider (\Cref{fig:decomposition}). Our main intellectual contribution is in providing a principled explanation for this breakdown. }

\twy{The key distinction lies in the type of the distribution shifts. $X$-shifts, which involves changes in covariate marginal distributions, behave fundamentally 
differently from $Y|X$-shifts, which involves changes in conditional relationships. Under $X$-shifts, model rankings remain stable and accuracy-on-the-line holds, justifying the standard ML practice of benchmarking out-of-distribution performance with the expectation of a single robust model across domains (e.g., images classified by humans). However, $Y|X$-shifts are common in our tabular data because of changes in unobserved covariates across environments. This introduces considerable 
performance variation, leading to different relationships between in-distribution and out-of-distribution performance across datasets, and ultimately causes the breakdown of accuracy-on-the-line.
}

\vspace{-10pt}
\paragraph{Evaluating robust algorithms: model class dominates algorithm choice.}
We investigate whether algorithms designed for \twy{controlled} distribution shifts perform well on the \twy{semi-synthetic} shifts we construct. Section \ref{sec:perform-comp}  presents our systematic evaluation across diverse learning algorithms and model classes—including DRO, imbalanced learning, and fairness-enhancing methods. Unsurprisingly, algorithms specialized to \twy{controlled} shifts have no clear advantage over carefully tuned vanilla ERM when applied to distributions that don't match their design assumptions.

To understand which aspects of algorithmic
design impacts performance, we conduct  extensive empirical analysis in Section \ref{sec:method-lr} where we carefully control for confounding factors such as the type of distribution shift. 
Our main finding is that implementation details often overlooked by methodology researchers can have outsize impact. For example, 
 the choice of the model class (e.g., neural network vs. LightGBM) has a significantly greater impact on the performance of DRO methods compared to the choice of the ambiguity set, its radius, and even the use of DRO itself. 

Our analysis suggests that the cultural divide between OR and ML has profound implications. If DRO methods are built and evaluated only with linear models, their comparatively poor performance versus more expressive tree-based ensemble methods can be attributed to the choice of model class, not to any inherent flaw in the DRO approach. In other words, restricting attention to ``theoretically convenient'' linear models masks the real potential of DRO with richer model families.

{ \color{black}
\vspace{-10pt}
\paragraph{Tailoring interventions with target data.} 
Beyond evaluation, our benchmark enables concrete guidance for practitioners. In Section \ref{sec:intervention}, we present case studies illustrating how a nuanced data-driven understanding of shifts from small amounts of target-domain data can enhance both algorithmic and data-centric interventions. First, we present an algorithmic intervention: by analyzing which features exhibit the largest shifts, we design tailored ambiguity sets that align DRO's worst-case distributions with actual shift patterns. This targeted approach substantially improves DRO robustness (\Cref{sec:alg_intervention}).
Then we present a data-centric approach: leveraging patterns across our  diverse shift pairs, we identify covariate regions suffering the largest $Y|X$-shifts (\Cref{subsubsec:riskregion}) and show this understanding enables efficient data and feature collection (\Cref{sec:model_intervention}).

Our results demonstrate data-driven modeling based on small amounts of out-of-distribution observations can yield substantial benefits, and acknowledge the practical limitations of purely
algorithm development on controlled shifts when these algorithms are applied on distributions that do not enjoy similar structures.
Our perspective
represents a conceptual shift from asking “what mathematical structure should we assume?” to “what
can the data tell us about the structure of realistic distribution shifts?”. 
In the parlance of ML, our
view is akin to meta-learning~\citep{hospedales2021meta} where modelers have observed some set of distribution shifts prior
to deployment and can use patterns across these shifts---rather than individual distributions---to
inform robust model design.

\vspace{10pt}
Taken together, our framework highlights the disparity between semi-synthetic shifts and controlled shifts that researchers analyze in the literature, and helps researchers diagnose and inform algorithm design through comprehensive empirical analysis and data-driven modeling.

}

\vspace{-10pt}
\paragraph{Related work.}
Distribution shifts is an age-old challenge~\citep{schlimmer1986incremental,gama2014survey,lu2018learning, quinonero2008dataset}, but standard empirical benchmarks in ML focus on the standard i.i.d. setting \citep{bischl2017openml} or synthetic shifts such as \texttt{MNIST-C}~\cite{mu2019mnist}.
To address this gap, several authors recently develop distribution shift benchmarks for computer vision and natural language  processing tasks~\citep{koh2021wilds,recht2019imagenet,santurkar2021breeds,hendrycks2021many,
  DBLP:conf/nips/YaoCC0KF22, subshift, gulrajanisearch}. However, these works do not investigate the type of shift they
address since $Y|X$-shifts are less likely to occur in vision and language: $Y$ is constructed from human knowledge given an input $X$ (e.g., pixels and words). 
Since existing tabular benchmarks also focus on $X$-shifts following this trend  \citep{robert2020, gardner2022subgroup}, we build a benchmark with diverse shift patterns
covering both $Y|X$- and $X$-shifts. \wty{Using image datasets, \citet{gulrajanisearch} illustrates that existing algorithm interventions based on deep neural networks often fail to handle all types of distribution shifts effectively. In contrast, tabular data tends to exhibit more $Y|X$-shifts. Therefore, our benchmark evaluates methods across a variety of model classes -- including linear and tree-based ensemble methods beyond deep neural networks -- given their strong performance on tabular data~\cite{grinsztajn2022why}. Furthermore, we also provide algorithmic and data-centric interventions, showing that a better understanding of shift patterns can lead to improved method performance. (See \Cref{app:relatedwork} for more comparison details).} 
\twy{Therefore, our work complements 
diagnostic tools that help explain performance gaps and refine methods~\cite{budhathoki2021why,pmlr-v202-kulinski23a,zhang2023why,namkoong2023diagnosing}.}

Robust methods, especially DRO methods, can provide benefits in terms of generalization, fairness and robustness in recent investigations. Variants of DRO methods are known to obtain better statistical performance
\citep{blanchet2021statistical,duchi2021statistics} and generalization error
than ERM counterparts with and without distribution shifts \citep{esfahani2018data,lee2018minimax,shafieezadeh2019regularization,gao2022wasserstein,duchi2019variance,iyengar2022hedging}. Besides, DRO
can contribute to the fairness and robustness of models under different
domains theoretically and empirically \citep{taskesen2020distributionally}. These algorithms are useful \twy{when the ambiguity set can capture the corresponding distribution shifts, such as variations in group distributions~\citep{sagawa2020distributionally}, demographic shifts~\citep{hashimoto2018fairness}, and adversarial perturbations~\citep{sinhaND18,bennouna2023certified}.} \tyw{These methods based on \twy{controlled} distribution shifts, while useful for insensitivity \citep{shafieezadeh2019regularization} and reliability \citep{duchi2021learning}, often overestimate the severity of shifts and make them less practical for real-world scenarios.}
In this work, we consider the setting of \twy{semi-synthetic} distribution shifts instead of \twy{controlled} distribution shifts
characterized by a distance measure or fairness metrics in the literature. 

\twy{Recent DRO work acknowledges that methods designed for general worst-case shifts can be 
overly pessimistic~\citep{hu2018does} and proposes more plausible worst-case distributions 
through refined distance notions~\citep{wang2021sinkhorn,bennouna2022holistic,blanchet2023unifying}. 
Several authors move beyond joint $(X,Y)$ distributions: \citet{duchi2023distributionally} 
study covariate subset shifts, while \citet{sahoo2022learning} study $Y|X$-shifts. However, our evaluations reveal these approaches still don't accurately reflect patterns in our \twy{semi-synthetic} benchmark. {This gap partly stems from how these methods 
are developed and evaluated. The DRO literature typically focuses on specific model classes 
for theoretical tractability: classic works use linear models, while recent extensions employ basic neural networks (See \Cref{tab:loss_choice} in Appendix~\ref{app:relatedwork} for more design details).} We argue for complementing algorithmic design with data-driven 
approaches tailored to target distributions, concretely demonstrating how careful understanding 
of $Y|X$- and $X$-shifts yields practical benefits.}

\paragraph{Notations.} Letting $(X, Y)$ be random variable supported on the
  space $\Xscr \times \Yscr$, we consider a model
$f:\mathcal X\rightarrow \mathcal Y$ (from the model class $\Fscr$) that predicts the outcome $Y \in \mathcal{Y}$
from the covariate $X\in \mathcal{X}$. 
We denote $\ell_{tr}(\cdot, \cdot): \Yscr \times \Yscr \mapsto [0,1]$ as the training loss (hinge loss for binary classification) and $\ell(\cdot,\cdot): \Yscr \times \Yscr \mapsto \{0,1\}$ as the accuracy evaluation loss (0-1 loss) in our context unless specified.

\section{Benchmark Setup}\label{sec:data}
As emphasized in \Cref{sec:intro}, this work mainly focuses on \wty{spatiotemporal} distribution shifts in tabular data, which are among the \emph{oldest} and \emph{most widely used} types of data in machine learning and operations research. 
In this section, we first illustrate $Y|X$-shift patterns are overlooked in tabular data. Following that, we introduce the \emph{data and method foundations} of our new benchmark, along with the \emph{key desiderata} that guided its design and setup in Sections~\ref{subsec:data-foundation} and~\ref{subsec:method-foundation} respectively. Our benchmarks confirm that $Y|X$-shifts are indeed prevalent in tabular data, addressing shift patterns that are underrepresented in existing datasets.

\begin{figure}[!htp]
  \centering\includegraphics[width=\textwidth]{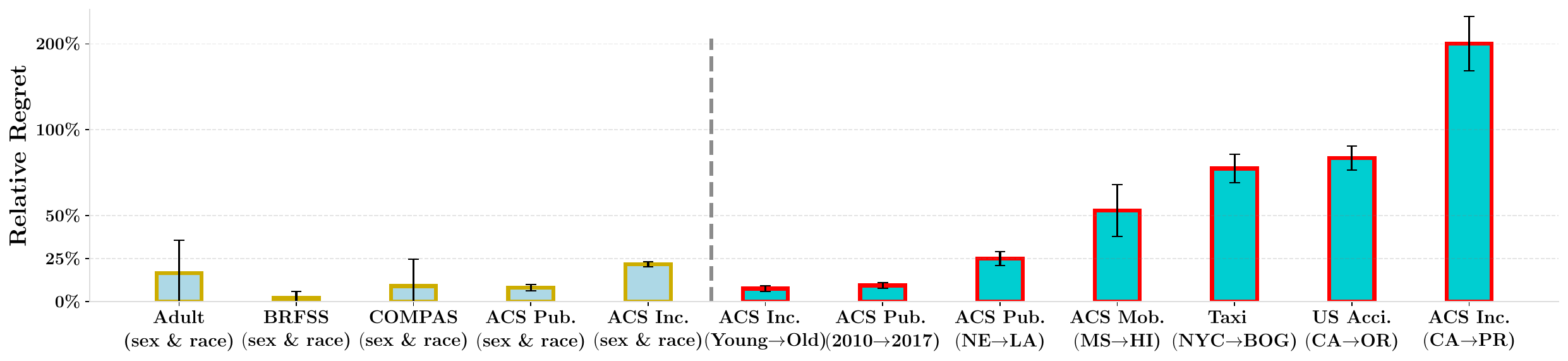}
    \caption{\textit{Relative regret} in typical benchmarks for algorithmic robustness~\citep{Dua:2019, gardner2022subgroup} (\emph{left 5 bars}) and seven settings designed in our benchmark in Settings 1 - 7 (\emph{right 7 bars}). We use XGBoost as $\Fscr$ here for illustration.}
    \label{fig:regret}
  \end{figure}

\vspace{-0.15in}
\paragraph{Lack of diverse shift patterns in existing tabular benchmarks.}
Before introducing our benchmark setup, we first demonstrate that existing tabular benchmarks for distribution shifts tend to implicitly focus on $X$-shifts~\citep{recht2019imagenet, koh2021wilds, zhang2023nico, santurkar2021breeds, DBLP:conf/nips/YaoCC0KF22}, leaving a gap in the empirical foundation for studying a broader range of distribution shifts. 
Specifically, common methods for evaluating algorithmic robustness typically concentrate on demographic subgroups within datasets such as \texttt{Adult}, \texttt{BRFSS}, \texttt{COMPAS}, \texttt{ACS Public Coverage}, and \texttt{ACS Income}~\citep{robert2020, yu2022, gardner2022subgroup}.
To examine the shift patterns among demographic groups used in previous studies, we treat the largest demographic group as the source distribution $P$ and the smallest as the target distribution $Q$, simulating demographic shifts. 
For example, in the \texttt{Adult} dataset, $P$ represents white men, while $Q$ represents non-white women. 
We then measure the model's performance gap between the source and target using \emph{relative regret}, defined as the optimality gap of a model $f_P$ trained on $P$ when evaluated on $Q$:
\begin{equation}
  \label{eqn:regret}
  \frac{\mathbb{E}_Q[\ell(f_P(X), Y)]}{\min_{f\in \Fscr} \mathbb{E}_Q[\ell(f(X), Y)]} - 1,
  ~~~\mbox{where}~~~
  f_P \in \argmin_{f\in \Fscr} \mathbb{E}_P [\ell(f(X), Y)].
\end{equation}
For these widely-used benchmarks, the relative regret is small (left 5 bars in \Cref{fig:regret}), indicating that the $Y|X$ distribution is \emph{largely transferable} across these demographic groups. 
This underscores once again that existing widely-used benchmarks are largely confined to $X$-shifts.

In response, we propose a more comprehensive benchmark that captures a wider range of distribution shift patterns, surpassing the limitations of traditional benchmarks that will not be further considered in this paper. 
As demonstrated below, our benchmark incorporates distribution shifts in tabular datasets covering socioeconomic and physical systems, with varying degrees of $Y|X$-shifts (\Cref{fig:regret}).

\subsection{Data Foundations}\label{subsec:data-foundation}

We design our benchmark to cover diverse prediction tasks in \emph{socioeconomic}, \emph{traffic}, \wty{\emph{education} and \wty{\emph{healthcare} systems}}, each of which inherently reflects \wty{spatiotemporal} distribution shifts. These tasks include:
\begin{itemize}[leftmargin=*]
	\vspace{-0.1in}
	\item \textbf{Socioeconomic systems}: These tasks focus on predicting important socioeconomic outcomes for individuals based on demographic, geographic, and employment features. The datasets are derived from the U.S. Census (nationwide ACS PUMS)~\cite{ding2022retiring} from all 50 states and Puerto Rico, covering various years and diverse demographic groups. 
		\begin{itemize}
			\vspace{-0.1in}
			\item \texttt{ACS Income} Dataset: Predict whether an individual’s income exceeds \$50K.
			\item \texttt{ACS Mobility} Dataset: Predict whether an individual has the same residential address as one year ago.
			\item \texttt{ACS Pub.Cov} Dataset: Predict whether an individual has public health insurance.
		\end{itemize}
	\vspace{-0.15in}
	\item \textbf{Traffic systems}: These tasks involve predicting outcomes related to transportation and road safety, emphasizing shifts in traffic patterns and accident severity across geographic regions.
		\begin{itemize}
			\vspace{-0.1in}
			\item \texttt{Taxi} Dataset~\citep{taxidata,taxidata1}: Predict whether the total ride duration time will exceed 30 minutes based on location and temporal features.
			\item \texttt{US Accident} Dataset~\citep{accident1, accident2}: Predict whether an accident is severe (long delay) or not (short delay) based on weather features and road condition features.
		\end{itemize}
        \vspace{-0.15in}
    \wty{\item \textbf{Education systems}: These tasks involve predicting student performance or institution's low degree completion rate using institutional and individual-level data. The datasets encompass records from multiple educational institutions, capturing distribution shifts that occur across different organizational contexts (e.g., schools and institution types).
    \begin{itemize}
        \item \texttt{ASSISTments} Dataset~\citep{assist}: Predict whether a student could answer a problem on the ASSISTments tutoring platform correctly based on student-, problem-, and school-level features.
        \item \texttt{College Scorecard} Dataset~\citep{college}: Predict whether an institution has a low completion rate, based on other characteristics of that institution.
    \end{itemize}
    \item \textbf{Healthcare systems}: The task involves predicting patients' diagnoses based on their features, where we focus on demographic shifts in order to investigate race-based disparities in treatment for diseases.
    \begin{itemize}
        \item \texttt{Diabetes} Dataset~\citep{diabetes}: Predict whether the respondent has ever been told they have diabetes based on known indicators for diabetes including general physical health, high cholesterol, BMI/obesity, smoking, chronic health conditions, and demographic indicators.
    \end{itemize}}
\end{itemize}
\subsubsection{Domain Selection}
\paragraph{Desiderata for domain selection.}
To comprehensively evaluate model performance across diverse \twy{semi-synthetic} distribution shifts, it is essential to establish clear criteria for selecting both source and target domains. Our benchmark prioritizes several key factors:
\begin{itemize}
	\vspace{-0.1in}
	\item[1.] Geographic diversity: 
 We assess models under various environmental contexts and urban-rural dynamics by including data from different regions.
	\vspace{-0.1in}
	\item[2.] Demographic diversity: We evaluate how well each model generalizes across different groups characterized by demographic factors, such as age and population composition.
	\vspace{-0.1in}
	\item[3.] Temporal changes: We examine the model performance influenced by evolving socioeconomic conditions through data from multiple time periods.
	\vspace{-0.1in}
	\item[4.] Socioeconomic diversity: We evaluate models from individual data representing varying economic conditions characterized by different income levels \wty{and education institutions}.
\end{itemize}
By adhering to these principles in our domain selection, we ensure that our evaluations are realistic and comprehensive, ultimately leading to robust insights into model behavior under diverse conditions.

\begin{table}[h]
\caption{Overview of datasets and \wty{10 settings}}
\label{table:overview}
\resizebox{\textwidth}{!}{
\begin{tabular}{@{}lllllcll@{}}
\toprule 
Desiderata & ID & \multicolumn{1}{l}{Dataset} & Source Domain& \multicolumn{1}{l}{\begin{tabular}[l]{@{}l@{}}Selected\\ Target Domain\end{tabular}} & \multicolumn{1}{c}{\begin{tabular}[l]{@{}l@{}}Total Number\\of Target Domains\end{tabular}}  & Shift Pattern & Outcome \\ \midrule
\multirow{5}{*}{Geographic} 
& 1  &  \texttt{ACS Income} & California (CA) &  Puerto Rico (PR)&  50                                                                             &   $Y|X\gg X$    & Income$\geq$50k         \\
& 2   & \texttt{ACS Mobility}  & Mississippi (MS) &   Hawaii (HI) & 50 &   $Y|X\gg X$ & Residential Address                                                                                     \\
& 3   &  \texttt{Taxi}  & New York City (NYC)  & Botogá (BOG) & 3                                                                                      &  $Y|X\gg X$ & Duration$\geq$ 30 min             \\ 
& 4   &  \texttt{ACS Pub.Cov}  &  Nebraska (NE)  & Louisiana (LA) & 50                                                                                      &  $Y|X> X$ & Public Ins. Coverage             \\ 
& 5   &  \texttt{US Accident}  & California (CA)  & Oregon (OR) & 12                                                                                       &  $Y|X> X$ & Severity of Accident             \\ \midrule
Temporal & 6 & \texttt{ACS Pub.Cov} & 2010 (NY) & 2017 (NY) & 3 & $Y|X<X$ & Public Ins. Coverage \\\midrule
\multirow{2}{*}{Demographic} & 7 & \texttt{ACS Income} & Younger & Older & 1 & $Y|X\ll X$ & Income$\geq$50k \\
&8 & \texttt{Diabetes} & White & Others & 1 & $Y|X\gg X$  & Been told have diabetes\\ \midrule 
\multirow{2}{*}{Socioeconomic} & 9 & \texttt{ASSISTments} & 700 Schools & 1 New School & 1 & $Y|X > X$ & Next Answer Correct \\ 
&10 &\texttt{College} & \twy{Mainstream Institutions} & \twy{Non-traditional Institutions} & 1 & $Y|X > X$ & Completion rate $\geq 50\%$ \\ \midrule
\textbf{Total} & & \textbf{\wty{8} datasets} &\textbf{\wty{10} settings} & \begin{tabular}[l]{@{}l@{}}\bf \wty{10} \textit{selected}\\ \bf source-target pairs\end{tabular} &\begin{tabular}[l]{@{}l@{}}\bf \wty{172} \textit{all}\\ \bf source-target pairs\end{tabular}  & & \textbf{8 prediction goals}\\\bottomrule
\end{tabular}
}
\vspace{-0.15in}
\end{table}

\paragraph{Source-target domain selection.}
In alignment with the desiderata, we carefully design seven settings, summarized in \Cref{table:overview}, to capture diverse distribution shift patterns:\\
(i) \emph{Geographic diversity}: In \textbf{Settings 1-5}, we select California (CA) and New York (NY) as source domains representing wealthier states, alongside Mississippi (MS) and Nebraska (NE) to capture regions with lower income levels. For target domains, we include other states and regions (e.g., 50 target domains for Settings 1, 2, and 4; 3 target domains for Setting 3; and 13 target domains for Setting 5). This selection spans a broad range of socioeconomic conditions and geographic diversity across the US.\\
(ii) \emph{Temporal diversity}: In \textbf{Setting 6}, using individuals from New York in the \texttt{ACS Public Coverage} dataset, we construct temporal shifts by considering source data from 2010 and target data from 2014, 2017, and 2021.\\
(iii) \emph{Demographic diversity}: In \textbf{Setting 7}, we sub-sample the dataset by age, focusing on individuals from California. We form two groups based on whether their age is $\geq 25$, with the source data heavily skewed towards the older group (80\% from Age $\geq 25$) and the target data reversing this distribution (20\% from Age $\geq 25$). \wty{In \textbf{Setting 8}, we sub-sample the dataset by race and form two groups based on racial categories, with the source data consisting of individuals identified as ``White non-Hispanic'' and the target data consisting of individuals from all other race groups.}\\
\twy{(iv) \emph{Socioeconomic diversity}: In \textbf{Settings 9–10}, using student-level data from the \texttt{ASSISTments} and \texttt{College} datasets, we illustrate socioeconomic diversity through institutional context, which serves as a proxy for differences in student background and access to educational resources.
In \textbf{Setting 9} (\texttt{ASSISTments}), we randomly partition schools to capture school-level heterogeneity, reflecting indirect socioeconomic variation across educational institutions.
In \textbf{Setting 10} (\texttt{College}), we separate students enrolled in mainstream degree-granting institutions (e.g., standard public or private colleges and universities) from those enrolled in special-focus or non-traditional institutions (e.g., special-focus institutions, associate’s colleges, or other non-traditional institution types), as defined by the Carnegie Commission on Higher Education. These institutional categories are known to be associated with systematic differences in student socioeconomic composition.}

Overall, we design \emph{\wty{10} settings} that encompass various types of shifts, each with one source domain and multiple target domains, resulting in \emph{\wty{172} source-target pairs}. 
For ease of presentation in some of the following analyses (\Cref{subsec:acc-on-the-line}), we also selected a single target domain in each setting that exhibits relatively large distribution shifts, referred to as the \emph{\wty{10} selected source-target pairs}.

\subsubsection{Shift Pattern Analysis}
\begin{figure}[!htp]
\centering\includegraphics[width=0.9\textwidth]{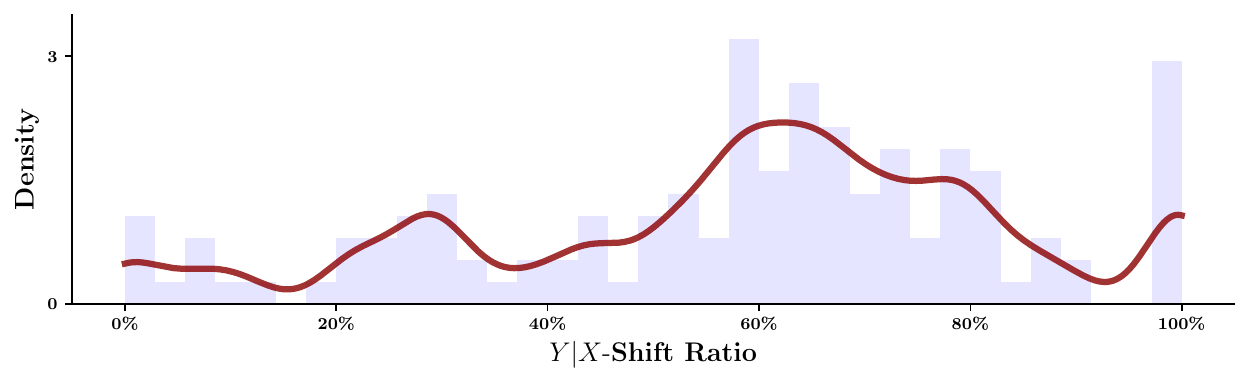}
	\caption{Histogram of the percentage of performance degradation from source to target pairs attributed to $Y|X$-shifts (for pairs with degradation larger than 5 percentage points).}
	\label{fig:decomposition}
 \vspace{-0.2in}
\end{figure}

Before proceeding to the method foundations in our benchmark, we first examine the distribution shift patterns in \Cref{table:overview} to assess their comprehensiveness.
\vspace{-0.1in}
\paragraph{Prevalence of $Y|X$-shifts.}
Considering all the \wty{172} source-target pairs in~\Cref{table:overview}, we find that performance degradation under shifts is overwhelmingly attributed to $Y|X$-shifts.
For each source-target pair, we use DISDE~\citep{namkoong2023diagnosing} to attribute the total performance drop to $X$-shifts and $Y|X$-shifts (more details of DISDE and calculations of the $Y|X$-Shift-Ratio can be found in Appendix~\ref{app:disde}).
Among source-target pairs whose performance degradation is larger than $8$ percentage points (67 out of 172 pairs), 87.2\% of them have over 50\% of the performance degradation attributed to $Y|X$-shifts (and 70.2\% of them have over 60\% of the gap attributed to $Y|X$-shifts).  
Among pairs with degradation larger than $5$ percentage points (133/172), we plot a histogram of the percentage of performance degradation attributed to $Y|X$-shifts in~\Cref{fig:decomposition}. 
This phenomenon illustrates the prevalence of $Y|X$-shifts in spatiotemporal distribution shifts, underscoring the importance of accounting for $Y|X$-shifts when evaluating model robustness to more accurately reflect real-world performance. \wty{Furthermore, \citet{zeng2024llm} examine 7,650 source-target pairs under Settings 1, 2, and 4 of the \texttt{WhyShift} dataset, treating each state in turn as the source domain. They find that most of the observed performance degradation stems from $Y|X$-shifts.}

\vspace{-0.1in}
\paragraph{Comprehensive coverage of $Y|X$-shift degrees.}
As for the 10 selected source-target pairs, as shown in \Cref{fig:regret}, we demonstrate that they exhibit varying levels of relative regret, indicating different degrees of $Y|X$-shifts. In contrast to widely used benchmarks that primarily focus on demographic shifts, our benchmark encompasses a much broader range of distribution shift patterns, making it a more robust testbed for evaluating model robustness.

\subsection{Method Foundations}\label{subsec:method-foundation}
We now introduce the methodological foundations of our benchmark.
\vspace{-0.15in}
\paragraph{Method setup and goal under distribution shifts.}
Our goal is to evaluate model performance under distribution shifts, specifically focusing on out-of-distribution generalization. Here, the model \(\hat f\) is trained on data from the source domain \(P\), \wty{validated on limited data from the target domain $Q$,} and tested on the target domain \(Q\). 

The objective of a machine learning model is to minimize the loss on the target distribution \(Q\):
\vspace{-0.1in}
\begin{equation*}\label{eq:tgt-src-shift}
    \E_{Q}[\ell(\hat f(X);Y)]. 
\vspace{-0.1in}
\end{equation*}
When \(\ell(\cdot,\cdot)\) represents the 0-1 loss, this is equivalent to maximizing target accuracy, i.e.,
\begin{equation}\label{eq:src-tgt-acc-def}
    \underbrace{1 - \mathbb E_{Q}[\ell(\hat f(X);Y)]}_{\text{Target Accuracy}} = \underbrace{(1 - \E_{P}[\ell(\hat f(X);Y)])}_{\text{Source Accuracy}} + \underbrace{(\E_{P}[\ell(\hat f(X);Y)] - \E_{Q}[\ell(\hat f(X);Y)])}_{\text{Performance Gap}}.
\end{equation}
\tyw{Here, evaluation based on the target accuracy is more suitable compared with the source accuracy or performance gap as it reflects the model's performance under distribution shifts in practice.}

\vspace{-0.15in}
\paragraph{Desiderata for method selection.} 
To comprehensively assess model performance under distribution shifts, our benchmark incorporates a variety of methods chosen from the following aspects. 
\begin{itemize}[leftmargin=*]
    \vspace{-0.1in}
    \item[1.] \emph{Diverse learning algorithms}: For illustration, we first discuss the standard ERM method:
 \begin{equation}\label{eqn:general-erm}
     \min_{f\in \Fscr} \E_{\widehat P}[\ell_{tr}(f(X),Y)],
 \end{equation}
where $\what{P}$ is the training empirical distribution,  $\Fscr$ is the model class, and $\ell_{tr}(\widehat y, y)$ is the loss used to fit the model, which is the hinge loss for binary classification unless specified. 

    Apart from ERM, we design different branches of algorithms to improve model's performance under distribution shifts, including distributionally robust optimization (DRO), imbalanced learning methods and fairness-enhancing methods. 
    We outline the key learning algorithms studied here:
    \begin{itemize}[leftmargin = *]
        \item[(\rm{i})] DRO methods:  DRO methods optimize the worst-case loss over an \emph{ambiguity set} $\mc{P}$:
        \vspace{-0.1in}
        \begin{equation}\label{eqn:general-dro}
            \min_{f \in \mc{F}}
            \sup_{Q \in \mc{P}} \E_Q[\loss_{tr}(f(X), Y)],\quad \text{where } \mc{P}(d, \epsilon) = \left\{Q: d(Q, \widehat P) \le \epsilon
            \right\}.
            \vspace{-0.1in}
        \end{equation}
        The ambiguity set is usually modeled as a perturbed version of the training distribution, $d(\cdot, \cdot)$ is a notion of distance between probability measures, and $\epsilon$ is the \emph{radius} of set.  
        \tyw{We reiterate that some existing DRO methods, designed for worst-case protections, are not specifically designed to handle \twy{semi-synthetic} distribution shifts. For example, the Holistic-DRO primarily focuses on defending against adversarial attacks \citep{bennouna2023certified}. However, these models remain the gold standard for training when data samples from the target domain are scarce. Understanding the performance -- when and why they fail -- offers valuable insights for further improvement.}
        We consider different choices of \emph{ambiguity sets} $\mc{P}$ (or distance functions $d(\cdot,\cdot)$) and different \emph{radii} $\epsilon$ of the set.
        \item[(\rm{ii})] Imbalanced learning methods: Imbalanced learning methods balance the data first via sample reweighting or sub-sampling and show competitive performance when group portions or labels shift \citep{idrissi2022simple}. We consider different {balancing objectives} and {procedures}.
        \item[(\rm{iii})] Fairness-enhancing methods: Fairness-enhancing methods adjust data or model outputs by applying fairness constraints and even help improve accuracy when the source data is biased \cite{blum2019recovering}. We consider different types of fairness constraints as well as various adjustment approaches.
    \end{itemize}
    \item[2.] \emph{Diverse model classes}: Besides learning algorithms, diverse model classes need to be incorporated. We set $\Fscr$ as linear models, tree-based ensemble models, and neural networks.
\end{itemize}
\vspace{-0.15in}
\paragraph{Method overview.} Building on the desiderata for diverse model classes and learning algorithms, we selected \wty{45} methods that cover a wide range of learning approaches and model classes for tabular data. These methods are organized into eight categories: basic ERM methods (6 methods), tree-based ensemble methods (4 methods), DRO applied to linear models (linear-DRO, 13 methods), \wty{DRO applied to kernel models (kernel-DRO, 4 methods), DRO applied to tree-based ensemble (tree-DRO, 4 methods)}, DRO applied to neural networks (NN-DRO, 8 methods), imbalanced learning methods (4 methods), and fairness-enhancing methods (2 methods).
\begin{enumerate}[leftmargin=*]
\vspace{-0.1in}
	\item \emph{Basic ERM methods}: 
    We include diverse model classes optimized via ERM~\eqref{eqn:general-erm}: Linear Support Vector Machine  (SVM), where $\Fscr$ denotes the linear class and $\ell_{tr}$ denotes the hinge loss for binary classification; \wty{Kernel Support Vector Machine (Kernel SVM), where $\Fscr$ denotes the kernel class and $\ell_{tr}$ denotes the hinge loss for binary classification;} Logistic Regression (LR), where $\Fscr$ denotes the linear class and $\ell_{tr}$ denotes the cross entropy loss for binary classification; fully-connected feedforward neural networks (NN, a.k.a. multi-layer perceptrons), where $\Fscr$ denotes the neural network class \wty{and NN-2, NN-3, NN-4 are three methods to distinguish the neural network class with the number of layers being 2, 3, 4.}  
\vspace{-0.1in}
 	\item \emph{Tree-based ensemble methods}: We include tree-based ensemble methods in~\eqref{eqn:general-erm} with $\Fscr$ set as Random Forest (RF)~\citep{breiman2001random}, Gradient-Boosting Machine (GBM)~\citep{natekin2013gradient}, Light Gradient-Boosting Machine (LGBM)~\citep{ke2017lightgbm}, XGBoost (XGB)~\citep{chen2016xgboost};
\vspace{-0.1in}
	\item \emph{Linear-DRO methods}:  We include DRO methods~\eqref{eqn:general-dro}, where $\Fscr$ denotes the linear class, under different ambiguity sets. Specifically, we include Wasserstein-DRO~\citep{blanchet2019data} (where $d$ is the Wasserstein distance with the associated distance allowing perturbations only in $X$), Augmented Wasserstein-DRO~\citep{shafieezadeh2019regularization} (where $d$ is the Wasserstein distance with the associated distance allowing perturbations in both $X$ and $Y$), Sinkhorn-DRO~\citep{wang2021sinkhorn} (where $d$ is the Sinkhorn distance), Unified-DRO~\citep{blanchet2023unifying} with $L_2$-norm, and with $L_{\text{inf}}$-norm~\citep{blanchet2023unifying} (where $d$ is the so-called optimal transport discrepancy with conditional constraints in Definition 2.1 there), CVaR-DRO~\citep{rockafellar2000optimization} (where $d$ is the so-called CVaR distance defined in \Cref{app:dro-setup}), $\chi^2$-DRO~\citep{duchi2021learning} (where $d$ is the $\chi^2$-divergence), TV-DRO~\citep{jiang2018risk} (where $d$ is the Total Variation distance), KL-DRO~\citep{hu2013kullback} (where $d$ is Kullback-Leibler divergence); we also include other DRO methods such as Satisficing Wasserstein-DRO~\citep{long2023robust}, Holistic-DRO~\citep{bennouna2022holistic}, \wty{Marginal-CVaR-DRO~\citep{duchi2023distributionally} and Conditional-CVaR-DRO~\citep{sahoo2022learning}} beyond the ambiguity set with the form of~\eqref{eqn:general-dro}. 
    \vspace{-0.1in}
    \wty{\item \emph{Kernel-DRO methods}: We include DRO methods~\eqref{eqn:general-dro}, where $\Fscr$ denotes the kernel class. Specifically, we set $d$ as Wasserstein distance, CVaR distance, $\chi^2$-divergence, Kullback-Leibler divergence.}
    \vspace{-0.1in}
    \wty{\item \emph{Tree-based DRO methods}: We include DRO methods~\eqref{eqn:general-dro}, where $\Fscr$ denotes the tree-based model class (LGBM and XGB). Specifically, we set $d$ as the CVaR distance and Kullback-Leibler divergence.}
\vspace{-0.1in} 
  \item \emph{NN-DRO methods}: We include DRO methods~\eqref{eqn:general-dro}, with $\Fscr$ denoting the neural network class \wty{while varying the number of layers being 2, 3, 4} and $\ell_{tr}$ denoting the cross-entropy loss for binary classification, under different ambiguity sets. Specifically, we include CVaR-DRO (NN) and $\chi^2$-DRO (NN) with $d$ being the CVaR and $\chi^2$-distance respectively~\citep{levy2020large}. We also include CVaR-DORO (NN) and $\chi^2$-DORO (NN) that are designed for outlier robustness~\citep{zhai2021doro}. 
\vspace{-0.1in}   
    \item \emph{Imbalanced learning methods}: We include Subsampling-$Y$/$G$ (SUBY / SUBG) and Reweighting-$Y$/$G$ (RWY / RWG)~\citep{idrissi2022simple}, which reweight or sub-sample data to balance the samples belonging to different outcomes ($Y$) or different demographic groups ($G$) and $\Fscr$ in these methods are set as XGB due to its superior performance on tabular data~\citep{grinsztajn2022why}. 
\vspace{-0.1in}
	\item \emph{Fairness-enhancing methods}: We incorporate different fairness constraints, including in-processing methods~\citep{agarwal2018reductions} with demographic parity, equal opportunity, and error parity as constraints, and post-processing methods~\citep{hardt2016equality} with exponential and threshold controls, where $\Fscr$ in these methods are set as XGB.
\end{enumerate}
We refer details of DRO methods to Appendix~\ref{app:dro-setup} and other methods to Appendix~\ref{app:alg-implement}. We do not implement DRO methods with tree-based ensembles as it is not clear how to adapt worst-case approaches under such a model class. We leave this as a direction of future work. For methods requiring demographic labels (SUBG, RWG, and fairness-enhancing methods), we only test them on Settings 1, 2, 4, 6, and 10 (based on \texttt{ACS Income, Pub.Cov, Mobility} and \texttt{College} datasets) and use the ``Sex'' feature as the group.

Equipped with the \wty{10} settings in \Cref{table:overview} that represent diverse distribution shift patterns across $X$- and $Y|X$-shifts, we conduct an extensive grid search across its hyperparameters for each method and evaluate the performance of over 60,000 different configurations across 28 methods in total. Each configuration represents a unique set of hyperparameter values. The whole list of hyperparameter grid choices can be seen from \Cref{tab:hparam-grids} in Appendix~\ref{app:hyper}. As examples, we showcase one configuration for $\chi^2$-DRO (linear) and GBM:
\begin{small}
\begin{lstlisting}[label = {config},caption = {One Specific  Configuration in each method}, language = json]
chi2-DRO: { "radius": 0.01} 
GBM: { "n_estimators": 512, "min_samples_split": 8, "min_samples_leaf": 2, "learning_rate": 1.0, "max_depth": 16, "max_features": "sqrt" }  
\end{lstlisting}
\end{small}

\paragraph{Desiderate for evaluation pipeline.} In each setting from \Cref{table:overview}, we assume that we have sufficient labeled samples from the source domain to train a model and \wty{128 labeled samples} from the target domain \wty{used for validating and selecting the optimal hyperparameter configuration}. Unless specified, in each method, we choose the best hyperparameter configuration with the highest out-of-distribution accuracy, as evaluated on these samples from the target domain. 
Note that 128 samples from the target domain are insufficient to train a new model or apply transfer learning procedures (e.g., \cite{bastani2021predicting}) due to high-dimensional features (e.g., 76 features in Setting 1). Therefore, we only utilize these samples for hyperparameter tuning after the initial training process. \wty{Empirically in Setting 1, our proposed evaluation pipeline outperforms alternative pipelines, such as directly training a separate model for each target domain using only the 128 target-domain samples.}

Nonetheless, our evaluation pipeline gives robust methods, such as DRO, an edge against non-robust methods by selecting the ambiguity set size based on the target samples instead of worst-case assumptions. \wty{For DRO methods with limited samples from the target domain, we compare our validation approach against other alternatives based on the upper confidence bound (UCB) approach—i.e., computing the radius of the ambiguity set to ensure that the target distribution falls into the ambiguity set with high probability~\citep{lee2018minimax}. We find that the UCB-based approach tends to underperform due to the overly conservative nature of the resulting ambiguity sets.} We provide training details \wty{and detailed comparison of different evaluation pipelines and DRO validation approaches} in \Cref{app:traindetail}.

\section{Primary Findings}\label{sec:perform-comp}
Based on the \wty{10} settings outlined in \Cref{sec:data}, we conduct a comprehensive investigation of the performance of \wty{45} methods under various degrees of \(Y|X\)-shifts. 
In this section, we first demonstrate that the ``accuracy-on-the-line'' phenomenon does not hold under \(Y|X\)-shifts in tabular data, contrasting sharply with its behavior under \(X\)-shifts. 
We then highlight the challenges faced by existing methods in addressing distribution shifts in practice, including inconsistent algorithmic performance across different methods and the limited improvements offered by robust learning techniques.

\begin{figure}[!htb]
 \centering\captionsetup[subfloat]{labelfont=scriptsize,textfont=scriptsize}
\stackunder[3pt]{\includegraphics[width=0.24\textwidth, height=0.24\textwidth]{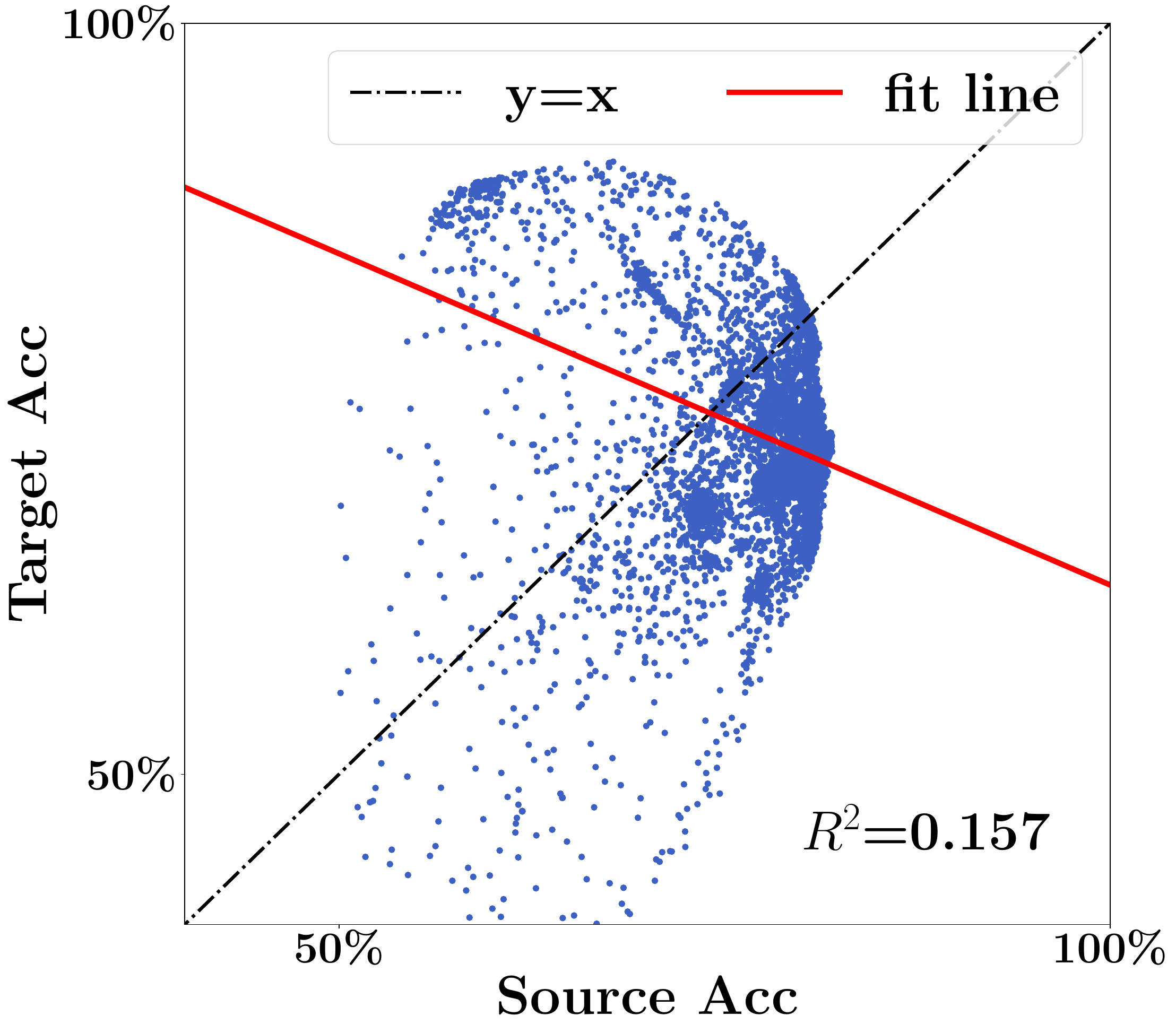}}{\scriptsize \parbox{0.24\textwidth}{\centering (a) \texttt{ACS Income} (CA$\rightarrow$PR)\\  $Y|X$-ratio: 85.4\% }}
\stackunder[3pt]{\includegraphics[width=0.24\textwidth, height=0.24\textwidth]{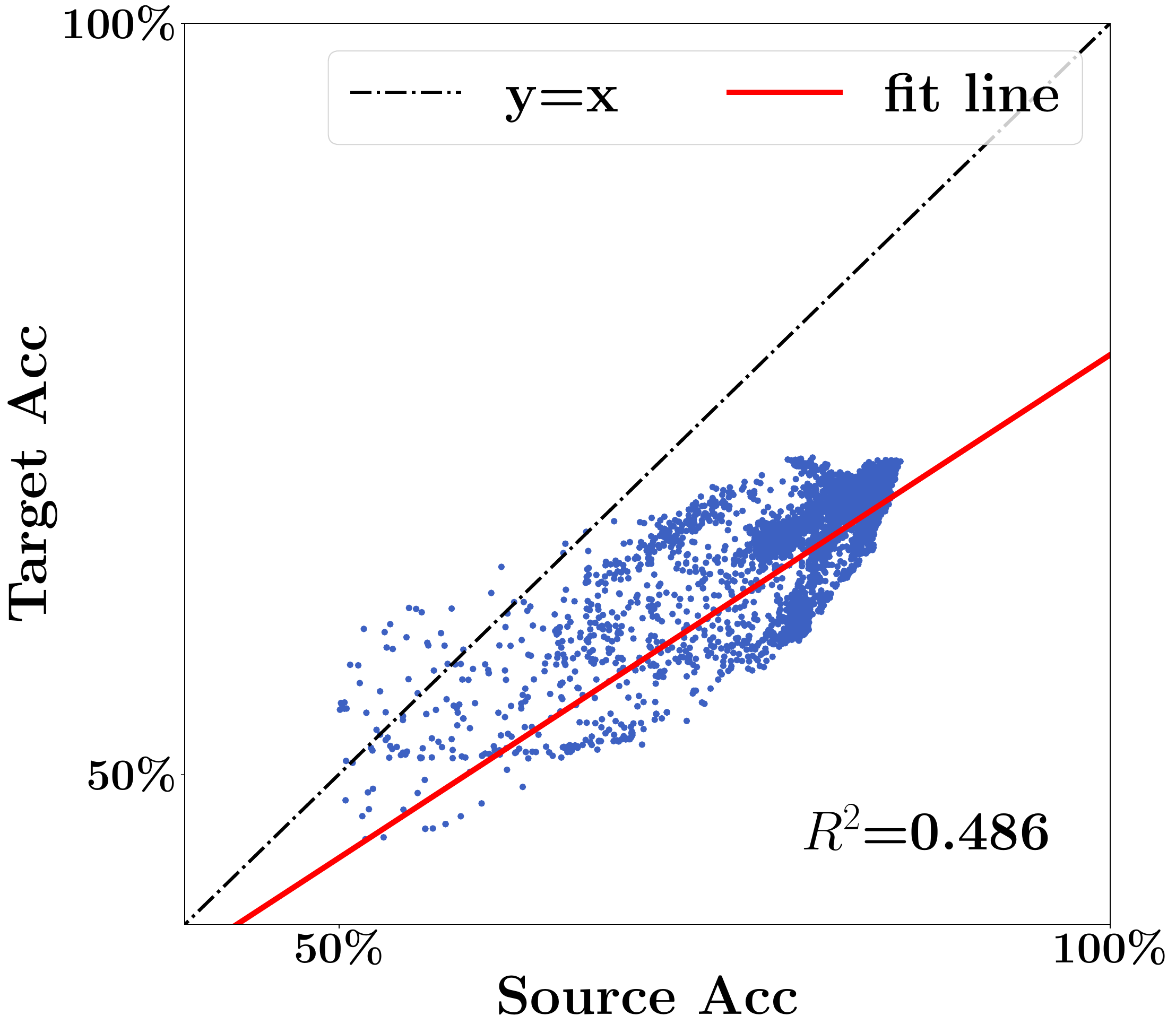}}{\scriptsize \parbox{0.24\textwidth}{\centering (b) \texttt{ACS Pub.Cov} (NE$\rightarrow$LA)\\  $Y|X$-ratio: 61.7\% }}
\stackunder[3pt]{\includegraphics[width=0.24\textwidth, height=0.24\textwidth]{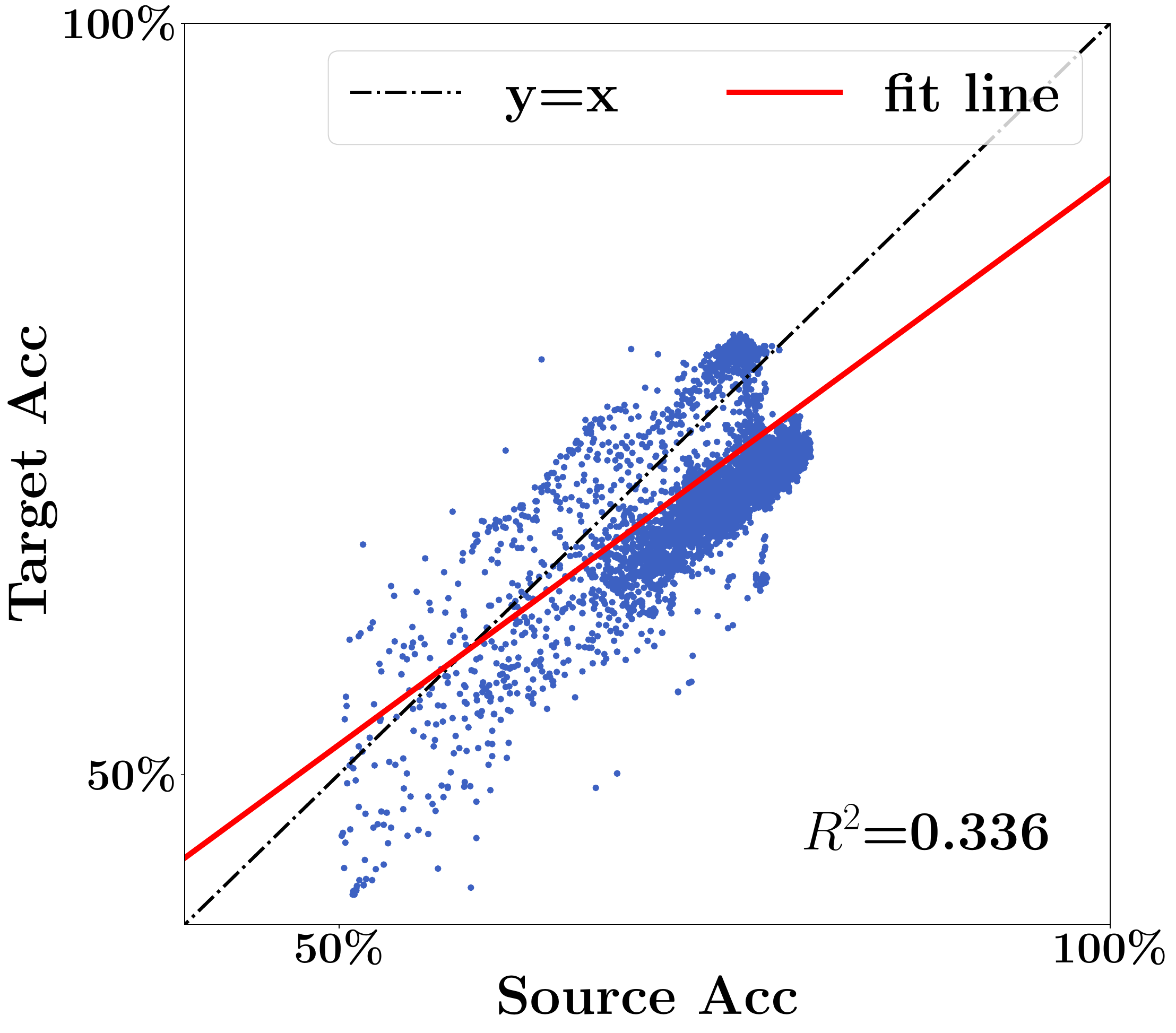}}{\scriptsize\parbox{0.24\textwidth}{\centering (c) \texttt{ACS Mobility} (MS$\rightarrow$HI)\\  $Y|X$-ratio: 68.7\% }}
\stackunder[3pt]{\includegraphics[width=0.24\textwidth, height=0.24\textwidth]{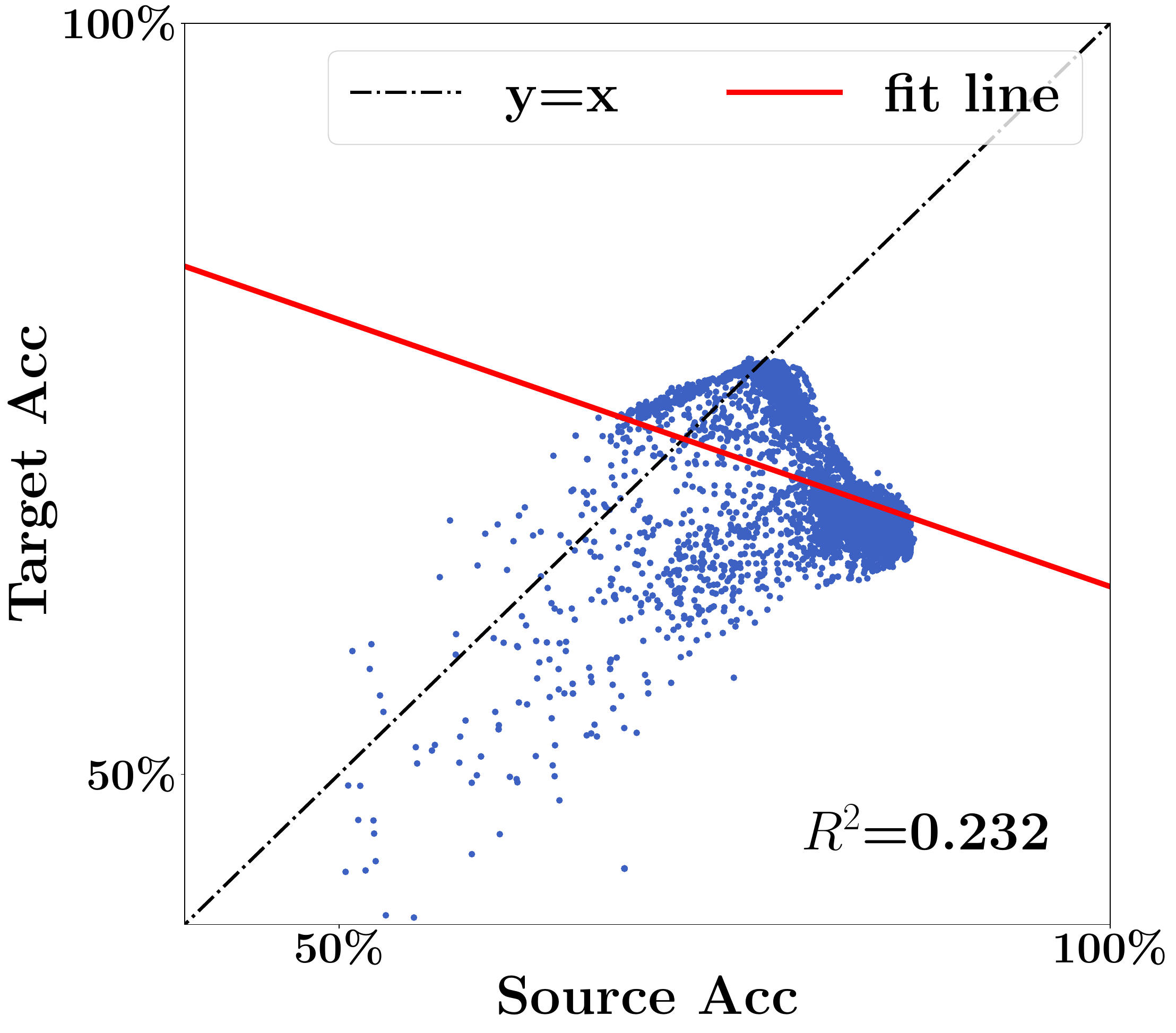}}{\scriptsize\parbox{0.24\textwidth}{\centering (d) \texttt{Accident} (CA$\rightarrow$OR)\\  $Y|X$-ratio: 61.3\% }}\\\vspace{0.1in}

\stackunder[3pt]{\includegraphics[width=0.24\textwidth, height=0.24\textwidth]{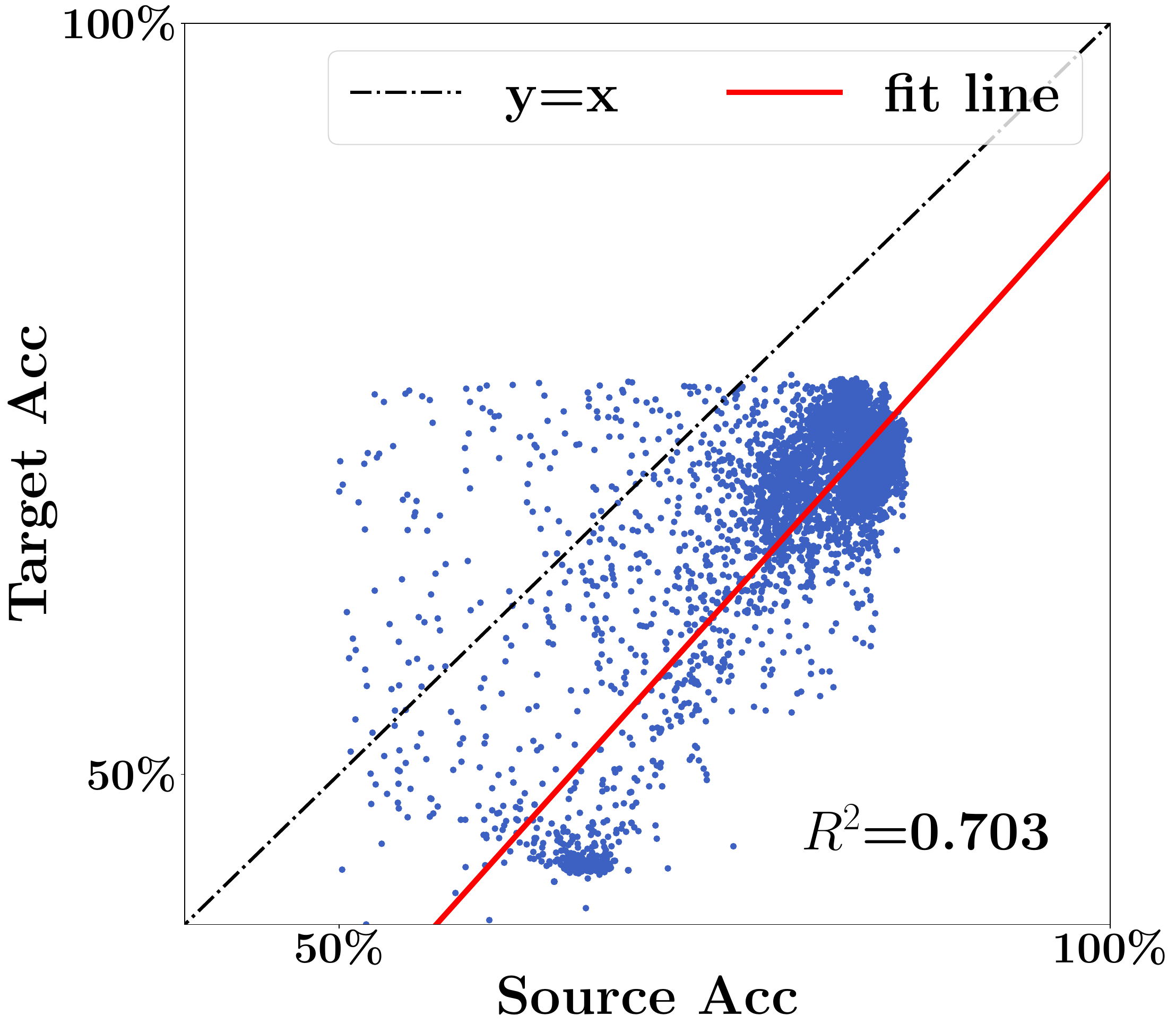}}{\scriptsize \parbox{0.24\textwidth}{\centering (e) \texttt{Taxi} (NYC$\rightarrow$BOG)\\  $Y|X$-ratio: 100\% }}
\stackunder[3pt]{\includegraphics[width=0.24\textwidth, height=0.24\textwidth]{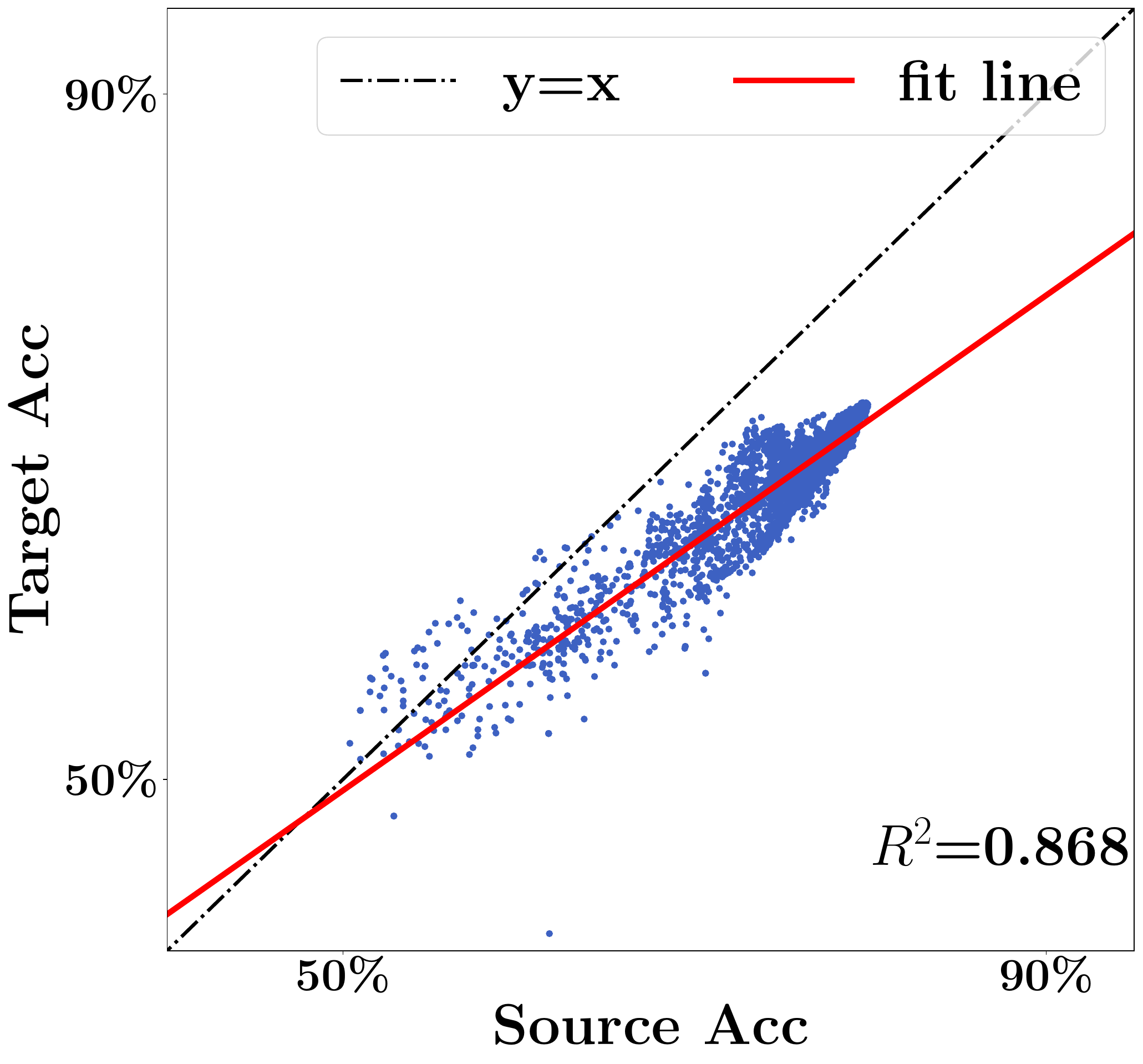}}{\scriptsize \parbox{0.24\textwidth}{\centering (f) \texttt{ACS Pub.Cov} (2010$\rightarrow$2017)\\  $Y|X$-ratio: 13\% }}
\stackunder[3pt]{\includegraphics[width=0.24\textwidth, height=0.24\textwidth]{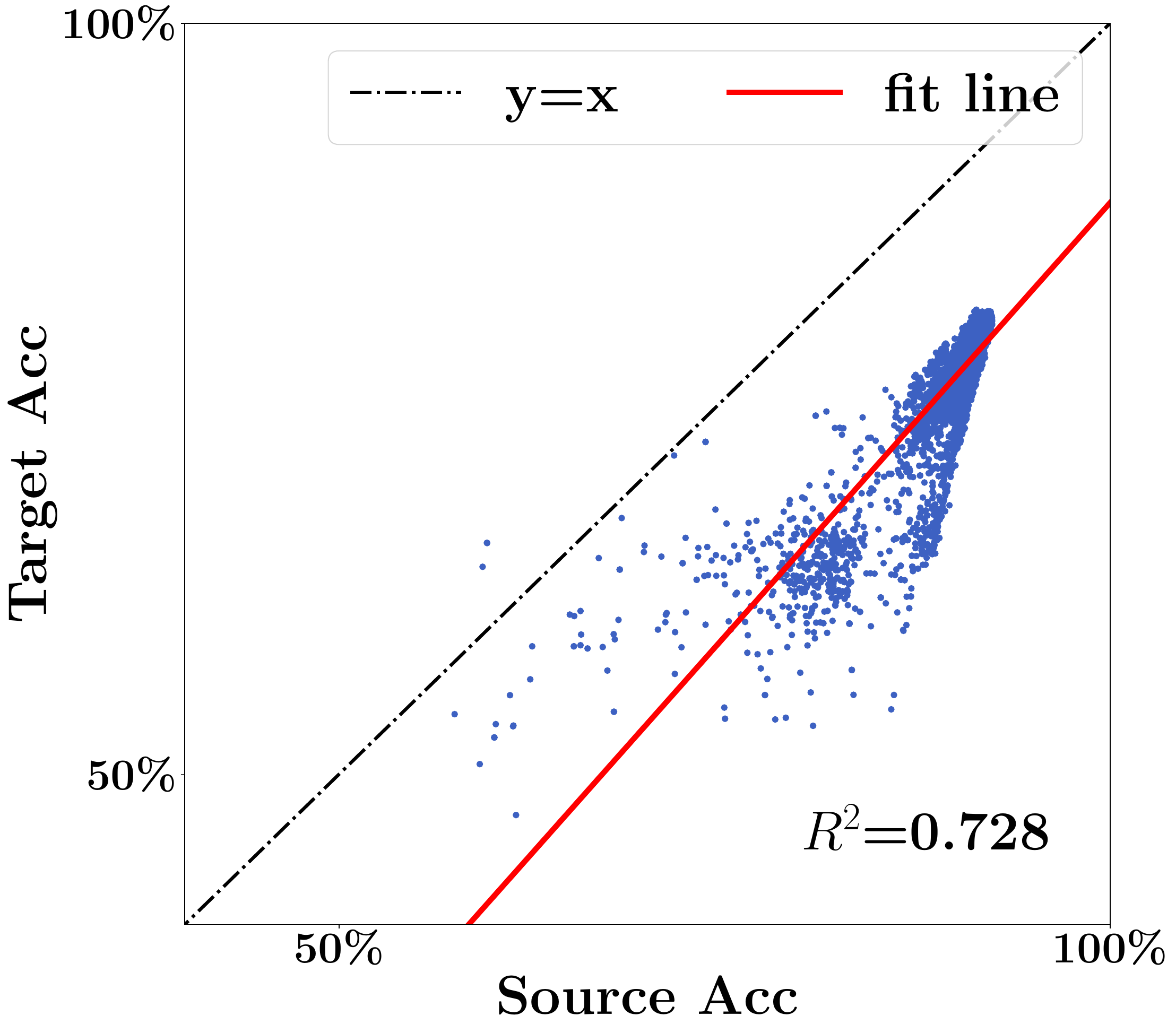}}{\scriptsize\parbox{0.24\textwidth}{\centering (g) \texttt{ACS Income} (Young$\rightarrow$Old)\\  $Y|X$-ratio: 0\% }}
\stackunder[3pt]{\includegraphics[width=0.24\textwidth, height=0.24\textwidth]{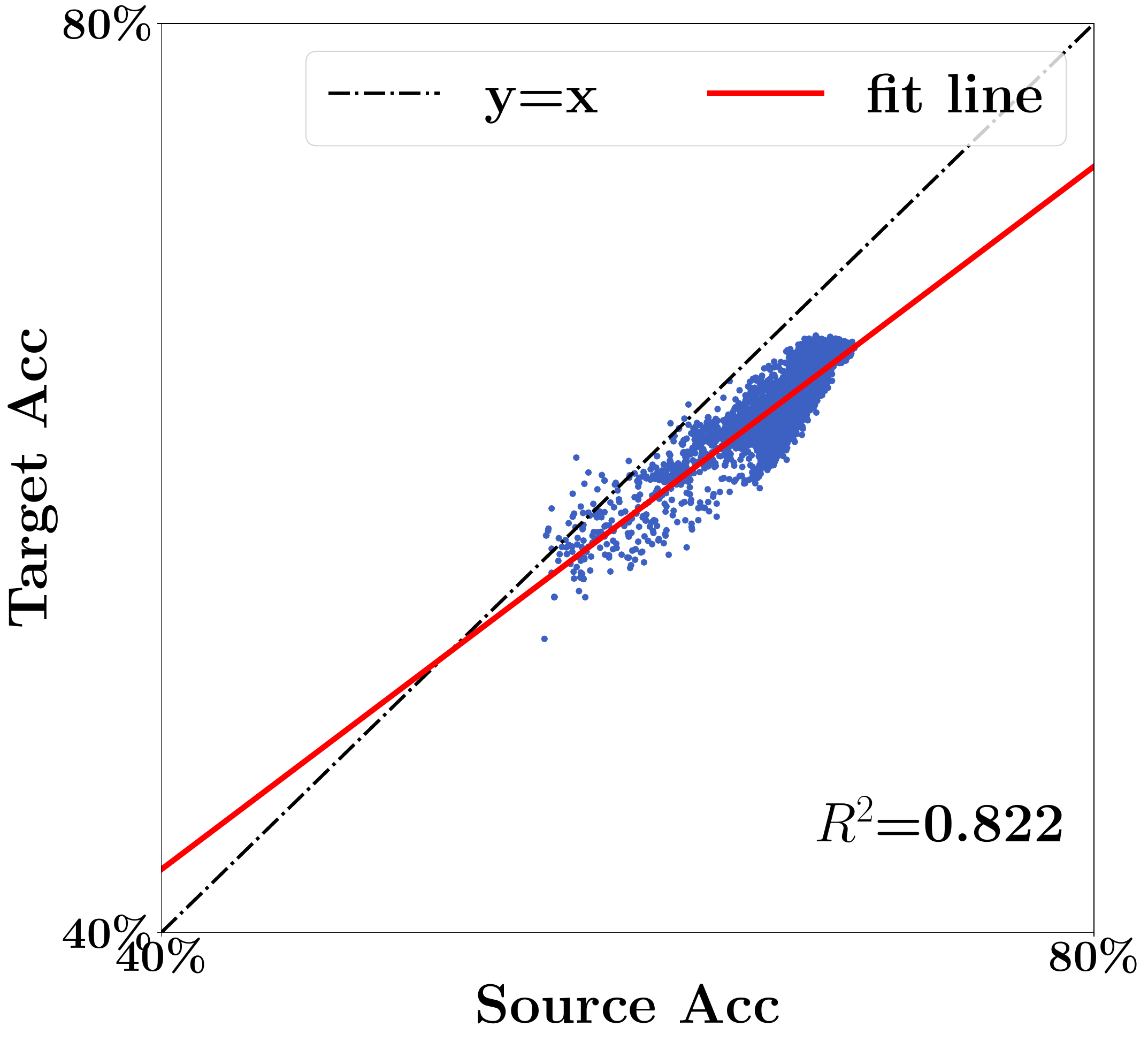}}{\scriptsize \parbox{0.24\textwidth}{\centering (h) \texttt{Diabete} (White$\rightarrow$Other Race Groups)\\  $Y|X$-ratio: 100\% }}\\
\stackunder[3pt]{\includegraphics[width=0.24\textwidth, height=0.24\textwidth]{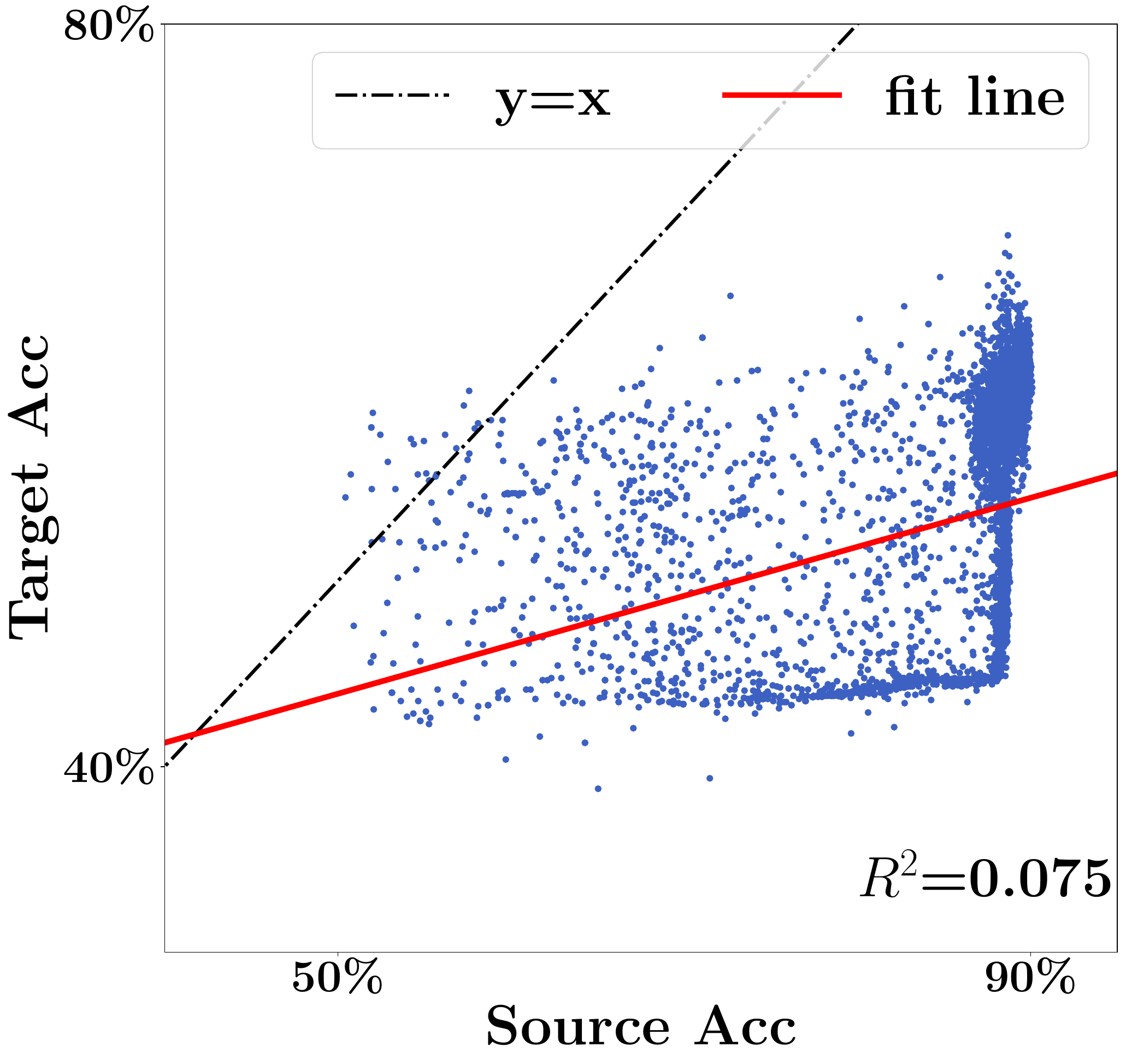}}{\scriptsize \parbox{0.24\textwidth}{\centering (i) \texttt{Assistment} (Source$\rightarrow$Target Schools)\\  $Y|X$-ratio: 56.2\% }}
\stackunder[3pt]{\includegraphics[width=0.24\textwidth, height=0.24\textwidth]{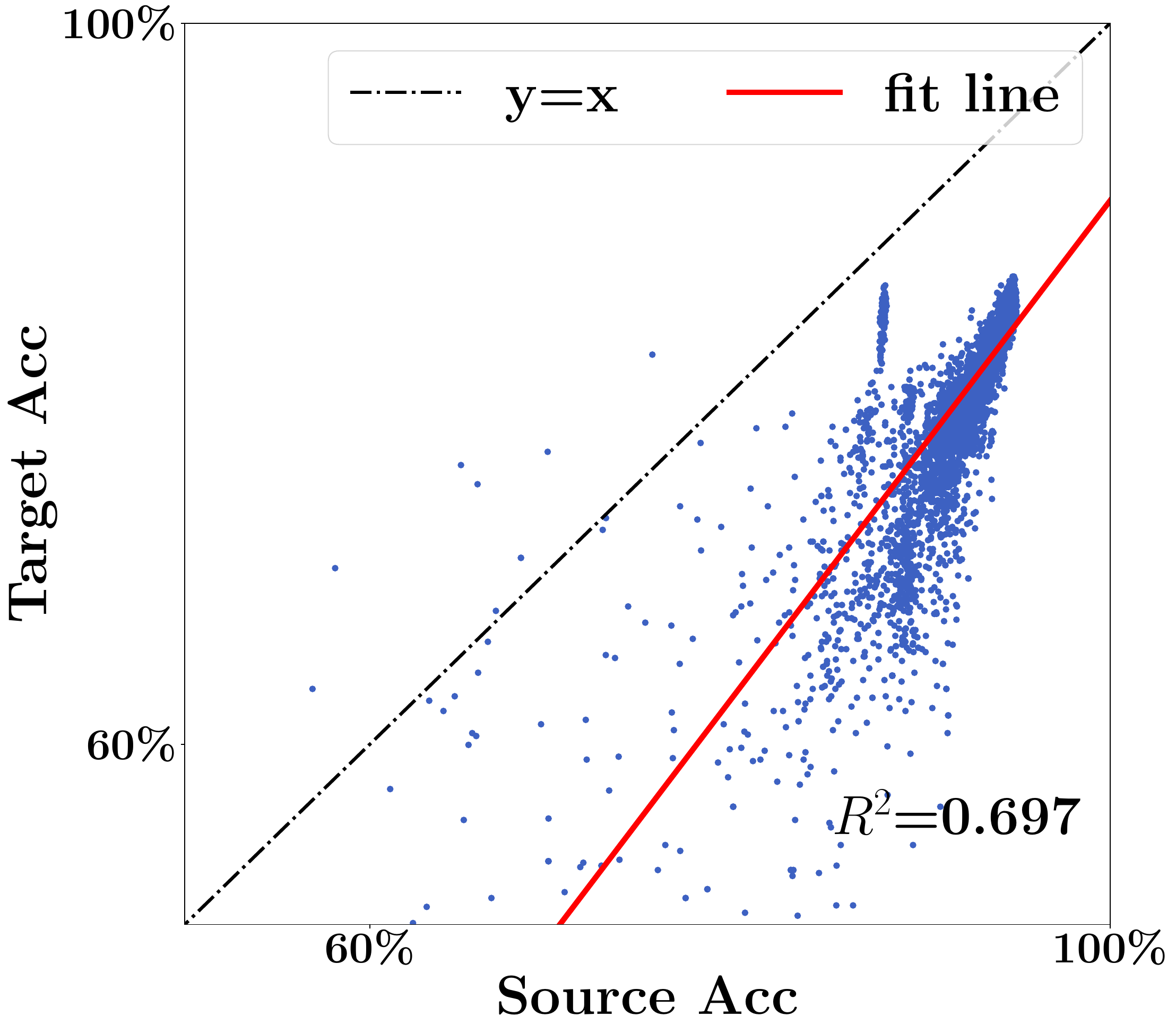}}{\scriptsize \parbox{0.24\textwidth}{\centering (j) \texttt{College} (Source$\rightarrow$Target Institutions)\\  $Y|X$-ratio: 63.3\% }}
\caption{Target ($y$-axis) vs. source ($x$-axis) accuracies for different kinds of methods and datasets. Each sub-figure corresponds to one selected source-target pair in \Cref{table:overview}. In each sub-figure, the corresponding $R^2$ value for a linear fit (red line) is reported on the bottom right. Each blue point represents one hyperparameter configuration of one method.}
 \label{fig:acc-on-the-line}  
\end{figure}

\subsection{Finding 1: Accuracy-on-the-line Fails to Hold Under $Y|X$-shifts}
\label{subsec:acc-on-the-line}

The ``accuracy-on-the-line''  phenomenon recently received  much attention  in the ML literature~\cite{recht2019imagenet, miller2021accuracy}, where there is a linear relationship (up to probit/logit transformations) between source and target performance across almost all prominent \emph{image classification} datasets. 
However, we confirm that the so-called ``accuracy-on-the-line'' phenomenon does not persist across strong $Y|X$-shifts observed in tabular datasets. To quantify the relationship between source and target accuracy, we fit a linear regression line \wty{by regressing target accuracy on source accuracy} using the data points and report the coefficient of determination, $R^2$ \wty{across all 172 pairs in 10 settings.} In~\Cref{fig:acc-on-the-line}(a)-(g), we plot the source accuracy (on the $x$-axis) against the target accuracy (on the $y$-axis) within \wty{one selected pair in} each setting.

\paragraph{Results.} (i) \emph{Strong $Y|X$-shifts}: As shown in~\Cref{fig:acc-on-the-line}(a)-(e), we tend to see poor linear fits (low $R^2$) when the bulk of the performance degradation is attributed to $Y|X$-shifts. 
\tyw{Notably, the $R^2$ of the linear trend on image datasets~\cite{recht2019imagenet, miller2021accuracy} is above 0.8, highlighting the contrast.}
The weak correlation between source and target performance across various methods highlights the challenges of improving performance solely based on source data under \(Y|X\)-shifts. \wty{Although in some settings, e.g., \Cref{fig:acc-on-the-line}(e), the $R^2$ may be large even when $Y|X$-ratio is large, we demonstrate that the variability of $R^2$ is high when $Y|X$-ratio is large.} \wty{In \Cref{tab:std-ratio}, we report the variability of $R^2$ across 133 pairs exhibiting more than $5\%$ degradation, measured as the mean and standard deviation of $R^2$ within bins grouped by $Y|X$-shift ratios of these pairs. We observe that pairs with larger $Y|X$-ratio tend to have more varied and lower $R^2$ values -- higher standard deviations $(>0.2)$ and smaller $R^2$ $(<0.7)$ on average, whereas pairs with smaller $Y|X$-ratio show more stable and higher $R^2$ outcomes, with lower standard deviation ($< 0.1$) and larger mean ($\approx 0.8$). This implies that when the $Y|X$-shift is large, the source accuracy becomes a less reliable predictor of the target accuracy}. (ii) \emph{Weak $Y|X$-shifts}: When $X$-shifts dominate (\Cref{fig:acc-on-the-line}(f)-(g)), we observe that the source and target performances are correlated. This corresponds with the ``accuracy-on-the-line'' phenomenon observed on image datasets, where the input features often contain most of the necessary and invariant information for predicting the outcome across domains, making $Y|X$-shifts relatively weak. 

\begin{table}[!htb]
    \centering
    \caption{Mean and standard deviations of $R^2$ within bins aggregated by $Y|X$-shift ratio bins (133 pairs with $>5\%$ degradation).}
    \label{tab:std-ratio}
\resizebox{\textwidth}{!}{\begin{tabular}{c|cccccccccc}
\toprule
Bin Quantile & 0-10\% & 10-20\% & 20-30\% & 30-40\% & 40-50\% & 50-60\% & 60-70\% & 70-80\% & 80-90\% & 90-100\% \\
\midrule
$Y|X$-ratio range & [0.03, 0.29] & [0.32, 0.51] & [0.51, 0.57] & [0.57, 0.59] & [0.59, 0.63] & [0.63, 0.66] & [0.67, 0.70] & [0.71, 0.75] & [0.76, 0.81] & [0.81, 1.00] \\
Mean & 0.793 & 0.821 & 0.795 & 0.690 & 0.610 & 0.617 & 0.645 & 0.633 & 0.712 & 0.667\\
Standard Deviation& 0.092 & 0.066 & 0.110 & 0.239 & 0.304 & 0.281 & 0.206 & 0.162 & 0.231 & 0.219 \\
\bottomrule
\end{tabular}}
\end{table}

\vspace{-0.15in}
\paragraph{Insights.}
(\rm{i}) \emph{For benchmark design}:
Our findings highlight that the ``accuracy-on-the-line'' phenomenon should be interpreted with caution, as its validity is contingent upon the specific distribution shift patterns in play. This underscores a crucial aspect of empirical studies: without accounting for different shift patterns, one may draw misleading conclusions about model performance. This also reinforces the necessity of characterizing shift patterns in our benchmark, as a nuanced understanding of distribution shift patterns is essential for developing effective \twy{data-driven} modeling approaches.
(ii) \emph{For algorithmic design:}  
The weak correlation between source and target accuracy presents a significant challenge for model validation and hyperparameter tuning. Traditional validation relies on a held-out dataset drawn i.i.d. from the source domain. However, under strong $Y|X$ shifts, a model with good performance on the source may not generalize well to the target domain. This renders traditional validation methods unsuitable in such scenarios. To address this, we take an initial step in this paper by validating methods using \emph{data drawn from the target domain}. Future work can further explore model validation techniques specifically tailored to distribution shifts.

\subsection{Finding 2: Algorithmic ranking fluctuates across settings and shift patterns}\label{subsec:benchmark-empirical-result}
Literature on distribution shifts often seeks a single method that dominates across all settings \citep{recht2019imagenet}. 
However, our benchmark, which covers a wide range of distribution shifts, reveals significant variation in algorithmic rankings across the \wty{172} source-target pairs. 
This suggests that no single method consistently outperforms the rest in the presence of distribution shifts. 
Instead, the effectiveness of one method often depends on the specific setting, indicating that empirical studies should focus on understanding particular shift patterns. 

\begin{table}[!htb]
\caption{Performance of methods across all 172 source-target pairs. The table presents the \emph{algorithmic ranking} and \emph{accuracy} for each method over all pairs. The min, max, mean, standard deviation, and median of these metrics are computed across all pairs. Results of all methods can be found in~\Cref{tab:ranking-all}.}
\label{tab:rank-main}
\resizebox{\textwidth}{!}{\centering\begin{tabular}{@{}llcccccccc@{}}
\toprule
\multirow{2}{*}{Category} & \multirow{2}{*}{Method} & \multicolumn{5}{c}{Ranking} & \multicolumn{3}{c}{Accuracy} \\ \cmidrule(l){3-7}\cmidrule(l){8-10} 
 &  & Min & Max & Mean & Std & Median & Mean & Std & Median \\ \midrule
\multirow{1}{*}{\begin{tabular}[c]{@{}l@{}}1. Basic ERM\end{tabular}} 
&NN2&1&43&18.63 & 8.88 & 17&76.56 & 5.09 & 77.19\\
\multirow{1}{*}{\begin{tabular}[c]{@{}l@{}}2. Tree-based Ensemble \end{tabular}} 
&XGB&1&44&15.05 & 11.40 & 12&76.70 & 5.09 & 76.56\\
\multirow{1}{*}{\begin{tabular}[c]{@{}l@{}}3. Linear-DRO\end{tabular}} 
&Conditional-DRO&1&45&20.77 & 8.82 & 20&76.10 & 5.36 & 76.32\\
\multirow{1}{*}{\begin{tabular}[c]{@{}l@{}}4. Kernel-DRO\end{tabular}} 
&Kernel-Wasserstein-DRO&1&44&24.91 & 8.62 & 26&75.59 & 5.65 & 76.07\\
\multirow{1}{*}{\begin{tabular}[c]{@{}l@{}}5. Tree-DRO\end{tabular}} 
&LGBM-CVaR-DRO&1&45&17.95 & 13.59 & 14&76.63 & 5.33 & 76.55\\
\multirow{1}{*}{\begin{tabular}[c]{@{}l@{}}6. NN-DRO\end{tabular}} 
&NN4-$\chi^2$-DRO&1&45&21.92 & 14.94 & 20&76.01 & 5.78 & 76.61\\
\multirow{1}{*}{\begin{tabular}[c]{@{}l@{}}7. Imbalanced Learning\end{tabular}} 
&RWG&1&45&14.69 & 11.62 & 10&76.90 & 5.37 & 76.63\\
\multirow{1}{*}{\begin{tabular}[c]{@{}l@{}}8. Fairness-enhancing\end{tabular}} 
&Fairness In-process&1&42&13.19 & 11.57 & 8&76.48 & 4.65 & 76.54\\
\bottomrule
\end{tabular}}
\end{table}
\vspace{-0.15in}
\paragraph{Results.}  
For all \wty{172} source-target pairs in~\Cref{table:overview}, we present the algorithmic ranking and accuracy of each method in~\Cref{tab:rank-main}. We select one representative method from each category and defer the results of all methods in~\Cref{tab:ranking-all}. The results reveal significant variation in algorithmic rankings across the shift pairs: (i) each method performs best in some cases (see ``Min Ranking'') and worst in others (see ``Max Ranking''); (ii) the relatively large standard deviation in rankings indicates substantial fluctuations in performance; and (iii) the similar ``Mean Ranking'' against a large standard deviation across each method demonstrates no clear winner among two different methods.

\paragraph{Insights.}
(\rm{i}) \emph{For benchmark design:} Our findings highlight the complexity of our proposed \twy{semi-synthetic shift} scenarios, as algorithmic rankings vary significantly across different settings. This emphasizes the importance of incorporating a \emph{diverse} range of settings when designing benchmarks for evaluating methods under distribution shifts, as illustrated in this work.
(ii) \emph{For algorithmic design:}  
The variation in algorithmic rankings shows that no single method consistently outperforms others across different distribution shifts. This suggests that, instead of seeking universal solutions, methods should be designed with a deep understanding of the specific shift patterns they aim to address. This also underscores the importance of collecting additional data from the target domain to guide critical design choices. Moreover, efficient data collection strategies can significantly improve performance. In \Cref{sec:intervention}, we will provide examples of tailored algorithmic interventions and data-centric approaches for various types of shifts.

\subsection{Finding 3: DRO shows limited improvements under distribution shifts}  
\label{subsec:finding3}
In addition to the algorithmic rankings presented in~\Cref{tab:rank-main}, it is noteworthy that DRO methods -- both Linear-DRO and NN-DRO -- do not demonstrate significant improvements over basic ERM methods and tree-based ensemble methods. 
Furthermore, we provide detailed results for each of the 10 settings in~\Cref{table:overview} to offer a more comprehensive overview.
\vspace{-0.15in}
\paragraph{Results.}  

\begin{figure}[!htb]
        \includegraphics[width = \textwidth]{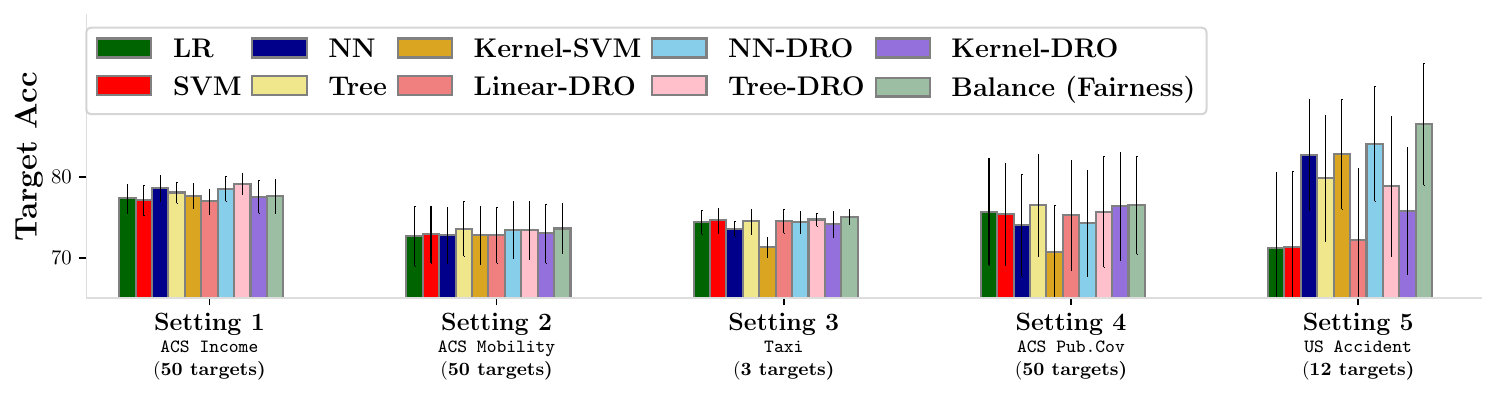}
	\includegraphics[width=\textwidth]{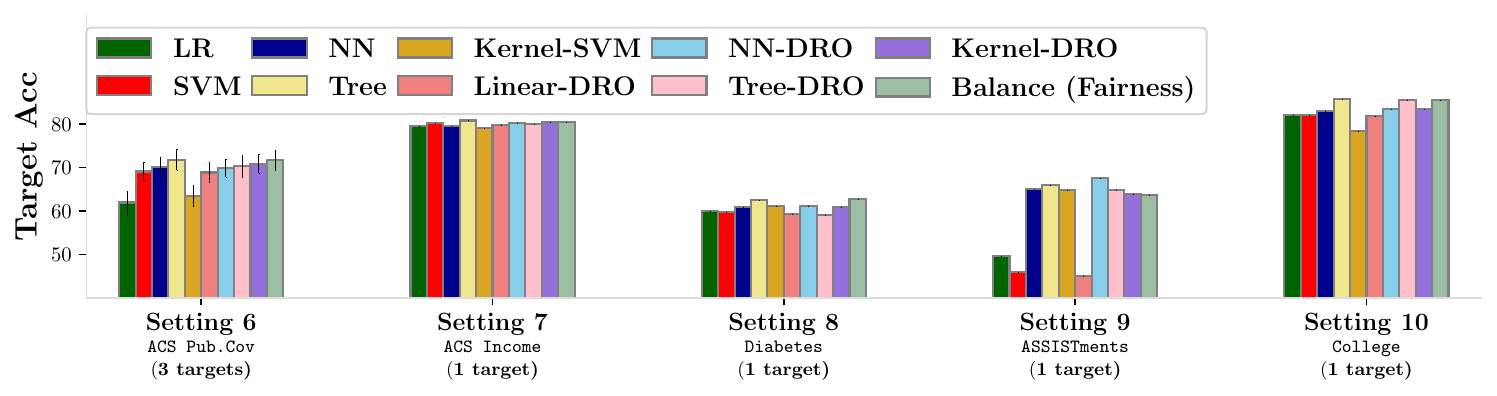}
	\caption{Overall performances of all methods on the target data in our
          \wty{10} settings in Table \ref{table:overview}. Here we calculate the average accuracy (as well as the standard deviation) among multiple target domains in each setting. 
          Note that for Settings 3, 5 and 9, we do not include fairness-enhancing methods since there are no sensitive attributes. \twy{Within each method class, we report only the method with median average accuracy.} Full numerical results can be found in Tables~\ref{table:selected_results_to_all_accuracy} and~\ref{table:selected_results_to_all_f1}.}
     \label{fig:one-to-all}
     \vspace{-0.15in}
\end{figure}

In \Cref{fig:one-to-all}, each setting represents a single source domain paired with \emph{multiple} target domains, resulting in a total of \wty{172} source-target shift pairs (see \Cref{table:overview}). For each method within each setting, we present the average target accuracy along with the standard deviation calculated across the multiple target domains. 
The results yield the following observations:  
(\rm{i}) The overall performance of DRO methods in each setting does not demonstrate significant improvement compared to other categories, particularly when compared with Basic ERM and tree-ensemble methods; \wty{In Settings 4 and \twy{6}, Kernel-DRO methods achieve significant improvements over the ERM counterparts (i.e., Kernel-SVM), averaged across target domains. This occurs because the regularization in DRO methods helps control the high variance that can arise in the projection of kernel methods. Nevertheless, switching to tree-based ensemble models or linear models yields even greater performance gains, suggesting that model class selection has a larger impact than the choice of learning methods.}  
(\rm{ii}) For settings with multiple target domains, the relatively large standard deviation indicates considerable variability in generalization performance across different target domains, with DRO methods showing limited enhancements in this regard.
See \Cref{app:setup} for more numerical results including other evaluation metrics (i.e., macro F1-score).

\paragraph{Insights.} 
The limited gains from DRO reveal the inherent limitations of worst-case distribution optimization, which can overestimate the severity of real-world shifts, reducing its practical relevance. This observation calls for further investigation into why optimizing for the worst-case distribution does not yield better generalization, an analysis we begin in~\Cref{subsec:worst-analysis}. 
Additionally, beyond the conventional ambiguity sets selected for mathematical convenience, it is crucial to explore how tailored shift patterns can be incorporated into their construction in a more data-driven manner. 
To address this, we provide examples of customized ambiguity set designs for DRO in~\Cref{sec:alg_intervention}, where we demonstrate that incorporating target-specific knowledge significantly enhances the performance of DRO methods.

\subsection{Finding 4: DRO performance is strongly correlated with its base model}  
\label{subsec:finding4} 
\begin{figure}[!htb]
	\includegraphics[width=\textwidth]{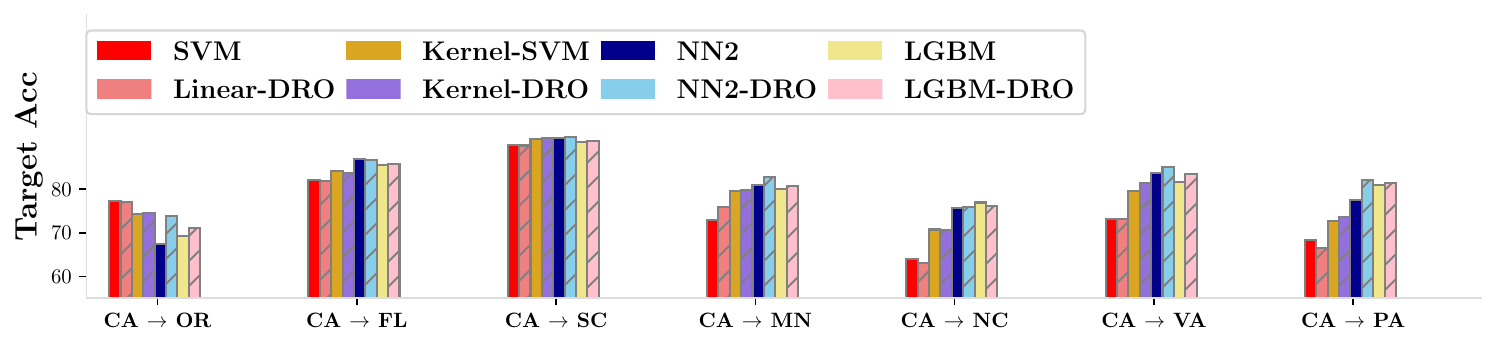}
	\caption{Overall performances on different source-target shift pairs in Setting 5. \twy{Within each DRO method class, we report only the method with median average accuracy.}}
 	\label{fig:one-to-multiple}
  \vspace{-0.15in}
\end{figure}

A notable observation from~\Cref{fig:one-to-all} is that the performance gap between Linear-DRO and NN-DRO (light red vs. light blue) is generally larger than the differences among methods within each category, especially in Setting 1 and Setting 5.
\vspace{-0.1in}
\paragraph{Results.}  
To deepen the analysis, we compare the performance of \wty{each DRO variant (i.e., Linear-DRO, NN2-DRO, Kernel-DRO, and LGBM-DRO)} with their respective base models \wty{(i.e., SVM, NN2, Kernel, and LGBM)} across various target domains in Setting 5, as shown in~\Cref{fig:one-to-multiple}. 
The results indicate that DRO performance is closely tied to the performance of its base model, with the base model class having a greater impact than the specific choice of ambiguity sets -- except for one DRO method that significantly underperforms.
\vspace{-0.15in}

\paragraph{Insights.}
This finding highlights an important but often overlooked aspect in the design of DRO methods: compatibility with the base model. Specifically, when base model performance varies significantly, it is unrealistic to expect DRO to bridge the performance gap. Instead, applying DRO to a stronger base model may yield better results.
Furthermore, a natural question arises: which design elements are critical for DRO's practical performance? A more nuanced understanding of this could drive the development of more practical and effective DRO methods, which we investigate through a detailed linear analysis in~\Cref{subsec:lr-empirical}.

\wty{\paragraph{Comparison with Existing Domain Selection.} Finally, we emphasize that our empirical findings are based on selected domains from spatiotemporal shifts, which give rise to a different class of distribution shifts compared to shifts through controlled perturbations that are commonly studied in the robust optimization literature. In settings where distribution shifts arise due to feature perturbations, tailored DRO approaches that incorporate Wasserstein ambiguity sets have demonstrated strong performance in literature~\citep{wang2021sinkhorn,bennouna2022holistic,justin2023learning}. In Setting 1, we also perturb the feature data of the source domain and observe similar results. However, our settings are significantly different from such \twy{controlled} shifts and typically induce larger changes to model coefficients. We defer corresponding experimental and visualization details to Appendices~\ref{app:synthetic} and~\ref{app:sythetic-real} respectively.}

\wty{\paragraph{Feasibility in Settings.} We assess the \emph{feasibility} of learning under our proposed distribution shift settings (i.e., abundant data from the source domain to train and limited data from the target domain to validate) by comparing each method's performance in the target domain $Q$ against two benchmarks: (1) Random guess (loss = 0.5); (2) The best oracle ERM model $f^*(X)$ trained on all target data among all model classes $\Fscr$. Concretely, for any method $\hat f$, we define its \emph{feasibility} as:
\[\frac{0.5 - \E_{Q}[\ell(\hat f(X), Y)]}{0.5 - \min_{\Fscr} \min_{f \in \Fscr}\E_{Q}[\ell(f(X), Y)]},\]
so that values of 0 and 1 correspond to random guessing and the best oracle ERM model on $Q$. 
We provide detailed setups and results in \Cref{app:feasibility}. Across all settings, the top-performing methods achieve feasibility scores between 0.4 and 0.8, and exceed 0.6 in most cases, demonstrating that our learning scenarios are both challenging and practically feasible. Finally, these findings reinforce the conclusions of performance comparisons reported in Section~\ref{sec:perform-comp}.}

\section{In-depth Investigations on DRO}\label{sec:method-lr}
Empirical findings in~\Cref{sec:perform-comp} reveal significant opportunities for exploring the design of DRO methods, which can be categorized into two key aspects: (i) identifying the critical design elements (e.g., base model class, type of ambiguity set, etc.) that influence DRO's practical performance (corresponding to Finding 4 in~\Cref{subsec:finding4}), and (ii) understanding why optimizing the worst-case distribution yields only limited improvements (corresponding to Finding 3 in~\Cref{subsec:finding3}).

In order to complement the existing theory-driven algorithm development paradigm with data-driven modeling-based empirical insights, we provide in-depth investigations on DRO from two aspects.
In~\Cref{subsec:lr-empirical}, we conduct the first systematic and comprehensive empirical analysis of typical DRO methods across different DRO ambiguity sets. 
Our analysis further underscores the need to go beyond the traditional focus on the
ambiguity set as the main modeling lever.  Factors previously regarded as
implementation details, such as the model class (e.g., XGB vs. NN vs. linear
models) or hyperparameter selection methods (e.g., based only on source domain
or a very small out-of-distribution data), have an outsize impact on distributional robustness. Then in~\Cref{subsec:worst-analysis}, we demonstrate that the hypothetical
worst-case distributions used by DRO methods are \twy{hard to fit and }too conservative, substantiating the standard critique for worst-case approaches.  

Taken together, these results highlight the importance of a \twy{data-driven} modeling approach to algorithm development, whose benefits we
substantiate in~\Cref{sec:intervention}.

\subsection{Identifying the critical design elements in DRO}\label{subsec:lr-empirical}  
The major design elements in DRO include: ({i}) the base model class \(\mathcal{F}\) (e.g., SVM, tree-based ensembles, and neural networks); ({ii}) the choice of ambiguity set (i.e., the distance function and the radius). 
To identify the critical algorithmic design elements that influence DRO's performance under distribution shifts, we collect experimental results from the benchmarks and conduct a linear regression analysis based on them. To the best of our knowledge, this represents the first attribution study for DRO that seeks to explain its empirical performance.

Specifically, we regress the \emph{target accuracy} on variables such as base model classes, types of distance functions, and other explanatory variables. 
\tyw{To further examine the utility of robust methods against model class, we also regress the \emph{performance gap}~\eqref{eq:src-tgt-acc-def} on these variables, along with additional interaction variables, and find similar results in Appendix~\ref{app:ablation}.} 
\vspace{-0.15in}

\paragraph{Empirical model setup.} 
In this study, one data point represents one configuration (such as \Cref{config}) of one method and its accuracy in a particular domain from one setting. \wty{Note that we only choose settings with more than one target domain (excluding Settings 7 - 10) to ensure the variety of target domains in the analysis.} \tyw{We consider two types of regression designs. (i) \texttt{Best Config}: we regress on the highest accuracy of each method achieved by the best configuration for each source-target domain pair; (ii) \texttt{Worst Domain}: we regress on the worst-case target accuracy across all target domains for each configuration per method and setting.}
In each type, we estimate the following equation via ordinary least squares (OLS):
\vspace{-0.1in}
\begin{equation}\label{eq:lr-model}
    \mbox{Accuracy}_{i,j,s,t} = \alpha + \beta_1^{\top} X_{i,j,s} + \beta_2^{\top} D_{i,j,s,t} + \beta_3 Z_{i,j} + \beta_4^{\top} V_{i,j} + \mu_{i} + \tau_j + \eta_{i,j,s,t},
\vspace{-0.1in}
\end{equation}
where $i,j,s,t$ represent the setting, target domain, model class, and the configuration ID of the method respectively. The explanatory variables in the linear regression model~\eqref{eq:lr-model} are separated into four groups, \tyw{where variables in the third and fourth groups only appear in \texttt{Best Config} setting}.
\begin{itemize}[leftmargin=*]
\vspace{-0.1in}
    \item[1.] Model types: $X_{i,j,s}$ includes dummy variables of model types indicating whether the model class of the method is \wty{LGBM, Kernel-SVM, NN-2, NN-3, NN-4} (with the default defined as the linear SVM class).
\vspace{-0.1in}
    \item[2.] Ambiguity set: $D_{i,j,s,t}$ includes dummy variables for the DRO hyperparameter, indicating whether the DRO methods use a particular distance metric, and one variable representing the scaled robustness radius with the detailed definition of the radius shown in Appendix~\ref{app:ambiguity-scale}. The default is the method without any robustness intervention, i.e., the ERM method.
\vspace{-0.1in}
    \item[3.] Distribution shift pattern: $Z_{i,j}$ denotes the $Y|X$-shift percentage for the target domain $j$ of the setting $i$. This allows us to control for the type of distribution shift.
\vspace{-0.1in}
    \item[4.] \tyw{Model validation type: $V_{i,j}$ denotes dummy variables for validation types, indicating whether, for each method and setting, the best configuration is selected based on the highest (i) average-case target accuracy, i.e., average performance from small samples across all target domains, or (ii) worst-case accuracy across all target domains, i.e., worst-case performance across target domains. The default is the method configuration selected by the highest in-distribution accuracy.} 
\end{itemize}
Besides these explanatory variables, we use $\mu_i$ and $\tau_j$ as control variables representing the setting and domain fixed effects, respectively.  
The full set of definitions of all variables can be found in~\Cref{tab:var_def} in~\Cref{app:var-def}.

\vspace{-0.15in}
\paragraph{Data setup.}
In \Cref{table:overview}, we run regression across all settings and include \wty{24} representative ERM and DRO methods from \Cref{sec:perform-comp}, encompassing basic ERM methods (SVM, Kernel-SVM, NN-2, NN-3, NN-4); Linear-DRO methods (Wasserstein-DRO, CVaR-DRO, $\chi^2$-DRO, TV-DRO, KL-DRO, Unified-DRO);  Tree-based ensemble methods (LGBM); \wty{Kernel-DRO methods (Wasserstein-DRO, CVaR-DRO, $\chi^2$-DRO, KL-DRO); Tree-DRO methods (LGBM-KL-DRO, LGBM-CVaR-DRO)}; NN-DRO methods (CVaR-DRO, $\chi^2$-DRO) \wty{varying the layer number as 2, 3, 4}. Our two designs ensure enough sample size in either scenario since \texttt{Best Config} measures performance over a wealth of target domains and \texttt{Worst Domain} measures performance over diverse hyperparameter configurations. In total, we generate \wty{12,527} data points in \texttt{Best Config} where each data point represents the target accuracy for the best configuration of each method and \wty{23,776} points in \texttt{Worst Domain} where each data point represents the worst-domain accuracy for one configuration of a method. In each setting, the source domain is the corresponding selected source domain in the ``Selected Source-Target'' column and the target domains are all the possible target domains there in \Cref{table:overview}. We train and evaluate models in a consistent manner across settings, as we detail in \Cref{app:traindetail}.

\begin{table}[htbp]
    \centering
    \caption{Regression results on algorithmic design components on the method performance}
    \label{tab:linear-analysis}
    \sisetup{
    table-format=-1.4, 
    add-integer-zero=false 
    }
    
	\resizebox{\textwidth}{!}{\begin{tabular}{llSSSSSS}
        \toprule
        & & \multicolumn{6}{c}{Dependent variable: Target accuracy}\\
        & & \multicolumn{3}{c}{\texttt{Best Config}} & \multicolumn{3}{c}{\texttt{Worst Domain}}\\\cmidrule(lr){3-8}
    \multicolumn{2}{c}{Variable Name} & \multicolumn{1}{c}{\ \ All} & \multicolumn{1}{c}{\ \ Setting 5} & \multicolumn{1}{c}{\ \ Setting 1} & \multicolumn{1}{c}{\ \ All} & \multicolumn{1}{c}{\ \ Setting 4} & \multicolumn{1}{c}{\ \ Setting 2}\\
         \midrule
\multirow{3}{*}{\parbox[c]{1.5cm}{Model\\Class}}&LGBM& .0081$^{***}$& .0500$^{***}$& .0156$^{***}$&-.0116$^{***}$&-.0345$^{***}$&-.0062$^{***}$\\
&&\multicolumn{1}{c}{ \scriptsize\hspace{0.35cm}(.0016)}&\multicolumn{1}{c}{ \scriptsize\hspace{0.35cm}(.0051)}&\multicolumn{1}{c}{ \scriptsize\hspace{0.35cm}(.0006)}&\multicolumn{1}{c}{ \scriptsize\hspace{0.35cm}(.0015)}&\multicolumn{1}{c}{ \scriptsize\hspace{0.35cm}(.0018)}&\multicolumn{1}{c}{ \scriptsize\hspace{0.35cm}(.0022)}\\
&NN2& .0077$^{***}$& .0730$^{***}$& .0115$^{***}$&-.0269$^{***}$&-.0322$^{***}$&-.0388$^{***}$\\
&&\multicolumn{1}{c}{ \scriptsize\hspace{0.35cm}(.0017)}&\multicolumn{1}{c}{ \scriptsize\hspace{0.35cm}(.0054)}&\multicolumn{1}{c}{ \scriptsize\hspace{0.35cm}(.0007)}&\multicolumn{1}{c}{ \scriptsize\hspace{0.35cm}(.0020)}&\multicolumn{1}{c}{ \scriptsize\hspace{0.35cm}(.0029)}&\multicolumn{1}{c}{ \scriptsize\hspace{0.35cm}(.0034)}\\
&NN3& .0028$^{*}$& .0774$^{***}$& .0108$^{***}$&-.0593$^{***}$&-.0428$^{***}$&-.0395$^{***}$\\
&&\multicolumn{1}{c}{ \scriptsize\hspace{0.35cm}(.0017)}&\multicolumn{1}{c}{ \scriptsize\hspace{0.35cm}(.0053)}&\multicolumn{1}{c}{ \scriptsize\hspace{0.35cm}(.0007)}&\multicolumn{1}{c}{ \scriptsize\hspace{0.35cm}(.0020)}&\multicolumn{1}{c}{ \scriptsize\hspace{0.35cm}(.0033)}&\multicolumn{1}{c}{ \scriptsize\hspace{0.35cm}(.0042)}\\
&NN4&-.0011& .0790$^{***}$& .0124$^{***}$&-.0536$^{***}$&-.0681$^{***}$&-.0520$^{***}$\\
&&\multicolumn{1}{c}{ \scriptsize\hspace{0.35cm}(.0017)}&\multicolumn{1}{c}{ \scriptsize\hspace{0.35cm}(.0052)}&\multicolumn{1}{c}{ \scriptsize\hspace{0.35cm}(.0007)}&\multicolumn{1}{c}{ \scriptsize\hspace{0.35cm}(.0021)}&\multicolumn{1}{c}{ \scriptsize\hspace{0.35cm}(.0039)}&\multicolumn{1}{c}{ \scriptsize\hspace{0.35cm}(.0048)}\\
&Kernel& .0000& .0397$^{***}$& .0012$^{**}$&-.0236$^{***}$&-.0422$^{***}$&-.0154$^{***}$\\
&&\multicolumn{1}{c}{ \scriptsize\hspace{0.35cm}(.0013)}&\multicolumn{1}{c}{ \scriptsize\hspace{0.35cm}(.0042)}&\multicolumn{1}{c}{ \scriptsize\hspace{0.35cm}(.0006)}&\multicolumn{1}{c}{ \scriptsize\hspace{0.35cm}(.0015)}&\multicolumn{1}{c}{ \scriptsize\hspace{0.35cm}(.0018)}&\multicolumn{1}{c}{ \scriptsize\hspace{0.35cm}(.0022)}\\
\midrule
\multirow{3}{*}{\parbox[c]{1.7cm}{Ambiguity\\Set}}&Wasserstein& .0020&-.0077& .0025$^{***}$&-.0222$^{***}$&-.0308$^{***}$&-.0046\\
&&\multicolumn{1}{c}{ \scriptsize\hspace{0.35cm}(.0018)}&\multicolumn{1}{c}{ \scriptsize\hspace{0.35cm}(.0057)}&\multicolumn{1}{c}{ \scriptsize\hspace{0.35cm}(.0008)}&\multicolumn{1}{c}{ \scriptsize\hspace{0.35cm}(.0027)}&\multicolumn{1}{c}{ \scriptsize\hspace{0.35cm}(.0040)}&\multicolumn{1}{c}{ \scriptsize\hspace{0.35cm}(.0067)}\\
&Chi-squared& .0010& .0018& .0010$^{**}$& .0091$^{***}$&-.0018& .0158$^{***}$\\
&&\multicolumn{1}{c}{ \scriptsize\hspace{0.35cm}(.0012)}&\multicolumn{1}{c}{ \scriptsize\hspace{0.35cm}(.0037)}&\multicolumn{1}{c}{ \scriptsize\hspace{0.35cm}(.0005)}&\multicolumn{1}{c}{ \scriptsize\hspace{0.35cm}(.0013)}&\multicolumn{1}{c}{ \scriptsize\hspace{0.35cm}(.0018)}&\multicolumn{1}{c}{ \scriptsize\hspace{0.35cm}(.0023)}\\
&Kullback-Leibler& .0007&-.0089$^{*}$& .0017$^{***}$& .0015& .0046$^{***}$& .0033$^{*}$\\
&&\multicolumn{1}{c}{ \scriptsize\hspace{0.35cm}(.0015)}&\multicolumn{1}{c}{ \scriptsize\hspace{0.35cm}(.0048)}&\multicolumn{1}{c}{ \scriptsize\hspace{0.35cm}(.0006)}&\multicolumn{1}{c}{ \scriptsize\hspace{0.35cm}(.0013)}&\multicolumn{1}{c}{ \scriptsize\hspace{0.35cm}(.0015)}&\multicolumn{1}{c}{ \scriptsize\hspace{0.35cm}(.0019)}\\
&Total Variation&-.0049$^{**}$&-.0019&-.0006&-.0369$^{***}$&-.0517$^{***}$&-.0150$^{***}$\\
&&\multicolumn{1}{c}{ \scriptsize\hspace{0.35cm}(.0024)}&\multicolumn{1}{c}{ \scriptsize\hspace{0.35cm}(.0079)}&\multicolumn{1}{c}{ \scriptsize\hspace{0.35cm}(.0009)}&\multicolumn{1}{c}{ \scriptsize\hspace{0.35cm}(.0027)}&\multicolumn{1}{c}{ \scriptsize\hspace{0.35cm}(.0034)}&\multicolumn{1}{c}{ \scriptsize\hspace{0.35cm}(.0037)}\\
&OT-Discrepancy&-.0022&-.0062&-.0021$^{**}$&-.0109$^{***}$&-.0455$^{***}$&-.0137$^{***}$\\
&&\multicolumn{1}{c}{ \scriptsize\hspace{0.35cm}(.0024)}&\multicolumn{1}{c}{ \scriptsize\hspace{0.35cm}(.0076)}&\multicolumn{1}{c}{ \scriptsize\hspace{0.35cm}(.0009)}&\multicolumn{1}{c}{ \scriptsize\hspace{0.35cm}(.0025)}&\multicolumn{1}{c}{ \scriptsize\hspace{0.35cm}(.0030)}&\multicolumn{1}{c}{ \scriptsize\hspace{0.35cm}(.0035)}\\
&Radius& .0025$^{**}$&-.0091& .0033$^{***}$&-.0097$^{***}$& .0039$^{***}$&-.0081$^{***}$\\
&&\multicolumn{1}{c}{ \scriptsize\hspace{0.35cm}(.0010)}&\multicolumn{1}{c}{ \scriptsize\hspace{0.35cm}(.0057)}&\multicolumn{1}{c}{ \scriptsize\hspace{0.35cm}(.0004)}&\multicolumn{1}{c}{ \scriptsize\hspace{0.35cm}(.0004)}&\multicolumn{1}{c}{ \scriptsize\hspace{0.35cm}(.0005)}&\multicolumn{1}{c}{ \scriptsize\hspace{0.35cm}(.0006)}\\
\midrule
\multirow{2}{*}{\parbox[c]{1.5cm}{Shift\\Pattern}}&$Y|X$-ratio&-.0044$^{***}$&-.0051$^{***}$& 0.0679$^{***}$&{-}&{-}&{-}\\
&&\multicolumn{1}{c}{ \scriptsize\hspace{0.35cm}(.0002)}&\multicolumn{1}{c}{ \scriptsize\hspace{0.35cm}(.0002)}&\multicolumn{1}{c}{ \scriptsize\hspace{0.35cm}(0.0013)}&&&\\
\midrule
\multirow{3}{*}{\parbox[c]{1.5cm}{Validation\\Type}}&Worst-case &-.0056$^{***}$&-.0067$^{**}$& .0050$^{***}$&{-}&{-}&{-}\\
&&\multicolumn{1}{c}{ \scriptsize\hspace{0.35cm}(.0010)}&\multicolumn{1}{c}{ \scriptsize\hspace{0.35cm}(.0033)}&\multicolumn{1}{c}{ \scriptsize\hspace{0.35cm}(.0004)}&&&\\
&Average-case& .0054$^{***}$& .0202$^{***}$& .0078$^{***}$&{-}&{-}&{-}\\
&&\multicolumn{1}{c}{ \scriptsize\hspace{0.35cm}(.0010)}&\multicolumn{1}{c}{ \scriptsize\hspace{0.35cm}(.0033)}&\multicolumn{1}{c}{ \scriptsize\hspace{0.35cm}(.0004)}&&&\\
\midrule
\multirow{2}{*}{\parbox[c]{1.5cm}{Fixed\\Effect}}&Setting&{\ \ Yes}&{\ \ No}&{\ \ No}&{\ \ Yes}&{\ \ No}&{\ \ No}\\
&Domain&{\ \ No}&{\ \ Yes}&{\ \ Yes}&{\ \ No}&{\ \ No}&{\ \ No}\\
\midrule
\multirow{2}{*}{\parbox[c]{1.5cm}{Overall}}&$N$&{\ \ 12527}&{\ \ 935}&{\ \ 3671}&{\ \ 23776}&{\ \ 3330}&{\ \ 3212}\\
&Adjusted $R^2$& .191& .777& .789& .441& .244& .160\\
    \bottomrule
    \end{tabular}}
    
    \smallskip
    \footnotesize{\emph{Notes.} $^{***}, ^{**}$ and $^{*}$ show statistical significance at the 1\%, 5\%, and 10\% levels using two-tailed tests, respectively. In \texttt{Worst Domain}, there are no effects of the validation types since we do not select the best configuration, and no effects of the shift patterns since we consider evaluating the domain with the worst accuracy. 
    }
\end{table}
\vspace{-0.15in}
\paragraph{Primary findings.} 
We report regression results for \texttt{Best Config} (left) and \texttt{Worst Domain} (right) in \Cref{tab:linear-analysis}. We also include regression results on four specific settings (i.e., Settings 1, 2, 4, and 5 in \Cref{table:overview}, two settings for each design) to show whether the findings are consistent across different settings. 
Results on other settings can be found in~Appendix~\ref{app:empirical-all}.

We observe the \emph{choice of the model class} has by far the greatest impact on robustness. In the \texttt{Best Config} design (left), the coefficients of the model class (including NN and LGBM) are statistically significant in all settings. Furthermore, in general, the model class effects have the largest positive coefficients when compared to the coefficients of the ambiguity set and validation type. This implies that changing the model class of the method, such as switching between linear models, tree-based ensemble methods and neural networks, results in the largest improvements in the method performance, compared to building distributional robustness counterparts or changing the validation types.  
This finding generalizes across different settings (i.e., Settings 1 and 4 here). In each setting, the coefficients of the model class are again largest with statistical significance.

The \emph{hyperparameter selection} mechanism also has relatively large effects on robustness against \twy{spatiotemporal} distribution shifts we construct in a semi-synthetic manner. In the \texttt{Best Config} design (left), the coefficient of the average-case accuracy is positive and usually larger than the coefficients of ambiguity set across different settings. 
This implies that selecting the best configuration using the average-case accuracy improves robustness more than using DRO.

In terms of the key components of DRO, \emph{the effects of ambiguity sets} are inconsistent and relatively small across different settings. This can be seen from the small and relatively unstable coefficients on the radius of the ambiguity set (\Cref{tab:linear-analysis} left). For example, while the impact of the radius of the ambiguity set is significantly positive in the \texttt{ACS Income} setting, its effect is minor compared to the choice of the model class. Furthermore, none of the distance types used in the ambiguity set is significantly positive across all the settings. This further validates that DRO does not have as much effect as the model class, making it less useful for ensuring the method generalizes well to other target domains. 

\twy{In Settings 1 and 5 of \Cref{tab:linear-analysis}, we observe that for certain fine-grained model class changes, such as varying the number of layers in neural networks, the resulting accuracy gains are comparable in magnitude to those obtained by applying DRO methods (e.g., adjusting the ambiguity radius). This indicates that DRO interventions yield significant improvements relative to incremental changes within the neural network model class. However, this effect is largely confined to neural networks, which achieve relatively lower absolute accuracy in our benchmark. For model classes with higher baseline performance, such as LightGBM (e.g., when varying the number of base estimators), the relative benefits of DRO largely vanish. To investigate this more rigorously, we conduct fine-grained experimental comparisons with NN / LGBM while varying the number of base estimators. Detailed analysis of these comparisons is provided in Appendix~\ref{app:fine-grain-model-class}.}

While we initially conjectured that the  \texttt{Worst Domain} design would show the benefits of DRO methods, such methods are unreliable even when we consider the worst-domain performance. In the right side of \Cref{tab:linear-analysis}, we observe coefficient sizes for distance types and radius in the ambiguity set $D_{i,j,s,t}$ are inconsistent across different settings. For example, in Setting 4, KL-DRO shows a significant performance improvement, but it does not perform as well in all settings. We hypothesize that the worst-case distribution calculated by KL-DRO aligns well with the source-target pair in this setting while it tends to be overly conservative in most other cases. To support the claim further, we analyze the worst-case distribution in the next subsection.
 \vspace{-0.15in}
\paragraph{Validity and robustness check.} \tyw{In this regression setup, we can assign values to most variables independently (i.e., $X_{i,j,s}, D_{i,j,s,t}, V_{i,j}$) by running through potential configurations for each base method and examining the resulting accuracy. These assignments are fully randomized within this controlled experiment, allowing us to avoid issues of endogeneity and multicollinearity \citep{angrist2009mostly}, which often undermine the validity of statistical results in observational studies. }

\tyw{We also conduct some further ablation studies in Appendix~\ref{app:ablation} and summarize them as follows:
\begin{itemize}[leftmargin=*]
\vspace{-0.1in}
\item [1.] \emph{Nonlinear effects of the ambiguity size}: We include a variable $\text{Radius}^2$ that denotes the square of the (scaled) radius of the ambiguity set such that the effect of the ambiguity size to the accuracy becomes 
nonlinear and the best ambiguity sizes are usually nonzero in Wasserstein and $f$-divergence DROs \citep{blanchet2019robust,gotoh2021calibration,iyengar2022hedging}. Despite this, we find that the effect of the ambiguity set is still relatively small. 
\vspace{-0.1in}
\item [2.] \emph{Other interaction variables}: \wty{To demonstrate the heterogeneity of DRO design across model classes, we introduce interaction terms between the model class and the radius in the ambiguity set into~\eqref{eq:lr-model} and perform regression under the \texttt{Best Config} design. We find that incorporating distributional robustness provides benefits primarily for linear and kernel models. Although the coefficient of the radius of the ambiguity set is significantly positive, its overall effect remains modest—likely due to the relatively small robustness radii selected and the more substantial performance differences attributable to model classes.} To further examine model performance degradation under $Y|X$-shifts, we introduce \emph{interaction terms between model class and shift patterns} into~\eqref{eq:lr-model}. We find no significant interaction effects, suggesting the model performance across different model classes is unstable against distribution shift patterns. This observation is consistent with results in \Cref{sec:perform-comp}. 
\vspace{-0.1in}
\item [3.] \emph{Attribution of the performance gap}: We change the dependent variable in our main specification~\eqref{eq:lr-model}, replacing the target accuracy with the \emph{performance gap}~\eqref{eq:src-tgt-acc-def} to isolate the effect of the model class on the target accuracy. Our findings suggest that DRO may not effectively mitigate the performance gap under \twy{semi-synthetic} distribution shifts.
\vspace{-0.1in}
\wty{\item [4.] \emph{Attribution using the nonlinear explanation model}: To improve the explanatory power when assessing feature importance, we modify the regression design in~\eqref{eq:lr-model} by replacing the linear regression over algorithmic design components with a nonlinear random forest regression. The main results in this section remain valid.}
\end{itemize}
}

Despite these robustness checks, we acknowledge that this empirical analysis remains preliminary and lacks full theoretical rigor, primarily due to the use of the same sets of training samples used across each method. However, it still provides a heuristic and transparent framework for understanding the importance of various algorithmic components in determining the accuracy. While previous works, such as \citep{lundberg2017unified,zhang2023why,namkoong2023diagnosing}, have focused on attributing performance degradation to differences in data distributions between domains (whether covariate or conditional), our study, to the best of our knowledge, is the first to attribute model generalization performance specifically to different algorithmic components rather than data-related factors.

\subsection{Worst-case Distribution Analysis}\label{subsec:worst-analysis}
To investigate the underlying reasons for the limited improvements of DRO, we proceed to analyze the worst-case distributions in typical DRO methods.
The worst-case distributions of typical DRO methods are computed as follows:
\vspace{-0.1in}
\begin{equation}
\label{eqn:worst-case}
	\what{P}^\star \in \argmax_{\tilde P\in\mc{P}}\mathbb{E}_{\tilde P}[\ell_{tr}(\what f_{\text{lin}}(X),Y)],
\vspace{-0.1in}
\end{equation}
where  $\what f_{\text{lin}} \in \Fscr_{\text{lin}}$ is the model trained using each DRO method, i.e., in~\eqref{eqn:general-dro} under the linear model class $\Fscr_{\text{lin}}$. \wty{We compute the worst-case distribution via solving the standard maximization problem over the probability simplex in the case of $f$-divergence and via the asymptotically optimal worst-case distribution for Wasserstein distance as described in \cite{shafieezadeh2019regularization}. In each case, we use the default interior-point algorithm from the MOSEK solver to compute the worst-case distribution for each DRO method. See \Cref{subsec:appendix-training-detail-4.2} for additional computation details. We acknowledge the limitation of our worst-case distribution analysis since the worst-case distribution $\what{P}^*$ may not be unique.}
\vspace{-0.15in}
\paragraph{Worst-case distributions are \twy{hard to learn}.}
As a heuristic barometer of the \twy{``difficulty'' of a distribution $\what{P}^\star$}, we consider the best performance we can achieve by fitting and cross-validating a powerful model class \wty{$\Fscr \neq \Fscr_{\text{lin}}$ (for example, $\Fscr$ is XGB)} on data generated by $\what{P}^\star$. That is, we compute the \emph{optimal in-distribution accuracy} as:
\vspace{-0.1in}
\begin{equation}\label{eq:optimal-f}
	1 - \E_{\what{P}^*}[\ell(f^*(X),Y)], ~\text{where}~f^* \in \argmin_{f\in\Fscr} \mathbb E_{\what{P}^\star}[\ell(f(X),Y)].
 \vspace{-0.1in}
\end{equation}
Intuitively, if the optimal in-distribution accuracy for $\what{P}^\star$ is low compared to that of the empirical distribution $\what{P}$ in~\Cref{table:overview}, then fitting the distribution $\what{P}^\star$ is very challenging, likely because it contains a higher proportion of noisy samples. 

\begin{table}[!htb]
\caption{Analysis of the worst-case distribution via optimal in-distribution accuracy}
\label{tab:optimal-iid}
\resizebox{\textwidth}{!}{
\begin{tabular}{@{}lcllllclll@{}}
\toprule
\multirow{2}{*}{\begin{tabular}[c]{@{}l@{}} Distribution \end{tabular}} & \multicolumn{1}{c}{\begin{tabular}[c]{@{}c@{}} Source \\ Domain \end{tabular}} & \multicolumn{2}{c}{\begin{tabular}[c]{@{}c@{}}Worst-Distribution \\ of KL-DRO\end{tabular}} & \multicolumn{2}{c}{\begin{tabular}[c]{@{}c@{}}Worst-Distribution \\ of $\chi^2$-DRO\end{tabular}} & \multicolumn{1}{c}{\begin{tabular}[c]{@{}c@{}}Worst-Distribution \\ of TV-DRO\end{tabular}} & \multicolumn{3}{c}{\begin{tabular}[c]{@{}c@{}}50 Target\\ Domains' Quantile\end{tabular}} \\ \cmidrule(l){2-10} 
& \multicolumn{1}{c}{$\epsilon=0$}                                                      & \multicolumn{1}{c}{$\epsilon=1e^{-2}$}       & \multicolumn{1}{c}{$\epsilon=1e^{-1}$}       & \multicolumn{1}{c}{$\epsilon=1e^{-1}$}          & \multicolumn{1}{c}{$\epsilon=5e^{-1}$}          & \multicolumn{1}{c}{$\epsilon=5e^{-2}$}         & \multicolumn{1}{c}{50\%}   & \multicolumn{1}{c}{25\%}   & \multicolumn{1}{c}{0\%}  \\ \midrule
LR  & 80.37  & 75.50  & 64.81  & 70.39    & 58.95    & 64.55      & 79.77   & 78.93   & 76.07  \\
SVM & 80.72  & 75.38  & 64.65    & 70.28    & 58.75    & 64.39         & 79.86   & 78.88   & 76.11  \\ 
NN    & 80.26  &  75.55    & 65.57                                             & 71.08    &  61.13       & 63.66                                            &   79.81  & 78.52    & 75.08   \\ 
RF  & 79.61  & 75.35  & 66.09   & 71.28   & 61.22    & 62.51    & 78.78   & 77.84   & 75.93  \\
LGBM  & 81.74  & 76.18  & 66.76     & 72.23   & 63.02   &    61.85    & 80.51   & 79.47   & 76.43                         \\
XGB & 81.29  & 75.84  & 66.31   &  71.92     & 62.73        & 61.45    & 80.13   & 79.13   & 75.08                         \\\bottomrule
\end{tabular}}
\vspace{-0.15in}
\end{table}

To assess the \twy{hardness of learning} on worst-case distributions~\eqref{eqn:worst-case}, we compare the optimal in-distribution accuracy~\eqref{eq:optimal-f} on our \twy{semi-synthetic} distribution shifts (target domains) vs. worst-case distributions posited by DRO methods.
We focus on divergence-based DRO methods, including KL-DRO, $\chi^2$-DRO, and TV-DRO.
Our empirical investigation utilizes the \texttt{ACS Income} dataset (Setting 1), with California (CA) serving as the source domain and the other 50 states as target domains. 
We evaluate a wide range of model classes $\mathcal{F}$, including SVM, LR, NN, RF, LGBM, and XGB.
In \Cref{tab:optimal-iid}, for each model class, we report the optimal in-distribution accuracy for the worst-case distributions obtained by DRO~\eqref{eqn:worst-case}, alongside that for the source and all available target domains. For target domains, we report the median (50\%), 25\%, and minimum (0\%) quantiles of the optimal in-distribution accuracy across the 50 target domains. \twy{We observe that worst-case distributions are significantly more challenging to fit compared to both the source and a broad range of 50 target domains. This finding reveals a critical insight: for commonly used moderate radii, the worst-case distributions DRO methods construct are intrinsically harder to fit than the actual distributions we observe in practice.}
\ljs{Since these worst-case distributions are harder to fit, achieving good out-of-distribution performance in practice necessitates selecting a very small radius \(\epsilon\). \wty{For instance, in Setting 1 of \Cref{fig:one-to-all}, the best \(\epsilon\) values are \(1e^{-1}\) for KL-DRO and \(1e^{-2}\) for TV-DRO.} As a result, DRO behaves similarly to ERM methods, leading to only limited improvements.}
\vspace{-0.15in}
\paragraph{Misalignment exists between worst-case distributions and target domains.}
We further illustrate how the worst-case distribution $\what{P}^\star$ (mis)aligns with the actual 50 target domains $\Qscr_{50} = \{\what{Q}_t, t \in [50]\}$.
For each target domain $\what{Q}_t\in \Qscr_{50}$, we propose to calculate the \emph{transfer accuracy} from $\what{P}^\star$ to $\what{Q}_t$, which is defined as:
\vspace{-0.1in}
\begin{equation}\label{eq:transfer-acc}
    \text{TAcc}(\what{P}^\star, \what{Q}_t) = 1 - \mathbb E_{\what{Q}_t}[\ell(f^\star(X),Y)],\quad \text{where}\quad f^\star\in\arg\min_{f\in\Fscr}\mathbb E_{\what{P}^\star}[\ell_{tr}(f(X),Y)].
\vspace{-0.1in}
\end{equation}
A higher transfer accuracy indicates a better alignment between $\what{P}^\star$ and $\what{Q}_t$. \wty{Note that the transfer accuracy is different from the accuracy of the corresponding linear DRO model on the target domain since we select a powerful model class $\Fscr(\neq \Fscr_{\text{lin}})$ in~\eqref{eq:transfer-acc}.}

We focus on Setting 1, where we use the \texttt{ACS Income} dataset and select CA as the source domain.
We take KL-DRO and Wasserstein-DRO (both based on SVM) as illustrative examples.
We vary the radius $\epsilon$ of the ambiguity set of DRO ($\epsilon \in \{0.05, 0.1, 0.2\}$ for KL-DRO and $\epsilon  \in \{0.001, 0.005, 0.01\}$ for Wasserstein-DRO), and obtain the corresponding worst-case distributions $\what{P}^\star$.
Then for each of the 50 target domains ($\what{Q}_t\in\Qscr_{50}$), we calculate the transfer accuracy from the training distribution $\what{P}$ and from the worst-case distribution $\what{P}^\star$, denoted by $\{\text{TAcc}(\what{P},\what{Q}_t)\}_{t\in[50]}$ and $\{\text{TAcc}(\what{P}^\star,\what{Q}_t)\}_{t\in[50]}$.

\begin{figure}[t]
 \centering\captionsetup[subfloat]{labelfont=scriptsize,textfont=scriptsize}  
 \stackunder[3pt]{\includegraphics[width=0.32\textwidth]{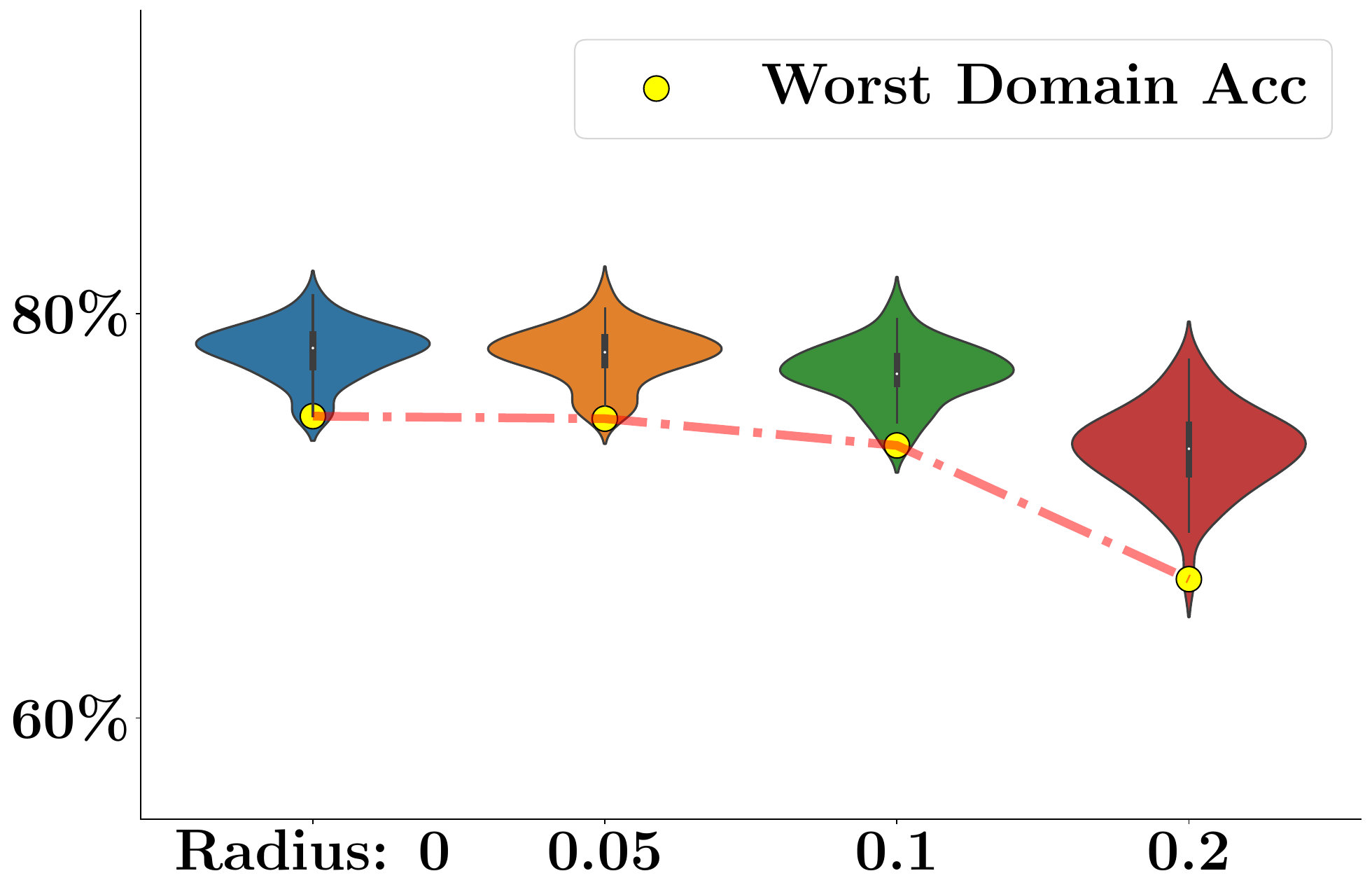}}{\scriptsize (a) \texttt{ACS Income}, KL-DRO}
 \stackunder[3pt]{\includegraphics[width=0.32\textwidth]{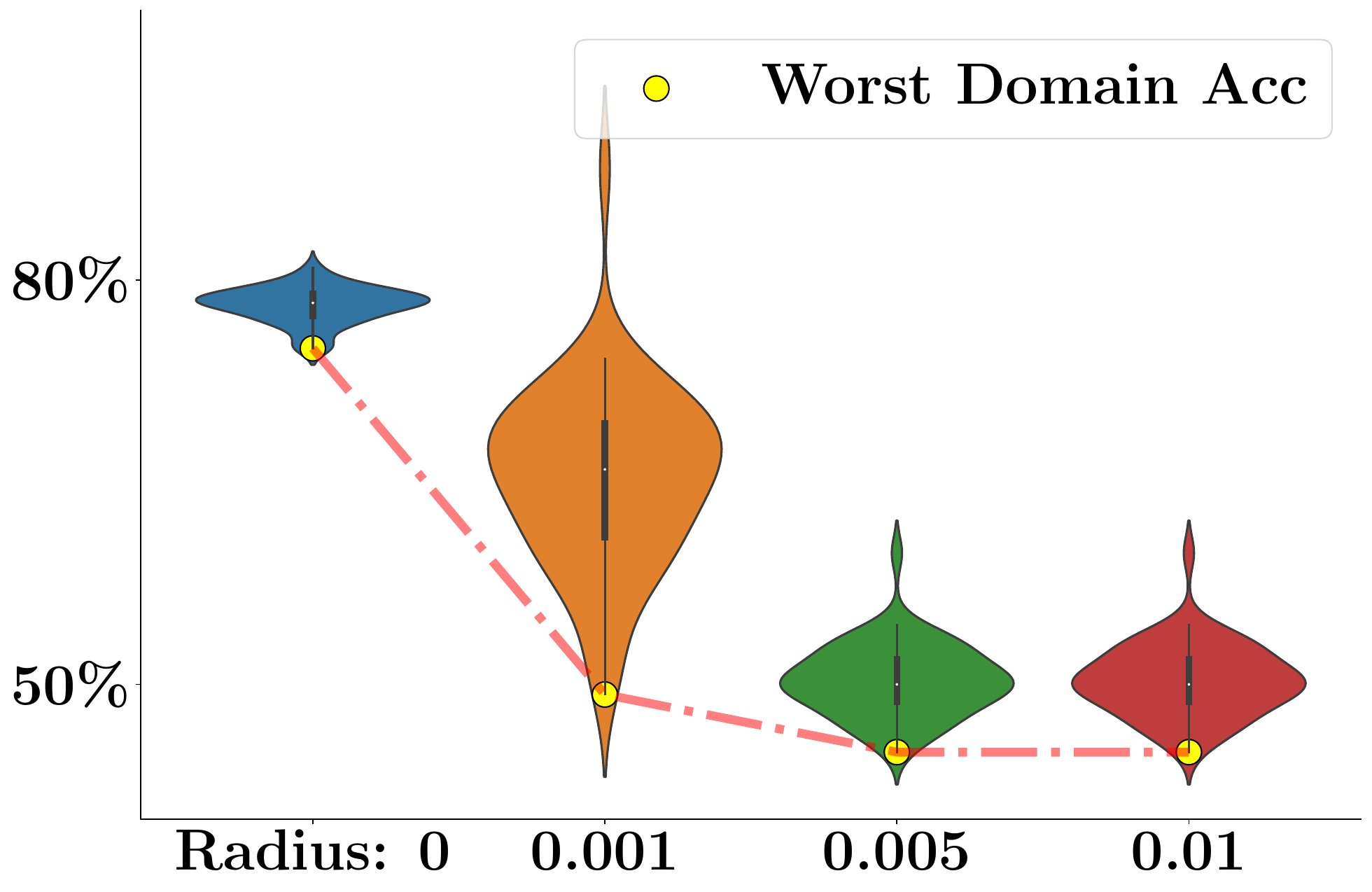}}{\scriptsize (b) \texttt{ACS Income}, Wasserstein-DRO}
 \stackunder[3pt]{\includegraphics[width=0.33\textwidth]{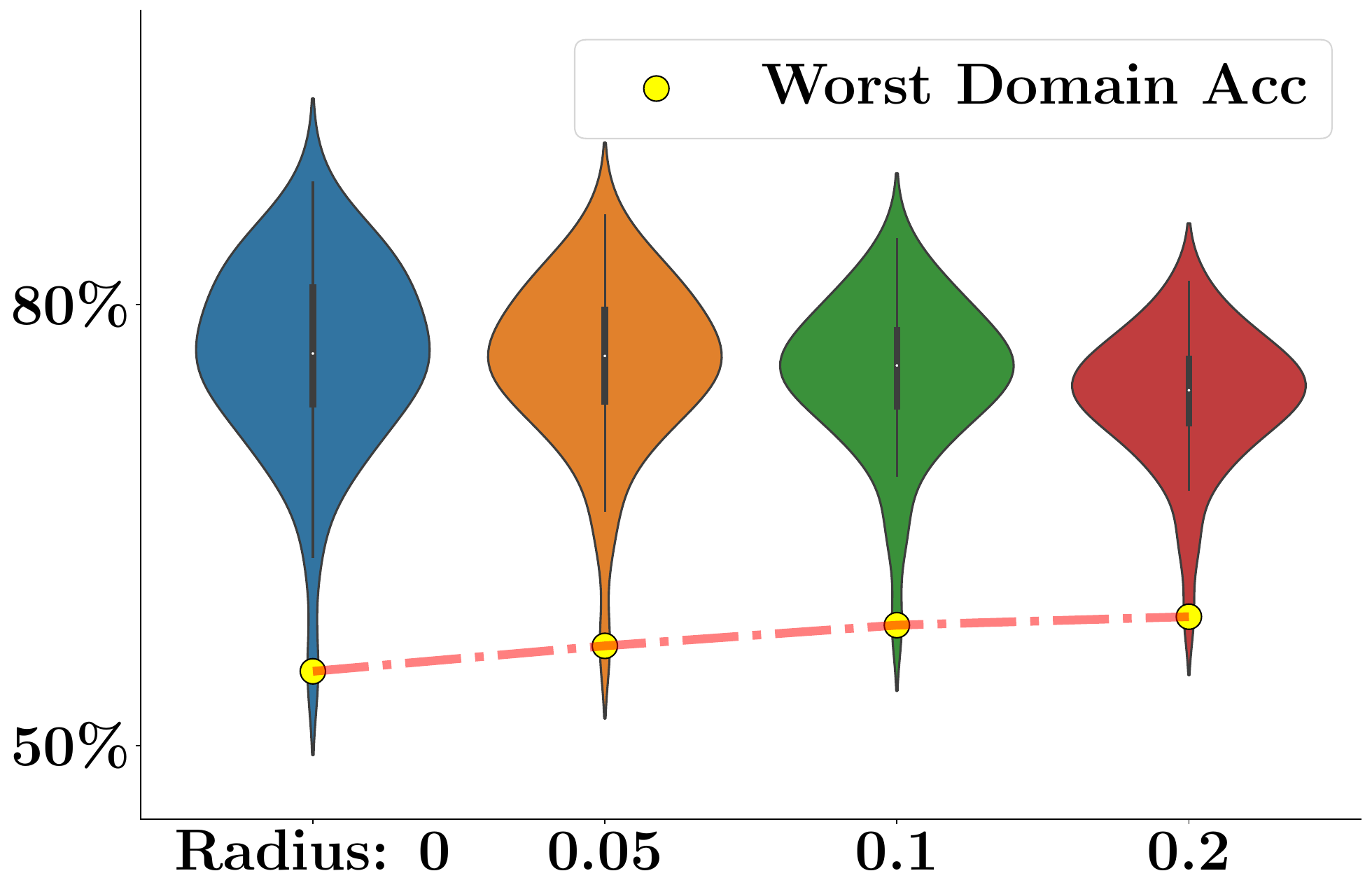}}{\scriptsize (c) \texttt{ACS Pub.Cov}, KL-DRO}
\vspace{0.1in}
\caption{Transfer accuracy~\eqref{eq:transfer-acc} on \texttt{ACS Income} (Setting 1) and \texttt{ACS Pub.Cov} datasets (Setting 4). In each figure, the first bar represents the transfer accuracies from the training distribution $\what{P}$ to each of the 50 target domains ($\{\text{TAcc}(\what{P},\what{Q}_t)\}_{t\in[50]}$), and the rest three bars represent the transfer accuracy from the worst-case distribution $\what{P}^\star$ to 50 target domains ($\{\text{TAcc}(\what{P}^\star,\what{Q}_t)\}_{t\in[50]}$). The model class $\Fscr$ used here is LGBM. Results of other model classes can be found in Appendix~\ref{subsec:more_results_worst}.}
 \label{fig:worst-distribution}  
 \vspace{-0.15in}
\end{figure}

In \Cref{fig:worst-distribution} (a)-(b), we present violin plots for $\{\text{TAcc}(\what{P}^\star,\what{Q}_t)\}_{t\in[50]}$, where we observe a consistent decrease in generalization performance across all 50 target domains as the radius $\epsilon$ increases.  
This trend suggests a fundamental misalignment between the constructed worst-case distributions $\what{P}^\star$ and the actual target domains $\Qscr_{50}$, challenging the practical validity of the current DRO approach under varying degrees of distributional robustness. 
This insight not only underscores the complexity of achieving distributional robustness but also signals the necessity for refined strategies to build a more reasonable ambiguity set in DRO.
See \Cref{subsec:appendix-training-detail-4.2} for training details. To check the generalizability of our findings, in in~\Cref{subsec:more_results_worst} we select a wide range of model classes $\Fscr$ including LR, RF, and XGB, and observe consistent trends.

\vspace{-0.15in}
\paragraph{DRO improves worst target performance in some cases, but is still pessimistic in general.}
In~\Cref{subsec:lr-empirical}, we observed that the impact of DRO methods was not as large as other implementation details such as the underlying model class and validation types. 
A notable exception is that KL-DRO significantly improves the worst-domain accuracy on the \texttt{ACS Pub.Cov} dataset (Setting 4, see ``\texttt{Worst Domain}'' column in \Cref{tab:linear-analysis}). 
To understand this improvement, we investigate the worst-case distribution of KL-DRO by calculating the transfer accuracy as done above.
Specifically, we vary the radius of KL-DRO ($\epsilon\in\{0.05,0.1,0.2\}$), obtain the corresponding worst-case distributions $\what{P}^\star$, and plot the violin bars of $\{\text{TAcc}(\what{P}^\star,\what{Q}_t)\}_{t\in[50]}$.
In \Cref{fig:worst-distribution} (c), we find a noticeable improvement in the worst-case accuracy across all 50 target domains under such setup, i.e., the lower bounds of the right three bars are marginally higher than that of the far-left bar (see red dotted line). 
This suggests that KL-DRO retains the potential to enhance worst-case performance in practice. 
Nonetheless, in such case, the overall performance across the 50 target domains remains conservative compared with the ERM counterpart (the leftmost bar with $\epsilon=0$).
This observation highlights the need for further work to develop ambiguity sets or worst-case distributions that better reflect \twy{semi-synthetic} target scenarios.

\section{Interventions based on modeling of
distribution shifts}
\label{sec:intervention}

Our empirical analysis so far underscores the importance
of complementing the current theory-driven algorithmic development paradigm with a \twy{data-driven} modeling approach grounded in \twy{semi-synthetic} distribution shifts.
In this section, we illustrate how a nuanced characterization of shifts can lead to more effective interventions. 
We go beyond the usual focus on training algorithms and advocate for a holistic approach that encompasses data-centric and algorithmic interventions. \tyw{Below, we demonstrate insights obtained from a small amount of target data. While the number of target samples is insufficient to retrain the entire method, we demonstrate even this small set can help modify the model training or data collection procedure to improve performance. This aligns with the empirical findings presented in Section~\ref{subsec:acc-on-the-line}.}

\subsection{Algorithmic Interventions}\label{sec:alg_intervention}
In this part, we examine how certain \twy{data-driven} modeling techniques, particularly specific features that suffer substantial distribution shifts, can significantly boost the practical robustness of DRO methods. \wty{We formulate this algorithmic intervention as a principled framework:
\begin{figure}[!htb]
    \centering
    \includegraphics[width=0.9\linewidth]{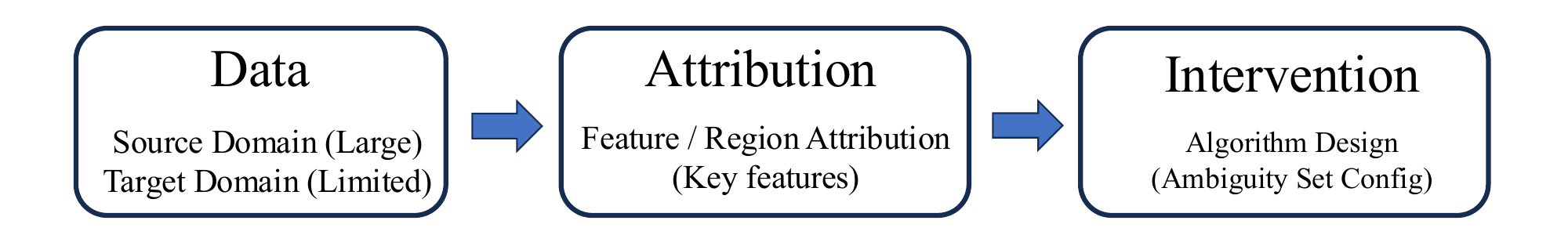}
\end{figure}}

\wty{In general, we attribute performance degradation to specific distribution shifts given abundant data from the source domain and limited data from the target domain. This allows us to identify critical features that need protection. Specifically, we design models focusing on the subset of features $X$ and $Y$ that are subject to distribution shifts. Finally, we incorporate this refined characterization of distribution shifts into the algorithmic intervention step, e.g., adjusting the ambiguity set for DRO approaches.
}

In the following, \wty{we first describe the algorithm intervention procedure to integrate refined shift patterns.} We explore three DRO approaches that integrate such refined distribution shift patterns into the ambiguity set, i.e., Marginal-DRO~\citep{duchi2023distributionally}, Wasserstein-DRO~\citep{blanchet2019data}, and Conditional-DRO~\citep{sahoo2022learning} given their flexibility in controlling features to modify the ambiguity set and illustrate that a small amount of data from the target domain can build such knowledge empirically. We hope to inspire future methodological work, and
only present highly preliminary results. 
In the following, we set $\widehat P, \widehat P_X, \widehat P_Z$ as the empirical distribution of the whole training set, all covariates $X$, subset $Z$ (i.e., subsets of covariate $X$ and $Y$), unless specified.
\vspace{-0.15in}
\paragraph{Marginal-DRO.} The objective function of Marginal-DRO is:
\vspace{-0.1in}
\begin{equation}\label{eq:mdro-formula}
\begin{aligned}
	\min_{f\in \Fscr} \sup_{Q_0\in \Pscr_{\alpha, Z}}\mathbb E_{Z\sim Q_0}\left[\mathbb E[\ell_{tr}(f(X), Y)|Z]\right],
\end{aligned}
\vspace{-0.1in}
\end{equation}
where $\Pscr_{\alpha, Z} \coloneqq \{Q_0: \widehat P_Z=\beta Q_0 + (1-\beta)Q_1 \text{ for some } \beta\geq \alpha \text{ and distribution }Q_1\text{ on }\Zscr\}$ denotes the ambiguity set of distributions on $\Zscr$, and $Z\subseteq \{(X, Y)\}$ denotes a subset of features $X$ and $Y$.
To improve practical applicability, we can pinpoint features associated with shifts and construct the distribution set focusing solely on these tailored features.
For the hyperparameter $\alpha$, we first choose the best $\alpha\in\{0.1, 0.2, \dots, 0.9\}$ according to 128 samples drawn from the target domain and fix it for the subsequent intervention.

\vspace{-0.15in}
\paragraph{Wasserstein-DRO.} Wasserstein-DRO builds the ambiguity set on the Wasserstein distance, taking the form of:
\begin{equation}\label{eq:wdro-formula}
	\begin{aligned}
		&\min_{f\in\Fscr} \sup_{Q: W_c(Q,\widehat P)\leq \epsilon} \mathbb E_{X,Y\sim Q}\left[\ell_{tr}( f(X),Y)\right],\\
		\text{where } & W_c(Q, P)= \min_{\pi\in \Pi(Q,P)} \mathbb E_{Z_1,Z_2 \sim \pi}\left[c(Z_1,Z_2)\right],
	\end{aligned}
\end{equation}
where $c(Z_1,Z_2)$ denotes the transport cost between samples $Z_1$ and $Z_2$.
Motivated by~\citet{blanchet2019data}, we select the transport cost as the Mahalanobis distance:
\vspace{-0.1in}
\begin{equation*}
	c(Z_1, Z_2) = (Z_1-Z_2)^{\top}\Sigma^{-1}(Z_1-Z_2),
 \vspace{-0.1in}
\end{equation*}
where $\Sigma \in \R^{d_z \times d_z}$ is a positive semidefinite matrix. Note that when $\Sigma = \mathbf I$, $c(\cdot, \cdot)$ becomes the squared $\ell_2$-metric. To incorporate the distribution shift patterns, 
we adjust the matrix $\Sigma$ to selectively perturb these specific features while leaving the rest unaffected.
Specifically, we select the top-$K$ features associated with specific distribution shifts, and set the matrix $\Sigma=\text{diag}(v)$ with $v\in\mathbb R^{d_z}$. 
If feature $i$ is among the top-$K$ features, $v_i=+\infty$; otherwise $v_i=1$. 
For the hyperparameter $\epsilon$, we choose the best $\epsilon$ according to the in-distribution accuracy obtained from the original Wasserstein-DRO and fix it during the subsequent intervention.
\vspace{-0.15in}
\paragraph{Conditional-DRO.} The objective function of Conditional-DRO is:
\vspace{-0.1in}
\begin{equation}\label{eq:cdro-formula}
	\min_{f\in\Fscr} \sup_{Q\in \Pscr_{\Gamma}(\widehat P_Z)} \mathbb E_{X,Y\sim Q}[\ell_{tr}(f(X), Y)],
 \vspace{-0.1in}
\end{equation}
where $\Pscr_{\Gamma}(P)=\{Q: \Gamma^{-1}\leq \frac{dQ_{Y|Z=z}(y)}{dP_{Y|Z=z}(y)}\leq \Gamma, \forall y, \forall z; \text{ and } \sup_{z\in\Zscr}\frac{dP_Z(z)}{dQ_Z(z)}<\infty \}$.
For the algorithmic intervention, we choose the feature $Z$ as the one with large $Y|Z$-shifts.
For the hyperparameter $\Gamma$, we choose the best $\Gamma\in\{1.0, 1.1, \dots, 1.9, 2, 3, \dots, 10\}$ according to 128 samples drawn $i.i.d$ from the target domain and fix it during the intervention.

We summarize the key components of different existing DRO methods in \Cref{tab:intervention}.
Note that the hyperparameters of these three DRO methods are fixed during the intervention, and we only modify the covariates used to build the distance.

\begin{table}[htb]
    \centering
    \caption{Interventions of existing DRO methods, where the two columns ``$X$-shift'' and ``$Y|X$-shift'' denote how the interventions in these DRO methods work.}
    \vspace{0.1in}
    \label{tab:intervention}
    \resizebox{\textwidth}{!}{\begin{tabular}{c|cccc}
    \toprule
         & Hyperparameter & Intervention Choice & $X$-shift & $Y|X$-shift\\
        \midrule
       Marginal-DRO \eqref{eq:mdro-formula} & $\alpha$ & $Z$ & subset of $X$ & $\{\text{subset of }X\}$ + $Y$\\
       Wasserstein-DRO \eqref{eq:wdro-formula} & $\epsilon$ & $\Sigma$ & diagonal element size of $\Sigma$ & - \\
       Conditional-DRO \eqref{eq:cdro-formula} & $\Gamma$ & $Z$ & - & subset of $X$ \\
    \bottomrule
    \end{tabular}}
\end{table}

In~\eqref{eq:mdro-formula},~\eqref{eq:wdro-formula} and~\eqref{eq:cdro-formula}, we set $\Fscr$ as the linear model class and $\ell_{tr}(\cdot, \cdot)$ as the hinge loss for binary classification, i.e., SVM as our model class. We comment that while our previous empirical analysis highlights the importance of the basic model class, it is unclear how to adapt powerful model classes for tabular data (e.g., XGBoost) for distributionally robust objectives and we leave this important problem to future work.
\wty{In the following Sections~\ref{sec:casestudy-xshift} and~\ref{sec:casestudy-yshift}, we focus on two case studies, i.e., Settings 7 and 4 with their specific attribution procedures and sensitivity analysis of the selected subset of features $Z$. And we provide results of other settings in Appendix~\ref{app:result_alg_intervention}.} 

\begin{figure}[!htb]
	\includegraphics[width=\textwidth]{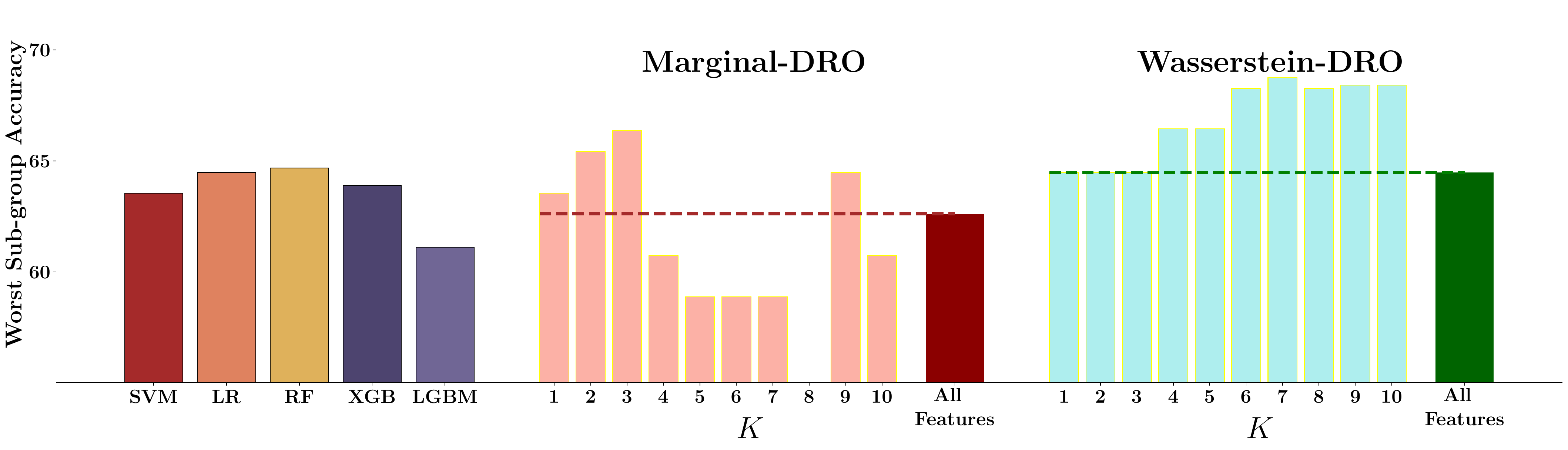}
	\caption{Algorithmic interventions under $X$-shifts. We report the worst subgroup accuracy for each method. }
	\label{fig:x-shift}
 \vspace{-0.15in}
\end{figure}

\subsubsection{Case Study: $X$-shifts}\label{sec:casestudy-xshift}
We begin with a case study on demographic shifts within the \texttt{ACS Income} dataset (refer to Setting 7 in \Cref{table:overview}), which mainly involves $X$-shifts. 
Recall in this setting, the source domain contains 80\% individuals aged 25 or older, and the ratio is reversed in the target domain.
We sample 2000 points to train a method, and another 2000 points as the validation set for hyperparameter selection from the source domain.
In the target domain, we divide age groups into intervals: [20,25), [25,30), [30,35), $\dots$, [95,100), and evaluate the worst-group accuracy for each method across these age groups.

\paragraph{Attribution.} To design a more reasonable ambiguity set, we consider the age group structure on the validation set $\Dscr_{\text{val}}$, and denote the data set of the worst-case age group within that validation set as $\Dscr_{\text{worst}}$. Recall the feature vector $X = (X^{(1)}, \ldots, X^{(d_x)})^{\top}$. As a preliminary indicator of the magnitude of covariate shift, we compare the mean difference of each feature component between $\Dscr_{\text{val}}$ and $\Dscr_{\text{worst}}$ via the following score:
\vspace{-0.1in}
\begin{equation}
	s_i = \frac{\left|\mathbb E_{\Dscr_{\text{val}}}[X^{(i)}] - \mathbb E_{\Dscr_{\text{worst}}}[X^{(i)}]\right|}{\min\{\mathbb E_{\Dscr_{\text{val}}}[X^{(i)}], \mathbb E_{\Dscr_{\text{worst}}}[X^{(i)}]\}},~\forall i \in [d_x].
 \vspace{-0.1in}
\end{equation}
A higher score indicates a more significant covariate shift associated with that feature $i$.

Consequently, we select $K$ features ($Z$) with the highest scores and set Marginal-DRO and Wasserstein-DRO only to perturb distributions over top-$K$ features in \Cref{tab:intervention}. As illustrated in \Cref{fig:x-shift}, this strategic intervention markedly improves the accuracy for the worst-case groups. 
For some choices of $K$, this ``targeted'' DRO approach even significantly outperforms tree-based ensemble methods.

\subsubsection{Case Study: $Y|X$-shifts}\label{sec:casestudy-yshift}
Building upon this foundation, we consider a case study within the \texttt{ACS Pub.Cov} dataset (refer to Setting 4 in \Cref{table:overview}) with the source domain being Nebraska (NE) and the target being Louisiana (LA) in this section. For the LR method, we observe a notable degradation in performance between the source and target domains, decreasing from 83.9\% to 66.4\%. 


\wty{\paragraph{Attribution.} We quantify feature importance under distribution shifts by computing the difference in aggregated Shapley values between the source and target domains~\citep{lundberg2017unified} (see Appendix~\ref{app:feature-attribution} for details).} We find that the conditional distribution $Y|\text{Income}$ exhibits significant shifts between the source and target domains. This observation is consistent with the findings presented in \citet[Section 6]{DBLP:journals/corr/abs-2402-14254}, which identified the ``Income'' feature with the largest shifts in $Y|X$ among all features. \wty{To demonstrate the stability of the attribution results, we select the Top-$K$ ($K \leq 5$) features with the highest scores. In Setting 4, these Top-5 features correspond to ``Income'', ``married'', ``never'', ``ESR\_employed'', ``divorced''. We then incorporate these most significant identified shifts into the ambiguity set: for each $K \leq 5$, in the Marginal-DRO approach, we set $Z$ as $\{\text{Top-$K$ features}, Y\}$; whereas in the Conditional-DRO approach, we set $Z$ to $\{\text{Top-$K$ features}\}$ only.}

\begin{figure}[!htb]
	\includegraphics[width=\textwidth]{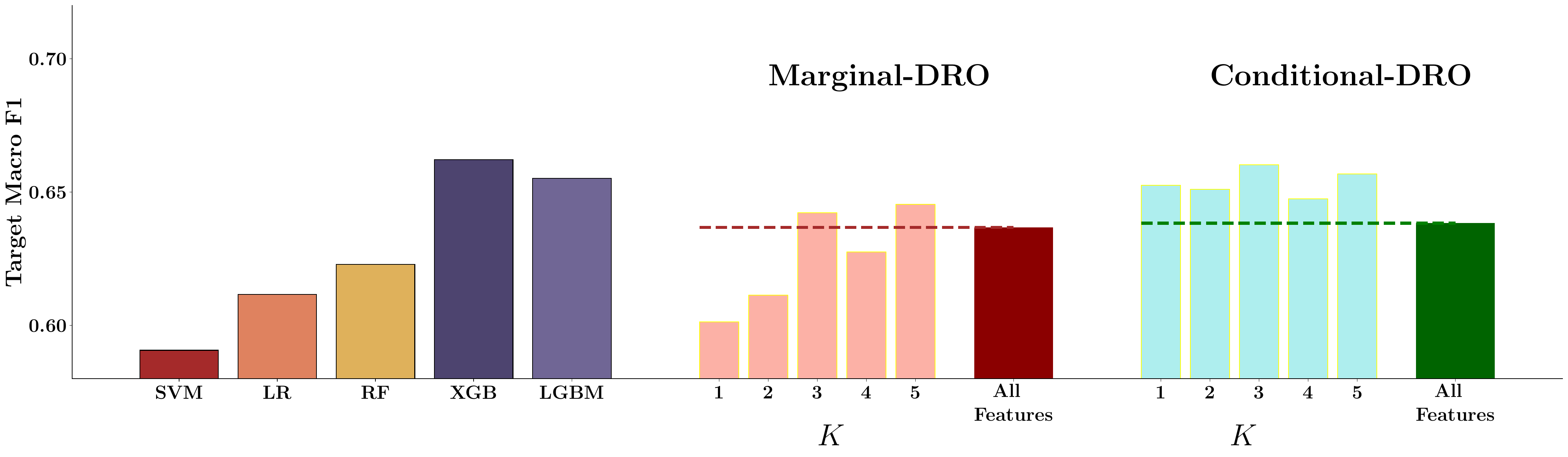}
	\caption{Algorithmic interventions under $Y|X$-shifts. We report the Macro-F1 score for each method.}
	\label{fig:Y|X}
\end{figure}

As illustrated in \Cref{fig:Y|X}, our algorithmic interventions significantly enhance the generalization performance of DRO methods \wty{for all $K$ in the Conditional-DRO approach and some $K$ in the Marginal-DRO approach. For some choices of $K$, the ``targeted'' DRO approach also achieves performance comparable to or even outperforms tree-based ensemble methods significantly. } 

We also comment that the improved performance trend may not always hold, e.g., we observe variability of performance when we increase $K$ in Marginal-DRO as in Figures~\ref{fig:x-shift} and~\ref{fig:Y|X}. However, this straightforward intervention indicates a promising direction for refining DRO approaches and these two case studies represent the first effort to integrate insights from distribution shifts into the development of DRO methodologies. 

{\color{black}\subsubsection{Case Study: A New Ambiguity Set to Generalize Across All Target Domains}\label{subsec:llm-shift}
Beyond the single source-target experiments in Sections~\ref{sec:casestudy-xshift} and~\ref{sec:casestudy-yshift}, we now turn to a richer case study on the \texttt{ACS Income} dataset (refer to Setting 1 in \Cref{table:overview}) and focus on the method performance over all target domains. To further mitigate DRO's well-known over-conservativeness, we introduce a \emph{Sliced-Marginal-DRO} that defines its ambiguity set in a low‐dimensional feature space:
\begin{equation}\label{eq:mdro-formula2}
\begin{aligned}
	\min_{f\in \Fscr} \sup_{Q_0\in \Pscr_{\alpha, Z}}\mathbb E_{Z\sim Q_0}\left[\mathbb E[\ell_{tr}(f(\Pi_{\tilde d}X), Y)|Z]\right],
\end{aligned}
\end{equation}
where $\Pscr_{\alpha, Z} \coloneqq \{Q_0: \widehat P_Z=\beta Q_0 + (1-\beta)Q_1 \text{ for some } \beta\geq \alpha \text{ and distribution }Q_1\text{ on }\Zscr\}$ denotes the ambiguity set of distributions on $\Zscr$, $\Pi_{\tilde d}$ projects $X$ onto its first $\tilde d$ principal components and $Z\subseteq \{(\Pi_{\tilde d}X, Y)\}$ denotes a subset of features $\Pi_{\tilde d}X$ and the label $Y$. We call this \emph{Sliced-Marginal-DRO} because, instead of forming an ambiguity set over the full feature space, we ``slice'' it along a low-dimensional subspace: first reducing to the top $\tilde d$ PCA components, and then further restricting attention to the subset of most important coordinates via limited target data. This two-step slicing focuses robustness on the subspace where shifts matter most.

\paragraph{Dimensionality Selection and Attribution.} First, we vary $\tilde d$ and select the value that minimizes the drop in target accuracy measured using target samples from all target domains. In Setting 1, this yields $\tilde d = 52$. Then for each reduced feature, we compute its importance:
\[\text{Score}_i = \Pi_{\tilde d}(w \Delta + (1 - w)\Delta_{\text{LLM}})_i, \forall i \in [\tilde d];\]
where $\Delta$ is the attribution from Appendix~\ref{app:feature-attribution} (averaged over all target domains), $\Delta_{\text{LLM}}$ comes from a Gemini-2.5-pro prompt and $w = 0.9$.  Finally, we select the top-$K$ features with the highest scores and build the ambiguity set of the Sliced-Marginal-DRO method only over these $K$ coordinates of $\Pi_{\tilde d}X$.

\Cref{fig:Y|X2} reports the average target accuracy across 50 target domains. We find that both Sliced-Marginal-DRO and {Marginal-DRO} methods on the top-$K$ features outperform the standard linear-SVM baseline and match the performance of tree-based ensembles (XGBoost or LightGBM) for some $K$ ($K = 4$ for Marginal-DRO and $K = 1$ for Sliced-Marginal-DRO). Furthermore, a two-sample test by comparing accuracies across all target domains, we find that the performance of the {Sliced-Marginal-DRO} over the top-1 feature yields a statistically significant improvement over both (i) {Sliced-Marginal-DRO} over all 52 reduced features and (ii) Marginal-DRO over the top-1 feature. This indicates the importance of incorporating both algorithmic design (dimension reduction and feature selection) in this case. 
\begin{figure}[!htb]
	\includegraphics[width=\textwidth]{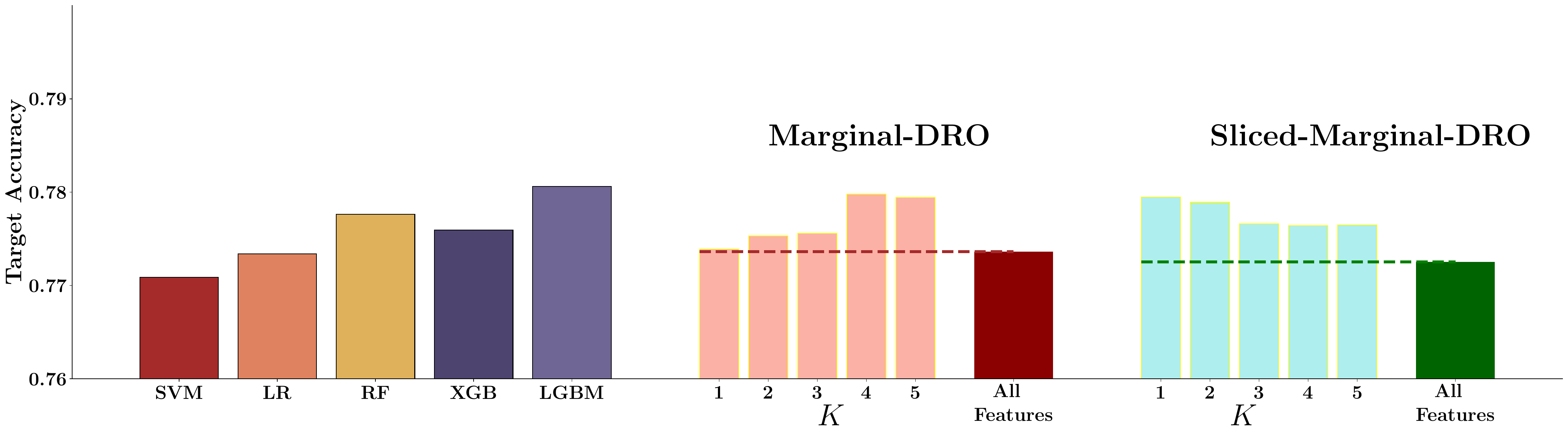}
	\caption{Algorithmic interventions across all target domains in Setting 1. We report the averaged accuracy across all target domains for each method.}
	\label{fig:Y|X2}
\end{figure}

Besides the linear-SVM, when we apply the same dimensionality‐reduction and slicing pipeline to XGBoost and LightGBM (i.e., their CVaR-DRO variants), we observe no further gains. This suggests that refining the ambiguity set on only a handful of features may be most effective for linear methods, and that extending this refinement sensibly to tree ensembles warrants further investigation.
}
\subsection{Data-centric Interventions}
\label{subsec:data-intervention}
Besides algorithmic interventions, collecting additional information (samples or features) is a natural intervention for addressing distribution shift. While data-centric interventions have received little attention, we showcase the potential for developing principled algorithms for target data collection. In particular, we
focus on $Y|X$-shifts due to its challenging nature. 
To illustrate the
promising nature of data-centric interventions and spur further research in this
direction, 
we show that concerted data collection in a small subset of the covariate region based on some knowledge of when distribution shifts degrade model performance can effectively address $Y|X$ changes.
Before that, we first describe how to identify covariate regions with strong $Y|X$-shifts.

\subsubsection{Identifying Covariate Regions with Large $Y|X$-shifts}\label{subsubsec:riskregion}

We study a simple yet effective algorithm that identifies covariate regions with strong $Y|X$-shifts, i.e., \emph{identifying a region
    $\mathcal R\subseteq \mathcal X$ where $P_{Y|X}$ differs a lot from
    $Q_{Y|X}$} given samples $(X,Y)$ drawn from the source
  and target distributions $P$ and $Q$.
Since $P_{Y|X}$ and $Q_{Y|X}$ are undefined outside the support of $P_X$
and $Q_X$, it only makes sense to compare their difference in their \emph{common support}. Without evaluating the performance on the shared distribution $S_X$, it is hard to distinguish the source of the model performance degradation, i.e., from $X$-shifts or $Y|X$-shifts. To facilitate this comparison, \citet{namkoong2023diagnosing} introduces a shared distribution $s_X(x)\propto \frac{p_X(x)q_X(x)}{p_X(x)+q_X(x)}$ that has
high density when both $p_X$ and $q_X$ are high, and low density whenever either is small. We provide more discussion on the choices and correctness of $s_X$ in \Cref{appendix:algorithm1}.
Since we do not have access to samples from the shared distribution $S_X$, we
reweight samples from $P_X$ and $Q_X$ using the likelihood ratios. Specifically, if we denote $\alpha^*$ as the proportion of the pooled data that comes from $Q_X$ and $\pi^*(x):= \P(X\text{ from }Q_X| X = x)$,  we can express the likelihood ratios as
\begin{align}
	\label{equ:region1}
  \frac{s_X}{p_X}(x) &\propto \frac{\pi^*(x)}{(1-\alpha^*)\pi^*(x)+\alpha^*(1-\pi^*(x))}=:w_P(\pi^*(x),\alpha^*),\\
  \label{equ:region2}
  \frac{s_X}{q_X}(x) &\propto \frac{1-\pi^*(x)}{(1-\alpha^*)\pi^*(x)+\alpha^*(1-\pi^*(x))}=:w_Q(\pi^*(x),\alpha^*).
\end{align}	
The ratio are seen as the probability that an input $x$ is from $P_X$ or $Q_X$ respectively. Equipped with Equations~\eqref{equ:region1} and~\eqref{equ:region2}, we can estimate the best prediction models under $P$ and $Q$ over the shared distribution $S_X$ for $\mu = P, Q$:
  \vspace{-0.1in}
  \begin{align}
    \label{equ:fp-fq}
    f_{\mu} := &\arg\min_{f\in \Fscr}\left\{\mathbb{E}_{S_X}\Paran{\mathbb{E}_\mu[\ell_{tr}(f(X),Y)|X]} \left(= \mathbb{E}_\mu\Paran{\ell_{tr}(f(X),Y) w_{\mu}(\pi^*(x),\alpha^*)}\right)\right\}.
    \vspace{-0.1in}
\end{align}	
Then, for any threshold $b\in [0, 1]$,
$\{x \in \Xscr : |f_P(x)-f_Q(x)| \geq b\}$ represents a region that suffers
model performance degradation with at least $b$ due to $Y|X$-shifts. 

Under our setting, given samples $\{(x_i^P,y_i^P)\}_{i \in [n_P]}$ from $P$ and $\{(x_j^Q,y_j^Q)\}_{j \in [n_Q]}$ from $Q$, we estimate $\hat{\alpha} = \frac{n_Q}{n_P+n_Q}$ and then train a binary ``domain'' classifier $\hat \pi(x)$ to approximate the ratio $\pi^*(x)$. Note that the ``domain'' classifier can be any black-box method, and we use XGBoost throughout. Then we plug these empirical estimands in to obtain the estimated likelihood ratios $w_{\mu}(\hat\pi(x),\hat\alpha)$ and learn prediction models $f_P$ and $f_Q$ in~\eqref{equ:fp-fq}. Then we learn a prediction model $h(x)$ to approximate $|f_P(x)-f_Q(x)|$ on the shared distribution $S_X$.
The pseudo-code is summarized in the Algorithm
\ref{algo}. To allow simple interpretation and efficient region identification, we use a shallow \emph{decision tree} $h(x)$ and consider the region $\Rscr$ corresponding to the feature range of a leaf node within the tree. Here we need the target data to fit a model in~\eqref{equ:fp-fq}. When the amount of target data is relatively small to fit a stable model $f_Q$, we propose a more sample-efficient alternative method that does not require to fit $f_Q$ in~\Cref{app:others}. 
More details could be found in \Cref{app:other-dataintervention}.

\begin{algorithm}[t]
    \KwIn{Source samples $\{(x_i^P, y_i^P)\}_{i \in [n_P]}\overset{\text{i.i.d}}{\sim} P$ and target samples
      $\{(x_j^Q, y_j^Q)\}_{j \in [n_Q]}\overset{\text{i.i.d}}{\sim} Q$. Model
      discrepancy threshold $b$.}
    Estimate $\hat{\pi}(x)\approx\mathbb{P}(\tilde{X}\sim Q_x|\tilde{X}=x)$ by training a classifier on the source and target samples.\\
    Calculate density ratios $w_{\mu}(\hat\pi(x), \hat\alpha)$ according to Equation \eqref{equ:region1} and \eqref{equ:region2} for $\mu = P, Q$.\\
    Fit prediction models $f_{\mu}$ according to Equations \eqref{equ:fp-fq} replacing $w_{\mu}(\pi^*(x),\alpha^*)$ there with $w_{\mu}(\hat\pi(x), \hat\alpha)$ for $\mu = P, Q$, where we set $\Fscr$ as XGBoost;\\
    Fit a model $h(x)$ to predict $|f_P(x)-f_Q(x)|$ using samples $\{(x_i^P, y_i^P)\}_{i \in [n_P]}$ weighted by $\lambda_i^P$ and $\{(x_j^Q, y_j^Q)\}_{j \in [n_Q]}$ weighted by $\lambda_j^Q$, where $\lambda_i^{\mu} = \frac{w_{\mu}(\hat\pi(x_i^{\mu}),\hat\alpha)}{\sum_{k \in [n_{\mu}]} w_{\mu}(\hat\pi(x_k^{\mu}),\hat\alpha)}, \forall i \in [n_{\mu}], \mu \in \{P, Q\}$.
    \BlankLine \KwOut{Region $\Rscr = \{x \in \Xscr: h(x)\geq b\}$.}
    \BlankLine
  \caption{Identify Regions with Strong $Y|X$-Shifts.}
  \label{algo}
\end{algorithm}

\subsubsection{Case Study: Efficient Data \& Feature Collection}
\label{sec:model_intervention}

Using the output region from the (oracle) Algorithm~\ref{algo}, we now demonstrate how a better understanding of
distribution shifts can inform resource-efficient data collection. 
We focus on the
\texttt{ACS Income} dataset (Setting 1) where the goal is to predict whether an
individual's income exceeds 50k ($Y$) based on their tabular census data
($X$).  We train an income classifier on 20,000 samples from California (CA,
source), and deploy the classifier in Puerto Rico and South Dakota (PR \& SD,
target), where we get 500 samples from PR and SD after deployment.  Given
the considerable disparities in the economy, job markets, and cost of living
between CA and PR/SD, we observe substantial performance degradation due to
distribution shifts.

\begin{figure}[!htb]

\subfloat[Raw setting: drop 10.3]{\includegraphics[width=0.255\textwidth, height=0.24\textwidth]{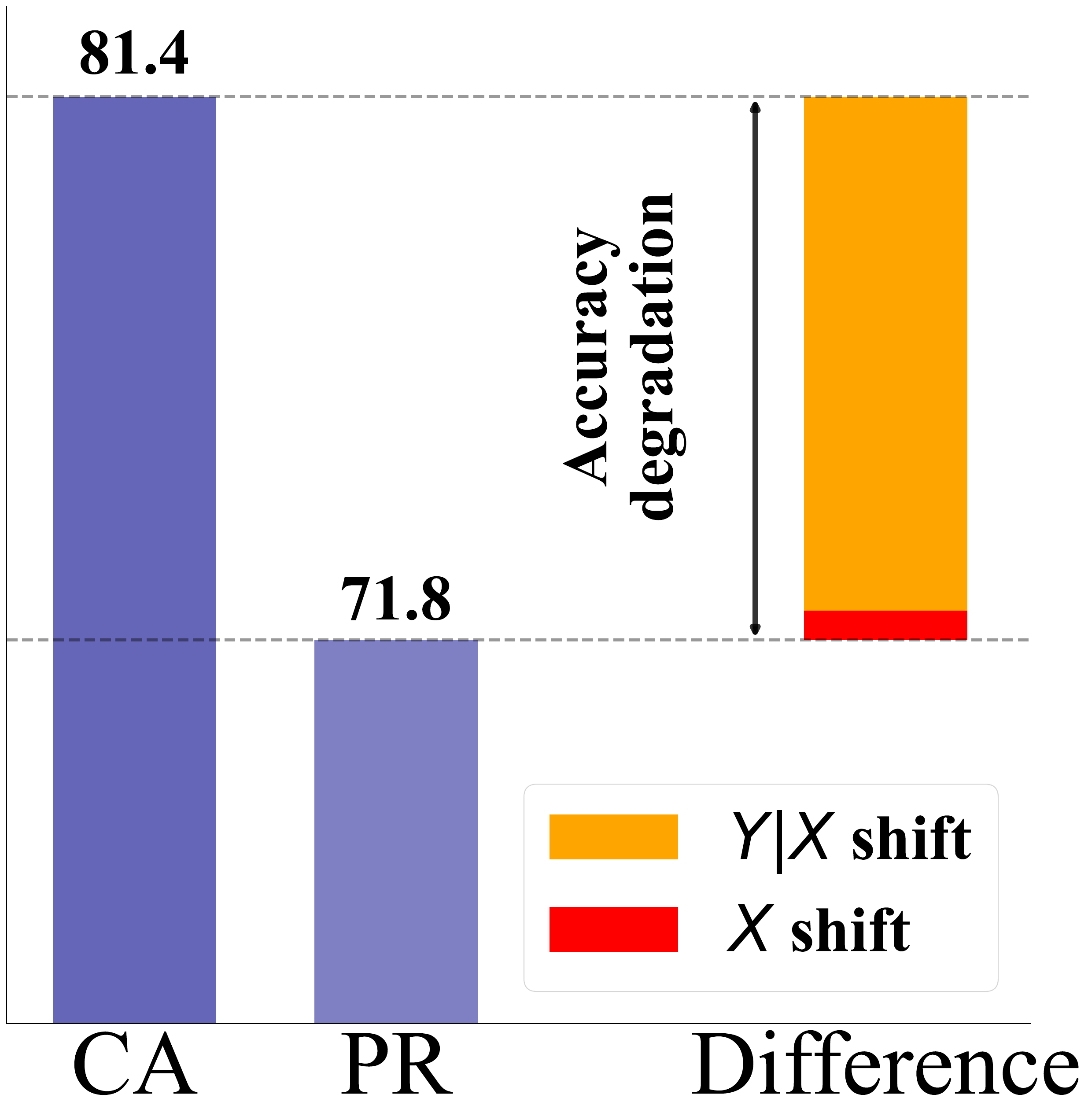}\label{fig:case_subfig1}}
 \hfill
 \subfloat[Add ENG: drop 2.1]{\includegraphics[width=0.235\textwidth, height=0.25\textwidth]{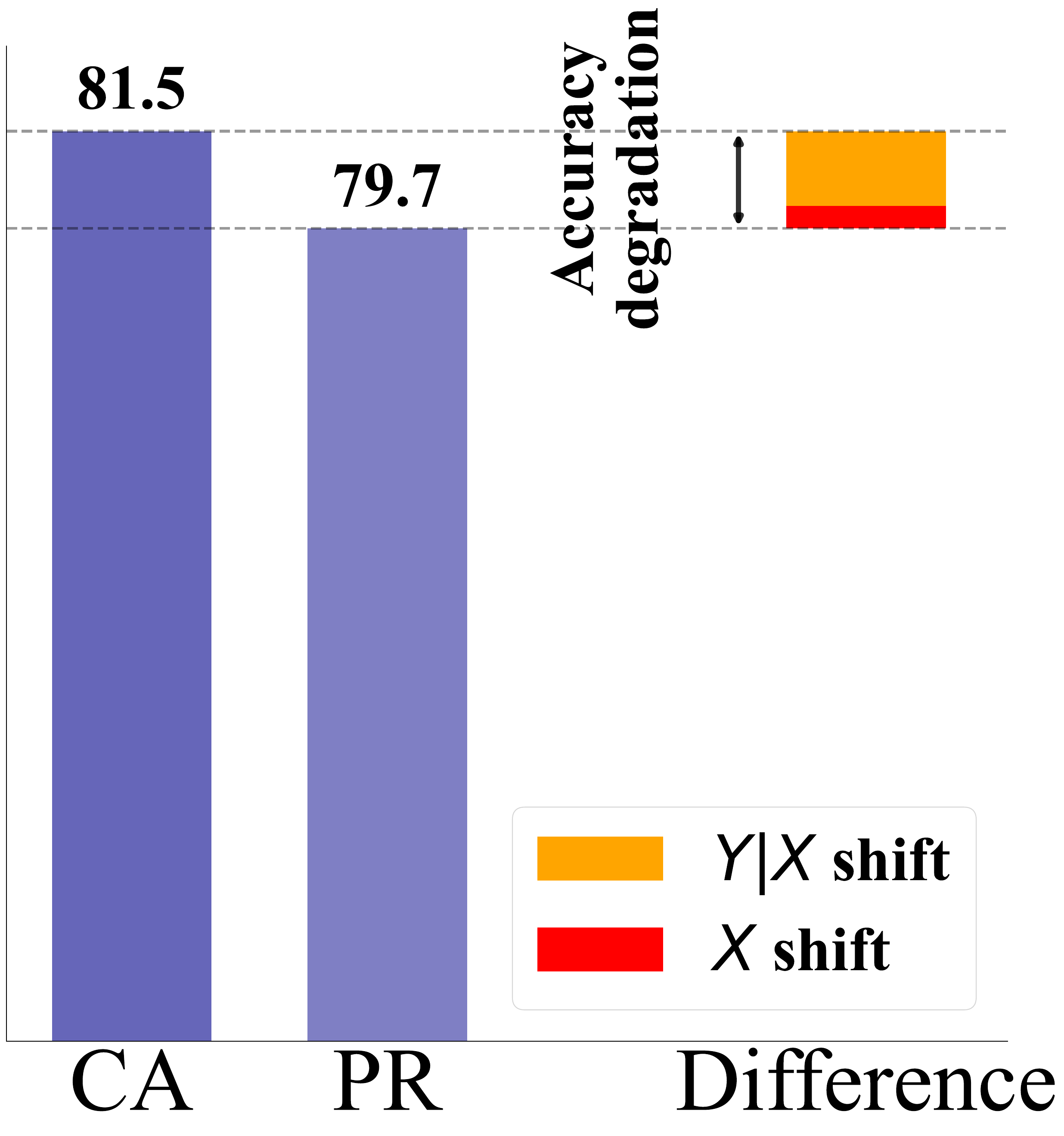}\label{fig:case_subfig2}}
 \hfill
\centering \hfill
 \subfloat[Region with  $Y|X$-shifts (XGB)]{\includegraphics[width=0.4\textwidth]{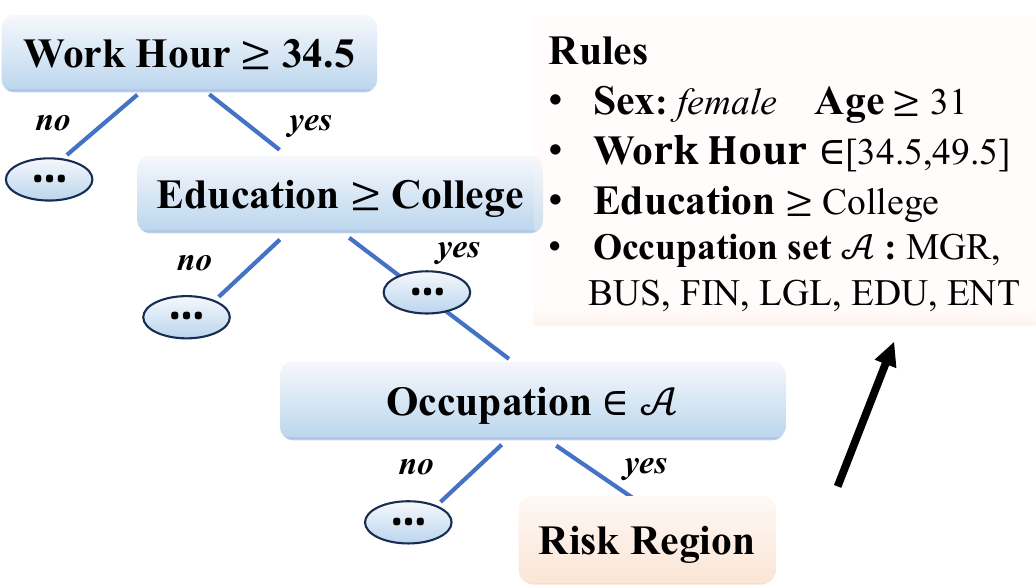}\label{fig:case_subfig5}}
 \hfill
 \subfloat[Region with  $Y|X$-shifts (NN)]{\includegraphics[width=0.4\textwidth]{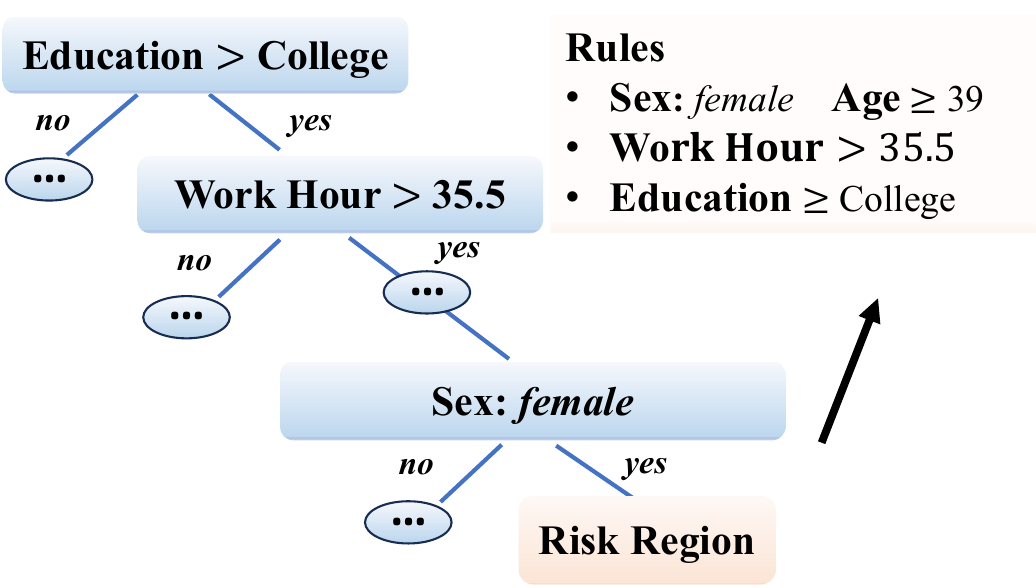}\label{fig:case_subfig6}}
 \hfill 
 \subfloat[Test accuracies of different ways to incorporate data.]{\includegraphics[width=0.55\textwidth, height=0.24\textwidth]{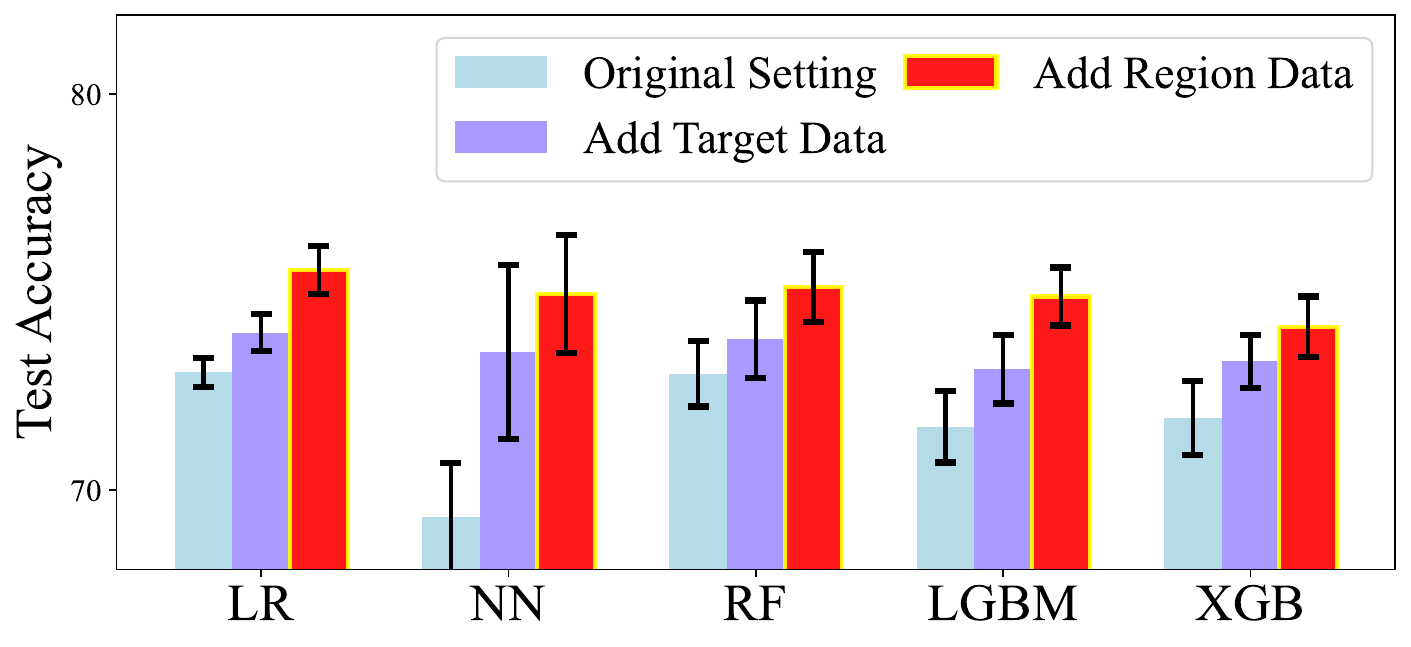}\label{fig:case_subfig7}}
 \hfill
 \subfloat[Add ENG (CA$\rightarrow$PR)]{\includegraphics[width=0.3\textwidth, height=0.3\textwidth]{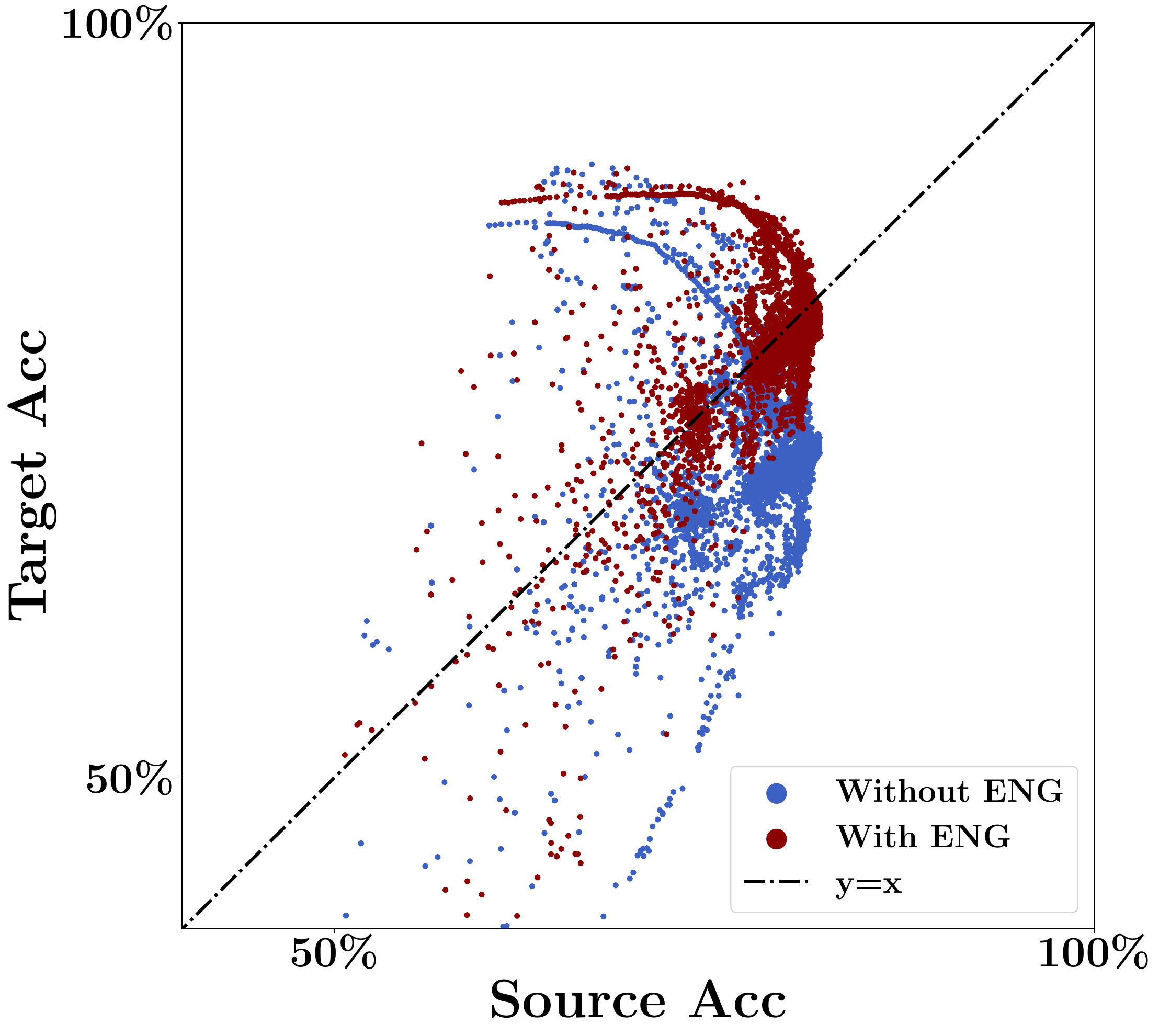}\label{fig:case_subfig3}}
 \hspace{0.3in}
\subfloat[Add ENG (CA$\rightarrow$ SD)]{\includegraphics[width=0.3\textwidth, height=0.3\textwidth]{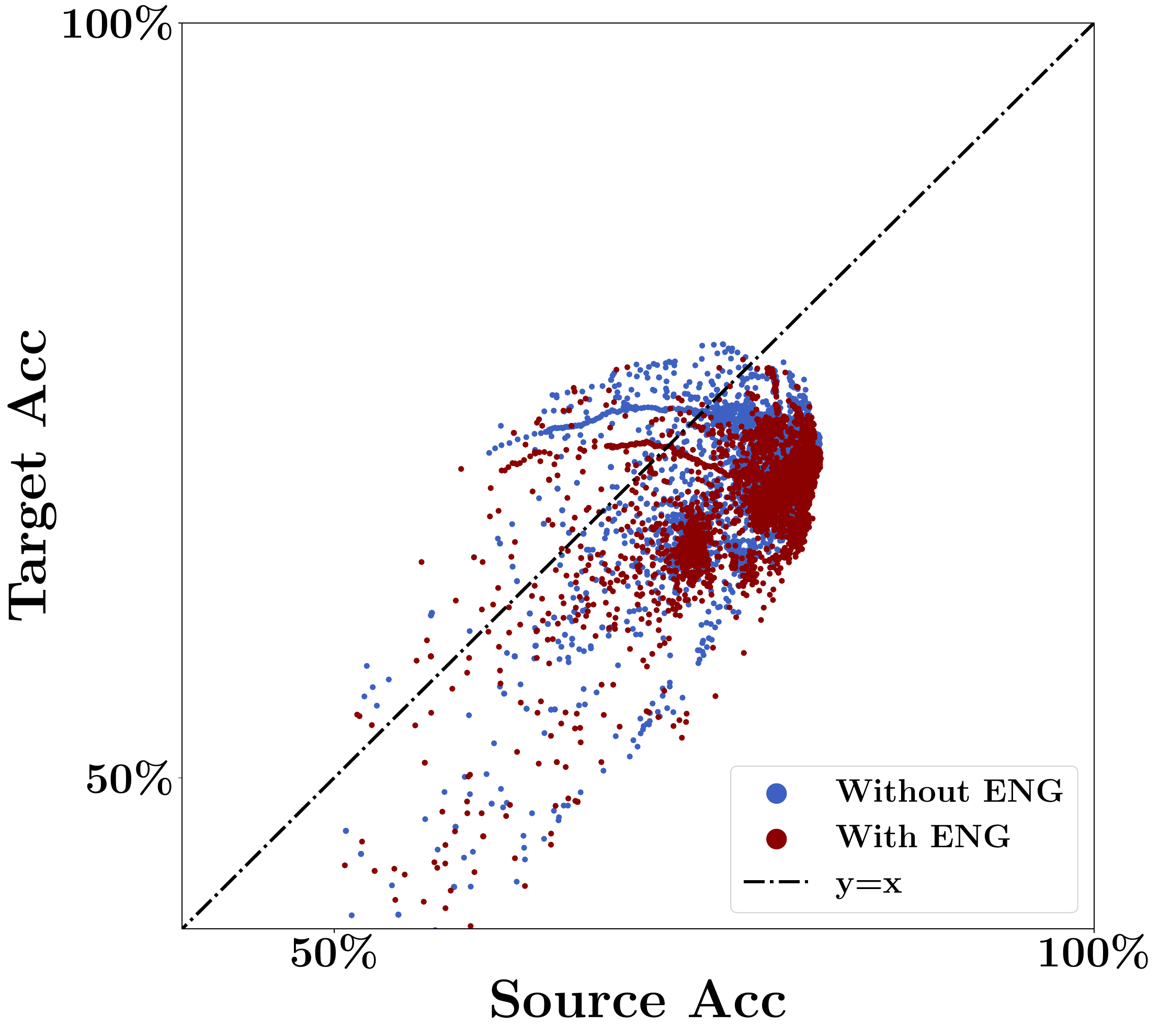}\label{fig:case_subfig4}}
       \caption{Case study illustrations. \textbf{(a)-(b)} Decomposition of performance degradation for the
         XGBoost classifier from CA to PR. Figure (a) is for the original
         setting and (b) corresponds to the results post-integration of the
         "ENG" feature. 
       \textbf{(c)-(d)} Demonstration of
         Algorithm \ref{algo}: an interpretable version of the region with
         strong $Y|X$-shifts for the XGB and NN methods, respectively.  \textbf{(e)} Test
         accuracies of five typical base methods trained on the source, post
         addition of 250 randomly selected \emph{target observations}, and
         250 observations from the identified \emph{covariate region}.
          \textbf{(f)-(g)} Performances of all algorithms
         prior to and following the addition of the "ENG" feature. Figure (f)
         corresponds to the CA to PR, and Figure (g) is CA to SD.}
\label{fig:case_study}  
\vspace{-0.15in}
\end{figure}

In \Cref{fig:case_subfig1}, we first decompose the performance degradation
from CA to PR to understand the shift and find $Y|X$-shifts are the largest. 
We dive deeper into the significant $Y|X$-shifts and identify from CA to PR for the XGB and NN Classifier. 
From the region shown in \Cref{fig:case_subfig5} and \Cref{fig:case_subfig6}, we find college-educated individuals in business and educational roles (such as management, business, and educational work) exhibit large $Y|X$ differences. 

We illustrate how an understanding of distributional differences can inspire subsequent data-centric interventions using two approaches.
\vspace{-0.15in}
\paragraph{Collecting specific data from the target.} To improve target performance, the most natural
operational intervention is to collect additional data from the target
distribution. Data collection is expensive and resource-intensive, and we
must allocate it optimally to maximize robustness.
\wty{Although active learning has extensively explored strategies for acquiring limited labels to improve model performance~\cite{settles2009active}, few studies address how to gather supervised data efficiently from a target distribution that differs from the training environment. Existing work typically handles only covariate shifts~\cite{rai2010domain,su2020active} or label shifts~\cite{zhao2021active}, whereas our focus is on $Y|X$-shifts. Detailed literature comparisons appear in Appendix~\ref{app:relatedwork}.}
To highlight the need for future research in this space, we
use the interpretable region identified by Algorithm~\ref{algo} as shown in
\Cref{fig:case_subfig5} to simulate a concerted data collection effort. 

Since indiscriminately collecting data from the target distribution can be
resource-intensive, we concentrate sampling efforts on the subpopulation that
may suffer from $Y|X$-shifts and selectively gather data on them.  For five
base methods (LR, NN, RF, LGBM, and XGB),
we compare two data collection procedures, the one randomly sampling from the whole target distribution and the other from the
identified region suffering prominent $Y|X$-shifts with 250 points respectively.  We report the test
accuracies in \Cref{fig:case_subfig7}, and observe that incorporating data
from this region is more effective in improving performance under distribution shifts. 
While preliminary, our results demonstrate the potential robustness benefits of efficiently allocating resources toward concerted data collection. 
Additionally, our proposed data collection method can be integrated in a sequential manner by dynamically collecting a small amount of target data and refining the covariate region.
Future methodological research in this direction may be fruitful, e.g., leveraging active learning algorithms~\citep{settles2009active, active1, active3}.

\paragraph{Adding more relevant features.} We also illustrate the potential
benefits of generating qualitative insights on
the distribution shift at hand. Our analysis in \Cref{fig:case_subfig5}
suggests educated individuals in financial, educational, and legal professions
tend to experience large $Y|X$-shifts from CA to PR.  These roles typically
need communication skills, and language barriers could potentially affect
their incomes.  In California (CA), English is the primary language, while in
Puerto Rico (PR), despite both English and Spanish being recognized as
official languages, Spanish is predominantly spoken. Consequently, for a model
trained on CA data and tested on PR data, incorporating a new feature that
denotes English language proficiency (hereafter denoted ``ENG'') might prove
beneficial in improving generalization performances. However, this feature is
not included in the ACS Income dataset \citep{ding2022retiring}.

To address this, we went back to the Census Bureau's American Community Survey
database to include the ENG feature in the set of covariates.  In
\Cref{fig:case_subfig2}, we observe that the inclusion of this feature
substantially reduces the degradation due to $Y|X$-shifts, verifying that the
originally missing ENG feature may be one cause of $Y|X$-shifts.
\Cref{fig:case_subfig3} contrasts the performances of 28 methods (each with
200 hyperparameter configurations) with original features with those that
additionally use the ENG feature. The new feature significantly improves
target performances across all algorithms; roughly speaking, we posit that we
have identified a variable $C$ such that $Y|X, C$ remains similar across CA
and PR.  However, when we extend this comparison to the source-target pair (CA
$\rightarrow$ SD), we observe no significant improvement
(\Cref{fig:case_subfig4}). This indicates that the selection of new features
should be undertaken judiciously depending on the target distributions of
interest. A feature that proves to be effective in one target distribution might not
yield similar results in another.

\section{Discussion}

\wty{We acknowledge several limitations of our study. First, our empirical findings are necessarily bounded by the specific datasets and application domains we examine, though we demonstrate consistency across diverse prediction tasks spanning socioeconomic, transportation, and safety-critical applications. Second, while our distribution shifts represent a different class from those typically studied in the DRO literature, they necessarily occupy a position on the spectrum between controlled feature perturbations and entirely naturalistic deployment scenarios. Finally, our data-driven modeling approach requires some target domain data to be effective, limiting applicability in scenarios where target distributions are completely unknown.}

Our study also leaves many open directions. Our case studies are necessarily preliminary, and the feature-based modeling we explore is just one approach to setting the ambiguity set. Other approaches consider the relationship between continuous and category-feature effects to improve DRO model performance \citep{belbasi2023s} and classification trees robust to general distribution shift \citep{justin2023learning}. Our benchmark only includes tabular datasets from particular domains and expanding the scope of the datasets (e.g., medicine or those involving feature embeddings) may reveal different types of distribution shifts. On the algorithmic interventions side, our design of ambiguity sets based on a subset of features is still preliminary and more refined ambiguity sets incorporating domain knowledge are expected to be developed. On the data-based intervention side, our region-identification algorithm requires some target data to identify regions with large $Y|X$-shifts and cannot be used where the target distribution is completely unknown. 

\wty{Despite these limitations, we believe this work provides a meaningful foundation for empirically grounded approaches to distributionally robust learning. By highlighting the gap between theoretical assumptions and empirical realities, we hope to inspire the development of more practical and effective robust learning methods that acknowledge the complexity of distribution shifts.}

\bibliography{main}

\begin{thebibliography}{112}
\providecommand{\natexlab}[1]{#1}
\providecommand{\url}[1]{\texttt{#1}}
\expandafter\ifx\csname urlstyle\endcsname\relax
  \providecommand{\doi}[1]{doi: #1}\else
  \providecommand{\doi}{doi: \begingroup \urlstyle{rm}\Url}\fi

\bibitem[ass()]{assist}
Assistments.
\newblock \url{https://www.assistments.org/}.
\newblock Accessed: July 2025.

\bibitem[col()]{college}
College scorecard.
\newblock \url{https://collegescorecard.ed.gov/}.
\newblock Accessed: July 2025.

\bibitem[dia()]{diabetes}
Diabetes.
\newblock \url{https://www.cdc.gov/brfss/index.html}.
\newblock Accessed: July 2025.

\bibitem[tax({\natexlab{a}})]{taxidata}
{US} taxi dataset.
\newblock
  \url{https://www.kaggle.com/competitions/nyc-taxi-trip-duration/data},
  {\natexlab{a}}.
\newblock Accessed: June 2024.

\bibitem[tax({\natexlab{b}})]{taxidata1}
Other taxi dataset.
\newblock
  \url{https://www.kaggle.com/datasets/mnavas/taxi-routes-for-mexico-city-and-quito},
  {\natexlab{b}}.
\newblock Accessed: June 2024.

\bibitem[Adragna et~al.(2020)Adragna, Creager, Madras, and Zemel]{robert2020}
R.~Adragna, E.~Creager, D.~Madras, and R.~Zemel.
\newblock Fairness and robustness in invariant learning: A case study in
  toxicity classification.
\newblock \emph{arXiv preprint arXiv:2011.06485}, 2020.

\bibitem[Agarwal et~al.(2018)Agarwal, Beygelzimer, Dud{\'\i}k, Langford, and
  Wallach]{agarwal2018reductions}
A.~Agarwal, A.~Beygelzimer, M.~Dud{\'\i}k, J.~Langford, and H.~Wallach.
\newblock A reductions approach to fair classification.
\newblock In \emph{International Conference on Machine Learning}, pages 60--69.
  PMLR, 2018.

\bibitem[Angrist and Pischke(2009)]{angrist2009mostly}
J.~D. Angrist and J.-S. Pischke.
\newblock \emph{Mostly harmless econometrics: An empiricist's companion}.
\newblock Princeton university press, 2009.

\bibitem[Baehrens et~al.(2010)Baehrens, Schroeter, Harmeling, Kawanabe, Hansen,
  and M{\"u}ller]{baehrens2010explain}
D.~Baehrens, T.~Schroeter, S.~Harmeling, M.~Kawanabe, K.~Hansen, and K.-R.
  M{\"u}ller.
\newblock How to explain individual classification decisions.
\newblock \emph{Journal of Machine Learning Research}, 11:\penalty0 1803--1831,
  2010.

\bibitem[Bastani(2021)]{bastani2021predicting}
H.~Bastani.
\newblock Predicting with proxies: Transfer learning in high dimension.
\newblock \emph{Management Science}, 67\penalty0 (5):\penalty0 2964--2984,
  2021.

\bibitem[Bayraksan and Love(2015)]{bayraksan2015data}
G.~Bayraksan and D.~K. Love.
\newblock Data-driven stochastic programming using phi-divergences.
\newblock In \emph{Tutorials in Operations Research}, pages 1--19. INFORMS,
  2015.

\bibitem[Belbasi et~al.(2026)Belbasi, Selvi, and Wiesemann]{belbasi2023s}
R.~Belbasi, A.~Selvi, and W.~Wiesemann.
\newblock It’s all in the mix: Wasserstein classification and regression with
  mixed features.
\newblock \emph{Manufacturing \& Service Operations Management}, 2026.

\bibitem[Bellamy et~al.(2019)Bellamy, Dey, Hind, Hoffman, Houde, Kannan, Lohia,
  Martino, Mehta, Mojsilovic, Nagar, Ramamurthy, Richards, Saha, Sattigeri,
  Singh, Varshney, and Zhang]{aif360}
R.~K.~E. Bellamy, K.~Dey, M.~Hind, S.~C. Hoffman, S.~Houde, K.~Kannan,
  P.~Lohia, J.~Martino, S.~Mehta, A.~Mojsilovic, S.~Nagar, K.~N. Ramamurthy,
  J.~T. Richards, D.~Saha, P.~Sattigeri, M.~Singh, K.~R. Varshney, and
  Y.~Zhang.
\newblock {AI} fairness 360: An extensible toolkit for detecting and mitigating
  algorithmic bias.
\newblock \emph{{IBM} J. Res. Dev.}, 63\penalty0 (4/5):\penalty0 4:1--4:15,
  2019.

\bibitem[Ben-Tal et~al.(2013)Ben-Tal, Den~Hertog, De~Waegenaere, Melenberg, and
  Rennen]{ben2013robust}
A.~Ben-Tal, D.~Den~Hertog, A.~De~Waegenaere, B.~Melenberg, and G.~Rennen.
\newblock Robust solutions of optimization problems affected by uncertain
  probabilities.
\newblock \emph{Management Science}, 59\penalty0 (2):\penalty0 341--357, 2013.

\bibitem[Bennouna and Van~Parys(2022)]{bennouna2022holistic}
A.~Bennouna and B.~Van~Parys.
\newblock Holistic robust data-driven decisions.
\newblock \emph{arXiv preprint arXiv:2207.09560}, 2022.

\bibitem[Bennouna et~al.(2023)Bennouna, Lucas, and
  Van~Parys]{bennouna2023certified}
A.~Bennouna, R.~Lucas, and B.~Van~Parys.
\newblock Certified robust neural networks: Generalization and corruption
  resistance.
\newblock In \emph{International Conference on Machine Learning}, pages
  2092--2112. PMLR, 2023.

\bibitem[Bird et~al.(2020)Bird, Dud{\'\i}k, Edgar, Horn, Lutz, Milan, Sameki,
  Wallach, and Walker]{bird2020fairlearn}
S.~Bird, M.~Dud{\'\i}k, R.~Edgar, B.~Horn, R.~Lutz, V.~Milan, M.~Sameki,
  H.~Wallach, and K.~Walker.
\newblock Fairlearn: A toolkit for assessing and improving fairness in ai.
\newblock \emph{Microsoft, Tech. Rep.}, 2020.

\bibitem[Bischl et~al.(2021)Bischl, Casalicchio, Feurer, Gijsbers, Hutter,
  Lang, Mantovani, van Rijn, and Vanschoren]{bischl2017openml}
B.~Bischl, G.~Casalicchio, M.~Feurer, P.~Gijsbers, F.~Hutter, M.~Lang, R.~G.
  Mantovani, J.~N. van Rijn, and J.~Vanschoren.
\newblock Openml benchmarking suites.
\newblock In \emph{Neural Information Processing Systems Datasets and
  Benchmarks Track}, volume~35, 2021.

\bibitem[Blanchet et~al.(2019{\natexlab{a}})Blanchet, Kang, and
  Murthy]{blanchet2019robust}
J.~Blanchet, Y.~Kang, and K.~Murthy.
\newblock Robust {W}asserstein profile inference and applications to machine
  learning.
\newblock \emph{Journal of Applied Probability}, 56\penalty0 (3):\penalty0
  830--857, 2019{\natexlab{a}}.

\bibitem[Blanchet et~al.(2019{\natexlab{b}})Blanchet, Kang, Murthy, and
  Zhang]{blanchet2019data}
J.~Blanchet, Y.~Kang, K.~Murthy, and F.~Zhang.
\newblock Data-driven optimal transport cost selection for distributionally
  robust optimization.
\newblock In \emph{Winter Simulation Conference}, pages 3740--3751. IEEE,
  2019{\natexlab{b}}.

\bibitem[Blanchet et~al.(2021)Blanchet, Murthy, and
  Nguyen]{blanchet2021statistical}
J.~Blanchet, K.~Murthy, and V.~A. Nguyen.
\newblock Statistical analysis of wasserstein distributionally robust
  estimators.
\newblock In \emph{Tutorials in Operations Research: Emerging optimization
  methods and modeling techniques with applications}, pages 227--254. INFORMS,
  2021.

\bibitem[Blanchet et~al.(2023)Blanchet, Kuhn, Li, and
  Taskesen]{blanchet2023unifying}
J.~Blanchet, D.~Kuhn, J.~Li, and B.~Taskesen.
\newblock Unifying distributionally robust optimization via optimal transport
  theory.
\newblock \emph{arXiv preprint arXiv:2308.05414}, 2023.

\bibitem[Blum and Stangl(2020)]{blum2019recovering}
A.~Blum and K.~Stangl.
\newblock Recovering from biased data: Can fairness constraints improve
  accuracy?
\newblock In \emph{Symposium on Foundations of Responsible Computing (FORC)},
  volume~1, 2020.

\bibitem[Breiman(2001)]{breiman2001random}
L.~Breiman.
\newblock Random forests.
\newblock \emph{Machine learning}, 45:\penalty0 5--32, 2001.

\bibitem[Budhathoki et~al.(2021)Budhathoki, Janzing, Bloebaum, and
  Ng]{budhathoki2021why}
K.~Budhathoki, D.~Janzing, P.~Bloebaum, and H.~Ng.
\newblock Why did the distribution change?
\newblock In \emph{International Conference on Artificial Intelligence and
  Statistics}, pages 1666--1674. {PMLR}, 2021.

\bibitem[Cai et~al.(2026)Cai, Namkoong, and Yadlowsky]{namkoong2023diagnosing}
T.~Cai, H.~Namkoong, and S.~Yadlowsky.
\newblock Diagnosing model performance under distribution shift.
\newblock \emph{Operations Research}, 74\penalty0 (2):\penalty0 898--916, 2026.

\bibitem[Chattopadhyay et~al.(2013)Chattopadhyay, Fan, Davidson, Panchanathan,
  and Ye]{chattopadhyay2013joint}
R.~Chattopadhyay, W.~Fan, I.~Davidson, S.~Panchanathan, and J.~Ye.
\newblock Joint transfer and batch-mode active learning.
\newblock In \emph{International Conference on Machine Learning}, pages
  253--261. PMLR, 2013.

\bibitem[Chen and Guestrin(2016)]{chen2016xgboost}
T.~Chen and C.~Guestrin.
\newblock {{XGBoost}}: {{A Scalable Tree Boosting System}}.
\newblock In \emph{ACM SIGKDD International Conference on Knowledge Discovery},
  pages 785--794. {ACM}, 2016.

\bibitem[Cohn et~al.(1996)Cohn, Ghahramani, and Jordan]{cohn1996active}
D.~A. Cohn, Z.~Ghahramani, and M.~I. Jordan.
\newblock Active learning with statistical models.
\newblock \emph{Journal of Artificial Intelligence Research}, 4:\penalty0
  129--145, 1996.

\bibitem[Dastile et~al.(2020)Dastile, Celik, and
  Potsane]{dastile2020statistical}
X.~Dastile, T.~Celik, and M.~Potsane.
\newblock Statistical and machine learning models in credit scoring: A
  systematic literature survey.
\newblock \emph{Applied Soft Computing}, 91:\penalty0 106263, 2020.

\bibitem[Ding et~al.(2021)Ding, Hardt, Miller, and Schmidt]{ding2022retiring}
F.~Ding, M.~Hardt, J.~Miller, and L.~Schmidt.
\newblock Retiring adult: New datasets for fair machine learning.
\newblock In \emph{Advances in Neural Information Processing Systems},
  volume~34, pages 6478--6490, 2021.

\bibitem[Dua and Graff(2017)]{Dua:2019}
D.~Dua and C.~Graff.
\newblock {UCI} machine learning repository, 2017.
\newblock URL \url{http://archive.ics.uci.edu/ml}.

\bibitem[Duchi and Namkoong(2019)]{duchi2019variance}
J.~Duchi and H.~Namkoong.
\newblock Variance-based regularization with convex objectives.
\newblock \emph{Journal of Machine Learning Research}, 20\penalty0
  (1):\penalty0 2450--2504, 2019.

\bibitem[Duchi et~al.(2023)Duchi, Hashimoto, and
  Namkoong]{duchi2023distributionally}
J.~Duchi, T.~Hashimoto, and H.~Namkoong.
\newblock Distributionally robust losses for latent covariate mixtures.
\newblock \emph{Operations Research}, 71\penalty0 (2):\penalty0 649--664, 2023.

\bibitem[Duchi and Namkoong(2021)]{duchi2021learning}
J.~C. Duchi and H.~Namkoong.
\newblock Learning models with uniform performance via distributionally robust
  optimization.
\newblock \emph{Annals of Statistics}, 49\penalty0 (3):\penalty0 1378--1406,
  2021.

\bibitem[Duchi et~al.(2021)Duchi, Glynn, and Namkoong]{duchi2021statistics}
J.~C. Duchi, P.~W. Glynn, and H.~Namkoong.
\newblock Statistics of robust optimization: A generalized empirical likelihood
  approach.
\newblock \emph{Mathematics of Operations Research}, 46\penalty0 (3):\penalty0
  946--969, 2021.

\bibitem[Esfahani and Kuhn(2018)]{esfahani2018data}
P.~M. Esfahani and D.~Kuhn.
\newblock Data-driven distributionally robust optimization using the
  {W}asserstein metric: Performance guarantees and tractable reformulations.
\newblock \emph{Mathematical Programming}, 171\penalty0 (1):\penalty0 115--166,
  2018.

\bibitem[Fatima and Pasha(2017)]{fatima2017survey}
M.~Fatima and M.~Pasha.
\newblock Survey of machine learning algorithms for disease diagnostic.
\newblock \emph{Journal of Intelligent Learning Systems and Applications},
  9\penalty0 (01):\penalty0 1--16, 2017.

\bibitem[Gama et~al.(2014)Gama, {\v{Z}}liobait{\.e}, Bifet, Pechenizkiy, and
  Bouchachia]{gama2014survey}
J.~Gama, I.~{\v{Z}}liobait{\.e}, A.~Bifet, M.~Pechenizkiy, and A.~Bouchachia.
\newblock A survey on concept drift adaptation.
\newblock \emph{ACM computing surveys (CSUR)}, 46\penalty0 (4):\penalty0 1--37,
  2014.

\bibitem[Gao et~al.(2024)Gao, Chen, and Kleywegt]{gao2022wasserstein}
R.~Gao, X.~Chen, and A.~J. Kleywegt.
\newblock Wasserstein distributionally robust optimization and variation
  regularization.
\newblock \emph{Operations Research}, 72\penalty0 (3):\penalty0 1177--1191,
  2024.

\bibitem[Gardner et~al.(2022)Gardner, Popovic, and
  Schmidt]{gardner2022subgroup}
J.~Gardner, Z.~Popovic, and L.~Schmidt.
\newblock Subgroup robustness grows on trees: An empirical baseline
  investigation.
\newblock In \emph{Advances in Neural Information Processing Systems},
  volume~35, pages 9939--9954, 2022.

\bibitem[Ghorbani et~al.(2022)Ghorbani, Zou, and Esteva]{ghorbani2022data}
A.~Ghorbani, J.~Zou, and A.~Esteva.
\newblock Data shapley valuation for efficient batch active learning.
\newblock In \emph{2022 56th Asilomar Conference on Signals, Systems, and
  Computers}, pages 1456--1462. IEEE, 2022.

\bibitem[Gotoh et~al.(2021)Gotoh, Kim, and Lim]{gotoh2021calibration}
J.-y. Gotoh, M.~J. Kim, and A.~E. Lim.
\newblock Calibration of distributionally robust empirical optimization models.
\newblock \emph{Operations Research}, 69\penalty0 (5):\penalty0 1630--1650,
  2021.

\bibitem[Grinsztajn et~al.(2022)Grinsztajn, Oyallon, and
  Varoquaux]{grinsztajn2022why}
L.~Grinsztajn, E.~Oyallon, and G.~Varoquaux.
\newblock Why do tree-based models still outperform deep learning on typical
  tabular data?
\newblock In \emph{Advances in Neural Information Processing Systems},
  volume~35, pages 507--520, 2022.

\bibitem[Gulrajani and Lopez-Paz(2021)]{gulrajanisearch}
I.~Gulrajani and D.~Lopez-Paz.
\newblock In search of lost domain generalization.
\newblock In \emph{International Conference on Learning Representations}, 2021.

\bibitem[Guo et~al.(2017)Guo, Tang, Ye, Li, and He]{guo2017deepfm}
H.~Guo, R.~Tang, Y.~Ye, Z.~Li, and X.~He.
\newblock Deepfm: a factorization-machine based neural network for ctr
  prediction.
\newblock \emph{arXiv preprint arXiv:1703.04247}, 2017.

\bibitem[Hardt et~al.(2016)Hardt, Price, and Srebro]{hardt2016equality}
M.~Hardt, E.~Price, and N.~Srebro.
\newblock Equality of opportunity in supervised learning.
\newblock In \emph{Advances in Neural Information Processing Systems},
  volume~29, pages 3315--3323, 2016.

\bibitem[Hashimoto et~al.(2018)Hashimoto, Srivastava, Namkoong, and
  Liang]{hashimoto2018fairness}
T.~Hashimoto, M.~Srivastava, H.~Namkoong, and P.~Liang.
\newblock Fairness without demographics in repeated loss minimization.
\newblock In \emph{International Conference on Machine Learning}, pages
  1929--1938. PMLR, 2018.

\bibitem[Hendrycks et~al.(2021)Hendrycks, Basart, Mu, Kadavath, Wang, Dorundo,
  Desai, Zhu, Parajuli, Guo, et~al.]{hendrycks2021many}
D.~Hendrycks, S.~Basart, N.~Mu, S.~Kadavath, F.~Wang, E.~Dorundo, R.~Desai,
  T.~Zhu, S.~Parajuli, M.~Guo, et~al.
\newblock The many faces of robustness: A critical analysis of
  out-of-distribution generalization.
\newblock In \emph{Proceedings of the IEEE/CVF International Conference on
  Computer Vision}, pages 8340--8349, 2021.

\bibitem[Hospedales et~al.(2021)Hospedales, Antoniou, Micaelli, and
  Storkey]{hospedales2021meta}
T.~Hospedales, A.~Antoniou, P.~Micaelli, and A.~Storkey.
\newblock Meta-learning in neural networks: A survey.
\newblock \emph{IEEE transactions on pattern analysis and machine
  intelligence}, 44\penalty0 (9):\penalty0 5149--5169, 2021.

\bibitem[Hu et~al.(2018)Hu, Niu, Sato, and Sugiyama]{hu2018does}
W.~Hu, G.~Niu, I.~Sato, and M.~Sugiyama.
\newblock Does distributionally robust supervised learning give robust
  classifiers?
\newblock In \emph{International Conference on Machine Learning}, pages
  2029--2037. PMLR, 2018.

\bibitem[Hu and Hong(2013)]{hu2013kullback}
Z.~Hu and L.~J. Hong.
\newblock Kullback-leibler divergence constrained distributionally robust
  optimization.
\newblock \emph{Available at Optimization Online}, 2013.

\bibitem[Idrissi et~al.(2022)Idrissi, Arjovsky, Pezeshki, and
  {Lopez-Paz}]{idrissi2022simple}
B.~Y. Idrissi, M.~Arjovsky, M.~Pezeshki, and D.~{Lopez-Paz}.
\newblock Simple data balancing achieves competitive worst-group-accuracy.
\newblock In \emph{{{Conference}} on {{Causal Learning}} and {{Reasoning}}},
  pages 336--351. {PMLR}, 2022.

\bibitem[Iyengar et~al.(2022)Iyengar, Lam, and Wang]{iyengar2022hedging}
G.~Iyengar, H.~Lam, and T.~Wang.
\newblock Hedging complexity in generalization via a parametric
  distributionally robust optimization framework.
\newblock \emph{arXiv preprint arXiv:2212.01518}, 2022.

\bibitem[Jia et~al.(2019)Jia, Dao, Wang, Hubis, Hynes, G{\"u}rel, Li, Zhang,
  Song, and Spanos]{jia2019towards}
R.~Jia, D.~Dao, B.~Wang, F.~A. Hubis, N.~Hynes, N.~M. G{\"u}rel, B.~Li,
  C.~Zhang, D.~Song, and C.~J. Spanos.
\newblock Towards efficient data valuation based on the shapley value.
\newblock In \emph{International Conference on Artificial Intelligence and
  Statistics}, pages 1167--1176. PMLR, 2019.

\bibitem[Jiang and Guan(2018)]{jiang2018risk}
R.~Jiang and Y.~Guan.
\newblock Risk-averse two-stage stochastic program with distributional
  ambiguity.
\newblock \emph{Operations Research}, 66\penalty0 (5):\penalty0 1390--1405,
  2018.

\bibitem[Johnson et~al.(2016)Johnson, Pollard, Shen, Lehman, Feng, Ghassemi,
  Moody, Szolovits, Anthony~Celi, and Mark]{johnson2016mimic}
A.~E. Johnson, T.~J. Pollard, L.~Shen, L.-w.~H. Lehman, M.~Feng, M.~Ghassemi,
  B.~Moody, P.~Szolovits, L.~Anthony~Celi, and R.~G. Mark.
\newblock Mimic-iii, a freely accessible critical care database.
\newblock \emph{Scientific data}, 3\penalty0 (1):\penalty0 1--9, 2016.

\bibitem[Justin et~al.(2023)Justin, Aghaei, G{\'o}mez, and
  Vayanos]{justin2023learning}
N.~Justin, S.~Aghaei, A.~G{\'o}mez, and P.~Vayanos.
\newblock Learning optimal classification trees robust to distribution shifts.
\newblock \emph{arXiv preprint arXiv:2310.17772}, 2023.

\bibitem[Ke et~al.(2017)Ke, Meng, Finley, Wang, Chen, Ma, Ye, and
  Liu]{ke2017lightgbm}
G.~Ke, Q.~Meng, T.~Finley, T.~Wang, W.~Chen, W.~Ma, Q.~Ye, and T.-Y. Liu.
\newblock Lightgbm: A highly efficient gradient boosting decision tree.
\newblock In \emph{Advances in Neural Information Processing Systems},
  volume~30, pages 3149--3157, 2017.

\bibitem[Knight(1921)]{knight1921risk}
F.~H. Knight.
\newblock \emph{Risk, uncertainty and profit}, volume~31.
\newblock Houghton Mifflin, 1921.

\bibitem[Koh et~al.(2021)Koh, Sagawa, Marklund, Xie, Zhang, Balsubramani, Hu,
  Yasunaga, Phillips, Gao, Lee, David, Stavness, Guo, Earnshaw, Haque, Beery,
  Leskovec, Kundaje, Pierson, Levine, Finn, and Liang]{koh2021wilds}
P.~W. Koh, S.~Sagawa, H.~Marklund, S.~M. Xie, M.~Zhang, A.~Balsubramani, W.~Hu,
  M.~Yasunaga, R.~L. Phillips, I.~Gao, T.~Lee, E.~David, I.~Stavness, W.~Guo,
  B.~Earnshaw, I.~Haque, S.~M. Beery, J.~Leskovec, A.~Kundaje, E.~Pierson,
  S.~Levine, C.~Finn, and P.~Liang.
\newblock {{WILDS}}: {{A Benchmark}} of in-the-{{Wild Distribution Shifts}}.
\newblock In \emph{International Conference on Machine Learning}, pages
  5637--5664. {PMLR}, 2021.

\bibitem[Kuhn et~al.(2025)Kuhn, Shafiee, and
  Wiesemann]{kuhn2025distributionally}
D.~Kuhn, S.~Shafiee, and W.~Wiesemann.
\newblock Distributionally robust optimization.
\newblock \emph{Acta Numerica}, 34:\penalty0 579--804, 2025.

\bibitem[Kulinski and Inouye(2023)]{pmlr-v202-kulinski23a}
S.~Kulinski and D.~I. Inouye.
\newblock Towards explaining distribution shifts.
\newblock In \emph{International Conference on Machine Learning}, volume 202,
  pages 17931--17952. PMLR, 2023.

\bibitem[Lee and Raginsky(2018)]{lee2018minimax}
J.~Lee and M.~Raginsky.
\newblock Minimax statistical learning with {W}asserstein distances.
\newblock In \emph{Advances in Neural Information Processing Systems},
  volume~31, 2018.

\bibitem[Levy et~al.(2020)Levy, Carmon, Duchi, and Sidford]{levy2020large}
D.~Levy, Y.~Carmon, J.~C. Duchi, and A.~Sidford.
\newblock Large-scale methods for distributionally robust optimization.
\newblock In \emph{Advances in Neural Information Processing Systems},
  volume~33, pages 8847--8860, 2020.

\bibitem[Liaw et~al.(2018)Liaw, Liang, Nishihara, Moritz, Gonzalez, and
  Stoica]{liaw2018tune}
R.~Liaw, E.~Liang, R.~Nishihara, P.~Moritz, J.~E. Gonzalez, and I.~Stoica.
\newblock Tune: A research platform for distributed model selection and
  training.
\newblock \emph{arXiv preprint arXiv:1807.05118}, 2018.

\bibitem[Liu et~al.(2023)Liu, Wang, Cui, and Namkoong]{liu2024need}
J.~Liu, T.~Wang, P.~Cui, and H.~Namkoong.
\newblock On the need for a language describing distribution shifts:
  Illustrations on tabular datasets.
\newblock In \emph{Advances in Neural Information Processing Systems}, pages
  51371--51408, 2023.

\bibitem[Liu et~al.(2021)Liu, Ding, Zhong, Li, Dai, and He]{active3}
Z.~Liu, H.~Ding, H.~Zhong, W.~Li, J.~Dai, and C.~He.
\newblock Influence selection for active learning.
\newblock In \emph{International Conference on Computer Vision}, pages
  9254--9263. {IEEE}, 2021.

\bibitem[Long et~al.(2023)Long, Sim, and Zhou]{long2023robust}
D.~Z. Long, M.~Sim, and M.~Zhou.
\newblock Robust satisficing.
\newblock \emph{Operations Research}, 71\penalty0 (1):\penalty0 61--82, 2023.

\bibitem[Lu et~al.(2018)Lu, Liu, Dong, Gu, Gama, and Zhang]{lu2018learning}
J.~Lu, A.~Liu, F.~Dong, F.~Gu, J.~Gama, and G.~Zhang.
\newblock Learning under concept drift: A review.
\newblock \emph{IEEE transactions on knowledge and data engineering},
  31\penalty0 (12):\penalty0 2346--2363, 2018.

\bibitem[Lundberg and Lee(2017)]{lundberg2017unified}
S.~M. Lundberg and S.-I. Lee.
\newblock A unified approach to interpreting model predictions.
\newblock In \emph{Proceedings of the 31st International Conference on Neural
  Information Processing Systems}, pages 4768--4777, 2017.

\bibitem[McMahan et~al.(2013)McMahan, Holt, Sculley, Young, Ebner, Grady, Nie,
  Phillips, Davydov, Golovin, et~al.]{mcmahan2013ad}
H.~B. McMahan, G.~Holt, D.~Sculley, M.~Young, D.~Ebner, J.~Grady, L.~Nie,
  T.~Phillips, E.~Davydov, D.~Golovin, et~al.
\newblock Ad click prediction: a view from the trenches.
\newblock In \emph{Proceedings of the 19th ACM SIGKDD international conference
  on Knowledge discovery and data mining}, pages 1222--1230, 2013.

\bibitem[Miller et~al.(2020)Miller, Krauth, Recht, and
  Schmidt]{miller2020effect}
J.~Miller, K.~Krauth, B.~Recht, and L.~Schmidt.
\newblock The effect of natural distribution shift on question answering
  models.
\newblock In \emph{International Conference on Machine Learning}, pages
  6905--6916. PMLR, 2020.

\bibitem[Miller et~al.(2021)Miller, Taori, Raghunathan, Sagawa, Koh, Shankar,
  Liang, Carmon, and Schmidt]{miller2021accuracy}
J.~P. Miller, R.~Taori, A.~Raghunathan, S.~Sagawa, P.~W. Koh, V.~Shankar,
  P.~Liang, Y.~Carmon, and L.~Schmidt.
\newblock Accuracy on the {{Line}}: On the {{Strong Correlation Between
  Out-of-Distribution}} and {{In-Distribution Generalization}}.
\newblock In \emph{International Conference on Machine Learning}, pages
  7721--7735. {PMLR}, 2021.

\bibitem[Moosavi et~al.(2019{\natexlab{a}})Moosavi, Samavatian, Parthasarathy,
  and Ramnath]{accident1}
S.~Moosavi, M.~H. Samavatian, S.~Parthasarathy, and R.~Ramnath.
\newblock A countrywide traffic accident dataset.
\newblock \emph{arXiv preprint arXiv:1906.05409}, 2019{\natexlab{a}}.

\bibitem[Moosavi et~al.(2019{\natexlab{b}})Moosavi, Samavatian, Parthasarathy,
  Teodorescu, and Ramnath]{accident2}
S.~Moosavi, M.~H. Samavatian, S.~Parthasarathy, R.~Teodorescu, and R.~Ramnath.
\newblock Accident risk prediction based on heterogeneous sparse data: New
  dataset and insights.
\newblock In \emph{International Conference on Advances in Geographic
  Information Systems}, pages 33--42. {ACM}, 2019{\natexlab{b}}.

\bibitem[Mu and Gilmer(2019)]{mu2019mnist}
N.~Mu and J.~Gilmer.
\newblock Mnist-c: A robustness benchmark for computer vision.
\newblock \emph{arXiv preprint arXiv:1906.02337}, 2019.

\bibitem[Natekin and Knoll(2013)]{natekin2013gradient}
A.~Natekin and A.~Knoll.
\newblock Gradient boosting machines, a tutorial.
\newblock \emph{Frontiers in neurorobotics}, 7:\penalty0 21, 2013.

\bibitem[Paszke et~al.(2019)Paszke, Gross, Massa, Lerer, Bradbury, Chanan,
  Killeen, Lin, Gimelshein, Antiga, Desmaison, K{\"{o}}pf, Yang, DeVito,
  Raison, Tejani, Chilamkurthy, Steiner, Fang, Bai, and Chintala]{pytorch}
A.~Paszke, S.~Gross, F.~Massa, A.~Lerer, J.~Bradbury, G.~Chanan, T.~Killeen,
  Z.~Lin, N.~Gimelshein, L.~Antiga, A.~Desmaison, A.~K{\"{o}}pf, E.~Z. Yang,
  Z.~DeVito, M.~Raison, A.~Tejani, S.~Chilamkurthy, B.~Steiner, L.~Fang,
  J.~Bai, and S.~Chintala.
\newblock Pytorch: An imperative style, high-performance deep learning library.
\newblock In \emph{Advances in Neural Information Processing Systems}, pages
  8024--8035, 2019.

\bibitem[Pedregosa et~al.(2011)Pedregosa, Varoquaux, Gramfort, Michel, Thirion,
  Grisel, Blondel, Prettenhofer, Weiss, Dubourg, et~al.]{pedregosa2011scikit}
F.~Pedregosa, G.~Varoquaux, A.~Gramfort, V.~Michel, B.~Thirion, O.~Grisel,
  M.~Blondel, P.~Prettenhofer, R.~Weiss, V.~Dubourg, et~al.
\newblock Scikit-learn: Machine learning in python.
\newblock \emph{Journal of Machine Learning Research}, 12:\penalty0 2825--2830,
  2011.

\bibitem[Quinonero-Candela et~al.(2008)Quinonero-Candela, Sugiyama,
  Schwaighofer, and Lawrence]{quinonero2008dataset}
J.~Quinonero-Candela, M.~Sugiyama, A.~Schwaighofer, and N.~D. Lawrence.
\newblock \emph{Dataset shift in machine learning}.
\newblock MIT Press, 2008.

\bibitem[Rahimian and Mehrotra(2019)]{rahimian2019distributionally}
H.~Rahimian and S.~Mehrotra.
\newblock Distributionally robust optimization: A review.
\newblock \emph{arXiv preprint arXiv:1908.05659}, 2019.

\bibitem[Rai et~al.(2010)Rai, Saha, Daum{\'e}~III, and
  Venkatasubramanian]{rai2010domain}
P.~Rai, A.~Saha, H.~Daum{\'e}~III, and S.~Venkatasubramanian.
\newblock Domain adaptation meets active learning.
\newblock In \emph{Proceedings of the NAACL HLT 2010 workshop on active
  learning for natural language processing}, pages 27--32, 2010.

\bibitem[Recht et~al.(2019)Recht, Roelofs, Schmidt, and
  Shankar]{recht2019imagenet}
B.~Recht, R.~Roelofs, L.~Schmidt, and V.~Shankar.
\newblock Do imagenet classifiers generalize to imagenet?
\newblock In \emph{International Conference on Machine Learning}, pages
  5389--5400. PMLR, 2019.

\bibitem[Rockafellar et~al.(2000)Rockafellar, Uryasev,
  et~al.]{rockafellar2000optimization}
R.~T. Rockafellar, S.~Uryasev, et~al.
\newblock Optimization of conditional value-at-risk.
\newblock \emph{Journal of Risk}, 2:\penalty0 21--42, 2000.

\bibitem[Sagawa et~al.(2020)Sagawa, Koh, Hashimoto, and
  Liang]{sagawa2020distributionally}
S.~Sagawa, P.~W. Koh, T.~B. Hashimoto, and P.~Liang.
\newblock Distributionally robust neural networks for group shifts: On the
  importance of regularization for worst-case generalization.
\newblock In \emph{International Conference on Learning Representations}, 2020.

\bibitem[Sahoo et~al.(2022)Sahoo, Lei, and Wager]{sahoo2022learning}
R.~Sahoo, L.~Lei, and S.~Wager.
\newblock Learning from a biased sample.
\newblock \emph{arXiv preprint arXiv:2209.01754}, 2022.

\bibitem[Santurkar et~al.(2021)Santurkar, Tsipras, and
  Madry]{santurkar2021breeds}
S.~Santurkar, D.~Tsipras, and A.~Madry.
\newblock Breeds: Benchmarks for subpopulation shift.
\newblock In \emph{International Conference on Learning Representations}, 2021.

\bibitem[Schlimmer and Granger(1986)]{schlimmer1986incremental}
J.~C. Schlimmer and R.~H. Granger.
\newblock Incremental learning from noisy data.
\newblock \emph{Machine learning}, 1:\penalty0 317--354, 1986.

\bibitem[Settles(2009)]{settles2009active}
B.~Settles.
\newblock Active learning literature survey.
\newblock 2009.

\bibitem[Shafieezadeh-Abadeh et~al.(2015)Shafieezadeh-Abadeh, Esfahani, and
  Kuhn]{shafieezadeh2015distributionally}
S.~Shafieezadeh-Abadeh, P.~M. Esfahani, and D.~Kuhn.
\newblock Distributionally robust logistic regression.
\newblock In \emph{Advances in Neural Information Processing Systems},
  volume~29, pages 1576--1584, 2015.

\bibitem[Shafieezadeh-Abadeh et~al.(2019)Shafieezadeh-Abadeh, Kuhn, and
  Esfahani]{shafieezadeh2019regularization}
S.~Shafieezadeh-Abadeh, D.~Kuhn, and P.~M. Esfahani.
\newblock Regularization via mass transportation.
\newblock \emph{Journal of Machine Learning Research}, 20\penalty0
  (103):\penalty0 1--68, 2019.

\bibitem[Shafieezadeh-Abadeh et~al.(2023)Shafieezadeh-Abadeh, Aolaritei,
  D{\"o}rfler, and Kuhn]{shafieezadeh2023new}
S.~Shafieezadeh-Abadeh, L.~Aolaritei, F.~D{\"o}rfler, and D.~Kuhn.
\newblock New perspectives on regularization and computation in optimal
  transport-based distributionally robust optimization.
\newblock \emph{arXiv preprint arXiv:2303.03900}, 2023.

\bibitem[Shapiro et~al.(2023)Shapiro, Zhou, and Lin]{shapiro2023bayesian}
A.~Shapiro, E.~Zhou, and Y.~Lin.
\newblock Bayesian distributionally robust optimization.
\newblock \emph{SIAM Journal on Optimization}, 33\penalty0 (2):\penalty0
  1279--1304, 2023.

\bibitem[Shwartz-Ziv and Armon(2022)]{shwartz2022tabular}
R.~Shwartz-Ziv and A.~Armon.
\newblock Tabular data: Deep learning is not all you need.
\newblock \emph{Information Fusion}, 81:\penalty0 84--90, 2022.

\bibitem[Singh et~al.(2024)Singh, Xia, Subbaswamy, Gossmann, and
  Feng]{DBLP:journals/corr/abs-2402-14254}
H.~Singh, F.~Xia, A.~Subbaswamy, A.~Gossmann, and J.~Feng.
\newblock A hierarchical decomposition for explaining ml performance
  discrepancies.
\newblock In \emph{Advances in Neural Information Processing Systems},
  volume~38, pages 128516--128555, 2024.

\bibitem[Sinha et~al.(2018)Sinha, Namkoong, and Duchi]{sinhaND18}
A.~Sinha, H.~Namkoong, and J.~C. Duchi.
\newblock Certifying some distributional robustness with principled adversarial
  training.
\newblock In \emph{International Conference on Learning Representations}, 2018.

\bibitem[Su et~al.(2020)Su, Tsai, Sohn, Liu, Maji, and
  Chandraker]{su2020active}
J.-C. Su, Y.-H. Tsai, K.~Sohn, B.~Liu, S.~Maji, and M.~Chandraker.
\newblock Active adversarial domain adaptation.
\newblock In \emph{Proceedings of the IEEE/CVF Winter Conference on
  Applications of Computer Vision}, pages 739--748, 2020.

\bibitem[Taskesen et~al.(2020)Taskesen, Nguyen, Kuhn, and
  Blanchet]{taskesen2020distributionally}
B.~Taskesen, V.~A. Nguyen, D.~Kuhn, and J.~Blanchet.
\newblock A distributionally robust approach to fair classification.
\newblock \emph{arXiv preprint arXiv:2007.09530}, 2020.

\bibitem[Ulmer et~al.(2020)Ulmer, Meijerink, and Cin{\`a}]{ulmer2020trust}
D.~Ulmer, L.~Meijerink, and G.~Cin{\`a}.
\newblock Trust issues: Uncertainty estimation does not enable reliable ood
  detection on medical tabular data.
\newblock In \emph{Machine Learning for Health}, pages 341--354. PMLR, 2020.

\bibitem[Wang et~al.(2025)Wang, Gao, and Xie]{wang2021sinkhorn}
J.~Wang, R.~Gao, and Y.~Xie.
\newblock Sinkhorn distributionally robust optimization.
\newblock \emph{Operations Research}, 2025.

\bibitem[Xie et~al.(2022)Xie, Yuan, Li, Liu, and Cheng]{active1}
B.~Xie, L.~Yuan, S.~Li, C.~H. Liu, and X.~Cheng.
\newblock Towards fewer annotations: Active learning via region impurity and
  prediction uncertainty for domain adaptive semantic segmentation.
\newblock In \emph{Conference on Computer Vision and Pattern Recognition},
  pages 8058--8068. {IEEE}, 2022.

\bibitem[Yan et~al.(2018)Yan, Chaudhuri, and Javidi]{yan2018active}
S.~Yan, K.~Chaudhuri, and T.~Javidi.
\newblock Active learning with logged data.
\newblock In \emph{International Conference on Machine Learning}, pages
  5521--5530. PMLR, 2018.

\bibitem[Yang et~al.(2023)Yang, Zhang, Katabi, and Ghassemi]{subshift}
Y.~Yang, H.~Zhang, D.~Katabi, and M.~Ghassemi.
\newblock Change is hard: {A} closer look at subpopulation shift.
\newblock In \emph{International Conference on Machine Learning}, volume 202,
  pages 39584--39622. {PMLR}, 2023.

\bibitem[Yao et~al.(2022)Yao, Choi, Cao, Lee, Koh, and
  Finn]{DBLP:conf/nips/YaoCC0KF22}
H.~Yao, C.~Choi, B.~Cao, Y.~Lee, P.~W. Koh, and C.~Finn.
\newblock Wild-time: {A} benchmark of in-the-wild distribution shift over time.
\newblock In \emph{Advances in Neural Information Processing Systems},
  volume~35, pages 10309--10324, 2022.

\bibitem[Yu et~al.(2023)Yu, Cui, He, Shen, Lin, Xu, and Zhang]{yu2022}
H.~Yu, P.~Cui, Y.~He, Z.~Shen, Y.~Lin, R.~Xu, and X.~Zhang.
\newblock Stable learning via sparse variable independence.
\newblock In \emph{Proceedings of the AAAI Conference on Artificial
  Intelligence}, pages 10998--11006, 2023.

\bibitem[Zeng et~al.(2024)Zeng, Liu, Lam, and Namkoong]{zeng2024llm}
Y.~Zeng, J.~Liu, H.~Lam, and H.~Namkoong.
\newblock {LLM} embeddings improve test-time adaptation to tabular $ y| x
  $-shifts.
\newblock \emph{arXiv preprint arXiv:2410.07395}, 2024.

\bibitem[Zhai et~al.(2021)Zhai, Dan, Kolter, and Ravikumar]{zhai2021doro}
R.~Zhai, C.~Dan, Z.~Kolter, and P.~Ravikumar.
\newblock {DORO}: Distributional and outlier robust optimization.
\newblock In \emph{International Conference on Machine Learning}, pages
  12345--12355. PMLR, 2021.

\bibitem[Zhang et~al.(2023{\natexlab{a}})Zhang, Singh, Ghassemi, and
  Joshi]{zhang2023why}
H.~Zhang, H.~Singh, M.~Ghassemi, and S.~Joshi.
\newblock "{W}hy did the model fail?": Attributing model performance changes to
  distribution shifts.
\newblock In \emph{International Conference on Machine Learning}, volume 202,
  pages 41550--41578. PMLR, 2023{\natexlab{a}}.

\bibitem[Zhang et~al.(2023{\natexlab{b}})Zhang, He, Wang, Qi, Yu, Wang, Peng,
  Xu, Shen, Niu, et~al.]{zhang2023nico}
X.~Zhang, Y.~He, T.~Wang, J.~Qi, H.~Yu, Z.~Wang, J.~Peng, R.~Xu, Z.~Shen,
  Y.~Niu, et~al.
\newblock Nico challenge: Out-of-distribution generalization for image
  recognition challenges.
\newblock In \emph{Computer Vision--ECCV 2022 Workshops}, pages 433--450.
  Springer, 2023{\natexlab{b}}.

\bibitem[Zhao et~al.(2021)Zhao, Liu, Anandkumar, and Yue]{zhao2021active}
E.~Zhao, A.~Liu, A.~Anandkumar, and Y.~Yue.
\newblock Active learning under label shift.
\newblock In \emph{International Conference on Artificial Intelligence and
  Statistics}, pages 3412--3420. PMLR, 2021.

\bibitem[Zhu et~al.(2021)Zhu, Jitkrittum, Diehl, and
  Sch{\"o}lkopf]{zhu2021kernel}
J.-J. Zhu, W.~Jitkrittum, M.~Diehl, and B.~Sch{\"o}lkopf.
\newblock Kernel distributionally robust optimization: Generalized duality
  theorem and stochastic approximation.
\newblock In \emph{International Conference on Artificial Intelligence and
  Statistics}, pages 280--288. PMLR, 2021.

\end{thebibliography}
\bibliographystyle{abbrvnat}

\newpage
\appendix
\appendixpage

\startcontents[sections]
\printcontents[sections]{l}{1}{\setcounter{tocdepth}{2}}

{}
\section{Other Relevant Work}\label{app:relatedwork}

\wty{\paragraph{Model classes in DRO methods.} In \Cref{tab:loss_choice}, we summarize the model class of recent distance-based DRO methods that are considered in literature:}
\begin{table}[!htb]
    \centering
    \caption{Model classes of distance-based DRO methods that are explicitly considered in literature for classification ML models, where the models are exact reformulation unless specified by ``(Apx)''.}
    \label{tab:loss_choice}
    \small
    \begin{tabular}{l|cccccc|ccc}
        \toprule
         Reference &Ambiguity Set & Model class\\
         \midrule
         \cite{esfahani2018data} & 1-Wasserstein  & Linear SVM\\
         \cite{shafieezadeh2015distributionally}, \cite{blanchet2019robust} & 1(2)-Wasserstein & LR \\
         \cite{shafieezadeh2019regularization} & 1-Wasserstein & Kernel SVM, NN (Apx) \\
         \cite{sinhaND18} & $p$-Wasserstein & NN (Apx)\\
         \cite{shafieezadeh2023new} & General (OT) Wasserstein & Linear SVM, LR\\
         \hline
         \cite{ben2013robust}, \cite{bayraksan2015data} & (standard) $f$-divergence & Linear SVM, LR\\
         \cite{jiang2018risk} / \cite{rockafellar2000optimization} / \cite{hu2013kullback} / 
         \cite{duchi2019variance} & TV / CVaR / KL / $\chi^2$-divergence & Linear SVM, LR \\
         \cite{levy2020large} & CVaR, $\chi^2$-divergence & NN (Apx) \\
         \hline
         \cite{wang2021sinkhorn} & Sinkhorn distance & Linear SVM, LR, NN (Apx)\\
        \cite{bennouna2022holistic} & Holistic Robust & Linear SVM\\
        \cite{bennouna2023certified} & Holistic Robust & NN (Apx) \\
        \cite{blanchet2023unifying} & OT-Discrepancy & Linear SVM\\
        \bottomrule         
    \end{tabular}
\end{table}

\wty{\paragraph{Active Learning.} Active learning aims to improve model performance by acquiring a limited number of labels from the target distribution. See the survey \cite{settles2009active} for a detailed reference. The challenge here is to quantify the value of unlabeled data so that we can select better samples. Selection criteria of existing approaches include estimated variance \cite{cohn1996active}, influence on the model performance \cite{active3}, Shapley value \cite{ghorbani2022data,jia2019towards}, or distance metrics through the importance weighting \cite{chattopadhyay2013joint}. Active learning has also been studied under joint distribution shifts and covariate shifts~\cite{yan2018active,su2020active}. These works usually assume restricted shift structures between the source and target distributions where only $X$-shifts~\cite{rai2010domain,su2020active} or label shifts~\cite{zhao2021active} occur, which may not hold in tabular data when $Y|X$-shifts occur. In contrast, we assume access to labeled target data but impose no restrictions on the two distributions. Then we integrate an efficient algorithm that identifies the covariate region suffering the largest $Y|X$-shifts between two distributions.}


\wty{\paragraph{Difference with \cite{gulrajanisearch}.} Compared with the benchmark paper \citep{gulrajanisearch}, we highlight three major differences in our work:
\begin{enumerate}
    \item Datasets and Domains: We focus on tabular datasets instead of the image datasets in \cite{gulrajanisearch}. The distribution shift patterns in the tabular datasets are more complicated due to $Y|X$-shifts from \twy{unobserved variables}. The existence of $Y|X$-shifts leads to more complex and volatile algorithmic behavior, as shown in Sections 3.1 and 3.2, whereas the shift patterns in image datasets are relatively simple since most of them are covariate shifts. 
    \item Methodologies: In terms of the training procedure, we consider DRO methods since they focus on worst-case distributions that may incorporate $Y|X$-shifts, and therefore, are of potential utility in the tabular data; In contrast, \citet{gulrajanisearch} consider various domain generalization methods that build on neural networks and assume invariance across domains, and therefore, these methods are best suited to covariate shifts in the image datasets. In terms of the model classes, \citet{gulrajanisearch} limit their analysis to deep networks with 9 methods in all, whereas our benchmark spans 45 methods -- including linear models, tree-based ensembles, and neural networks -- to reflect the reality that simpler models often outperform very deep architectures on tabular data.
    \item Results and Insights: By including this broader range of model classes, we can disentangle the relative impact of algorithmic interventions (such as ambiguity-set design) from model-class effects. While we corroborate findings in \citep{gulrajanisearch} regarding which design components matter most, our expanded benchmark also uncovers simple, shift-aware interventions -- both algorithmic and data-centric -- that yield additional performance gains.
\end{enumerate}
}

\section{Details in \Cref{sec:data}}
\subsection{Details in \Cref{subsec:data-foundation}}\label{app:disde}
\paragraph{DISDE to $X$-shifts and $Y|X$-shifts.} When facing performance degradation under distribution shifts, one direct idea is to figure out the reasons why the performance drop.
To this end, Cai et al. \cite{namkoong2023diagnosing} propose DIstribution Shift DEcomposition (DISDE) to attribute the total performance degradation to $Y|X$-shifts and $X$-shifts.
Specifically, given samples $(X,Y)$ from distributions $P$ and $Q$, to quantify the discrepancy between $P_{Y|X}$ and $Q_{Y|X}$, they first control the marginal distribution on $\mathcal X$ by introducing the shared distribution $S_X$. From that, we can estimate the performance degradation caused by $Y|X$-shifts and that caused by $X$-shifts could also be estimated by comparing $S_X$ with $P_X$ and $Q_X$, respectively.
Note that DISDE could be used in image datasets (see Section 4.2 in \cite{namkoong2023diagnosing}). The official code for DISDE could be found at \url{https://github.com/namkoong-lab/disde}.
We specify the formula of DISDE as follows:
\begin{align*}
\E_Q[\ell(f_P(X), Y)] - \E_P[\ell(f_P(X), Y)] = &\quad \E_{S_X}[R_P(X)] - \E_P[R_P(X)] \tag{I}\\
& + \E_{S_X}[R_Q(X) - R_P(X)] \tag{II}\\
& + \E_{Q}[R_Q(X)] - \E_{S_X}[R_Q(X)] \tag{III},
\end{align*}
where $R_{\mu}(x):= \E_{\mu}[\ell(f_P(X), Y)| X = x]$ for $\mu = P, Q$ is denoted as the conditional risks on $P$ and $Q$. $S_X$ is the share distribution with support contained in both $P_X$ and $Q_X$. \emph{Then we see the sum of the two terms (I) and (III) as the performance drop attributed to $X$-shifts and the term (II) as the performance drop attributed to $Y|X$-shifts.}
Besides, we also implement DISDE in our released package named \textsc{WhyShift}, which could be found at \url{https://github.com/namkoong-lab/whyshift}.

\paragraph{Attribution of Performance Drop.} We use XGBoost classifier and calculate the decomposition of performance degradation via DISDE. We calculate the \textbf{$Y|X$-shift ratio} from the source distribution $P$ to the target distribution $Q$:
\[Y|X\text{-shift ratio} = \frac{\E_{S_X}[R_Q(X) - R_P(X)]}{\E_{Q}[\ell(f_P(X), Y)] - \E_{P}[\ell(f_P(X), Y)]},\]
where $R_{\mu}(X)$ and $S_X$ are defined above.

\subsection{Details in Section~\ref{subsec:method-foundation}}\label{app:method}
In this section, we provide the details of the methods in our benchmark.

\subsubsection{Further Details in DRO Methods}\label{app:dro-setup}
Recall we consider the general DRO problem in~\eqref{eqn:general-dro} in \Cref{sec:perform-comp} with the ambiguity set $\Pscr(d,\epsilon)$. Here, unless specified, $\Pscr(d,\epsilon):=\{P: d(P, \widehat P) \leq \epsilon\}$ denotes the ambiguity set around the training empirical distribution $\widehat P$ and $d(\cdot, \cdot)$ is a notion of distance between probability measures.

We describe the DRO methods into three main categories of distances used in DRO models: Wasserstein, Generalized $f$-divergence, and mixed distances due to their wide usage.

\paragraph{\bf Wasserstein Distance.}~In the standard \textbf{Wasserstein-DRO} method \citep{blanchet2019data}, we apply $d(P, Q) = W_c(P, Q)$, where $W_c$ is the Wasserstein distance:
\begin{equation}\label{eq:was-dist}
    W_c(Q, P)= \min_{\pi\in \Pi(Q,P)} \mathbb E_{Z_1,Z_2 \sim \pi}\left[c(Z_1,Z_2)\right],
\end{equation}
where the induced cost function inside the distance is given by: 
\[c((X_1, Y_1), (X_2, Y_2)) = (X_1-X_2)^{\top} \Lambda (X_1 - X_2).\]
Unless specified, $\Lambda$ is taken as the unit matrix and $c(\cdot, \cdot)$ then becomes squared Euclidean distance. That is to say, we only allow the perturbation of $X$ but not $Y$, where Theorem 1 in \cite{blanchet2019data} provides the corresponding tractable reformulations.

For the \textbf{Augmented Wasserstein-DRO}  (Aug.Wass.-DRO) method \citep{shafieezadeh2019regularization},
we still use $d(P, Q)$ being the Wasserstein distance $W_c$ in~\eqref{eq:was-dist} but now we allow the changes in $y$. More specifically, we now set:
\[c((X_1, Y_1), (X_2, Y_2)) = (X_1-X_2)^{\top}\Lambda (X_1 - X_2) + \kappa|Y_1 - Y_2|,\]
where $\kappa$ (set as 1) is another hyperparameter to control the change in $Y$. Note that when $\kappa = \infty$, it reduces to the previous standard case, where \cite{shafieezadeh2019regularization} provides various loss functions to reformulate the problem.

For the \textbf{Satisficing Wasserstein-DRO} (Satis.Wass.-DRO) method \cite{long2023robust},  we solve the following constrained optimization problem, where DRO is set in the constraint counterpart:
$$\max \paran{\|\theta\|_{\Sigma^{-1/2}},\quad \text{s.t.}~\E_{(X,Y) \sim P}[\ell_{tr}(\theta;(X, Y))] \leq \tau + \epsilon W_c(P, \widehat P), \forall P}.$$
In the formulation above, we do not set $\epsilon$ as the hyperparameter but as an optimization goal such that the minimized worst-case performance is still no worse than some target quantity $\tau$, which is a new hyperparameter and selected as the multiplication (i.e., the so-called \emph{target ratio}, $>1$) of the best empirical performance with $E_{(X, Y)\sim \widehat P}[\ell(\widehat\theta;(X, Y))]$ with $\widehat\theta$ obtained in the corresponding~\eqref{eqn:general-erm}. To solve this optimization problem, we prefix $\epsilon \in [0, 100]$  and compute the corresponding left-hand objective value at each binary search to see if it reaches the goal $\tau$ and then reduce the potential range of $\epsilon$ half by half.

\paragraph{\bf Generalized $f$-divergence.} When $d$ is set as the generalized $f$-divergence (including CVaR), all distances there can be formulated as follows:
\[d(P, Q) = E_{Q}\Paran{f\Para{\frac{dP}{dQ}}}.\] 

For the \textbf{KL-DRO} method \citep{hu2013kullback}, we apply $f(x) = x \log x - (x - 1)$;

For the \textbf{$\chi^2$-DRO} method \citep{duchi2019variance},  we apply $f(x) = (x - 1)^2$;

For the \textbf{TV-DRO} method \citep{jiang2018risk},  we apply $f(x) = |x - 1|$;

In each of the three methods above, the hyperparameter is $\epsilon$ as the uncertainty set size.

For the (joint) \textbf{CVaR-DRO} problem \citep{rockafellar2000optimization},
we apply $f(x) = 0$ if $x \in [\frac{1}{\alpha}, \alpha]$ and $\infty$ otherwise (an augmented definition of the standard $f$-DRO problem). Here the hyperparameter $\alpha$ denotes the worst-case ratio. And we denote such induced $d$ as the so-called \emph{CVaR distance}. 

Besides these standard (generalized) $f$-divergence DRO models, we also include DRO models modeling partial distribution shifts on $(X, Y)$ based on CVaR metrics. Here, we directly use $\mathcal{P}(\alpha)$ as the ambiguity set where $\alpha$ is the CVaR parameter.

If we only consider the shifts in the marginal distribution $X$, we formulate the \textbf{Marginal-CVaR-DRO} model:
\[\mathcal{P}(\alpha) = \{Q_0: P_X = \alpha Q_0 + (1-\alpha) Q_1, \text{for some}~\alpha \geq \alpha_0~\text{and distribution}~Q_1~\text{and}~\mathcal{X}\}.\]
Specifically, we follow the formulation of (27) in \cite{duchi2023distributionally} to fit the model.

If we consider the shift in the conditional distribution $Y|X$,  we formulate the \textbf{Conditional-CVaR-DRO} model:
\[\mathcal{P}(\alpha) = \{Q_0: P_{Y|X} = \alpha Q_0 + (1-\alpha)Q_1, \text{for some}~\alpha \geq \alpha_0~\text{and distribution}~Q_1~\text{and}~\mathcal{Y}\}.\] 
Specifically, we follow the formulation of Theorem 2 in \cite{sahoo2022learning} to fit the model where approximating $\alpha(x) = \theta^{\top}x$. 


\paragraph{\bf Mixed Distances.}~We include other ambiguity set design $\Pscr$ induced by some mixed distances including \textbf{Sinkhorn-DRO} \citep{wang2021sinkhorn}, \textbf{Holistic-DRO} \citep{bennouna2022holistic} (Theorem 3.4 there), \textbf{Unified-DRO} (with $L_2$ and $L_{\infty}$ norm corresponding to $p = 2, \infty$ norm in their Theorem 5.2) \citep{blanchet2023unifying}. We follow their initial github codebases and hyperparameter selection when implementing these methods. We also consider a class of DORO methods~\citep{zhai2021doro}, which discards a proportion $\epsilon$ of the largest error points in each iteration to mitigate the outliers in DRO. Specifically in DORO, we consider the ambiguity set as: 
\begin{equation}\label{eq:doro}
    \Pscr = \{P: d(P, Q)\leq f_d(1/\alpha), d_{TV}(Q, \widehat P) \leq \epsilon\}.
\end{equation}
Here in DORO, we consider \textbf{CVaR-DORO} and \textbf{$\chi^2$-DORO} where $d$ in~\eqref{eq:doro} is set as CVaR-distance and $\chi^2$-divergence respectively, where $f_d(t) = (t - 1)^2/2$ when $d$ is $\chi^2$-divergence and $f_d(t) = 1$ when $d$ is CVaR distance.

\paragraph{\bf Other DRO Setups.}~Besides these methods with performance reported in the paper, in our codebase, we also implement Bayesian-based DRO \citep{shapiro2023bayesian}, Parametric DRO \citep{iyengar2022hedging} where we replace the reference measure $Q$ to be parametric distribution (Gaussian mixture distribution); MMD-DRO \citep{zhu2021kernel}, where we set $d(P, Q)$ as the kernel distance with the Gaussian kernel. However, the performance of these methods shows similar behavior compared with the benchmarks we use, so we do not report them here.

\subsubsection{Further Details in Other Methods}\label{app:alg-implement}

\paragraph{Basic ERM methods.}
For LR and SVM, we use the standard implementation in \texttt{scikit-learn}~\citep{pedregosa2011scikit} and train them on CPUs.
For NN, we implement it via \texttt{PyTorch}~\citep{pytorch} and train it on GPUs.

\paragraph{Tree-based ensemble methods.} 
As shown by Gardner~\textit{et al.}~\citep{gardner2022subgroup}, several tree-based methods achieve good performances on tabular datasets.
And gradient-boosted trees (e.g., XGB, LGBM, GBM) are widely considered as the state-of-the-art methods on tabular data.
Therefore, we compare XGB, LGBM, and GBM in this work.
Also, we add RF to incorporate the performance of tree bagging methods.
For RF and GBM, we use the standard implementations in \texttt{scikit-learn}~\citep{pedregosa2011scikit}.
For XGB and LGBM, we use the standard implementations in the \texttt{xgboost} package\footnote{\url{https://pypi.org/project/xgboost/}} and the \texttt{lightgbm} package\footnote{\url{https://pypi.org/project/lightgbm/}}.
All these methods are trained on CPUs.  \wty{For tree-based ensemble DRO methods, we directly apply the DRO-kind loss functions to XGBoost and LightGBM by changing the empirical losses implemented in \texttt{xgboost.train} and \texttt{lightgbm.train} to distributionally robust losses with CVaR and KL divergences respectively.}

 \paragraph{Imbalanced learning methods.} 
 Recently, some simple data balancing methods~\citep{idrissi2022simple} have shown good worst-group performances under distribution shifts.
 In our benchmark, we implement 4 typical balancing methods, namely Sub-Sampling $Y$ (SUBY), Reweighting $Y$ (RWY), Sub-Sampling Group (SUBG), and Reweighting Group (RWG).
 For these imbalanced learning methods, we use XGB as the backbone model due to its superiority on tabular data and adjust sample weight or training procedure accordingly for each of the methods.

\paragraph{Fairness-enhancing methods.} 
Following \citet{ding2022retiring} and \citet{gardner2022subgroup}, fairness-enhancing methods have the potential to mitigate the performance degradation under distribution shifts.
In our benchmark, we evaluate the in-processing and post-processing intervention methods.
The in-processing method~\citep{agarwal2018reductions} minimizes the prediction error subject to some fairness constraints, and in our benchmark, we choose three typical fairness constraints, including demographic parity (DP), equal opportunity (EO), error parity (EP).
And the post-processing method~\citep{hardt2016equality} randomizes the predictions of a fixed classifier to satisfy equalized odds criterion, and we use exponential and threshold controls in our benchmark.
We use the implementations of \texttt{aif360}~\citep{aif360} and \texttt{fairlearn}~\citep{bird2020fairlearn}.

\subsubsection{Parameter Search Space}\label{app:hyper}
We provide the hyperparameter grids in \Cref{tab:hparam-grids}.
We mainly use the hyperparameter grids proposed in~\citep{gardner2022subgroup}, and restrict the grid size of each method in each setting to 200 due to computational costs.
In each setting, we randomly pick 200 configurations for each algorithm for a fair comparison.
For methods incorporating the underlying model class (e.g., NN2 / NN3 / NN4 / XGB), we choose the top 10 best configurations for that model class to reduce the search space, making the searched best configuration represent its best performance more accurately.
Moreover, to accelerate the grid search process, we utilize \texttt{Ray}~\citep{liaw2018tune} to run experiments in parallel.

\begin{tiny}    
\begin{longtable}{lccc}
\caption{Hyperparameter grids for methods in \Cref{sec:perform-comp} used in all experiments. $\diamond$: for methods with the total grid size above 200, we randomly sample \emph{200} configurations for fair comparisons. For methods incorporating backbone models (e.g., NN/XGB), we choose top-10 best configurations for that backbone model to reduce the search space, making the searched best configuration represent its best performance more accurately.}
\label{tab:hparam-grids} \\

\toprule \textbf{Model} & \textbf{Total Grid Size} & \textbf{Hyperparameter} & \textbf{Value Range} \\  \hline 
\endfirsthead

\multicolumn{4}{c}%
{{\bfseries \tablename\ \thetable{} -- continued from previous page}} \\
\hline \textbf{Model} & \textbf{Total Grid Size} & \textbf{Hyperparameter} & \textbf{Value Range} \\  \hline  
\endhead

\hline \multicolumn{3}{r}{{Continued on next page}} \\ \hline
\endfoot

\bottomrule
\endlastfoot

\multicolumn{4}{c}{\textbf{Basic ERM methods}} \\ \toprule 
\multirow{5}{*}{NN2/NN3/NN4} & \multirow{5}{*}{324$^\diamond$} & Learning Rate & {$\{1e^{-3}, 3e^{-3}, 5e^{-3}, 1e^{-2} \}$} \\
 & & Batch Size & $\{64, 128, 256 \}$ \\
 & & Hidden Units & $\{16, 32, 64\}$ \\
 & & Dropout Ratio& $\{0,1e^{-1},5e^{-1}\}$ \\ 
 & & Train Epoch& $\{50, 100, 200\}$ \\\midrule
\multirow{2}{*}{SVM} & \multirow{2}{*}{96} & C & { $\{ 1e^{-2}, 1e^{-1}, 1,1e^1,1e^2, 1e^3 \}$} \\
&& penalty & $\{L_1, L_2\}$\\
\midrule
\multirow{4}{*}{Kernel-SVM} & \multirow{4}{*}{324$^\diamond$}& Nystroem components & \{50, 100, 200\} \\
& & kernel & \{rbf, sigmoid, polynomial\}\\
& & C & { $\{1e^{-1}, 3e^{-1}, 1,3,1e^1,3e^1, 1e^2, 3e^2, 1e^3 \}$} \\
 & & $\gamma$ & { $\{0.01, 0.1, 1, \text{scale}\}$} \\
 \midrule
 \multirow{2}{*}{LR} & \multirow{2}{*}{23} & \multirow{3}{*}{$L_2$ penalty} & $\{1e^{-3}, 3e^{-3}, 5e^{-3}, 7e^{-3}, 1e^{-2},$ \\ 
 &  &  & $3e^{-2},5e^{-2},\dots, 1.3,1.7,5\}$\\
  &  &  & $1e^1,5e^1, 1e^2, 5e^2,1e^3,5e^3, 1e^4\}$\\\bottomrule
 \multicolumn{4}{c}{\textbf{Tree-Base Ensemble Methods}} \\ \toprule
\multirow{5}{*}{RF} & \multirow{5}{*}{640$^\diamond$} & Num. Estimators & $\{ 32, 64, 128, 256, 512\}$  \\
 & & Max Features & $\{ \textrm{sqrt}, \textrm{log2}\}$ \\
 & & Min. Samples Split & $\{ 2, 4, 8, 16 \}$ \\
 & & Min. Samples Leaf & $\{ 1, 2, 4, 8\}$ \\
  & & Cost-Complexity $\alpha$ & $\{ 0, 1e^{-3}, 1e^{-2}, 1e^{-1} \}$ \\ \midrule
 \multirow{5}{*}{GBM} & \multirow{5}{*}{1680$^\diamond$} & Learning Rate & $\{ 1e^{-2}, 1e^{-1}, 5e^{-1} , 1 \}$ \\
 & & Num. Estimators & $\{ 32, 64, 128, 256\}$  \\
 & & Max Depth & $\{ 2,4,8,16\}$ \\
 & & Min. Child Samples & $\{ 1, 2, 4, 8\}$ \\ \midrule
 \multirow{5}{*}{LGBM} & \multirow{5}{*}{1680$^\diamond$} & Learning Rate & $\{ 1e^{-2}, 1e^{-1}, 5e^{-1} , 1 \}$ \\
 & & Num. Estimators & $\{ 64, 128, 256, 512 \}$  \\
 & & $L_2$-reg. & $\{ 0, 1e^{-3}, 1e^{-2}, 1e^{-1}, 1 \}$ \\
 & & Min. Child Samples & $\{ 1, 2, 4, 8, 16, 32, 64 \}$ \\
  & & { Column Subsample Ratio (tree)} & $\{ 0.5, 0.8, 1. \}$ \\ \midrule
\multirow{7}{*}{XGB} & \multirow{7}{*}{1944$^\diamond$} & Learning Rate & $\{ 0.1, 0.3, 1.0, 2.0\}$ \\
 & & Min. Split Loss & $\{ 0, 0.1, 0.5 \}$  \\
 & & Max. Depth & $\{ 4, 6, 8 \}$ \\
 & & { Column Subsample Ratio (tree)} & $\{ 0.7, 0.9, 1 \}$ \\
 & & { Column Subsample Ratio (level)} & $\{ 0.7, 0.9, 1 \}$ \\
 & & Max. Bins & $\{ 128, 256, 512 \}$ \\ 
 & & Growth Policy & {$\{ \textrm{Depthwise}, \textrm{Loss Guide} \}$} \\ \bottomrule
  \multicolumn{4}{c}{\textbf{Linear-DRO Methods} (Underlying Model Class: SVM)} \\ \toprule 
CVaR-DRO &117 & Worst-case Ratio $\alpha$ & $\{1e^{-4}, \dots, 0.01, \dots, 0.99\}$\\
Conditional(-CVaR)-DRO &117 & Worst-case Ratio $\alpha$ & $\{1e^{-4}, \dots, 0.01, \dots, 0.99\}$\\
KL-DRO &117 & { Uncertainty Set Size $\epsilon$} & $\{1e^{-4}, \dots, 0.01, \dots, 0.99\}$\\ 
$\chi^2$-DRO &117 & { Uncertainty Set Size $\epsilon$} & $\{1e^{-4}, \dots, 0.01, \dots, 0.99\}$\\ 
TV-DRO &117 & { Uncertainty Set Size $\epsilon$}  & $\{1e^{-4}, \dots, 0.01, \dots, 0.99\}$\\
Wasserstein-DRO & 138 & { Uncertainty Set Size $\epsilon$} & { $\{1e^{-4}, \dots, 0.01, \dots, 0.99, \dots, 3\}$}\\
Aug.Wass.-DRO & 138 & { Uncertainty Set Size $\epsilon$} & { $\{1e^{-4}, \dots, 0.01, \dots, 0.99, \dots, 3\}$}\\\midrule
\multirow{3}{*}{Marginal(-CVaR)-DRO} &\multirow{3}{*}{162}& Worst-case Ratio $\alpha$ & $\{0.1,0.2,\ldots,0.8,0.9\}$\\
&&$L$ & {$\{1e^{-2}, 3e^{-2}, 5e^{-2}, 0.1, 0.3, 0.5, 1,2,5,10\}$}\\
&&$p$ & \{1.5, 2\}\\
\midrule
\multirow{2}{*}{Satis.Wass.-DRO} &\multirow{2}{*}{112} & Target Ratio & $\{1.1,\dots, 2.5, 3\}$\\
& &$\kappa$ &{ $\{1e^{-2}, 5e^{-2}, 1e^{-1}, 5e^{-1}, 1, 5, 1e^7\}$}\\\midrule
\multirow{3}{*}{Sinkhorn-DRO} &\multirow{3}{*}{1156$^\diamond$} & Regularizer & { $\{1e^{-3},\dots, 7e^{-2}, 1e^{-1},\dots, 9e^{-1}\}$}\\
& &$k$ &$\{2,3,4,5\}$\\
& &$\lambda$ &{ $\{1e^{-3},\dots, 7e^{-2}, 1e^{-1},\dots, 9e^{-1}\}$}\\\midrule
\multirow{4}{*}{Holistic-DRO} &\multirow{4}{*}{3969$^\diamond$} &$\alpha$ &$\{1e^{-1}, \dots,9e^{-1}\}$\\
& &$r$ &$\{1e^{-1}, \dots,9e^{-1}\}$ \\
& &$\epsilon$ &$\{1e^{-3}, 1e^{-2}, 1e^{-1}, \dots, 9e^{-1}\}$\\
& &$\epsilon'$ &$\{1e^{-3}, 1e^{-2}, 1e^{-1}, \dots, 9e^{-1}\}$\\\midrule
\multirow{3}{*}{Unified-DRO} &\multirow{3}{*}{180} & Distance Type & \{$L_2, L_{\text{inf}}$\}\\
& & { Uncertainty Set Size $\epsilon$} & $\{1e^{-3}, \dots, 9e^{-1}\}$\\
& &$\theta_1$ & \tiny $\{1.001, 1.01, 1.1, 1.5, 2,3,5,10,50,100\}$\\\midrule
\multicolumn{4}{c}{\textbf{Kernel-DRO Methods} (Underlying Model Class: Kernel-SVM)} \\ \toprule 
\multirow{2}{*}{CVaR-DRO} &\multirow{2}{*}{4212$^\diamond$} & Worst-case Ratio $\alpha$ & $\{1e^{-4}, \dots, 0.01, \dots, 0.99\}$\\ 
& & Underlying model class & Kernel-SVM \\
\midrule
\multirow{2}{*}{KL-DRO} &\multirow{2}{*}{4212$^\diamond$} & { Uncertainty Set Size $\epsilon$} & $\{1e^{-4}, \dots, 0.01, \dots, 0.99\}$\\ 
& & Underlying model class & Kernel-SVM \\
\midrule
\multirow{2}{*}{$\chi^2$-DRO} &\multirow{2}{*}{4212$^\diamond$}& { Uncertainty Set Size $\epsilon$} & $\{1e^{-4}, \dots, 0.01, \dots, 0.99\}$\\ 
& & Underlying model class & Kernel-SVM \\
\midrule
\multirow{2}{*}{Wasserstein-DRO} & \multirow{2}{*}{4968$^\diamond$} & {Uncertainty Set Size $\epsilon$} & { $\{1e^{-4}, \dots, 0.01, \dots, 0.99, \dots, 3\}$}\\
& & Underlying model class & Kernel-SVM \\
\midrule
\multicolumn{4}{c}{\textbf{Tree-based DRO Methods} (Underlying Model Class: LGBM / XGB)} \\ \toprule 
\multirow{2}{*}{LGBM-CVaR-DRO} &\multirow{2}{*}{15120$^\diamond$} & Worst-case Ratio $\alpha$ & $\{0.1,0.2,\ldots, 0.8, 0.9\}$\\ 
& & Underlying model class & LGBM\\
\midrule
\multirow{2}{*}{LGBM-KL-DRO} &\multirow{2}{*}{15120$^\diamond$} & {Uncertainty Set Size $\epsilon$} & $\{0.1, 0.2,\ldots,0.8, 0.9\}$\\ 
& & Underlying model class & LGBM \\
\midrule
\multirow{2}{*}{XGB-CVaR-DRO} &\multirow{2}{*}{17496$^\diamond$} & Worst-case Ratio $\alpha$ & $\{0.1,0.2,\ldots, 0.8, 0.9\}$\\ 
& & Underlying model class & XGB\\
\midrule
\multirow{2}{*}{XGB-KL-DRO} &\multirow{2}{*}{17496$^\diamond$} & { Uncertainty Set Size $\epsilon$} & $\{0.1, 0.2,\ldots,0.8, 0.9\}$\\ 
& & Underlying model class & XGB \\
\midrule
\multicolumn{4}{c}{\textbf{NN-DRO Methods} (Underlying Model Class: NN)} \\ \toprule 
\multirow{2}{*}{ CVaR-DRO} &\multirow{2}{*}{1620$^\diamond$} & Worst-case Ratio $\alpha$ & $\{0.01, 0.1, 0.2, 0.3, 0.5, 1.0\}$\\
& & Underlying Model Class & NN2/NN3/NN4\\ \midrule
\multirow{2}{*}{ $\chi^2$-DRO} &\multirow{2}{*}{1620$^\diamond$} & { Uncertainty Set Size $\epsilon$} & $\{0.01, 0.1, 0.2, 0.3, 0.5, 1.0\}$\\
& & Underlying Model Class & NN2/NN3/NN4\\ \midrule
\multirow{3}{*}{ CVaR-DORO} &\multirow{3}{*}{8100$^\diamond$} & Worst-case Ratio $\alpha$ & $\{0.1, 0.2, 0.3, 0.4, 0.5, 0.6\}$\\
& & $\epsilon$ & $\{0.001, 0.01, 0.1 , 0.2, 0.3\}$\\
& & Underlying Model Class & NN2\\\midrule
\multirow{3}{*}{ $\chi^2$-DORO} &\multirow{3}{*}{8100$^\diamond$} & Worst-case Ratio $\alpha$ & $\{0.1, 0.2, 0.3, 0.4, 0.5, 0.6\}$\\
& & $\epsilon$ & $\{0.001, 0.01, 0.1 , 0.2, 0.3\}$\\
& & Underlying Model Class & NN2\\\bottomrule
 \multicolumn{4}{c}{\textbf{Imbalanced Learning Methods}} \\ \toprule 
SUBY, RWY & 1944$^\diamond$ & Underlying Model Class & XGB \\ 
SUBG, RWG & 1944$^\diamond$ & Underlying Model Class & XGB \\ \midrule
\multicolumn{4}{c}{\textbf{Fairness-Enhancing Methods}} \\ \toprule 
\multirow{2}{*}{In-processing} & \multirow{2}{*}{1944$^\diamond$} & Constraint Type & $\{ \text{DP}, \text{EO}, \text{Error Parity} \}$ \\ 
& & Underlying Model Class & XGB \\\midrule
\multirow{2}{*}{Post-processing} & \multirow{2}{*}{1944$^\diamond$} & Constraint Type & $\{ \text{Exp}, \text{Threshold}\}$ \\ 
& & Underlying Model Class & XGB \\
\end{longtable}
\end{tiny}

\subsubsection{Training and Validation Details in~\Cref{sec:perform-comp}}\label{app:traindetail}
\paragraph{Training.} In each setting, we randomly sample 20,000 samples from the source domain for training and 20,000 samples from the target domain for testing. For \texttt{US Accident} and \texttt{Taxi} datasets, we only randomly sample 8,000 samples for the source/target domain due to fewer samples involved in each setting. In settings where the source domain does not have enough samples, we use 80\% samples from the source domain for training. Experiments of all settings were run on a server using 48 cores from two AMD EPYC 7402 24-Core Processors.
NN-DRO methods were trained on GPU, NVIDIA GeForce RTX 3090, while the other methods were trained only on CPUs. Note that all experiments are small-scale and could be run efficiently.
Besides, during training, we found that the bi-search in $\chi^2$-DRO (NN) sometimes failed to converge, and therefore we set the maximal iteration number to 2500. 

\paragraph{Hyperparameter Tuning. }Unless specified, we selected the best configuration according to the performance on 128 samples selected from the target domain, i.e., the so-called \textbf{out-of-distribution accuracy}, where the validation set is identically distributed ($i.d.$) with the test set.

\wty{\paragraph{Comparison of Evaluation Pipelines.} To highlight the suitability of our evaluation pipeline, we compare it with other alternative evaluation pipelines that utilize source and target data for basic ERM methods across different model classes, including LR, linear-SVM, XGB, LGBM:
\begin{enumerate}[(a),leftmargin=*]
    \item We train each ERM method on all samples from the source domain and tune the best hyperparameter of each ERM method based on the highest accuracy on samples from the source domain (in-distribution accuracy);
    \item We train each ERM method on all samples from the source domain and tune the best hyperparameter of each ERM method based on the highest accuracy on 128 samples from the target domain;
    \item We train each ERM method on all samples from the source domain and tune the best hyperparameter of each ERM method based on the highest accuracy on all samples from the target domain;
    \item We train each ERM method on 128 samples from the target domain.
\end{enumerate}
where $(a), (b), (d)$ are based on the realistic setting with sufficient samples from the source domain and 128 samples from the target domain. We rename $(b)$ and $(d)$ as $(b)$-128 and $(d)$-128 respectively and also consider $(b)$-32 and $(b)$-64 ($(d)$-32 and $(d)$-64 respectively) that use 32 and 64 samples from the target domain to indicate the robustness of the selection criterion with respect to the number of samples from the target domain.
\begin{table}[!htb]
    \centering
    \caption{Accuracy and Radius of ERM methods with different training methods, where the accuracy and radius are computed averaged over all target domains in the setting.}
    \label{tab:erm_validation}
    \resizebox{\textwidth}{!}{\begin{tabular}{c|cccc|cccc}
    \toprule
& \multicolumn{8}{c}{Method Performance}\\
& \multicolumn{4}{c}{Accuracy} & \multicolumn{4}{c}{Macro F1-score}\\
\midrule
Evaluation Approach & LR & SVM & XGB & LGBM & LR & SVM & XGB & LGBM\\
\midrule
(a)&77.3\scriptsize$\pm 2.0$&77.0\scriptsize$\pm 2.1$&77.7\scriptsize$\pm 2.3$&77.8\scriptsize$\pm 2.1$&75.6\scriptsize$\pm 3.2$&75.4\scriptsize$\pm 3.3$&76.2\scriptsize$\pm 3.5$&76.2\scriptsize$\pm 3.5$\\
(b)-128&77.2\scriptsize$\pm 1.8$&77.1\scriptsize$\pm 1.9$&77.9\scriptsize$\pm 2.8$&77.8\scriptsize$\pm 2.7$&75.5\scriptsize$\pm 3.0$&75.5\scriptsize$\pm 3.2$&76.4\scriptsize$\pm 4.0$&76.2\scriptsize$\pm 3.9$\\
(b)-64&77.2\scriptsize$\pm 1.8$&77.1\scriptsize$\pm 2.0$&74.5\scriptsize$\pm 3.5$&74.9\scriptsize$\pm 2.9$&75.5\scriptsize$\pm 3.0$&75.4\scriptsize$\pm 3.3$&72.8\scriptsize$\pm 4.5$&73.2\scriptsize$\pm 3.9$\\
(b)-32&77.2\scriptsize$\pm 2.0$&77.1\scriptsize$\pm 1.9$&73.7\scriptsize$\pm 3.9$&74.3\scriptsize$\pm 4.4$&75.5\scriptsize$\pm 3.2$&75.4\scriptsize$\pm 3.2$&72.0\scriptsize$\pm 5.1$&72.1\scriptsize$\pm 5.3$\\
(c)&\textbf{77.6}\scriptsize$\pm 1.8$&\textbf{77.2}\scriptsize$\pm 1.9$&\textbf{78.3}\scriptsize$\pm 2.5$&\textbf{79.2}\scriptsize$\pm 2.0$&\textbf{75.9}\scriptsize$\pm 3.1$&\textbf{75.5}\scriptsize$\pm 3.2$&\textbf{76.5}\scriptsize$\pm 3.2$&\textbf{76.7}\scriptsize$\pm 2.9$\\
(d)-128&71.8\scriptsize$\pm 3.0$&71.1\scriptsize$\pm 2.7$&72.8\scriptsize$\pm 2.9$&73.3\scriptsize$\pm 2.8$&75.5\scriptsize$\pm 3.0$&75.5\scriptsize$\pm 3.2$&74.4\scriptsize$\pm 4.0$&74.2\scriptsize$\pm 3.9$\\
(d)-64&68.7\scriptsize$\pm 4.7$&66.8\scriptsize$\pm 5.1$&70.7\scriptsize$\pm 3.6$&71.2\scriptsize$\pm 3.2$&75.5\scriptsize$\pm 3.0$&75.4\scriptsize$\pm 3.3$&72.8\scriptsize$\pm 4.5$&73.2\scriptsize$\pm 3.9$\\
(d)-32&66.9\scriptsize$\pm 4.9$&65.0\scriptsize$\pm 5.2$&68.8\scriptsize$\pm 4.6$&68.9\scriptsize$\pm 4.4$&75.5\scriptsize$\pm 3.2$&75.4\scriptsize$\pm 3.2$&72.0\scriptsize$\pm 5.1$&72.1\scriptsize$\pm 5.3$\\
\hline
\bottomrule
\end{tabular}}
\end{table}}

\wty{We evaluate the model performance on the \texttt{ACS Income} (Setting 1) dataset in \Cref{tab:erm_validation}. Rule (b)—training on the full source dataset and using 128 target samples for hyperparameter selection—outperforms alternative practical rules that train directly on 128 (or 64, 32) samples or the method that is validated using samples from the source domain. Moreover, its selected method performance closely matches that chosen by rule (c), which relies on oracle access to target-domain labels. These results validate our proposed learning scenario. In Appendix \ref{app:feasibility}, we further compare every method from Section \ref{sec:data} across all settings against the oracle ERM trained on the full target dataset, demonstrating that our experimental design is both challenging and practically achievable.}

\wty{\paragraph{Comparison of DRO Validation Approaches.} Furthermore, to highlight the suitability of our tuning method for DRO approaches, we compare validation approaches in several Wasserstein and $f$-divergence-based DRO methods under linear models since there are no other hyperparameter configurations and the required radii that contain the target distribution are easy to compute there compared with DRO under hybrid distances.}

\wty{We compare the following five rules to select the radius in each DRO approach:
\begin{enumerate}[(a),leftmargin=*]
    \item We tune the best radius based on the highest accuracy on validation samples from the source domain;
    \item We tune the best radius based on the highest accuracy on 128 validation samples from the target domain, which is how we evaluate models in the main body;
    \item We tune the best radius  based on the highest accuracy on all (20,000) samples from the target domain;
    \item We compute the required radius to ensure the target distribution is in the ambiguity set with a high probability using 128 validation samples from the target domain;
    \item We compute the required radius to ensure the target distribution is in the ambiguity set with a high probability using all samples from the target domain.
\end{enumerate}
Above, the set of all the possible radii in $(a)$ and $(b)$ are from a fixed grid in \Cref{tab:hparam-grids} of Appendix~\ref{app:hyper}; and validation approaches $(c)$ and $(e)$ are impractical since they require access to all samples from the target domain, and are included only for comparison. Then we detail how we set the radius to ensure the target distribution is in the ambiguity set with a high probability (i.e., the upper confidence bound principle). We compute the required radius of each method as follows:
\begin{itemize}
    \item For Wasserstein-DRO, we follow the rule in Section 4 of \cite{lee2018minimax};
    \item For $f$-divergence, besides training and validation samples, we assume access to a likelihood ratio function $r((x, y)) := \frac{d P((x, y))}{d Q((x, y))}$, which can be computed in advance using all samples from the source and the target domain. For CVaR, we set $\alpha = \frac{1}{\max_{i \in [m]}r((\tilde x_i, \tilde y_i))}$.  Recall training samples $\{(x_i, y_i)\}_{i \in [n]}$ from the source domain $P$ and $\{(\tilde x_j, \tilde y_j)\}_{j \in [m]}$ from the target domain $Q$. For other $f$-divergences (KL-divergence, $\chi^2$-divergence and TV-Distance), we set the required radius that contains the true target distribution with probability at least $1-\delta$ by the central limit theorem approximation:
\[\epsilon = \frac{1}{m}\sum_{i \in [m]}f(r((\tilde x_i, \tilde y_i))) + z_{1-\frac{\delta}{2}}\frac{\sqrt{\sum_{i \in [m]}(f(r((\tilde x_i, \tilde y_i)) - \sum_{i \in [m]}f(r((\tilde x_i, \tilde y_i)))/m)^2}}{m},\]
where $z_{\cdot}$ is the normal quantile. We set $\delta =0.2$ in our setting to ensure the target distribution is in the ambiguity set with 80\% probability.
\end{itemize}}
\wty{Since we focus on problems with distribution shifts where a large number of samples are available from the source domain, the radius selection principles for problems without distribution shifts and limited training samples are no longer suitable~\citep{esfahani2018data,blanchet2019robust}. Therefore, we only compare the proper validation approaches under distribution shifts.  }

\wty{We report the selected radius and corresponding accuracy of each validation approach in \Cref{tab:dro_validation} across settings with more than 10 target domains (i.e., Settings 1, 2, 4, 5). Unsurprisingly, the validation approach $(c)$ achieves the highest accuracy across each setting. Compared with that, our proposed rule $(b)$ yields a comparable accuracy and similar radius across each setting, indicating the stability of using 128 target samples to select the radius with the highest accuracy. In contrast, the validation approach $(d)$ based on the upper confidence bound principle usually selects overly large radii, which explains its underperformance. Even when full access to samples from the target domain is granted in the approach $(e)$, the overestimation issue persists. This is because choosing the radius to contain the true distribution for worst-case performance guarantees may yield an overly large ambiguity set and thus be overly conservative, particularly in the presence of distribution shifts. Finally, the in-sample validation approach $(a)$ consistently leads to poor accuracy, underscoring the importance of incorporating samples from the target domain in the validation procedure.}
\begin{table}[!htb]
    \centering
    \caption{Accuracy and Radius of Linear DRO with different validation rules, where the accuracy and radius are computed averaged over all target domains in the setting and boldfaced values in each setting denote the one with the highest accuracy and corresponding radius.}
    \label{tab:dro_validation}
\resizebox{\textwidth}{!}{
    \begin{tabular}{c|ccccc|ccccc}
    \toprule
& \multicolumn{5}{c}{Accuracy} & \multicolumn{5}{c}{Radius}\\
\midrule
Setting / Tuning Approach & (a) & (b) & (c) & (d) & (e) & (a) & (b) & (c) & (d) & (e)\\
\midrule
\multicolumn{11}{c}{KL-DRO} \\
\midrule
1 & 76.8 & 77.9 & \textbf{78.3} & 76.7 & 76.7 & 0.060 & 0.730 & \textbf{0.782} & 3.265 & 2.773 \\
2 & 73.3 & 74.1 & \textbf{74.6} & 73.2 & 73.2 & 0.220 & 0.197 & \textbf{0.268} & 0.669 & 0.411 \\
4 & 75.7 & 76.6 & \textbf{77.1} & 75.0 & 75.1 & 0.310 & 0.289 & \textbf{0.403} & 0.634 & 0.512 \\
5 & 72.6 & 72.9 & \textbf{73.0} & 68.9 & 69.0 & 0.002 & 0.016 & \textbf{0.016} & 6.217 & 5.023 \\
\midrule
\multicolumn{11}{c}{CVaR-DRO} \\
\midrule
1 & 75.8 & 78.2 & \textbf{79.1} & 77.2 & 77.1 & 0.010 & 0.087 & \textbf{0.121} & 0.228 & 0.142 \\
2 & 73.3 & 74.5 & \textbf{75.4} & 71.7 & 60.6 & 0.310 & 0.166 & \textbf{0.178} & 0.174 & 0.035 \\
4 & 74.4 & 76.9 & \textbf{77.9} & 73.6 & 74.0 & 0.120 & 0.172 & \textbf{0.225} & 0.149 & 0.132 \\
5 & 72.9 & 73.5 & \textbf{73.9} & 68.8 & 68.8 & 0.390 & 0.342 & \textbf{0.319} & 0.127 & 0.066 \\
\midrule
\multicolumn{11}{c}{$\chi^2$-divergence-DRO} \\
\midrule
1 & 76.8 & 77.9 & \textbf{78.2} & 77.2 & 77.3 & 0.040 & 0.954 & \textbf{1.049} & 5.501 & 5.319 \\
2 & 73.4 & 74.6 & \textbf{75.0} & 73.4 & 73.4 & 0.780 & 0.635 & \textbf{0.739} & 1.492 & 1.583 \\
4 & 75.8 & 76.8 & \textbf{77.3} & 75.2 & 75.2 & 1.200 & 0.826 & \textbf{0.862} & 1.486 & 1.412 \\
5 & 72.7 & 72.8 & \textbf{73.2} & 68.5 & 68.5 & 0.005 & 0.485 & \textbf{0.229} & 8.933 & 8.859 \\
\midrule
\multicolumn{11}{c}{TV-DRO} \\
\midrule
1 & 76.3 & 78.0 & \textbf{78.8} & 73.0 & 72.5 & 0.860 & 0.426 & \textbf{0.401} & 0.585 & 0.539 \\
2 & 73.1 & 74.2 & \textbf{74.9} & 73.2 & 73.1 & 0.840 & 0.624 & \textbf{0.586} & 0.418 & 0.404 \\
4 & 75.2 & 76.1 & \textbf{76.6} & 72.4 & 72.7 & 0.001 & 0.458 & \textbf{0.467} & 0.451 & 0.439 \\
5 & 72.6 & 73.4 & \textbf{73.6} & 68.4 & 68.2 & 0.006 & 0.193 & \textbf{0.221} & 0.875 & 0.816 \\
\midrule
\multicolumn{11}{c}{Wasserstein-DRO} \\
\midrule
1 & 77.3 & 78.5 & \textbf{79.0} & 65.2 & 65.2 & 0.000 & 0.203 & \textbf{0.200} & 7.558 & 4.927 \\
2 & 73.2 & 73.8 & \textbf{74.0} & 72.7 & 72.7 & 0.000 & 0.016 & \textbf{0.013} & 6.466 & 6.466 \\
4 & 75.3 & 75.4 & \textbf{75.6} & 70.1 & 70.1 & 0.001 & 0.001 & \textbf{0.000} & 5.054 & 3.628 \\
5 & 72.3 & 73.0 & \textbf{73.5} & 57.1 & 57.1 & 0.001 & 0.010 & \textbf{0.016} & 6.651 & 5.047 \\
\bottomrule
    \end{tabular}}
\end{table}

\section{Details in Section~\ref{sec:perform-comp}}\label{app:benchmark}
\subsection{Detailed Results of 10 Settings in \Cref{table:overview}}\label{app:setup}
In the main body, due to space limitations, we only visualize the target performances of different methods.
Here we provide the detailed results of all algorithms on the 10 selected pairs in \Cref{table:overview}.
For each method, we select the top-10 configurations according to the validation set from the corresponding target domain in each setting and report the mean accuracy (or macro F1-score) as well as the standard deviation in \Cref{table:selected_results} (or \Cref{table:selected_results_f1} respectively) with respect to both $i.d.$ and $o.o.d.$ (out-of-distribution, i.e., the corresponding target) data.


    \begin{landscape}
\begin{table}[!htb]
\caption{Results (Accuracy) of 10 selected pairs in \Cref{subsec:benchmark-empirical-result}, where we run each method with its top-10 configurations (according to the target F1-score) and report its mean accuracy and standard deviation.}
\vspace{5pt}
\label{table:selected_results}
\resizebox{21cm}{!}{
\begin{tabular}{@{}llcccccccccccccccccccc@{}}
\toprule\toprule
\multicolumn{2}{l}{\large \textbf{Dataset}}                                                                   & \multicolumn{2}{c}{\large \texttt{ACS Income}}                                                                    & \multicolumn{2}{c}{\large \texttt{ACS Mobility}}                                                                  & \multicolumn{2}{c}{\large \texttt{US Taxi}}                 & \multicolumn{2}{c}{\large \texttt{ACS Pub.Cov}}                                                                   & \multicolumn{2}{c}{\large \texttt{US Accident}}                                                                   & \multicolumn{2}{c}{\large \texttt{ACS Time}}                                                                      & \multicolumn{2}{c}{\large \texttt{Sub-Sampling}} & \multicolumn{2}{c}{\large \texttt{diabetes}} &  \multicolumn{2}{c}{\large \texttt{assistments}} & \multicolumn{2}{c}{\large \texttt{college}}\\ 
\multicolumn{2}{l}{\large \textbf{Shift Pattern}}                                                            & \multicolumn{2}{c}{$Y|X$ dominates}                                                                              & \multicolumn{2}{c}{ $Y|X$ dominates}                                                                              & \multicolumn{2}{c}{ $Y|X$ dominates}                       & \multicolumn{2}{c}{ $Y|X$ more}                                                                            & \multicolumn{2}{c}{$Y|X$ more}                                                                              & \multicolumn{2}{c}{$X$ more}                                                                            & \multicolumn{2}{c}{$X$ dominates}  & \multicolumn{2}{c}{$Y|X$ dominates} & \multicolumn{2}{c}{$Y|X$ more} & \multicolumn{2}{c}{$Y|X$ more} \\
\multicolumn{2}{l}{\large Source $\rightarrow$ Target Pair}                                                       & \multicolumn{2}{c}{CA$\rightarrow$PR}                                                                        & \multicolumn{2}{c}{MS$\rightarrow$HI}                                                                        & \multicolumn{2}{c}{NYC$\rightarrow$BOG}                & \multicolumn{2}{c}{NE$\rightarrow$LA}                                                                        & \multicolumn{2}{c}{CA$\rightarrow$OR}                                                                        & \multicolumn{2}{c}{2010$\rightarrow$2017}                                                                    & \multicolumn{2}{c}{Young$\rightarrow$Old}  & \multicolumn{2}{c}{White$\rightarrow$Others} & \multicolumn{2}{c}{700$\rightarrow$1} & \multicolumn{2}{c}{Normal $\rightarrow$ Special} \\
\multicolumn{2}{l}{}                                                                          & $i.d.$                                         & $o.o.d$                                         & $i.d.$                                         & $o.o.d$                                         & $i.d.$              & $o.o.d$              & $i.d.$                                         & $o.o.d$                                         & $i.d.$                                         & $o.o.d$                                         & $i.d.$                                         & $o.o.d$                                         & $i.d.$                                         & $o.o.d$  & $i.d.$                                         & $o.o.d$ &  $i.d.$                                         & $o.o.d$ &  $i.d.$                                         & $o.o.d$\\\midrule
\multirow{6}{*}{\begin{tabular}[c]{@{}l@{}}Basic \\ Methods\end{tabular}} &LR    &80.4\scriptsize$\pm 0.6$ & 73.5\scriptsize$\pm 0.5$
&76.8\scriptsize$\pm 0.2$ & 70.7\scriptsize$\pm 0.5$
&84.3\scriptsize$\pm 0.1$ & 74.9\scriptsize$\pm 0.2$
&83.5\scriptsize$\pm 0.1$ & 68.3\scriptsize$\pm 0.4$
&78.6\scriptsize$\pm 0.1$ & 77.0\scriptsize$\pm 0.2$
&72.1\scriptsize$\pm 0.0$ & 61.9\scriptsize$\pm 0.0$
&91.7\scriptsize$\pm 0.0$ & 79.4\scriptsize$\pm 0.0$
&65.0\scriptsize$\pm 0.0$ & 59.9\scriptsize$\pm 0.1$
&87.2\scriptsize$\pm 1.7$ & 46.5\scriptsize$\pm 1.2$
&93.4\scriptsize$\pm 0.0$ & 81.5\scriptsize$\pm 0.2$
\\
&SVM    &80.7\scriptsize$\pm 0.0$ & 72.8\scriptsize$\pm 0.0$
&77.9\scriptsize$\pm 0.1$ & 71.2\scriptsize$\pm 0.1$
&83.3\scriptsize$\pm 0.1$ & 75.6\scriptsize$\pm 0.1$
&83.5\scriptsize$\pm 0.1$ & 67.8\scriptsize$\pm 0.1$
&78.8\scriptsize$\pm 0.1$ & 77.2\scriptsize$\pm 0.1$
&77.1\scriptsize$\pm 0.0$ & 68.7\scriptsize$\pm 0.1$
&91.5\scriptsize$\pm 0.1$ & 79.6\scriptsize$\pm 0.2$
&64.9\scriptsize$\pm 0.0$ & 59.8\scriptsize$\pm 0.0$
&88.3\scriptsize$\pm 0.0$ & 45.7\scriptsize$\pm 0.1$
&93.4\scriptsize$\pm 0.0$ & 81.8\scriptsize$\pm 0.1$
\\
&Kernel-SVM    &74.6\scriptsize$\pm 4.1$ & {74.1}\scriptsize$\pm 5.0$
&65.7\scriptsize$\pm 7.3$ & 66.7\scriptsize$\pm 5.0$
&66.5\scriptsize$\pm 11.4$ & 69.4\scriptsize$\pm 1.7$
&66.5\scriptsize$\pm 7.1$ & 61.0\scriptsize$\pm 2.8$
&79.8\scriptsize$\pm 2.6$ & 69.6\scriptsize$\pm 4.6$
&60.4\scriptsize$\pm 4.9$ & 58.5\scriptsize$\pm 1.9$
&90.9\scriptsize$\pm 1.0$ & 77.3\scriptsize$\pm 1.1$
&64.6\scriptsize$\pm 0.5$ & 60.5\scriptsize$\pm 0.3$
&88.3\scriptsize$\pm 7.9$ & 61.2\scriptsize$\pm 2.0$
&87.4\scriptsize$\pm 3.9$ & 72.5\scriptsize$\pm 3.9$
\\
&NN2    &80.8\scriptsize$\pm 0.3$ & 74.1\scriptsize$\pm 2.4$
&72.7\scriptsize$\pm 4.6$ & 67.4\scriptsize$\pm 5.1$
&78.5\scriptsize$\pm 11.7$ & 72.1\scriptsize$\pm 1.9$
&81.6\scriptsize$\pm 1.2$ & 68.5\scriptsize$\pm 1.1$
&82.2\scriptsize$\pm 0.8$ & 66.0\scriptsize$\pm 1.2$
&75.3\scriptsize$\pm 2.4$ & 68.9\scriptsize$\pm 1.2$
&91.1\scriptsize$\pm 0.2$ & 79.1\scriptsize$\pm 0.3$
&63.2\scriptsize$\pm 0.9$ & 59.4\scriptsize$\pm 0.5$
&88.3\scriptsize$\pm 0.6$ & 58.8\scriptsize$\pm 3.4$
&93.3\scriptsize$\pm 0.2$ & 80.4\scriptsize$\pm 0.8$
\\
&NN3    &80.3\scriptsize$\pm 0.6$ & 73.0\scriptsize$\pm 3.7$
&75.9\scriptsize$\pm 0.5$ & 77.9\scriptsize$\pm 0.8$
&84.6\scriptsize$\pm 0.6$ & 73.0\scriptsize$\pm 0.9$
&81.2\scriptsize$\pm 1.0$ & 66.3\scriptsize$\pm 1.0$
&82.2\scriptsize$\pm 1.7$ & 66.1\scriptsize$\pm 1.0$
&77.1\scriptsize$\pm 0.6$ & 69.3\scriptsize$\pm 0.9$
&90.9\scriptsize$\pm 0.4$ & 79.1\scriptsize$\pm 0.6$
&63.0\scriptsize$\pm 0.9$ & 59.2\scriptsize$\pm 0.4$
&88.6\scriptsize$\pm 0.2$ & 59.8\scriptsize$\pm 2.9$
&93.4\scriptsize$\pm 0.2$ & 81.5\scriptsize$\pm 0.8$
\\
&NN4    &80.2\scriptsize$\pm 2.1$ & 73.2\scriptsize$\pm 2.3$
&75.6\scriptsize$\pm 0.5$ & 78.2\scriptsize$\pm 0.4$
&83.9\scriptsize$\pm 0.7$ & 72.8\scriptsize$\pm 2.1$
&81.2\scriptsize$\pm 0.5$ & 64.3\scriptsize$\pm 2.0$
&81.9\scriptsize$\pm 1.9$ & 66.4\scriptsize$\pm 0.9$
&75.1\scriptsize$\pm 3.1$ & 68.4\scriptsize$\pm 1.5$
&90.3\scriptsize$\pm 0.4$ & 77.7\scriptsize$\pm 0.7$
&62.5\scriptsize$\pm 1.3$ & 59.1\scriptsize$\pm 1.3$
&87.4\scriptsize$\pm 2.8$ & 57.3\scriptsize$\pm 2.5$
&93.2\scriptsize$\pm 0.4$ & 80.8\scriptsize$\pm 1.3$
\\\midrule
\multirow{4}{*}{\begin{tabular}[c]{@{}l@{}}Tree-based \\ Ensemble \\ Methods\end{tabular}}   &RF    &75.5\scriptsize$\pm 0.2$ & 76.0\scriptsize$\pm 0.8$
&79.9\scriptsize$\pm 0.4$ & 71.7\scriptsize$\pm 0.4$
&85.6\scriptsize$\pm 0.1$ & 73.2\scriptsize$\pm 0.2$
&85.4\scriptsize$\pm 0.3$ & 68.5\scriptsize$\pm 0.2$
&85.3\scriptsize$\pm 0.4$ & 67.9\scriptsize$\pm 0.3$
&78.3\scriptsize$\pm 0.9$ & 70.7\scriptsize$\pm 0.4$
&92.1\scriptsize$\pm 0.1$ & 80.5\scriptsize$\pm 0.2$
&65.0\scriptsize$\pm 0.6$ & 61.3\scriptsize$\pm 0.6$
&88.6\scriptsize$\pm 0.5$ & 64.7\scriptsize$\pm 1.3$
&94.3\scriptsize$\pm 0.1$ & 84.2\scriptsize$\pm 0.4$
\\
&XGB    &78.6\scriptsize$\pm 1.0$ & 69.8\scriptsize$\pm 1.1$
&71.2\scriptsize$\pm 1.2$ & 66.1\scriptsize$\pm 0.5$
&83.9\scriptsize$\pm 2.9$ & 71.4\scriptsize$\pm 0.8$
&83.9\scriptsize$\pm 1.5$ & 69.3\scriptsize$\pm 0.7$
&81.8\scriptsize$\pm 0.3$ & 66.8\scriptsize$\pm 0.2$
&76.8\scriptsize$\pm 1.5$ & 68.8\scriptsize$\pm 1.5$
&92.2\scriptsize$\pm 0.1$ & 80.2\scriptsize$\pm 0.1$
&61.7\scriptsize$\pm 1.0$ & 57.8\scriptsize$\pm 0.5$
&89.2\scriptsize$\pm 0.4$ & 60.5\scriptsize$\pm 0.6$
&94.7\scriptsize$\pm 0.0$ & 84.9\scriptsize$\pm 0.4$
\\
&GBM    &80.2\scriptsize$\pm 1.7$ & 73.8\scriptsize$\pm 2.6$
&65.3\scriptsize$\pm 8.4$ & 64.9\scriptsize$\pm 3.6$
&84.5\scriptsize$\pm 1.1$ & 72.3\scriptsize$\pm 0.7$
&84.9\scriptsize$\pm 0.9$ & 70.1\scriptsize$\pm 0.7$
&84.0\scriptsize$\pm 3.7$ & 67.0\scriptsize$\pm 1.7$
&77.4\scriptsize$\pm 1.6$ & 70.1\scriptsize$\pm 1.2$
&92.1\scriptsize$\pm 0.1$ & 80.6\scriptsize$\pm 0.1$
&63.7\scriptsize$\pm 0.0$ & 60.9\scriptsize$\pm 0.0$
&89.1\scriptsize$\pm 0.4$ & 62.3\scriptsize$\pm 1.3$
&94.6\scriptsize$\pm 0.1$ & 84.7\scriptsize$\pm 0.5$
\\
&LGBM    &79.5\scriptsize$\pm 0.7$ & 75.7\scriptsize$\pm 3.9$
&74.7\scriptsize$\pm 0.8$ & 67.6\scriptsize$\pm 1.1$
&81.2\scriptsize$\pm 2.4$ & 72.8\scriptsize$\pm 2.9$
&84.6\scriptsize$\pm 0.6$ & 69.9\scriptsize$\pm 0.3$
&79.7\scriptsize$\pm 3.2$ & 67.1\scriptsize$\pm 1.2$
&77.0\scriptsize$\pm 1.3$ & 69.2\scriptsize$\pm 1.1$
&92.1\scriptsize$\pm 0.1$ & 80.2\scriptsize$\pm 0.1$
&62.1\scriptsize$\pm 1.7$ & 58.4\scriptsize$\pm 1.2$
&88.1\scriptsize$\pm 0.4$ & 62.4\scriptsize$\pm 2.1$
&94.7\scriptsize$\pm 0.1$ & 84.4\scriptsize$\pm 0.7$
\\\midrule
\multirow{13}{*}{\begin{tabular}[c]{@{}l@{}}Linear-DRO \\  Methods \\ (base: SVM)\end{tabular}}   &KL-DRO    &80.1\scriptsize$\pm 0.1$ & 73.6\scriptsize$\pm 0.1$
&77.7\scriptsize$\pm 0.1$ & 70.9\scriptsize$\pm 0.1$
&83.3\scriptsize$\pm 0.3$ & 74.4\scriptsize$\pm 0.4$
&83.4\scriptsize$\pm 0.1$ & 68.3\scriptsize$\pm 0.3$
&79.2\scriptsize$\pm 0.0$ & 76.3\scriptsize$\pm 0.1$
&77.1\scriptsize$\pm 0.1$ & 68.8\scriptsize$\pm 0.2$
&91.4\scriptsize$\pm 0.1$ & 79.8\scriptsize$\pm 0.1$
&64.5\scriptsize$\pm 0.4$ & 59.5\scriptsize$\pm 0.4$
&88.3\scriptsize$\pm 0.0$ & 44.9\scriptsize$\pm 0.0$
&93.6\scriptsize$\pm 0.0$ & 81.6\scriptsize$\pm 0.2$
\\
&CVaR-DRO    &78.8\scriptsize$\pm 2.4$ & 72.9\scriptsize$\pm 1.1$
&77.6\scriptsize$\pm 0.0$ & 70.6\scriptsize$\pm 0.0$
&83.2\scriptsize$\pm 1.0$ & 75.2\scriptsize$\pm 1.7$
&82.5\scriptsize$\pm 1.0$ & 68.5\scriptsize$\pm 0.6$
&79.3\scriptsize$\pm 0.1$ & 76.1\scriptsize$\pm 0.1$
&76.8\scriptsize$\pm 0.2$ & 69.1\scriptsize$\pm 0.2$
&91.6\scriptsize$\pm 0.0$ & 80.2\scriptsize$\pm 0.1$
&63.7\scriptsize$\pm 0.4$ & 58.5\scriptsize$\pm 0.3$
&88.0\scriptsize$\pm 0.0$ & 44.9\scriptsize$\pm 0.3$
&93.1\scriptsize$\pm 0.4$ & 80.7\scriptsize$\pm 1.0$
\\
&$\chi^2$-DRO    &80.6\scriptsize$\pm 0.1$ & 73.3\scriptsize$\pm 0.4$
&77.6\scriptsize$\pm 0.1$ & 70.7\scriptsize$\pm 0.2$
&83.2\scriptsize$\pm 0.1$ & 76.0\scriptsize$\pm 0.1$
&83.3\scriptsize$\pm 0.2$ & 68.8\scriptsize$\pm 0.2$
&79.1\scriptsize$\pm 0.0$ & 76.6\scriptsize$\pm 0.3$
&77.1\scriptsize$\pm 0.0$ & 68.8\scriptsize$\pm 0.0$
&91.7\scriptsize$\pm 0.1$ & 79.7\scriptsize$\pm 0.0$
&64.8\scriptsize$\pm 0.2$ & 59.5\scriptsize$\pm 0.4$
&88.3\scriptsize$\pm 0.0$ & 44.9\scriptsize$\pm 0.0$
&93.6\scriptsize$\pm 0.0$ & 81.8\scriptsize$\pm 0.1$
\\
&TV-DRO    &78.4\scriptsize$\pm 2.0$ & 72.9\scriptsize$\pm 0.6$
&77.4\scriptsize$\pm 0.3$ & 70.4\scriptsize$\pm 0.6$
&81.4\scriptsize$\pm 2.3$ & 73.1\scriptsize$\pm 2.7$
&82.1\scriptsize$\pm 0.8$ & 66.5\scriptsize$\pm 0.9$
&79.1\scriptsize$\pm 0.4$ & 76.1\scriptsize$\pm 0.7$
&76.3\scriptsize$\pm 0.9$ & 68.7\scriptsize$\pm 0.3$
&91.0\scriptsize$\pm 0.1$ & 79.9\scriptsize$\pm 0.1$
&63.2\scriptsize$\pm 0.5$ & 57.9\scriptsize$\pm 0.7$
&88.3\scriptsize$\pm 0.0$ & 44.9\scriptsize$\pm 0.0$
&93.6\scriptsize$\pm 0.0$ & 81.6\scriptsize$\pm 0.1$
\\
&Wasserstein-DRO    &76.2\scriptsize$\pm 4.3$ & 74.2\scriptsize$\pm 2.4$
&77.5\scriptsize$\pm 0.0$ & 71.3\scriptsize$\pm 0.1$
&84.1\scriptsize$\pm 0.0$ & 75.3\scriptsize$\pm 0.1$
&82.0\scriptsize$\pm 0.3$ & 65.7\scriptsize$\pm 0.7$
&78.5\scriptsize$\pm 0.5$ & 76.8\scriptsize$\pm 0.3$
&76.0\scriptsize$\pm 0.0$ & 67.8\scriptsize$\pm 0.0$
&91.5\scriptsize$\pm 0.0$ & 77.2\scriptsize$\pm 0.0$
&64.0\scriptsize$\pm 0.3$ & 59.2\scriptsize$\pm 0.2$
&88.2\scriptsize$\pm 0.0$ & 44.9\scriptsize$\pm 0.0$
&93.6\scriptsize$\pm 0.1$ & 81.4\scriptsize$\pm 0.2$
\\
&Aug. Wass.-DRO    &78.5\scriptsize$\pm 3.9$ & 74.1\scriptsize$\pm 2.3$
&72.8\scriptsize$\pm 9.8$ & 69.9\scriptsize$\pm 2.8$
&81.7\scriptsize$\pm 0.8$ & 71.8\scriptsize$\pm 1.7$
&77.5\scriptsize$\pm 4.6$ & 65.5\scriptsize$\pm 2.3$
&79.9\scriptsize$\pm 0.3$ & 75.1\scriptsize$\pm 0.4$
&73.9\scriptsize$\pm 4.4$ & 66.6\scriptsize$\pm 2.8$
&91.2\scriptsize$\pm 0.2$ & 80.1\scriptsize$\pm 0.4$
&62.6\scriptsize$\pm 2.9$ & 58.0\scriptsize$\pm 1.9$
&70.3\scriptsize$\pm 10.8$ & 54.9\scriptsize$\pm 2.6$
&93.5\scriptsize$\pm 0.1$ & 81.5\scriptsize$\pm 0.2$
\\
&Satis. Wass.-DRO    &73.8\scriptsize$\pm 2.8$ & 73.7\scriptsize$\pm 0.5$
&75.8\scriptsize$\pm 0.0$ & 78.5\scriptsize$\pm 0.0$
&79.0\scriptsize$\pm 1.3$ & 69.4\scriptsize$\pm 2.5$
&80.0\scriptsize$\pm 0.0$ & 59.4\scriptsize$\pm 0.0$
&77.1\scriptsize$\pm 0.0$ & 77.2\scriptsize$\pm 0.0$
&71.7\scriptsize$\pm 0.0$ & 60.8\scriptsize$\pm 0.0$
&91.2\scriptsize$\pm 0.2$ & 78.2\scriptsize$\pm 1.1$
&58.5\scriptsize$\pm 0.0$ & 50.6\scriptsize$\pm 0.0$
&87.8\scriptsize$\pm 0.0$ & 44.8\scriptsize$\pm 0.0$
&92.9\scriptsize$\pm 0.2$ & 81.4\scriptsize$\pm 0.2$
\\
&Sinkhorn-DRO    &73.5\scriptsize$\pm 1.9$ & 85.5\scriptsize$\pm 1.7$
&68.7\scriptsize$\pm 3.9$ & 69.6\scriptsize$\pm 5.0$
&69.7\scriptsize$\pm 7.3$ & 75.0\scriptsize$\pm 1.3$
&81.7\scriptsize$\pm 1.4$ & 67.8\scriptsize$\pm 1.3$
&79.1\scriptsize$\pm 0.4$ & 76.2\scriptsize$\pm 0.5$
&56.7\scriptsize$\pm 12.2$ & 58.2\scriptsize$\pm 7.2$
&91.4\scriptsize$\pm 0.1$ & 79.7\scriptsize$\pm 0.1$
&62.8\scriptsize$\pm 0.4$ & 58.2\scriptsize$\pm 0.7$
&85.8\scriptsize$\pm 2.9$ & 44.6\scriptsize$\pm 0.3$
&91.6\scriptsize$\pm 1.6$ & 79.1\scriptsize$\pm 1.7$
\\
&Unified-DRO($L_2$)    &80.6\scriptsize$\pm 0.0$ & 72.7\scriptsize$\pm 0.0$
&77.5\scriptsize$\pm 0.0$ & 71.3\scriptsize$\pm 0.0$
&83.3\scriptsize$\pm 0.0$ & 75.2\scriptsize$\pm 0.1$
&82.0\scriptsize$\pm 0.0$ & 65.8\scriptsize$\pm 0.0$
&77.9\scriptsize$\pm 1.0$ & 76.9\scriptsize$\pm 0.2$
&76.0\scriptsize$\pm 0.0$ & 67.8\scriptsize$\pm 0.0$
&91.2\scriptsize$\pm 0.0$ & 79.7\scriptsize$\pm 0.0$
&62.6\scriptsize$\pm 0.0$ & 57.8\scriptsize$\pm 0.0$
&87.6\scriptsize$\pm 1.7$ & 44.8\scriptsize$\pm 0.2$
&93.5\scriptsize$\pm 0.0$ & 81.5\scriptsize$\pm 0.0$
\\
&Unified-DRO($L_\text{inf}$)    &79.9\scriptsize$\pm 0.9$ & 72.8\scriptsize$\pm 0.1$
&77.5\scriptsize$\pm 0.0$ & 71.3\scriptsize$\pm 0.1$
&83.6\scriptsize$\pm 0.0$ & 75.5\scriptsize$\pm 0.0$
&82.2\scriptsize$\pm 0.0$ & 66.0\scriptsize$\pm 0.0$
&77.6\scriptsize$\pm 0.7$ & 77.2\scriptsize$\pm 0.2$
&76.0\scriptsize$\pm 0.0$ & 67.8\scriptsize$\pm 0.0$
&91.9\scriptsize$\pm 0.0$ & 79.3\scriptsize$\pm 0.0$
&63.2\scriptsize$\pm 0.0$ & 58.7\scriptsize$\pm 0.0$
&86.6\scriptsize$\pm 4.6$ & 44.7\scriptsize$\pm 0.4$
&93.6\scriptsize$\pm 0.0$ & 81.5\scriptsize$\pm 0.0$
\\
&Marginal-DRO    &80.7\scriptsize$\pm 0.1$ & 73.2\scriptsize$\pm 0.2$
&77.9\scriptsize$\pm 0.3$ & 70.8\scriptsize$\pm 0.2$
&82.7\scriptsize$\pm 0.1$ & 76.0\scriptsize$\pm 0.1$
&83.5\scriptsize$\pm 0.3$ & 69.0\scriptsize$\pm 0.3$
&79.2\scriptsize$\pm 0.1$ & 76.0\scriptsize$\pm 0.2$
&77.2\scriptsize$\pm 0.0$ & 69.0\scriptsize$\pm 0.0$
&91.3\scriptsize$\pm 0.1$ & 79.0\scriptsize$\pm 0.3$
&64.7\scriptsize$\pm 0.2$ & 60.4\scriptsize$\pm 0.4$
&88.3\scriptsize$\pm 0.0$ & 56.9\scriptsize$\pm 0.0$
&93.4\scriptsize$\pm 0.0$ & 81.3\scriptsize$\pm 0.1$
\\
&Conditional-DRO    &80.2\scriptsize$\pm 0.3$ & 70.0\scriptsize$\pm 2.5$
&77.4\scriptsize$\pm 0.2$ & 70.8\scriptsize$\pm 0.5$
&82.4\scriptsize$\pm 0.3$ & 76.1\scriptsize$\pm 0.1$
&82.4\scriptsize$\pm 0.7$ & 69.2\scriptsize$\pm 0.3$
&82.0\scriptsize$\pm 0.6$ & 71.1\scriptsize$\pm 1.8$
&76.4\scriptsize$\pm 1.3$ & 69.0\scriptsize$\pm 0.5$
&91.5\scriptsize$\pm 0.0$ & 79.0\scriptsize$\pm 0.0$
&64.2\scriptsize$\pm 0.5$ & 60.0\scriptsize$\pm 0.7$
&88.2\scriptsize$\pm 0.0$ & 44.9\scriptsize$\pm 0.0$
&93.6\scriptsize$\pm 0.0$ & 81.6\scriptsize$\pm 0.2$
\\
&Holistic-DRO    &75.6\scriptsize$\pm 1.7$ & 80.2\scriptsize$\pm 1.0$
&77.5\scriptsize$\pm 0.1$ & 71.3\scriptsize$\pm 0.2$
&79.7\scriptsize$\pm 0.7$ & 70.8\scriptsize$\pm 0.6$
&82.6\scriptsize$\pm 0.2$ & 67.4\scriptsize$\pm 0.3$
&77.1\scriptsize$\pm 0.5$ & 77.2\scriptsize$\pm 0.1$
&76.9\scriptsize$\pm 0.1$ & 68.3\scriptsize$\pm 0.1$
&90.6\scriptsize$\pm 0.3$ & 76.8\scriptsize$\pm 0.3$
&63.8\scriptsize$\pm 0.1$ & 58.6\scriptsize$\pm 0.2$
&76.4\scriptsize$\pm 5.1$ & 43.8\scriptsize$\pm 0.4$
&91.9\scriptsize$\pm 0.5$ & 79.5\scriptsize$\pm 0.7$
\\\midrule
\multirow{8}{*}{\begin{tabular}[c]{@{}l@{}}NN-DRO\\ Methods\\ (base: NN)\end{tabular}}   &NN2-CVaR-DRO    &80.8\scriptsize$\pm 0.1$ & 70.3\scriptsize$\pm 3.3$
&62.0\scriptsize$\pm 4.3$ & 66.5\scriptsize$\pm 2.5$
&78.3\scriptsize$\pm 7.9$ & 74.2\scriptsize$\pm 0.9$
&75.9\scriptsize$\pm 9.8$ & 66.2\scriptsize$\pm 4.6$
&76.2\scriptsize$\pm 2.9$ & 73.3\scriptsize$\pm 2.3$
&75.4\scriptsize$\pm 1.4$ & 68.5\scriptsize$\pm 1.4$
&91.4\scriptsize$\pm 0.2$ & 80.7\scriptsize$\pm 0.3$
&62.9\scriptsize$\pm 1.0$ & 59.5\scriptsize$\pm 0.9$
&75.5\scriptsize$\pm 17.6$ & 60.3\scriptsize$\pm 4.1$
&93.7\scriptsize$\pm 0.4$ & 81.5\scriptsize$\pm 1.1$
\\
&NN2-$\chi^2$-DRO    &79.9\scriptsize$\pm 3.2$ & 75.1\scriptsize$\pm 5.1$
&72.5\scriptsize$\pm 4.7$ & 71.7\scriptsize$\pm 4.2$
&77.7\scriptsize$\pm 9.6$ & 74.2\scriptsize$\pm 1.4$
&77.4\scriptsize$\pm 4.2$ & 67.5\scriptsize$\pm 1.7$
&83.2\scriptsize$\pm 0.2$ & 67.6\scriptsize$\pm 0.7$
&76.5\scriptsize$\pm 1.0$ & 69.1\scriptsize$\pm 0.9$
&91.7\scriptsize$\pm 0.2$ & 79.8\scriptsize$\pm 0.2$
&62.7\scriptsize$\pm 1.8$ & 59.2\scriptsize$\pm 1.2$
&87.5\scriptsize$\pm 1.6$ & 61.8\scriptsize$\pm 3.4$
&93.4\scriptsize$\pm 0.5$ & 81.0\scriptsize$\pm 1.4$
\\
&NN2-CVaR-DORO    &76.1\scriptsize$\pm 4.7$ & 79.1\scriptsize$\pm 5.9$
&68.4\scriptsize$\pm 8.8$ & 69.0\scriptsize$\pm 5.5$
&80.5\scriptsize$\pm 2.7$ & 73.6\scriptsize$\pm 0.9$
&71.3\scriptsize$\pm 12.0$ & 63.1\scriptsize$\pm 5.3$
&82.2\scriptsize$\pm 0.9$ & 67.5\scriptsize$\pm 1.7$
&68.9\scriptsize$\pm 7.5$ & 63.5\scriptsize$\pm 4.1$
&91.6\scriptsize$\pm 0.2$ & 79.2\scriptsize$\pm 0.3$
&63.6\scriptsize$\pm 0.7$ & 60.3\scriptsize$\pm 0.6$
&87.9\scriptsize$\pm 0.7$ & 58.9\scriptsize$\pm 3.4$
&93.3\scriptsize$\pm 0.3$ & 80.7\scriptsize$\pm 0.8$
\\
&NN2-$\chi^2$-DORO    &73.2\scriptsize$\pm 4.2$ & 79.2\scriptsize$\pm 5.9$
&67.3\scriptsize$\pm 1.8$ & 72.9\scriptsize$\pm 0.9$
&71.6\scriptsize$\pm 11.6$ & 70.6\scriptsize$\pm 3.7$
&62.5\scriptsize$\pm 8.6$ & 57.7\scriptsize$\pm 3.6$
&82.9\scriptsize$\pm 0.5$ & 69.9\scriptsize$\pm 1.6$
&62.7\scriptsize$\pm 4.2$ & 58.1\scriptsize$\pm 3.7$
&91.3\scriptsize$\pm 0.2$ & 78.6\scriptsize$\pm 0.3$
&63.9\scriptsize$\pm 0.0$ & 60.6\scriptsize$\pm 0.0$
&68.8\scriptsize$\pm 18.4$ & 58.8\scriptsize$\pm 6.7$
&92.7\scriptsize$\pm 0.8$ & 80.1\scriptsize$\pm 2.3$
\\
&NN3-CVaR-DRO    &79.3\scriptsize$\pm 5.1$ & 73.7\scriptsize$\pm 5.8$
&62.4\scriptsize$\pm 8.8$ & 67.4\scriptsize$\pm 6.0$
&78.5\scriptsize$\pm 12.3$ & 73.2\scriptsize$\pm 1.6$
&77.2\scriptsize$\pm 6.8$ & 66.1\scriptsize$\pm 2.4$
&81.9\scriptsize$\pm 2.9$ & 69.6\scriptsize$\pm 3.0$
&76.8\scriptsize$\pm 0.4$ & 69.0\scriptsize$\pm 1.0$
&91.5\scriptsize$\pm 0.2$ & 79.7\scriptsize$\pm 0.4$
&63.0\scriptsize$\pm 1.8$ & 59.4\scriptsize$\pm 1.2$
&81.7\scriptsize$\pm 14.3$ & 59.0\scriptsize$\pm 3.2$
&93.4\scriptsize$\pm 0.5$ & 81.3\scriptsize$\pm 1.0$
\\
&NN3-$\chi^2$-DRO    &80.6\scriptsize$\pm 1.4$ & 74.9\scriptsize$\pm 4.0$
&74.6\scriptsize$\pm 4.1$ & 73.6\scriptsize$\pm 5.2$
&76.9\scriptsize$\pm 11.0$ & 73.1\scriptsize$\pm 1.2$
&78.3\scriptsize$\pm 3.3$ & 66.4\scriptsize$\pm 2.0$
&83.7\scriptsize$\pm 0.3$ & 66.9\scriptsize$\pm 0.4$
&76.5\scriptsize$\pm 0.9$ & 69.6\scriptsize$\pm 1.0$
&91.5\scriptsize$\pm 0.2$ & 79.5\scriptsize$\pm 0.3$
&64.0\scriptsize$\pm 1.5$ & 60.0\scriptsize$\pm 1.3$
&88.3\scriptsize$\pm 1.0$ & 60.7\scriptsize$\pm 3.1$
&93.2\scriptsize$\pm 0.4$ & 81.3\scriptsize$\pm 1.0$
\\
&NN4-CVaR-DRO    &80.9\scriptsize$\pm 0.4$ & 72.4\scriptsize$\pm 3.6$
&61.3\scriptsize$\pm 7.1$ & 66.2\scriptsize$\pm 5.1$
&72.5\scriptsize$\pm 14.6$ & 71.7\scriptsize$\pm 1.6$
&70.2\scriptsize$\pm 11.2$ & 63.7\scriptsize$\pm 2.9$
&83.4\scriptsize$\pm 1.1$ & 68.2\scriptsize$\pm 1.8$
&76.1\scriptsize$\pm 0.8$ & 68.2\scriptsize$\pm 1.4$
&91.4\scriptsize$\pm 0.2$ & 79.9\scriptsize$\pm 0.2$
&62.9\scriptsize$\pm 1.5$ & 59.3\scriptsize$\pm 1.4$
&78.6\scriptsize$\pm 17.4$ & 59.4\scriptsize$\pm 5.5$
&93.3\scriptsize$\pm 0.3$ & 81.8\scriptsize$\pm 0.5$
\\
&NN4-$\chi^2$-DRO    &78.0\scriptsize$\pm 4.1$ & 80.0\scriptsize$\pm 6.2$
&68.7\scriptsize$\pm 4.4$ & 66.0\scriptsize$\pm 6.0$
&79.7\scriptsize$\pm 7.6$ & 71.1\scriptsize$\pm 1.0$
&68.9\scriptsize$\pm 12.3$ & 60.3\scriptsize$\pm 5.1$
&83.9\scriptsize$\pm 0.2$ & 66.9\scriptsize$\pm 0.5$
&74.1\scriptsize$\pm 3.9$ & 68.6\scriptsize$\pm 2.5$
&90.9\scriptsize$\pm 0.4$ & 79.5\scriptsize$\pm 0.3$
&63.5\scriptsize$\pm 1.4$ & 60.0\scriptsize$\pm 1.1$
&88.7\scriptsize$\pm 0.1$ & 61.4\scriptsize$\pm 4.3$
&93.0\scriptsize$\pm 0.4$ & 80.5\scriptsize$\pm 1.0$
\\\midrule
\multirow{4}{*}{\begin{tabular}[c]{@{}l@{}}Tree-DRO\\ Methods\\ (base: XGB/LGBM)\end{tabular}}   &XGB-CVaR-DRO    &75.8\scriptsize$\pm 3.3$ & 71.5\scriptsize$\pm 3.6$
&66.5\scriptsize$\pm 2.9$ & 65.9\scriptsize$\pm 3.3$
&82.1\scriptsize$\pm 2.5$ & 72.5\scriptsize$\pm 1.6$
&83.7\scriptsize$\pm 2.5$ & 68.3\scriptsize$\pm 0.9$
&80.1\scriptsize$\pm 2.2$ & 65.9\scriptsize$\pm 1.1$
&72.4\scriptsize$\pm 5.6$ & 66.1\scriptsize$\pm 2.6$
&91.7\scriptsize$\pm 0.2$ & 80.0\scriptsize$\pm 0.3$
&62.3\scriptsize$\pm 1.9$ & 57.8\scriptsize$\pm 1.0$
&85.5\scriptsize$\pm 3.9$ & 59.8\scriptsize$\pm 1.3$
&94.5\scriptsize$\pm 0.3$ & 83.7\scriptsize$\pm 1.1$
\\
&XGB-KL-DRO    &76.4\scriptsize$\pm 2.4$ & 74.6\scriptsize$\pm 3.8$
&69.5\scriptsize$\pm 4.6$ & 66.5\scriptsize$\pm 4.7$
&81.8\scriptsize$\pm 4.6$ & 73.0\scriptsize$\pm 1.9$
&84.0\scriptsize$\pm 0.6$ & 68.8\scriptsize$\pm 0.3$
&82.4\scriptsize$\pm 6.1$ & 66.9\scriptsize$\pm 3.3$
&76.5\scriptsize$\pm 1.9$ & 68.3\scriptsize$\pm 1.1$
&91.3\scriptsize$\pm 0.3$ & 79.6\scriptsize$\pm 0.2$
&62.5\scriptsize$\pm 1.4$ & 58.1\scriptsize$\pm 0.5$
&87.6\scriptsize$\pm 0.4$ & 59.6\scriptsize$\pm 1.0$
&94.6\scriptsize$\pm 0.2$ & 84.0\scriptsize$\pm 1.1$
\\
&LGBM-CVaR-DRO    &75.2\scriptsize$\pm 6.0$ & 76.7\scriptsize$\pm 5.5$
&65.1\scriptsize$\pm 3.2$ & 66.3\scriptsize$\pm 3.7$
&78.9\scriptsize$\pm 5.0$ & 71.6\scriptsize$\pm 2.4$
&84.2\scriptsize$\pm 0.6$ & 68.4\scriptsize$\pm 0.4$
&78.2\scriptsize$\pm 2.9$ & 66.8\scriptsize$\pm 2.7$
&77.0\scriptsize$\pm 1.3$ & 68.5\scriptsize$\pm 1.0$
&91.9\scriptsize$\pm 0.1$ & 79.1\scriptsize$\pm 0.1$
&61.9\scriptsize$\pm 1.2$ & 57.5\scriptsize$\pm 0.5$
&89.2\scriptsize$\pm 0.6$ & 62.8\scriptsize$\pm 1.5$
&94.6\scriptsize$\pm 0.2$ & 83.9\scriptsize$\pm 0.6$
\\
&LGBM-KL-DRO    &73.5\scriptsize$\pm 6.1$ & 77.5\scriptsize$\pm 5.6$
&70.6\scriptsize$\pm 5.0$ & 69.6\scriptsize$\pm 5.0$
&83.0\scriptsize$\pm 3.8$ & 71.1\scriptsize$\pm 1.3$
&83.2\scriptsize$\pm 2.1$ & 68.2\scriptsize$\pm 1.1$
&84.1\scriptsize$\pm 3.1$ & 67.3\scriptsize$\pm 1.3$
&76.8\scriptsize$\pm 1.4$ & 68.8\scriptsize$\pm 0.8$
&91.9\scriptsize$\pm 0.2$ & 79.3\scriptsize$\pm 0.2$
&62.5\scriptsize$\pm 0.8$ & 57.7\scriptsize$\pm 0.5$
&86.2\scriptsize$\pm 5.0$ & 60.2\scriptsize$\pm 1.7$
&94.6\scriptsize$\pm 0.2$ & 84.4\scriptsize$\pm 0.6$
\\\midrule
\multirow{4}{*}{\begin{tabular}[c]{@{}l@{}}Kernel-DRO\\ Methods\\ (base: Kernel)\end{tabular}}   &Kernel-$\chi2$-DRO    &81.2\scriptsize$\pm 0.5$ & 71.0\scriptsize$\pm 0.5$
&73.1\scriptsize$\pm 2.1$ & 67.0\scriptsize$\pm 1.9$
&85.3\scriptsize$\pm 0.2$ & 70.9\scriptsize$\pm 1.7$
&84.7\scriptsize$\pm 0.2$ & 69.8\scriptsize$\pm 0.4$
&81.3\scriptsize$\pm 0.7$ & 69.9\scriptsize$\pm 1.9$
&76.0\scriptsize$\pm 1.6$ & 69.6\scriptsize$\pm 1.1$
&91.7\scriptsize$\pm 0.1$ & 79.6\scriptsize$\pm 0.2$
&64.8\scriptsize$\pm 0.5$ & 60.2\scriptsize$\pm 0.3$
&88.6\scriptsize$\pm 0.1$ & 57.6\scriptsize$\pm 2.3$
&93.6\scriptsize$\pm 0.2$ & 81.3\scriptsize$\pm 0.9$
\\
&Kernel-CVaR-DRO    &81.1\scriptsize$\pm 0.6$ & 71.0\scriptsize$\pm 0.6$
&72.7\scriptsize$\pm 5.3$ & 67.4\scriptsize$\pm 3.5$
&86.3\scriptsize$\pm 0.3$ & 71.7\scriptsize$\pm 0.9$
&83.8\scriptsize$\pm 0.5$ & 69.4\scriptsize$\pm 0.5$
&81.6\scriptsize$\pm 1.4$ & 68.6\scriptsize$\pm 2.1$
&77.0\scriptsize$\pm 1.3$ & 69.1\scriptsize$\pm 0.7$
&91.6\scriptsize$\pm 0.2$ & 80.1\scriptsize$\pm 0.2$
&63.9\scriptsize$\pm 2.0$ & 59.6\scriptsize$\pm 0.7$
&88.5\scriptsize$\pm 0.4$ & 61.2\scriptsize$\pm 1.5$
&93.6\scriptsize$\pm 0.2$ & 81.8\scriptsize$\pm 0.7$
\\
&Kernel-KL-DRO    &81.0\scriptsize$\pm 0.5$ & 71.2\scriptsize$\pm 0.8$
&67.3\scriptsize$\pm 5.1$ & 64.7\scriptsize$\pm 4.0$
&84.7\scriptsize$\pm 0.3$ & 73.0\scriptsize$\pm 1.2$
&84.1\scriptsize$\pm 0.3$ & 69.9\scriptsize$\pm 0.4$
&81.2\scriptsize$\pm 0.7$ & 69.8\scriptsize$\pm 2.3$
&74.9\scriptsize$\pm 3.1$ & 69.1\scriptsize$\pm 1.9$
&91.5\scriptsize$\pm 0.1$ & 80.0\scriptsize$\pm 0.2$
&65.0\scriptsize$\pm 0.4$ & 60.3\scriptsize$\pm 0.3$
&87.8\scriptsize$\pm 2.0$ & 60.5\scriptsize$\pm 3.3$
&93.4\scriptsize$\pm 0.3$ & 80.5\scriptsize$\pm 1.2$
\\
&Kernel-Wasserstein-DRO    &78.7\scriptsize$\pm 3.7$ & 72.7\scriptsize$\pm 2.7$
&77.4\scriptsize$\pm 0.2$ & 71.3\scriptsize$\pm 0.8$
&83.2\scriptsize$\pm 0.9$ & 73.9\scriptsize$\pm 1.2$
&84.5\scriptsize$\pm 0.3$ & 67.0\scriptsize$\pm 0.4$
&79.8\scriptsize$\pm 1.2$ & 74.2\scriptsize$\pm 2.1$
&77.1\scriptsize$\pm 0.4$ & 68.5\scriptsize$\pm 0.7$
&91.5\scriptsize$\pm 0.1$ & 79.3\scriptsize$\pm 0.2$
&63.8\scriptsize$\pm 0.9$ & 59.1\scriptsize$\pm 0.4$
&88.6\scriptsize$\pm 0.2$ & 58.4\scriptsize$\pm 3.7$
&93.3\scriptsize$\pm 0.3$ & 81.3\scriptsize$\pm 1.3$
\\\midrule
\multirow{6}{*}{\begin{tabular}[c]{@{}l@{}}Imbalanced\\ Learning\\ \& Fairness\\ Methods \\(base: XGB)\end{tabular}} &SUBY    &78.2\scriptsize$\pm 3.2$ & 64.9\scriptsize$\pm 1.4$
&72.0\scriptsize$\pm 1.0$ & 67.8\scriptsize$\pm 2.7$
&80.8\scriptsize$\pm 2.2$ & 73.4\scriptsize$\pm 1.0$
&73.2\scriptsize$\pm 2.3$ & 67.2\scriptsize$\pm 1.6$
&85.5\scriptsize$\pm 0.3$ & 64.6\scriptsize$\pm 0.6$
&72.0\scriptsize$\pm 1.7$ & 67.6\scriptsize$\pm 1.7$
&91.8\scriptsize$\pm 0.1$ & 80.6\scriptsize$\pm 0.1$
&62.3\scriptsize$\pm 2.1$ & 60.3\scriptsize$\pm 1.7$
&87.5\scriptsize$\pm 0.4$ & 60.1\scriptsize$\pm 1.5$
&88.2\scriptsize$\pm 2.0$ & 72.3\scriptsize$\pm 2.9$
\\
&RWY    &80.2\scriptsize$\pm 1.2$ & 66.7\scriptsize$\pm 0.5$
&71.5\scriptsize$\pm 2.6$ & 66.4\scriptsize$\pm 2.5$
&80.2\scriptsize$\pm 4.8$ & 73.5\scriptsize$\pm 1.9$
&81.1\scriptsize$\pm 0.7$ & 70.4\scriptsize$\pm 0.3$
&84.3\scriptsize$\pm 1.8$ & 65.4\scriptsize$\pm 0.9$
&73.7\scriptsize$\pm 0.7$ & 68.7\scriptsize$\pm 1.5$
&91.8\scriptsize$\pm 0.1$ & 80.6\scriptsize$\pm 0.1$
&62.2\scriptsize$\pm 1.7$ & 59.3\scriptsize$\pm 1.6$
&86.9\scriptsize$\pm 0.4$ & 58.0\scriptsize$\pm 1.8$
&94.0\scriptsize$\pm 0.2$ & 82.5\scriptsize$\pm 1.1$
\\
&SUBG    &77.4\scriptsize$\pm 6.0$ & 74.4\scriptsize$\pm 7.0$
&71.9\scriptsize$\pm 3.6$ & 66.3\scriptsize$\pm 2.5$
&-&-
&83.9\scriptsize$\pm 1.0$ & 69.5\scriptsize$\pm 0.5$
&-&-
&75.7\scriptsize$\pm 2.8$ & 68.2\scriptsize$\pm 3.3$
&91.8\scriptsize$\pm 0.1$ & 80.6\scriptsize$\pm 0.1$
&54.7\scriptsize$\pm 2.8$ & 53.5\scriptsize$\pm 1.8$
&-&-
&94.0\scriptsize$\pm 0.2$ & 82.9\scriptsize$\pm 0.9$
\\
&RWG    &80.5\scriptsize$\pm 0.9$ & 70.8\scriptsize$\pm 0.6$
&72.9\scriptsize$\pm 2.4$ & 67.7\scriptsize$\pm 0.9$
&-&-
&82.5\scriptsize$\pm 2.1$ & 68.7\scriptsize$\pm 1.3$
&-&-
&78.2\scriptsize$\pm 1.4$ & 70.4\scriptsize$\pm 1.2$
&91.8\scriptsize$\pm 0.1$ & 80.6\scriptsize$\pm 0.1$
&62.5\scriptsize$\pm 1.6$ & 59.2\scriptsize$\pm 1.3$
&-&-
&94.5\scriptsize$\pm 0.3$ & 83.2\scriptsize$\pm 0.6$
\\
&In-processing    &77.9\scriptsize$\pm 3.6$ & 71.2\scriptsize$\pm 1.5$
&70.1\scriptsize$\pm 0.7$ & 65.0\scriptsize$\pm 1.5$
&-&-
&83.7\scriptsize$\pm 0.9$ & 69.5\scriptsize$\pm 0.7$
&-&-
&78.9\scriptsize$\pm 1.6$ & 71.3\scriptsize$\pm 1.1$
&92.3\scriptsize$\pm 0.0$ & 80.4\scriptsize$\pm 0.1$
&64.7\scriptsize$\pm 1.8$ & 61.3\scriptsize$\pm 1.6$
&-&-
&94.4\scriptsize$\pm 0.3$ & 84.0\scriptsize$\pm 0.6$
\\
&Post-processing    &74.5\scriptsize$\pm 4.4$ & 76.3\scriptsize$\pm 6.5$
&74.0\scriptsize$\pm 2.3$ & 68.4\scriptsize$\pm 1.9$
&-&-
&82.5\scriptsize$\pm 1.6$ & 68.3\scriptsize$\pm 1.3$
&-&-
&79.3\scriptsize$\pm 0.3$ & 71.4\scriptsize$\pm 0.2$
&91.8\scriptsize$\pm 0.0$ & 80.0\scriptsize$\pm 0.1$
&62.1\scriptsize$\pm 1.4$ & 58.9\scriptsize$\pm 1.0$
&-&-
&87.7\scriptsize$\pm 0.1$ & 83.7\scriptsize$\pm 0.7$
\\\bottomrule
\end{tabular}}
\end{table}
\end{landscape}
\begin{landscape}
\begin{table}[htbp]
\caption{Results (Macro F1-Score) of 10 selected pairs in \Cref{subsec:benchmark-empirical-result}, where we run each method with its top-10 configurations (according to the target F1-score) and report its mean accuracy and standard deviation. }
\vspace{5pt}
\label{table:selected_results_f1}
\resizebox{21cm}{!}{
\begin{tabular}{@{}llcccccccccccccccccccc@{}}
\toprule\toprule
\multicolumn{2}{l}{\large \textbf{Dataset}}                                                                   & \multicolumn{2}{c}{\large \texttt{ACS Income}}                                                                    & \multicolumn{2}{c}{\large \texttt{ACS Mobility}}                                                                  & \multicolumn{2}{c}{\large \texttt{US Taxi}}                 & \multicolumn{2}{c}{\large \texttt{ACS Pub.Cov}}                                                                   & \multicolumn{2}{c}{\large \texttt{US Accident}}                                                                   & \multicolumn{2}{c}{\large \texttt{ACS Time}}                                                                      & \multicolumn{2}{c}{\large \texttt{Sub-Sampling}} & \multicolumn{2}{c}{\large \texttt{diabetes}} &  \multicolumn{2}{c}{\large \texttt{assistments}} & \multicolumn{2}{c}{\large \texttt{college}}\\ 
\multicolumn{2}{l}{\large \textbf{Shift Pattern}}                                                            & \multicolumn{2}{c}{$Y|X$ dominates}                                                                              & \multicolumn{2}{c}{ $Y|X$ dominates}                                                                              & \multicolumn{2}{c}{ $Y|X$ dominates}                       & \multicolumn{2}{c}{ $Y|X$ more}                                                                            & \multicolumn{2}{c}{$Y|X$ more}                                                                              & \multicolumn{2}{c}{$X$ more}                                                                            & \multicolumn{2}{c}{$X$ dominates}  & \multicolumn{2}{c}{$Y|X$ dominates} & \multicolumn{2}{c}{$Y|X$ more} & \multicolumn{2}{c}{$Y|X$ more} \\
\multicolumn{2}{l}{\large Source $\rightarrow$ Target Pair}                                                       & \multicolumn{2}{c}{CA$\rightarrow$PR}                                                                        & \multicolumn{2}{c}{MS$\rightarrow$HI}                                                                        & \multicolumn{2}{c}{NYC$\rightarrow$BOG}                & \multicolumn{2}{c}{NE$\rightarrow$LA}                                                                        & \multicolumn{2}{c}{CA$\rightarrow$OR}                                                                        & \multicolumn{2}{c}{2010$\rightarrow$2017}                                                                    & \multicolumn{2}{c}{Young$\rightarrow$Old}  & \multicolumn{2}{c}{White$\rightarrow$Others} & \multicolumn{2}{c}{700$\rightarrow$1} & \multicolumn{2}{c}{Normal $\rightarrow$ Special} \\
\multicolumn{2}{l}{}                                                                          & $i.d.$                                         & $o.o.d$                                         & $i.d.$                                         & $o.o.d$                                         & $i.d.$              & $o.o.d$              & $i.d.$                                         & $o.o.d$                                         & $i.d.$                                         & $o.o.d$                                         & $i.d.$                                         & $o.o.d$                                         & $i.d.$                                         & $o.o.d$  & $i.d.$                                         & $o.o.d$ &  $i.d.$                                         & $o.o.d$ &  $i.d.$                                         & $o.o.d$\\\midrule
\multirow{6}{*}{\begin{tabular}[c]{@{}l@{}}Basic \\ Methods\end{tabular}} &LR    &79.7\scriptsize$\pm 0.6$ & 61.3\scriptsize$\pm 0.4$
&53.7\scriptsize$\pm 1.0$ & 53.8\scriptsize$\pm 0.3$
&80.9\scriptsize$\pm 0.3$ & 74.8\scriptsize$\pm 0.2$
&65.0\scriptsize$\pm 0.2$ & 60.2\scriptsize$\pm 0.9$
&73.1\scriptsize$\pm 0.2$ & 59.4\scriptsize$\pm 1.2$
&46.8\scriptsize$\pm 0.0$ & 43.0\scriptsize$\pm 0.0$
&79.6\scriptsize$\pm 0.1$ & 77.7\scriptsize$\pm 0.0$
&61.0\scriptsize$\pm 0.1$ & 58.0\scriptsize$\pm 0.2$
&83.6\scriptsize$\pm 0.0$ & 41.1\scriptsize$\pm 0.6$
&84.3\scriptsize$\pm 0.0$ & 76.2\scriptsize$\pm 0.2$
\\
&SVM    &80.0\scriptsize$\pm 0.0$ & 60.7\scriptsize$\pm 0.0$
&59.7\scriptsize$\pm 0.7$ & 52.9\scriptsize$\pm 0.5$
&79.6\scriptsize$\pm 0.1$ & 75.6\scriptsize$\pm 0.1$
&64.7\scriptsize$\pm 0.4$ & 59.6\scriptsize$\pm 0.3$
&73.4\scriptsize$\pm 0.1$ & 60.9\scriptsize$\pm 0.1$
&65.2\scriptsize$\pm 0.2$ & 59.8\scriptsize$\pm 0.2$
&78.6\scriptsize$\pm 0.1$ & 77.4\scriptsize$\pm 0.2$
&60.7\scriptsize$\pm 0.0$ & 57.8\scriptsize$\pm 0.0$
&83.5\scriptsize$\pm 0.0$ & 35.8\scriptsize$\pm 0.0$
&84.0\scriptsize$\pm 0.0$ & 76.2\scriptsize$\pm 0.2$
\\
&Kernel-SVM    &69.8\scriptsize$\pm 7.0$ & 66.7\scriptsize$\pm 2.6$
&53.7\scriptsize$\pm 3.1$ & 51.7\scriptsize$\pm 1.3$
&63.6\scriptsize$\pm 10.3$ & 69.2\scriptsize$\pm 1.9$
&54.3\scriptsize$\pm 2.3$ & 57.9\scriptsize$\pm 1.4$
&76.5\scriptsize$\pm 2.3$ & 61.0\scriptsize$\pm 1.2$
&53.2\scriptsize$\pm 2.1$ & 54.4\scriptsize$\pm 1.8$
&78.2\scriptsize$\pm 1.1$ & 75.4\scriptsize$\pm 0.7$
&60.8\scriptsize$\pm 1.1$ & 58.9\scriptsize$\pm 0.8$
&66.0\scriptsize$\pm 6.7$ & 60.0\scriptsize$\pm 1.6$
&74.7\scriptsize$\pm 3.3$ & 69.4\scriptsize$\pm 3.2$
\\
&NN2    &79.8\scriptsize$\pm 0.3$ & 61.4\scriptsize$\pm 1.5$
&58.7\scriptsize$\pm 4.7$ & 54.4\scriptsize$\pm 2.8$
&75.5\scriptsize$\pm 11.8$ & 72.1\scriptsize$\pm 1.7$
&63.6\scriptsize$\pm 3.3$ & 62.3\scriptsize$\pm 3.6$
&80.5\scriptsize$\pm 0.7$ & 62.1\scriptsize$\pm 0.3$
&66.2\scriptsize$\pm 2.1$ & 62.7\scriptsize$\pm 3.2$
&79.0\scriptsize$\pm 0.6$ & 77.6\scriptsize$\pm 0.3$
&60.2\scriptsize$\pm 0.7$ & 58.5\scriptsize$\pm 0.7$
&83.9\scriptsize$\pm 0.9$ & 57.8\scriptsize$\pm 4.0$
&84.0\scriptsize$\pm 0.3$ & 76.3\scriptsize$\pm 0.8$
\\
&NN3    &79.4\scriptsize$\pm 0.6$ & 60.5\scriptsize$\pm 2.7$
&51.6\scriptsize$\pm 2.5$ & 48.3\scriptsize$\pm 0.8$
&82.1\scriptsize$\pm 0.7$ & 72.9\scriptsize$\pm 1.0$
&62.3\scriptsize$\pm 3.2$ & 59.4\scriptsize$\pm 2.9$
&80.4\scriptsize$\pm 1.5$ & 62.1\scriptsize$\pm 0.6$
&66.0\scriptsize$\pm 2.0$ & 61.3\scriptsize$\pm 2.5$
&78.6\scriptsize$\pm 0.6$ & 77.3\scriptsize$\pm 0.5$
&60.5\scriptsize$\pm 1.0$ & 58.6\scriptsize$\pm 0.4$
&84.2\scriptsize$\pm 0.3$ & 58.9\scriptsize$\pm 3.3$
&84.0\scriptsize$\pm 0.5$ & 77.6\scriptsize$\pm 0.8$
\\
&NN4    &79.2\scriptsize$\pm 2.5$ & 60.5\scriptsize$\pm 1.2$
&50.9\scriptsize$\pm 1.6$ & 48.4\scriptsize$\pm 0.5$
&81.0\scriptsize$\pm 1.3$ & 72.7\scriptsize$\pm 2.2$
&56.7\scriptsize$\pm 4.3$ & 54.2\scriptsize$\pm 4.5$
&80.0\scriptsize$\pm 1.9$ & 62.2\scriptsize$\pm 0.5$
&65.1\scriptsize$\pm 1.7$ & 61.4\scriptsize$\pm 1.9$
&77.4\scriptsize$\pm 0.6$ & 76.2\scriptsize$\pm 0.7$
&60.6\scriptsize$\pm 1.5$ & 58.6\scriptsize$\pm 1.5$
&83.2\scriptsize$\pm 2.4$ & 55.7\scriptsize$\pm 3.5$
&84.3\scriptsize$\pm 0.6$ & 77.3\scriptsize$\pm 0.7$
\\\midrule
\multirow{4}{*}{\begin{tabular}[c]{@{}l@{}}Tree-based \\ Ensemble \\ Methods\end{tabular}}   &RF    &73.3\scriptsize$\pm 0.2$ & 62.0\scriptsize$\pm 0.6$
&63.3\scriptsize$\pm 1.5$ & 51.0\scriptsize$\pm 0.8$
&83.0\scriptsize$\pm 0.1$ & 73.1\scriptsize$\pm 0.2$
&70.7\scriptsize$\pm 0.3$ & 61.9\scriptsize$\pm 0.4$
&83.0\scriptsize$\pm 0.5$ & 59.8\scriptsize$\pm 0.3$
&69.3\scriptsize$\pm 0.7$ & 64.6\scriptsize$\pm 1.0$
&77.1\scriptsize$\pm 0.5$ & 78.1\scriptsize$\pm 0.2$
&61.3\scriptsize$\pm 0.7$ & 60.2\scriptsize$\pm 0.7$
&85.1\scriptsize$\pm 0.3$ & 63.9\scriptsize$\pm 1.5$
&86.0\scriptsize$\pm 0.2$ & 79.7\scriptsize$\pm 0.5$
\\
&XGB    &77.9\scriptsize$\pm 1.0$ & 57.9\scriptsize$\pm 1.0$
&61.8\scriptsize$\pm 1.0$ & 52.5\scriptsize$\pm 0.5$
&81.9\scriptsize$\pm 3.0$ & 71.3\scriptsize$\pm 0.8$
&71.9\scriptsize$\pm 1.6$ & 65.6\scriptsize$\pm 0.5$
&78.4\scriptsize$\pm 0.4$ & 60.4\scriptsize$\pm 0.4$
&69.3\scriptsize$\pm 1.1$ & 64.7\scriptsize$\pm 0.9$
&79.6\scriptsize$\pm 0.2$ & 78.4\scriptsize$\pm 0.1$
&60.2\scriptsize$\pm 0.8$ & 57.7\scriptsize$\pm 0.5$
&86.0\scriptsize$\pm 0.4$ & 60.4\scriptsize$\pm 0.6$
&87.7\scriptsize$\pm 0.1$ & 81.5\scriptsize$\pm 0.4$
\\
&GBM    &79.1\scriptsize$\pm 2.3$ & 61.3\scriptsize$\pm 1.7$
&56.9\scriptsize$\pm 5.9$ & 52.5\scriptsize$\pm 1.5$
&82.4\scriptsize$\pm 1.2$ & 72.2\scriptsize$\pm 0.7$
&72.0\scriptsize$\pm 1.1$ & 66.0\scriptsize$\pm 0.8$
&81.4\scriptsize$\pm 4.1$ & 59.8\scriptsize$\pm 1.1$
&69.4\scriptsize$\pm 1.3$ & 65.3\scriptsize$\pm 1.3$
&79.9\scriptsize$\pm 0.2$ & 79.1\scriptsize$\pm 0.1$
&61.1\scriptsize$\pm 0.0$ & 60.5\scriptsize$\pm 0.0$
&85.8\scriptsize$\pm 0.4$ & 61.8\scriptsize$\pm 1.5$
&87.1\scriptsize$\pm 0.3$ & 80.8\scriptsize$\pm 0.6$
\\
&LGBM    &78.1\scriptsize$\pm 1.3$ & 62.4\scriptsize$\pm 2.7$
&63.7\scriptsize$\pm 1.0$ & 53.2\scriptsize$\pm 0.4$
&79.5\scriptsize$\pm 2.3$ & 72.6\scriptsize$\pm 3.0$
&71.6\scriptsize$\pm 0.7$ & 65.4\scriptsize$\pm 0.2$
&76.2\scriptsize$\pm 3.7$ & 59.7\scriptsize$\pm 1.2$
&69.3\scriptsize$\pm 1.0$ & 65.1\scriptsize$\pm 0.7$
&79.5\scriptsize$\pm 0.2$ & 77.8\scriptsize$\pm 0.1$
&60.4\scriptsize$\pm 1.5$ & 58.1\scriptsize$\pm 1.1$
&84.9\scriptsize$\pm 0.4$ & 62.2\scriptsize$\pm 2.1$
&87.7\scriptsize$\pm 0.1$ & 80.6\scriptsize$\pm 0.8$
\\\midrule
\multirow{13}{*}{\begin{tabular}[c]{@{}l@{}}Linear-DRO \\  Methods \\ (base: SVM)\end{tabular}}   &KL-DRO    &79.3\scriptsize$\pm 0.1$ & 61.2\scriptsize$\pm 0.1$
&61.0\scriptsize$\pm 0.1$ & 53.2\scriptsize$\pm 0.1$
&79.9\scriptsize$\pm 0.7$ & 74.2\scriptsize$\pm 0.4$
&65.3\scriptsize$\pm 0.7$ & 60.8\scriptsize$\pm 0.9$
&74.2\scriptsize$\pm 0.0$ & 61.8\scriptsize$\pm 0.0$
&65.2\scriptsize$\pm 0.2$ & 60.2\scriptsize$\pm 0.4$
&77.6\scriptsize$\pm 0.3$ & 77.4\scriptsize$\pm 0.1$
&60.2\scriptsize$\pm 0.7$ & 57.3\scriptsize$\pm 0.7$
&83.6\scriptsize$\pm 0.0$ & 38.4\scriptsize$\pm 0.0$
&84.1\scriptsize$\pm 0.1$ & 76.3\scriptsize$\pm 0.2$
\\
&CVaR-DRO    &77.8\scriptsize$\pm 2.7$ & 60.4\scriptsize$\pm 1.1$
&57.5\scriptsize$\pm 0.0$ & 52.4\scriptsize$\pm 0.0$
&80.4\scriptsize$\pm 2.1$ & 75.1\scriptsize$\pm 1.9$
&66.5\scriptsize$\pm 2.4$ & 63.9\scriptsize$\pm 1.2$
&74.5\scriptsize$\pm 0.2$ & 61.8\scriptsize$\pm 0.2$
&66.1\scriptsize$\pm 0.7$ & 61.5\scriptsize$\pm 1.0$
&79.4\scriptsize$\pm 0.0$ & 78.6\scriptsize$\pm 0.1$
&59.0\scriptsize$\pm 0.5$ & 56.1\scriptsize$\pm 0.6$
&83.6\scriptsize$\pm 0.2$ & 42.7\scriptsize$\pm 1.6$
&83.8\scriptsize$\pm 0.6$ & 75.7\scriptsize$\pm 0.9$
\\
&$\chi^2$-DRO    &79.8\scriptsize$\pm 0.1$ & 61.1\scriptsize$\pm 0.3$
&58.0\scriptsize$\pm 1.1$ & 52.6\scriptsize$\pm 0.3$
&80.5\scriptsize$\pm 0.2$ & 76.0\scriptsize$\pm 0.1$
&66.8\scriptsize$\pm 0.5$ & 63.8\scriptsize$\pm 0.2$
&74.1\scriptsize$\pm 0.1$ & 61.7\scriptsize$\pm 0.1$
&65.3\scriptsize$\pm 0.1$ & 60.2\scriptsize$\pm 0.1$
&78.5\scriptsize$\pm 0.1$ & 77.0\scriptsize$\pm 0.0$
&60.3\scriptsize$\pm 0.6$ & 57.1\scriptsize$\pm 0.8$
&83.5\scriptsize$\pm 0.0$ & 39.9\scriptsize$\pm 0.3$
&84.2\scriptsize$\pm 0.0$ & 76.5\scriptsize$\pm 0.1$
\\
&TV-DRO    &77.3\scriptsize$\pm 2.4$ & 60.2\scriptsize$\pm 0.7$
&58.0\scriptsize$\pm 1.2$ & 52.6\scriptsize$\pm 0.5$
&75.9\scriptsize$\pm 4.6$ & 72.8\scriptsize$\pm 2.9$
&62.9\scriptsize$\pm 1.6$ & 57.8\scriptsize$\pm 1.8$
&74.1\scriptsize$\pm 0.6$ & 61.5\scriptsize$\pm 0.6$
&66.3\scriptsize$\pm 0.7$ & 62.2\scriptsize$\pm 1.3$
&77.1\scriptsize$\pm 0.2$ & 77.4\scriptsize$\pm 0.1$
&58.1\scriptsize$\pm 0.9$ & 55.4\scriptsize$\pm 1.0$
&83.3\scriptsize$\pm 0.2$ & 34.2\scriptsize$\pm 3.4$
&84.1\scriptsize$\pm 0.1$ & 76.2\scriptsize$\pm 0.1$
\\
&Wasserstein-DRO    &74.8\scriptsize$\pm 5.1$ & 60.6\scriptsize$\pm 0.8$
&57.1\scriptsize$\pm 0.1$ & 51.9\scriptsize$\pm 0.0$
&80.6\scriptsize$\pm 0.0$ & 75.3\scriptsize$\pm 0.1$
&63.5\scriptsize$\pm 1.3$ & 55.5\scriptsize$\pm 1.7$
&73.0\scriptsize$\pm 0.9$ & 59.6\scriptsize$\pm 1.7$
&62.9\scriptsize$\pm 0.0$ & 58.8\scriptsize$\pm 0.0$
&77.5\scriptsize$\pm 0.1$ & 75.3\scriptsize$\pm 0.0$
&56.9\scriptsize$\pm 0.9$ & 55.6\scriptsize$\pm 0.4$
&83.4\scriptsize$\pm 0.0$ & 33.1\scriptsize$\pm 0.0$
&84.0\scriptsize$\pm 0.2$ & 76.3\scriptsize$\pm 0.4$
\\
&Aug. Wass.-DRO    &77.3\scriptsize$\pm 4.8$ & 60.9\scriptsize$\pm 0.5$
&55.2\scriptsize$\pm 4.0$ & 51.7\scriptsize$\pm 0.4$
&76.5\scriptsize$\pm 1.8$ & 71.2\scriptsize$\pm 1.9$
&65.0\scriptsize$\pm 3.2$ & 62.7\scriptsize$\pm 2.8$
&75.4\scriptsize$\pm 0.5$ & 62.6\scriptsize$\pm 0.5$
&62.0\scriptsize$\pm 3.3$ & 58.6\scriptsize$\pm 1.7$
&78.5\scriptsize$\pm 0.3$ & 78.1\scriptsize$\pm 0.1$
&57.4\scriptsize$\pm 1.5$ & 55.2\scriptsize$\pm 1.6$
&61.4\scriptsize$\pm 8.1$ & 53.7\scriptsize$\pm 1.2$
&83.9\scriptsize$\pm 0.2$ & 76.4\scriptsize$\pm 0.4$
\\
&Satis. Wass.-DRO    &72.2\scriptsize$\pm 3.2$ & 59.9\scriptsize$\pm 0.4$
&43.1\scriptsize$\pm 0.0$ & 44.0\scriptsize$\pm 0.0$
&71.8\scriptsize$\pm 2.6$ & 68.5\scriptsize$\pm 2.9$
&44.5\scriptsize$\pm 0.0$ & 37.3\scriptsize$\pm 0.0$
&70.0\scriptsize$\pm 0.0$ & 56.3\scriptsize$\pm 0.0$
&41.7\scriptsize$\pm 0.0$ & 37.8\scriptsize$\pm 0.0$
&75.3\scriptsize$\pm 1.2$ & 74.7\scriptsize$\pm 1.7$
&36.9\scriptsize$\pm 0.0$ & 33.6\scriptsize$\pm 0.0$
&82.6\scriptsize$\pm 0.0$ & 32.9\scriptsize$\pm 0.0$
&81.7\scriptsize$\pm 0.9$ & 75.9\scriptsize$\pm 0.4$
\\
&Sinkhorn-DRO    &69.6\scriptsize$\pm 2.9$ & 68.9\scriptsize$\pm 0.8$
&55.9\scriptsize$\pm 0.8$ & 53.1\scriptsize$\pm 0.6$
&69.2\scriptsize$\pm 7.0$ & 74.9\scriptsize$\pm 1.3$
&66.9\scriptsize$\pm 2.2$ & 63.5\scriptsize$\pm 2.4$
&74.2\scriptsize$\pm 0.6$ & 58.5\scriptsize$\pm 0.6$
&53.4\scriptsize$\pm 8.6$ & 56.2\scriptsize$\pm 5.5$
&79.3\scriptsize$\pm 0.2$ & 78.1\scriptsize$\pm 0.1$
&53.5\scriptsize$\pm 1.6$ & 53.5\scriptsize$\pm 1.7$
&82.4\scriptsize$\pm 1.2$ & 33.9\scriptsize$\pm 2.9$
&81.6\scriptsize$\pm 2.2$ & 75.2\scriptsize$\pm 1.6$
\\
&Unified-DRO($L_2$)    &79.8\scriptsize$\pm 0.0$ & 60.5\scriptsize$\pm 0.0$
&57.1\scriptsize$\pm 0.1$ & 51.9\scriptsize$\pm 0.0$
&79.9\scriptsize$\pm 0.0$ & 75.1\scriptsize$\pm 0.1$
&63.1\scriptsize$\pm 0.1$ & 55.6\scriptsize$\pm 0.0$
&71.8\scriptsize$\pm 2.1$ & 58.3\scriptsize$\pm 2.3$
&62.9\scriptsize$\pm 0.0$ & 58.8\scriptsize$\pm 0.0$
&78.7\scriptsize$\pm 0.0$ & 77.5\scriptsize$\pm 0.0$
&52.1\scriptsize$\pm 0.0$ & 52.5\scriptsize$\pm 0.0$
&83.4\scriptsize$\pm 0.0$ & 33.1\scriptsize$\pm 0.0$
&84.0\scriptsize$\pm 0.0$ & 76.4\scriptsize$\pm 0.0$
\\
&Unified-DRO($L_\text{inf}$)    &79.1\scriptsize$\pm 1.0$ & 60.7\scriptsize$\pm 0.1$
&57.1\scriptsize$\pm 0.1$ & 51.9\scriptsize$\pm 0.0$
&80.1\scriptsize$\pm 0.0$ & 75.4\scriptsize$\pm 0.1$
&64.4\scriptsize$\pm 0.0$ & 56.3\scriptsize$\pm 0.0$
&71.3\scriptsize$\pm 1.7$ & 57.3\scriptsize$\pm 1.0$
&62.9\scriptsize$\pm 0.0$ & 58.8\scriptsize$\pm 0.0$
&78.9\scriptsize$\pm 0.0$ & 77.3\scriptsize$\pm 0.0$
&55.2\scriptsize$\pm 0.0$ & 54.8\scriptsize$\pm 0.0$
&83.4\scriptsize$\pm 0.0$ & 33.1\scriptsize$\pm 0.0$
&84.1\scriptsize$\pm 0.0$ & 76.3\scriptsize$\pm 0.1$
\\
&Marginal-DRO    &79.9\scriptsize$\pm 0.1$ & 61.0\scriptsize$\pm 0.1$
&59.2\scriptsize$\pm 1.7$ & 53.0\scriptsize$\pm 0.6$
&79.9\scriptsize$\pm 0.1$ & 76.0\scriptsize$\pm 0.1$
&67.9\scriptsize$\pm 0.6$ & 64.0\scriptsize$\pm 1.0$
&74.4\scriptsize$\pm 0.2$ & 61.9\scriptsize$\pm 0.2$
&65.7\scriptsize$\pm 0.0$ & 60.6\scriptsize$\pm 0.0$
&78.1\scriptsize$\pm 0.2$ & 76.7\scriptsize$\pm 0.3$
&61.8\scriptsize$\pm 0.1$ & 59.4\scriptsize$\pm 0.5$
&83.9\scriptsize$\pm 0.0$ & 53.8\scriptsize$\pm 0.0$
&84.3\scriptsize$\pm 0.0$ & 76.4\scriptsize$\pm 0.2$
\\
&Conditional-DRO    &79.6\scriptsize$\pm 0.3$ & 58.6\scriptsize$\pm 1.8$
&59.3\scriptsize$\pm 1.3$ & 53.2\scriptsize$\pm 0.5$
&80.0\scriptsize$\pm 0.5$ & 76.1\scriptsize$\pm 0.1$
&67.6\scriptsize$\pm 1.2$ & 65.3\scriptsize$\pm 2.0$
&79.0\scriptsize$\pm 0.9$ & 62.4\scriptsize$\pm 0.9$
&66.5\scriptsize$\pm 0.5$ & 62.2\scriptsize$\pm 1.3$
&79.8\scriptsize$\pm 0.1$ & 77.4\scriptsize$\pm 0.0$
&60.9\scriptsize$\pm 1.0$ & 58.6\scriptsize$\pm 1.3$
&83.4\scriptsize$\pm 0.0$ & 33.4\scriptsize$\pm 0.0$
&84.1\scriptsize$\pm 0.0$ & 76.3\scriptsize$\pm 0.2$
\\
&Holistic-DRO    &73.2\scriptsize$\pm 2.3$ & 65.3\scriptsize$\pm 0.3$
&58.1\scriptsize$\pm 2.0$ & 52.0\scriptsize$\pm 0.6$
&73.5\scriptsize$\pm 1.3$ & 70.2\scriptsize$\pm 0.7$
&62.4\scriptsize$\pm 0.6$ & 58.6\scriptsize$\pm 0.9$
&70.4\scriptsize$\pm 1.1$ & 56.5\scriptsize$\pm 0.7$
&64.2\scriptsize$\pm 0.2$ & 59.1\scriptsize$\pm 0.2$
&74.8\scriptsize$\pm 0.3$ & 73.8\scriptsize$\pm 0.1$
&59.1\scriptsize$\pm 0.2$ & 56.1\scriptsize$\pm 0.3$
&83.4\scriptsize$\pm 0.0$ & 33.2\scriptsize$\pm 0.1$
&79.2\scriptsize$\pm 1.5$ & 72.7\scriptsize$\pm 1.0$
\\\midrule
\multirow{8}{*}{\begin{tabular}[c]{@{}l@{}}NN-DRO\\ Methods\\ (base: NN)\end{tabular}}   &NN2-CVaR-DRO    &80.1\scriptsize$\pm 0.1$ & 58.9\scriptsize$\pm 2.2$
&53.3\scriptsize$\pm 1.3$ & 51.1\scriptsize$\pm 0.7$
&75.1\scriptsize$\pm 8.0$ & 74.1\scriptsize$\pm 0.9$
&61.3\scriptsize$\pm 5.5$ & 61.5\scriptsize$\pm 4.1$
&66.3\scriptsize$\pm 7.9$ & 57.3\scriptsize$\pm 4.8$
&65.9\scriptsize$\pm 1.3$ & 61.9\scriptsize$\pm 2.1$
&79.2\scriptsize$\pm 0.4$ & 78.5\scriptsize$\pm 0.2$
&60.5\scriptsize$\pm 0.7$ & 58.9\scriptsize$\pm 0.9$
&72.8\scriptsize$\pm 15.9$ & 59.6\scriptsize$\pm 4.5$
&85.0\scriptsize$\pm 0.7$ & 78.0\scriptsize$\pm 1.0$
\\
&NN2-$\chi^2$-DRO    &78.3\scriptsize$\pm 5.3$ & 61.6\scriptsize$\pm 2.4$
&55.9\scriptsize$\pm 5.2$ & 51.9\scriptsize$\pm 1.7$
&75.1\scriptsize$\pm 8.6$ & 74.1\scriptsize$\pm 1.5$
&64.2\scriptsize$\pm 1.9$ & 64.0\scriptsize$\pm 2.1$
&81.0\scriptsize$\pm 0.2$ & 62.5\scriptsize$\pm 0.3$
&66.3\scriptsize$\pm 1.5$ & 61.8\scriptsize$\pm 2.3$
&79.7\scriptsize$\pm 0.4$ & 78.2\scriptsize$\pm 0.2$
&60.5\scriptsize$\pm 1.4$ & 58.5\scriptsize$\pm 1.2$
&83.3\scriptsize$\pm 1.6$ & 61.2\scriptsize$\pm 3.6$
&84.5\scriptsize$\pm 0.7$ & 77.5\scriptsize$\pm 1.2$
\\
&NN2-CVaR-DORO    &73.0\scriptsize$\pm 8.1$ & 63.3\scriptsize$\pm 2.4$
&53.0\scriptsize$\pm 4.4$ & 52.0\scriptsize$\pm 1.4$
&77.6\scriptsize$\pm 1.9$ & 73.4\scriptsize$\pm 1.0$
&57.6\scriptsize$\pm 7.7$ & 59.2\scriptsize$\pm 4.3$
&79.6\scriptsize$\pm 1.3$ & 60.9\scriptsize$\pm 0.4$
&60.5\scriptsize$\pm 4.0$ & 57.9\scriptsize$\pm 2.4$
&79.2\scriptsize$\pm 0.6$ & 76.9\scriptsize$\pm 0.4$
&61.4\scriptsize$\pm 0.5$ & 59.6\scriptsize$\pm 0.6$
&83.4\scriptsize$\pm 1.0$ & 57.0\scriptsize$\pm 4.0$
&84.3\scriptsize$\pm 0.4$ & 76.9\scriptsize$\pm 0.9$
\\
&NN2-$\chi^2$-DORO    &69.3\scriptsize$\pm 6.9$ & 62.2\scriptsize$\pm 3.0$
&51.0\scriptsize$\pm 1.7$ & 51.3\scriptsize$\pm 0.6$
&68.6\scriptsize$\pm 10.9$ & 70.2\scriptsize$\pm 4.0$
&51.6\scriptsize$\pm 3.1$ & 54.7\scriptsize$\pm 3.7$
&80.3\scriptsize$\pm 0.8$ & 62.1\scriptsize$\pm 0.3$
&56.6\scriptsize$\pm 4.0$ & 55.1\scriptsize$\pm 3.8$
&79.0\scriptsize$\pm 0.4$ & 77.1\scriptsize$\pm 0.3$
&59.9\scriptsize$\pm 0.0$ & 59.3\scriptsize$\pm 0.0$
&65.6\scriptsize$\pm 17.3$ & 58.2\scriptsize$\pm 6.4$
&82.1\scriptsize$\pm 2.0$ & 76.1\scriptsize$\pm 2.1$
\\
&NN3-CVaR-DRO    &77.1\scriptsize$\pm 8.9$ & 59.7\scriptsize$\pm 2.0$
&53.5\scriptsize$\pm 3.6$ & 51.6\scriptsize$\pm 1.3$
&74.9\scriptsize$\pm 12.1$ & 73.0\scriptsize$\pm 1.6$
&61.0\scriptsize$\pm 2.7$ & 60.6\scriptsize$\pm 2.8$
&77.6\scriptsize$\pm 6.4$ & 60.6\scriptsize$\pm 3.3$
&65.3\scriptsize$\pm 2.2$ & 60.6\scriptsize$\pm 2.8$
&79.5\scriptsize$\pm 0.3$ & 78.1\scriptsize$\pm 0.3$
&60.4\scriptsize$\pm 1.5$ & 58.7\scriptsize$\pm 1.2$
&78.2\scriptsize$\pm 12.7$ & 57.6\scriptsize$\pm 3.6$
&84.4\scriptsize$\pm 0.8$ & 77.5\scriptsize$\pm 0.8$
\\
&NN3-$\chi^2$-DRO    &79.4\scriptsize$\pm 2.2$ & 61.9\scriptsize$\pm 2.7$
&56.2\scriptsize$\pm 3.4$ & 51.2\scriptsize$\pm 1.3$
&74.5\scriptsize$\pm 10.0$ & 73.0\scriptsize$\pm 1.2$
&63.2\scriptsize$\pm 2.1$ & 61.5\scriptsize$\pm 2.3$
&81.4\scriptsize$\pm 0.3$ & 62.1\scriptsize$\pm 0.2$
&67.2\scriptsize$\pm 1.7$ & 63.4\scriptsize$\pm 2.7$
&78.9\scriptsize$\pm 0.4$ & 77.7\scriptsize$\pm 0.4$
&61.9\scriptsize$\pm 1.3$ & 59.4\scriptsize$\pm 1.4$
&84.1\scriptsize$\pm 0.9$ & 59.7\scriptsize$\pm 3.7$
&84.2\scriptsize$\pm 0.6$ & 77.5\scriptsize$\pm 1.2$
\\
&NN4-CVaR-DRO    &80.0\scriptsize$\pm 0.7$ & 60.3\scriptsize$\pm 2.6$
&52.5\scriptsize$\pm 2.5$ & 51.0\scriptsize$\pm 0.7$
&70.4\scriptsize$\pm 13.9$ & 71.6\scriptsize$\pm 1.6$
&57.1\scriptsize$\pm 5.2$ & 59.3\scriptsize$\pm 3.8$
&80.6\scriptsize$\pm 1.8$ & 62.0\scriptsize$\pm 0.5$
&64.7\scriptsize$\pm 2.6$ & 60.0\scriptsize$\pm 3.7$
&78.8\scriptsize$\pm 0.4$ & 78.3\scriptsize$\pm 0.1$
&60.5\scriptsize$\pm 1.5$ & 58.7\scriptsize$\pm 1.3$
&75.4\scriptsize$\pm 15.6$ & 57.4\scriptsize$\pm 6.3$
&84.8\scriptsize$\pm 0.3$ & 78.0\scriptsize$\pm 0.6$
\\
&NN4-$\chi^2$-DRO    &75.2\scriptsize$\pm 6.8$ & 64.6\scriptsize$\pm 3.2$
&54.4\scriptsize$\pm 2.4$ & 52.3\scriptsize$\pm 0.7$
&77.1\scriptsize$\pm 6.7$ & 70.8\scriptsize$\pm 1.2$
&54.0\scriptsize$\pm 4.8$ & 55.5\scriptsize$\pm 3.9$
&81.6\scriptsize$\pm 0.3$ & 62.1\scriptsize$\pm 0.2$
&66.2\scriptsize$\pm 2.6$ & 63.7\scriptsize$\pm 1.6$
&78.9\scriptsize$\pm 0.4$ & 77.7\scriptsize$\pm 0.3$
&61.3\scriptsize$\pm 1.2$ & 59.4\scriptsize$\pm 1.1$
&84.3\scriptsize$\pm 0.3$ & 59.8\scriptsize$\pm 5.2$
&84.0\scriptsize$\pm 1.0$ & 76.6\scriptsize$\pm 1.0$
\\\midrule
\multirow{4}{*}{\begin{tabular}[c]{@{}l@{}}Tree-DRO\\ Methods\\ (base: XGB/LGBM)\end{tabular}}   &XGB-CVaR-DRO    &74.6\scriptsize$\pm 3.3$ & 58.5\scriptsize$\pm 2.9$
&54.8\scriptsize$\pm 4.3$ & 52.6\scriptsize$\pm 1.9$
&79.5\scriptsize$\pm 2.2$ & 72.4\scriptsize$\pm 1.7$
&69.5\scriptsize$\pm 2.5$ & 62.8\scriptsize$\pm 0.7$
&76.5\scriptsize$\pm 2.7$ & 60.1\scriptsize$\pm 0.8$
&64.7\scriptsize$\pm 3.8$ & 61.8\scriptsize$\pm 1.5$
&78.0\scriptsize$\pm 0.2$ & 77.4\scriptsize$\pm 0.2$
&58.2\scriptsize$\pm 0.9$ & 56.1\scriptsize$\pm 0.6$
&83.0\scriptsize$\pm 3.4$ & 59.7\scriptsize$\pm 1.3$
&86.9\scriptsize$\pm 0.5$ & 79.9\scriptsize$\pm 1.3$
\\
&XGB-KL-DRO    &74.8\scriptsize$\pm 2.7$ & 60.5\scriptsize$\pm 2.5$
&60.4\scriptsize$\pm 3.6$ & 53.1\scriptsize$\pm 1.6$
&79.9\scriptsize$\pm 3.8$ & 72.8\scriptsize$\pm 1.9$
&70.1\scriptsize$\pm 0.7$ & 63.7\scriptsize$\pm 0.7$
&79.2\scriptsize$\pm 5.8$ & 59.9\scriptsize$\pm 0.9$
&67.6\scriptsize$\pm 0.9$ & 63.0\scriptsize$\pm 0.5$
&78.2\scriptsize$\pm 0.3$ & 77.4\scriptsize$\pm 0.1$
&58.4\scriptsize$\pm 1.1$ & 56.4\scriptsize$\pm 0.7$
&84.7\scriptsize$\pm 0.5$ & 59.6\scriptsize$\pm 1.1$
&87.1\scriptsize$\pm 0.5$ & 79.9\scriptsize$\pm 1.2$
\\
&LGBM-CVaR-DRO    &72.9\scriptsize$\pm 7.0$ & 61.2\scriptsize$\pm 4.4$
&53.5\scriptsize$\pm 4.1$ & 51.8\scriptsize$\pm 1.6$
&76.3\scriptsize$\pm 4.5$ & 71.4\scriptsize$\pm 2.5$
&69.8\scriptsize$\pm 1.1$ & 62.7\scriptsize$\pm 1.7$
&73.9\scriptsize$\pm 3.2$ & 59.2\scriptsize$\pm 1.2$
&66.7\scriptsize$\pm 1.6$ & 62.2\scriptsize$\pm 1.4$
&78.4\scriptsize$\pm 0.3$ & 76.6\scriptsize$\pm 0.2$
&58.0\scriptsize$\pm 0.9$ & 55.9\scriptsize$\pm 0.9$
&86.3\scriptsize$\pm 0.6$ & 62.4\scriptsize$\pm 1.8$
&87.1\scriptsize$\pm 0.4$ & 79.9\scriptsize$\pm 0.9$
\\
&LGBM-KL-DRO    &70.0\scriptsize$\pm 8.9$ & 60.5\scriptsize$\pm 3.2$
&57.3\scriptsize$\pm 5.0$ & 53.2\scriptsize$\pm 1.8$
&79.9\scriptsize$\pm 3.4$ & 70.8\scriptsize$\pm 1.5$
&70.0\scriptsize$\pm 1.3$ & 63.5\scriptsize$\pm 0.8$
&80.9\scriptsize$\pm 3.8$ & 60.4\scriptsize$\pm 0.4$
&67.6\scriptsize$\pm 1.0$ & 63.1\scriptsize$\pm 0.4$
&78.2\scriptsize$\pm 0.4$ & 76.5\scriptsize$\pm 0.1$
&58.1\scriptsize$\pm 1.0$ & 55.8\scriptsize$\pm 0.8$
&83.6\scriptsize$\pm 4.7$ & 59.7\scriptsize$\pm 1.8$
&87.1\scriptsize$\pm 0.3$ & 80.4\scriptsize$\pm 0.8$
\\\midrule
\multirow{4}{*}{\begin{tabular}[c]{@{}l@{}}Kernel-DRO\\ Methods\\ (base: Kernel)\end{tabular}}   &Kernel-$\chi2$-DRO    &80.5\scriptsize$\pm 0.6$ & 59.4\scriptsize$\pm 0.3$
&59.4\scriptsize$\pm 0.9$ & 53.0\scriptsize$\pm 0.8$
&82.6\scriptsize$\pm 0.2$ & 70.8\scriptsize$\pm 1.7$
&67.9\scriptsize$\pm 0.4$ & 63.6\scriptsize$\pm 0.9$
&77.8\scriptsize$\pm 1.2$ & 61.5\scriptsize$\pm 0.6$
&67.6\scriptsize$\pm 0.7$ & 65.7\scriptsize$\pm 0.9$
&80.1\scriptsize$\pm 0.5$ & 78.3\scriptsize$\pm 0.2$
&61.0\scriptsize$\pm 0.8$ & 58.5\scriptsize$\pm 0.5$
&84.3\scriptsize$\pm 0.3$ & 55.7\scriptsize$\pm 2.8$
&84.3\scriptsize$\pm 0.5$ & 77.4\scriptsize$\pm 1.0$
\\
&Kernel-CVaR-DRO    &80.4\scriptsize$\pm 0.6$ & 59.4\scriptsize$\pm 0.4$
&57.1\scriptsize$\pm 3.3$ & 52.6\scriptsize$\pm 1.3$
&84.3\scriptsize$\pm 0.4$ & 71.6\scriptsize$\pm 0.9$
&68.3\scriptsize$\pm 0.6$ & 64.0\scriptsize$\pm 1.3$
&78.2\scriptsize$\pm 2.2$ & 61.4\scriptsize$\pm 0.5$
&65.7\scriptsize$\pm 1.0$ & 61.7\scriptsize$\pm 1.5$
&78.6\scriptsize$\pm 0.4$ & 78.3\scriptsize$\pm 0.2$
&60.3\scriptsize$\pm 1.6$ & 57.9\scriptsize$\pm 1.1$
&83.9\scriptsize$\pm 0.5$ & 59.8\scriptsize$\pm 1.5$
&84.3\scriptsize$\pm 0.4$ & 77.6\scriptsize$\pm 1.0$
\\
&Kernel-KL-DRO    &80.3\scriptsize$\pm 0.6$ & 59.6\scriptsize$\pm 0.5$
&56.5\scriptsize$\pm 1.8$ & 52.9\scriptsize$\pm 1.5$
&81.3\scriptsize$\pm 0.4$ & 72.9\scriptsize$\pm 1.3$
&67.8\scriptsize$\pm 1.0$ & 64.8\scriptsize$\pm 0.8$
&77.6\scriptsize$\pm 1.2$ & 61.2\scriptsize$\pm 0.5$
&66.5\scriptsize$\pm 2.2$ & 65.0\scriptsize$\pm 1.9$
&78.7\scriptsize$\pm 0.6$ & 78.5\scriptsize$\pm 0.2$
&61.3\scriptsize$\pm 0.5$ & 58.6\scriptsize$\pm 0.5$
&83.6\scriptsize$\pm 1.9$ & 60.2\scriptsize$\pm 3.3$
&83.9\scriptsize$\pm 0.7$ & 76.7\scriptsize$\pm 1.2$
\\
&Kernel-Wasserstein-DRO    &77.6\scriptsize$\pm 4.5$ & 60.0\scriptsize$\pm 0.9$
&56.1\scriptsize$\pm 2.1$ & 51.0\scriptsize$\pm 0.4$
&79.6\scriptsize$\pm 1.6$ & 73.7\scriptsize$\pm 1.3$
&66.2\scriptsize$\pm 0.8$ & 56.5\scriptsize$\pm 0.8$
&75.2\scriptsize$\pm 2.1$ & 60.1\scriptsize$\pm 1.5$
&63.3\scriptsize$\pm 1.7$ & 58.7\scriptsize$\pm 2.0$
&77.8\scriptsize$\pm 0.3$ & 77.3\scriptsize$\pm 0.3$
&56.6\scriptsize$\pm 1.9$ & 55.6\scriptsize$\pm 0.8$
&84.1\scriptsize$\pm 0.3$ & 56.9\scriptsize$\pm 3.5$
&83.3\scriptsize$\pm 1.0$ & 76.9\scriptsize$\pm 1.4$
\\\midrule
\multirow{6}{*}{\begin{tabular}[c]{@{}l@{}}Imbalanced\\ Learning\\ \& Fairness\\ Methods \\(base: XGB)\end{tabular}} &SUBY    &77.8\scriptsize$\pm 3.4$ & 54.8\scriptsize$\pm 1.1$
&60.4\scriptsize$\pm 5.0$ & 53.7\scriptsize$\pm 1.0$
&79.5\scriptsize$\pm 2.3$ & 73.4\scriptsize$\pm 1.0$
&66.4\scriptsize$\pm 2.1$ & 67.0\scriptsize$\pm 1.5$
&84.3\scriptsize$\pm 0.3$ & 62.4\scriptsize$\pm 0.6$
&69.0\scriptsize$\pm 1.5$ & 67.2\scriptsize$\pm 1.6$
&80.7\scriptsize$\pm 0.2$ & 79.2\scriptsize$\pm 0.1$
&62.0\scriptsize$\pm 1.9$ & 60.0\scriptsize$\pm 1.8$
&84.8\scriptsize$\pm 0.6$ & 60.0\scriptsize$\pm 1.5$
&80.0\scriptsize$\pm 2.6$ & 71.3\scriptsize$\pm 2.7$
\\
&RWY    &79.7\scriptsize$\pm 1.3$ & 56.2\scriptsize$\pm 0.2$
&61.5\scriptsize$\pm 4.3$ & 53.4\scriptsize$\pm 1.1$
&78.6\scriptsize$\pm 4.6$ & 73.4\scriptsize$\pm 1.9$
&71.6\scriptsize$\pm 0.3$ & 68.8\scriptsize$\pm 0.7$
&82.4\scriptsize$\pm 2.4$ & 61.7\scriptsize$\pm 0.6$
&69.7\scriptsize$\pm 1.3$ & 67.4\scriptsize$\pm 2.2$
&80.7\scriptsize$\pm 0.2$ & 79.2\scriptsize$\pm 0.1$
&61.3\scriptsize$\pm 1.8$ & 59.3\scriptsize$\pm 1.5$
&84.2\scriptsize$\pm 0.4$ & 57.9\scriptsize$\pm 1.7$
&87.0\scriptsize$\pm 0.5$ & 79.7\scriptsize$\pm 1.1$
\\
&SUBG    &74.4\scriptsize$\pm 10.6$ & 59.7\scriptsize$\pm 1.9$
&61.0\scriptsize$\pm 2.8$ & 52.9\scriptsize$\pm 1.0$
&-&-
&70.9\scriptsize$\pm 0.9$ & 65.3\scriptsize$\pm 0.6$
&-&-
&68.3\scriptsize$\pm 2.2$ & 64.0\scriptsize$\pm 2.3$
&80.7\scriptsize$\pm 0.2$ & 79.2\scriptsize$\pm 0.1$
&53.4\scriptsize$\pm 2.2$ & 53.1\scriptsize$\pm 1.8$
&-&-
&85.8\scriptsize$\pm 0.4$ & 78.7\scriptsize$\pm 1.0$
\\
&RWG    &79.7\scriptsize$\pm 0.9$ & 59.0\scriptsize$\pm 0.6$
&61.8\scriptsize$\pm 2.7$ & 52.8\scriptsize$\pm 1.0$
&-&-
&70.4\scriptsize$\pm 2.0$ & 65.0\scriptsize$\pm 1.3$
&-&-
&70.2\scriptsize$\pm 0.9$ & 65.3\scriptsize$\pm 0.7$
&80.7\scriptsize$\pm 0.2$ & 79.2\scriptsize$\pm 0.1$
&60.5\scriptsize$\pm 1.3$ & 58.9\scriptsize$\pm 1.2$
&-&-
&87.2\scriptsize$\pm 0.5$ & 78.8\scriptsize$\pm 0.8$
\\
&In-processing    &76.7\scriptsize$\pm 4.3$ & 58.6\scriptsize$\pm 0.8$
&60.6\scriptsize$\pm 0.5$ & 51.8\scriptsize$\pm 0.6$
&-&-
&70.8\scriptsize$\pm 0.7$ & 65.3\scriptsize$\pm 0.5$
&-&-
&67.7\scriptsize$\pm 3.0$ & 63.5\scriptsize$\pm 2.7$
&80.1\scriptsize$\pm 0.1$ & 78.7\scriptsize$\pm 0.1$
&62.2\scriptsize$\pm 1.5$ & 61.1\scriptsize$\pm 1.6$
&-&-
&86.9\scriptsize$\pm 0.4$ & 80.3\scriptsize$\pm 0.7$
\\
&Post-processing    &71.6\scriptsize$\pm 6.7$ & 60.9\scriptsize$\pm 3.0$
&62.1\scriptsize$\pm 1.6$ & 52.8\scriptsize$\pm 0.9$
&-&-
&69.1\scriptsize$\pm 1.7$ & 64.1\scriptsize$\pm 1.3$
&-&-
&69.1\scriptsize$\pm 2.1$ & 64.6\scriptsize$\pm 1.2$
&79.2\scriptsize$\pm 0.1$ & 78.7\scriptsize$\pm 0.1$
&60.2\scriptsize$\pm 1.2$ & 58.8\scriptsize$\pm 1.0$
&-&-
&82.5\scriptsize$\pm 0.2$ & 80.0\scriptsize$\pm 0.8$
\\\bottomrule
\end{tabular}}
\end{table}
\end{landscape}


\begin{table}[]
\caption{Performance of methods across all 172 source-target pairs. The table presents the algorithmic ranking and accuracy for each method over all pairs. The min, max, mean, standard deviation, and median of these metrics are computed across all pairs. }
\label{tab:ranking-all}
\centering
\resizebox{\textwidth}{10cm}{
\begin{tabular}{@{}llcccccccc@{}}
\toprule
\multirow{2}{*}{Category} & \multirow{2}{*}{Method} & \multicolumn{5}{c}{Rank} & \multicolumn{3}{c}{Accuracy} \\ \cmidrule(l){3-7}\cmidrule(l){8-10} 
 &  &Min & Max & Mean & Std & Median & Mean & Std & Median \\ \midrule
\multirow{6}{*}{\begin{tabular}[c]{@{}l@{}}Basic \\ Methods\end{tabular}}&LR&5&44&28.58 & 8.20 & 30&74.86 & 5.99 & 75.78\\
&SVM&7&45&29.68 & 8.58 & 32&74.86 & 5.71 & 75.56\\
&Kernel-SVM&2&44&31.13 & 12.64 & 36&75.04 & 5.94 & 75.78\\
&NN2&1&43&18.63 & 8.88 & 17&76.56 & 5.09 & 77.19\\
&NN3&1&43&23.22 & 12.58 & 26&76.19 & 5.61 & 77.03\\
&NN4&3&43&27.23 & 12.50 & 34&75.84 & 5.77 & 76.54\\\midrule
\multirow{4}{*}{\begin{tabular}[c]{@{}l@{}}Tree-based \\ Ensemble \\ Methods\end{tabular}}  &RF&1&39&14.56 & 8.59 & 15&76.51 & 5.21 & 76.74\\
&XGB&1&44&15.05 & 11.40 & 12&76.70 & 5.09 & 76.56\\
&GBM&1&35&9.53 & 7.89 & 6&76.95 & 4.93 & 77.15\\
&LGBM&1&29&8.80 & 6.55 & 8&77.23 & 4.97 & 77.45\\\midrule
\multirow{13}{*}{\begin{tabular}[c]{@{}l@{}}Linear-DRO \\  Methods \\ (base: SVM)\end{tabular}}  &KL-DRO&3&45&29.47 & 7.66 & 30&74.98 & 5.48 & 75.55\\
&CVaR-DRO&1&40&23.44 & 8.65 & 23&75.39 & 5.26 & 75.89\\
&$\chi^2$-DRO&5&45&26.11 & 8.35 & 27&75.18 & 5.40 & 75.82\\
&TV-DRO&4&43&31.14 & 8.07 & 32&74.86 & 5.51 & 75.38\\
&Wasserstein-DRO&11&43&33.63 & 6.07 & 35&74.66 & 5.95 & 75.70\\
&Aug. Wass.-DRO&2&40&25.45 & 9.02 & 25&75.98 & 5.06 & 76.30\\
&Satis. Wass.-DRO&13&45&41.88 & 5.16 & 43&72.45 & 7.05 & 73.77\\
&Sinkhorn-DRO&2&44&25.60 & 12.29 & 29&75.49 & 5.37 & 76.46\\
&Unified-DRO($L_2$)&14&45&35.91 & 5.86 & 37&74.44 & 5.91 & 75.33\\
&Unified-DRO($L_\text{inf}$)&4&45&34.95 & 7.06 & 36&74.46 & 5.89 & 75.52\\
&Marginal-DRO&2&45&24.03 & 9.58 & 24&75.46 & 5.08 & 76.15\\
&Conditional-DRO&1&45&20.77 & 8.82 & 20&76.10 & 5.36 & 76.32\\
&Holistic-DRO&2&45&31.49 & 10.17 & 34&74.52 & 5.95 & 75.64\\\midrule
\multirow{8}{*}{\begin{tabular}[c]{@{}l@{}}NN-DRO\\ Methods\\ (base: NN)\end{tabular}} &NN2-CVaR-DRO&1&38&16.51 & 11.29 & 13&76.89 & 5.29 & 77.28\\
&NN2-$\chi^2$-DRO&1&36&15.17 & 9.87 & 12&76.87 & 5.25 & 77.38\\
&NN2-CVaR-DORO&2&40&21.84 & 8.79 & 21&76.34 & 5.01 & 76.95\\
&NN2-$\chi^2$-DORO&2&45&31.12 & 11.74 & 34&74.86 & 5.66 & 75.41\\
&NN3-CVaR-DRO&1&41&18.76 & 11.84 & 14&76.74 & 5.70 & 77.14\\
&NN3-$\chi^2$-DRO&1&40&18.09 & 12.47 & 14&76.69 & 5.67 & 77.22\\
&NN4-CVaR-DRO&1&42&23.99 & 12.85 & 22&76.12 & 5.45 & 76.25\\
&NN4-$\chi^2$-DRO&1&45&21.92 & 14.94 & 20&76.01 & 5.78 & 76.61\\\midrule
\multirow{4}{*}{\begin{tabular}[c]{@{}l@{}}Tree-DRO\\ Methods\\ (base: XGB/LGBM)\end{tabular}}  &XGB-CVaR-DRO&1&41&16.48 & 10.17 & 15&76.81 & 5.21 & 77.31\\
&XGB-KL-DRO&1&44&17.24 & 11.83 & 16&76.98 & 5.50 & 77.79\\
&LGBM-CVaR-DRO&1&45&17.95 & 13.59 & 14&76.63 & 5.33 & 76.55\\
&LGBM-KL-DRO&1&43&16.99 & 11.88 & 15&76.91 & 5.33 & 77.48\\\midrule
\multirow{4}{*}{\begin{tabular}[c]{@{}l@{}}Kernel-DRO\\ Methods\\ (base: Kernel)\end{tabular}} &Kernel-$\chi^2$-DRO&1&36&16.40 & 8.16 & 18&76.17 & 4.97 & 76.39\\
&Kernel-CVaR-DRO&1&42&18.08 & 8.29 & 19&76.12 & 4.96 & 76.41\\
&Kernel-KL-DRO&1&35&17.15 & 8.61 & 17&76.15 & 4.89 & 76.32\\
&Kernel-Wasserstein-DRO&1&44&24.91 & 8.62 & 26&75.59 & 5.65 & 76.07\\\midrule
\multirow{6}{*}{\begin{tabular}[c]{@{}l@{}}Imbalanced\\ Learning\\ \& Fairness\\ Methods \\(base: XGB)\end{tabular}}&SUBY&1&45&33.69 & 14.43 & 42&75.40 & 5.27 & 74.50\\
&RWY&1&45&31.42 & 16.44 & 40&75.72 & 5.43 & 75.02\\
&SUBG&1&44&14.49 & 10.93 & 10&76.92 & 5.55 & 76.70\\
&RWG&1&45&14.69 & 11.62 & 10&76.90 & 5.37 & 76.63\\
&In-processing&1&42&13.19 & 11.57 & 8&76.48 & 4.65 & 76.54\\
&Post-processing&1&43&19.19 & 14.01 & 13&76.14 & 4.66 & 76.08\\
\bottomrule
\end{tabular}
}
\end{table}


\paragraph{Analysis.}
Based on results in Tables~\ref{table:selected_results} and~\ref{table:selected_results_f1}, we provide a more detailed analysis compared to our main body.
\begin{itemize}
	\item Different methods do not exhibit consistent rankings over different distribution shift patterns. The results show that the algorithmic rankings across different settings are quite different. And even the rankings of algorithms within the same method class vary a lot. This further demonstrates the complexity of $Y|X$-shifts. 
	\item Tree-based ensemble methods show competitive performances but do not significantly eliminate the generalization error between source and target data, characterized by the difference between \textit{o.o.d.} and \textit{i.d.} method performance. However, the performance degradation between source and target is still large.
  	\item Imbalance methods and fairness methods show similar performance with the base model class (XGBoost). We find these methods do not have a significant improvement over the basic model class (XGBoost). This is because they are not designed specifically for tree-based ensemble methods.
\end{itemize}

Additionally, we report the \emph{oracle} accuracy (Macro F1-Score) of each method across all target domains in each setting in~\Cref{table:selected_results_to_all_accuracy} and \Cref{table:selected_results_to_all_f1}, where we select the best configuration for each method according to the \emph{target} performance.
The observed phenomena are consistent with our previous findings.

\begin{landscape}
\begin{table}[!htb]
\caption{Results (Accuracy) of 10 settings in \Cref{subsec:benchmark-empirical-result} with all 172 target domains, where we report the source accuracy, the mean target accuracy and standard deviation across multiple target domains in each setting. We run each method with its best configuration (according to the target F1-score), and boldface the best target performance within each class of methods in each setting.}
\vspace{5pt}
\label{table:selected_results_to_all_accuracy}
\resizebox{21cm}{!}{
\begin{tabular}{@{}llcccccccccccccccccccc@{}}
\toprule\toprule
\multicolumn{2}{l}{\large \textbf{Dataset}}                                                                   & \multicolumn{2}{c}{\large \texttt{ACS Income}}                                                                    & \multicolumn{2}{c}{\large \texttt{ACS Mobility}}                                                                  & \multicolumn{2}{c}{\large \texttt{US Taxi}}                 & \multicolumn{2}{c}{\large \texttt{ACS Pub.Cov}}                                                                   & \multicolumn{2}{c}{\large \texttt{US Accident}}                                                                   & \multicolumn{2}{c}{\large \texttt{ACS Time}}                                                                      & \multicolumn{2}{c}{\large \texttt{Sub-Sampling}}   & \multicolumn{2}{c}{\large \texttt{diabetes}} &  \multicolumn{2}{c}{\large \texttt{assistments}} & \multicolumn{2}{c}{\large \texttt{college}}             \\ 
\multicolumn{2}{l}{\large \textbf{Shift Pattern}}                                                            & \multicolumn{2}{c}{$Y|X$ dominates}                                                                              & \multicolumn{2}{c}{ $Y|X$ dominates}                                                                              & \multicolumn{2}{c}{ $Y|X$ dominates}                       & \multicolumn{2}{c}{ $Y|X$ more}                                                                            & \multicolumn{2}{c}{$Y|X$ more}                                                                              & \multicolumn{2}{c}{$X$ more}                                                                            & \multicolumn{2}{c}{$X$ dominates}  & \multicolumn{2}{c}{$Y|X$ dominates} & \multicolumn{2}{c}{$Y|X$ more} & \multicolumn{2}{c}{$Y|X$ more} \\
\multicolumn{2}{l}{\large Source $\rightarrow$ Target Pair}                                                       & \multicolumn{2}{c}{CA$\rightarrow$ 50 Domains}                                                                        & \multicolumn{2}{c}{MS$\rightarrow$50 Domains}                                                                        & \multicolumn{2}{c}{NYC$\rightarrow$3 Domains}                & \multicolumn{2}{c}{NE$\rightarrow$50 Domains}                                                                        & \multicolumn{2}{c}{CA$\rightarrow$13 Domains}                                                                        & \multicolumn{2}{c}{2010$\rightarrow$3 Domains}                                                                    & \multicolumn{2}{c}{Young$\rightarrow$Old}   & \multicolumn{2}{c}{White$\rightarrow$Others} & \multicolumn{2}{c}{700$\rightarrow$1} & \multicolumn{2}{c}{Normal $\rightarrow$ Special}  \\
\multicolumn{2}{l}{}                                                                          & $i.d.$                                         & $o.o.d$                                         & $i.d.$                                         & $o.o.d$                                         & $i.d.$              & $o.o.d$              & $i.d.$                                         & $o.o.d$                                         & $i.d.$                                         & $o.o.d$                                         & $i.d.$                                         & $o.o.d$                                         & $i.d.$                                         & $o.o.d$                           & $i.d.$                                         & $o.o.d$                                         & $i.d.$                                         & $o.o.d$                                         & $i.d.$                                         & $o.o.d$                                         \\\midrule
\multirow{6}{*}{\begin{tabular}[c]{@{}l@{}}Basic \\ Methods\end{tabular}} &LR    &80.6 & 77.3\scriptsize$\pm 1.8$
&75.8 & 72.7\scriptsize$\pm 3.7$
&84.4 & 74.4\scriptsize$\pm 1.5$
&84.0 & \textbf{75.7}\scriptsize$\pm 6.6$
&78.7 & 71.2\scriptsize$\pm 9.5$
&72.1 & 62.0\scriptsize$\pm 2.6$
&91.9 & 79.5\scriptsize$\pm 0.0$
&65.1 & 60.0\scriptsize$\pm 0.0$
&88.3 & 49.7\scriptsize$\pm 0.0$
&93.4 & 81.9\scriptsize$\pm 0.0$
\\
&SVM    &80.7 & 77.1\scriptsize$\pm 1.9$
&77.8 & 72.9\scriptsize$\pm 3.5$
&83.3 & \textbf{74.6}\scriptsize$\pm 1.5$
&83.5 & 75.4\scriptsize$\pm 6.3$
&78.8 & 71.3\scriptsize$\pm 9.4$
&77.2 & 69.1\scriptsize$\pm 2.1$
&91.7 & \textbf{80.1}\scriptsize$\pm 0.0$
&64.9 & 59.9\scriptsize$\pm 0.0$
&88.3 & 46.0\scriptsize$\pm 0.0$
&93.4 & 81.9\scriptsize$\pm 0.0$
\\
&Kernel-SVM    &80.1 & 77.6\scriptsize$\pm 1.6$
&77.0 & 72.8\scriptsize$\pm 3.6$
&79.9 & 71.3\scriptsize$\pm 1.3$
&78.2 & 70.7\scriptsize$\pm 5.8$
&78.4 & 82.8\scriptsize$\pm 6.8$
&72.5 & 63.5\scriptsize$\pm 2.4$
&91.9 & 79.0\scriptsize$\pm 0.0$
&64.0 & 61.1\scriptsize$\pm 0.0$
&66.1 & 64.7\scriptsize$\pm 0.0$
&88.6 & 78.3\scriptsize$\pm 0.0$
\\
&NN2    &81.0 & 78.6\scriptsize$\pm 1.7$
&77.6 & \textbf{72.9}\scriptsize$\pm 3.4$
&81.5 & 73.6\scriptsize$\pm 0.9$
&82.9 & 75.3\scriptsize$\pm 6.2$
&82.3 & 82.6\scriptsize$\pm 6.9$
&76.8 & \textbf{70.8}\scriptsize$\pm 1.9$
&91.2 & 79.5\scriptsize$\pm 0.0$
&64.2 & 60.4\scriptsize$\pm 0.0$
&88.9 & \textbf{65.6}\scriptsize$\pm 0.0$
&93.7 & 82.9\scriptsize$\pm 0.0$
\\
&NN3    &80.0 & \textbf{79.1}\scriptsize$\pm 1.6$
&76.6 & 72.8\scriptsize$\pm 3.4$
&85.0 & 73.1\scriptsize$\pm 0.9$
&82.5 & 74.0\scriptsize$\pm 6.3$
&82.9 & 82.7\scriptsize$\pm 6.9$
&77.6 & 70.2\scriptsize$\pm 2.3$
&91.5 & 80.0\scriptsize$\pm 0.0$
&64.5 & \textbf{60.9}\scriptsize$\pm 0.0$
&88.8 & 65.0\scriptsize$\pm 0.0$
&93.7 & \textbf{83.1}\scriptsize$\pm 0.0$
\\
&NN4    &81.0 & 78.5\scriptsize$\pm 1.6$
&75.8 & 72.7\scriptsize$\pm 3.7$
&83.8 & 74.3\scriptsize$\pm 1.3$
&82.1 & 74.0\scriptsize$\pm 6.0$
&84.4 & \textbf{84.6}\scriptsize$\pm 7.2$
&75.9 & 69.9\scriptsize$\pm 2.1$
&90.7 & 78.8\scriptsize$\pm 0.0$
&63.3 & \textbf{60.9}\scriptsize$\pm 0.0$
&88.6 & 60.7\scriptsize$\pm 0.0$
&93.7 & 82.7\scriptsize$\pm 0.0$
\\\midrule
\multirow{4}{*}{\begin{tabular}[c]{@{}l@{}}Tree-based \\ Ensemble \\ Methods\end{tabular}}   &RF    &81.3 & 77.8\scriptsize$\pm 1.9$
&80.0 & 73.6\scriptsize$\pm 3.4$
&83.6 & \textbf{75.0}\scriptsize$\pm 1.5$
&85.5 & 76.0\scriptsize$\pm 6.5$
&86.0 & 78.3\scriptsize$\pm 7.6$
&79.4 & 71.2\scriptsize$\pm 2.3$
&92.1 & \textbf{80.8}\scriptsize$\pm 0.0$
&65.6 & 62.5\scriptsize$\pm 0.0$
&88.1 & \textbf{66.6}\scriptsize$\pm 0.0$
&94.3 & 84.8\scriptsize$\pm 0.0$
\\
&XGB    &81.8 & 77.6\scriptsize$\pm 2.2$
&79.7 & 73.5\scriptsize$\pm 2.9$
&83.1 & 73.9\scriptsize$\pm 1.1$
&85.7 & 76.5\scriptsize$\pm 6.3$
&87.0 & \textbf{80.2}\scriptsize$\pm 7.8$
&79.5 & 71.8\scriptsize$\pm 2.4$
&92.3 & 80.3\scriptsize$\pm 0.0$
&66.3 & \textbf{62.6}\scriptsize$\pm 0.0$
&89.8 & 63.1\scriptsize$\pm 0.0$
&94.8 & \textbf{85.6}\scriptsize$\pm 0.0$
\\
&GBM    &81.0 & \textbf{78.2}\scriptsize$\pm 1.6$
&80.0 & 73.6\scriptsize$\pm 3.0$
&66.1 & 74.2\scriptsize$\pm 1.0$
&85.6 & \textbf{76.7}\scriptsize$\pm 6.0$
&85.7 & 79.0\scriptsize$\pm 7.3$
&79.7 & \textbf{71.9}\scriptsize$\pm 2.3$
&92.1 & \textbf{80.8}\scriptsize$\pm 0.0$
&63.7 & 60.9\scriptsize$\pm 0.0$
&72.7 & 65.4\scriptsize$\pm 0.0$
&94.7 & \textbf{85.6}\scriptsize$\pm 0.0$
\\
&LGBM    &80.4 & 78.1\scriptsize$\pm 1.3$
&80.0 & \textbf{73.8}\scriptsize$\pm 3.3$
&79.3 & 74.5\scriptsize$\pm 1.5$
&85.9 & 76.5\scriptsize$\pm 6.4$
&87.0 & 79.8\scriptsize$\pm 7.8$
&79.6 & 71.8\scriptsize$\pm 2.4$
&92.3 & 80.3\scriptsize$\pm 0.0$
&66.1 & \textbf{62.6}\scriptsize$\pm 0.0$
&87.5 & 65.9\scriptsize$\pm 0.0$
&94.9 & 85.8\scriptsize$\pm 0.0$
\\\midrule
\multirow{13}{*}{\begin{tabular}[c]{@{}l@{}}Linear-DRO \\  Methods \\ (base: SVM)\end{tabular}}   &KL-DRO    &80.7 & 77.0\scriptsize$\pm 2.2$
&77.8 & 72.7\scriptsize$\pm 3.5$
&83.4 & 74.4\scriptsize$\pm 1.6$
&83.6 & 75.4\scriptsize$\pm 5.9$
&79.2 & 72.2\scriptsize$\pm 8.8$
&77.0 & 68.8\scriptsize$\pm 2.3$
&91.6 & 79.9\scriptsize$\pm 0.0$
&64.8 & 59.9\scriptsize$\pm 0.0$
&88.2 & 47.4\scriptsize$\pm 0.0$
&93.6 & 81.9\scriptsize$\pm 0.0$
\\
&CVaR-DRO    &80.8 & 77.4\scriptsize$\pm 1.9$
&78.1 & 72.8\scriptsize$\pm 3.4$
&83.6 & 75.2\scriptsize$\pm 1.4$
&84.2 & 75.8\scriptsize$\pm 5.9$
&79.5 & 72.4\scriptsize$\pm 8.8$
&76.7 & 69.5\scriptsize$\pm 2.1$
&91.7 & \textbf{80.4}\scriptsize$\pm 0.0$
&64.9 & 60.5\scriptsize$\pm 0.0$
&88.3 & \textbf{50.7}\scriptsize$\pm 0.0$
&93.5 & 81.8\scriptsize$\pm 0.0$
\\
&$\chi^2$-DRO    &80.7 & 77.4\scriptsize$\pm 1.9$
&77.9 & 72.8\scriptsize$\pm 3.5$
&83.3 & 75.2\scriptsize$\pm 1.4$
&83.7 & 75.6\scriptsize$\pm 5.6$
&79.2 & 72.2\scriptsize$\pm 8.8$
&76.5 & 69.5\scriptsize$\pm 2.1$
&91.7 & 79.8\scriptsize$\pm 0.0$
&65.1 & 60.2\scriptsize$\pm 0.0$
&88.3 & 48.4\scriptsize$\pm 0.0$
&93.6 & 81.9\scriptsize$\pm 0.0$
\\
&TV-DRO    &80.6 & 76.8\scriptsize$\pm 2.0$
&77.5 & 72.6\scriptsize$\pm 3.6$
&82.9 & 74.0\scriptsize$\pm 1.6$
&82.4 & 74.8\scriptsize$\pm 6.6$
&79.2 & 72.3\scriptsize$\pm 8.9$
&76.6 & 69.4\scriptsize$\pm 2.0$
&91.0 & 80.1\scriptsize$\pm 0.0$
&64.0 & 59.3\scriptsize$\pm 0.0$
&88.3 & 44.9\scriptsize$\pm 0.0$
&93.6 & 81.7\scriptsize$\pm 0.0$
\\
&Wasserstein-DRO    &80.5 & 76.9\scriptsize$\pm 1.9$
&75.8 & 72.7\scriptsize$\pm 3.7$
&84.1 & 74.5\scriptsize$\pm 1.5$
&82.3 & 74.8\scriptsize$\pm 6.7$
&79.2 & 72.3\scriptsize$\pm 8.8$
&76.0 & 67.7\scriptsize$\pm 2.4$
&91.5 & 77.3\scriptsize$\pm 0.0$
&64.0 & 59.3\scriptsize$\pm 0.0$
&88.2 & 44.9\scriptsize$\pm 0.0$
&93.5 & 81.9\scriptsize$\pm 0.0$
\\
&Aug. Wass.-DRO    &80.2 & 77.0\scriptsize$\pm 1.6$
&77.6 & 73.0\scriptsize$\pm 3.6$
&82.0 & 73.9\scriptsize$\pm 1.4$
&83.5 & 75.3\scriptsize$\pm 6.8$
&69.8 & 79.6\scriptsize$\pm 8.5$
&76.8 & 68.8\scriptsize$\pm 2.3$
&91.4 & \textbf{80.4}\scriptsize$\pm 0.0$
&62.9 & 59.8\scriptsize$\pm 0.0$
&66.4 & 58.7\scriptsize$\pm 0.0$
&93.5 & 81.8\scriptsize$\pm 0.0$
\\
&Satis. Wass.-DRO    &79.4 & 76.3\scriptsize$\pm 1.8$
&75.8 & 72.7\scriptsize$\pm 3.7$
&79.7 & 70.1\scriptsize$\pm 1.2$
&80.0 & 69.9\scriptsize$\pm 8.1$
&77.1 & 68.9\scriptsize$\pm 10.0$
&71.7 & 61.0\scriptsize$\pm 2.8$
&91.3 & 78.7\scriptsize$\pm 0.0$
&58.5 & 50.6\scriptsize$\pm 0.0$
&87.8 & 44.8\scriptsize$\pm 0.0$
&93.0 & 81.5\scriptsize$\pm 0.0$
\\
&Sinkhorn-DRO    &77.5 & \textbf{78.1}\scriptsize$\pm 1.4$
&75.7 & 72.7\scriptsize$\pm 3.7$
&82.5 & 75.2\scriptsize$\pm 1.3$
&83.2 & \textbf{75.8}\scriptsize$\pm 5.0$
&80.1 & 72.2\scriptsize$\pm 8.6$
&76.1 & \textbf{70.3}\scriptsize$\pm 1.6$
&91.5 & 79.8\scriptsize$\pm 0.0$
&63.6 & 59.3\scriptsize$\pm 0.0$
&85.5 & 50.0\scriptsize$\pm 0.0$
&91.8 & \textbf{82.4}\scriptsize$\pm 0.0$
\\
&Unified-DRO($L_2$)    &80.6 & 76.9\scriptsize$\pm 1.9$
&77.5 & 72.6\scriptsize$\pm 3.6$
&83.3 & 74.3\scriptsize$\pm 1.5$
&82.1 & 74.7\scriptsize$\pm 6.6$
&79.0 & 71.9\scriptsize$\pm 9.0$
&76.0 & 67.7\scriptsize$\pm 2.4$
&91.3 & 79.7\scriptsize$\pm 0.0$
&62.6 & 57.8\scriptsize$\pm 0.0$
&88.3 & 44.9\scriptsize$\pm 0.0$
&93.6 & 81.6\scriptsize$\pm 0.0$
\\
&Unified-DRO($L_\text{inf}$)    &80.6 & 76.9\scriptsize$\pm 1.9$
&76.5 & 72.9\scriptsize$\pm 3.6$
&83.6 & 74.6\scriptsize$\pm 1.5$
&82.2 & 74.8\scriptsize$\pm 6.6$
&78.2 & 70.8\scriptsize$\pm 9.2$
&76.0 & 67.7\scriptsize$\pm 2.4$
&92.0 & 79.3\scriptsize$\pm 0.0$
&64.0 & 59.3\scriptsize$\pm 0.0$
&88.2 & 44.9\scriptsize$\pm 0.0$
&93.6 & 81.7\scriptsize$\pm 0.0$
\\
&Marginal-DRO    &80.8 & 77.4\scriptsize$\pm 1.9$
&78.2 & 72.8\scriptsize$\pm 3.4$
&82.7 & 75.2\scriptsize$\pm 1.4$
&83.5 & 76.3\scriptsize$\pm 5.7$
&79.5 & 72.8\scriptsize$\pm 8.7$
&76.2 & 69.8\scriptsize$\pm 1.9$
&91.1 & 79.3\scriptsize$\pm 0.0$
&64.7 & 60.6\scriptsize$\pm 0.0$
&88.3 & 56.9\scriptsize$\pm 0.0$
&93.6 & 81.9\scriptsize$\pm 0.0$
\\
&Conditional-DRO    &80.7 & 77.5\scriptsize$\pm 2.3$
&77.6 & \textbf{73.0}\scriptsize$\pm 3.3$
&82.6 & \textbf{75.4}\scriptsize$\pm 1.2$
&83.3 & 75.7\scriptsize$\pm 5.4$
&76.3 & \textbf{79.9}\scriptsize$\pm 8.1$
&76.4 & 69.7\scriptsize$\pm 2.1$
&91.6 & 79.3\scriptsize$\pm 0.0$
&64.1 & \textbf{60.8}\scriptsize$\pm 0.0$
&88.2 & 44.9\scriptsize$\pm 0.0$
&93.5 & 81.7\scriptsize$\pm 0.0$
\\
&Holistic-DRO    &78.5 & 76.6\scriptsize$\pm 1.4$
&77.6 & 72.9\scriptsize$\pm 3.6$
&80.4 & 70.9\scriptsize$\pm 1.4$
&82.8 & 74.7\scriptsize$\pm 6.5$
&78.1 & 69.9\scriptsize$\pm 9.8$
&77.0 & 68.3\scriptsize$\pm 2.5$
&91.1 & 77.3\scriptsize$\pm 0.0$
&63.9 & 58.8\scriptsize$\pm 0.0$
&88.3 & 45.0\scriptsize$\pm 0.0$
&92.9 & 80.6\scriptsize$\pm 0.0$
\\\midrule
\multirow{8}{*}{\begin{tabular}[c]{@{}l@{}}NN-DRO\\ Methods\\ (base: NN)\end{tabular}} &NN2-CVaR-DRO    &79.2 & 78.8\scriptsize$\pm 1.4$
&77.3 & \textbf{73.6}\scriptsize$\pm 3.3$
&83.9 & 74.4\scriptsize$\pm 1.4$
&80.9 & 74.3\scriptsize$\pm 5.1$
&80.8 & 85.1\scriptsize$\pm 7.3$
&77.9 & 69.8\scriptsize$\pm 2.4$
&91.3 & \textbf{80.9}\scriptsize$\pm 0.0$
&64.1 & 61.2\scriptsize$\pm 0.0$
&88.5 & 67.7\scriptsize$\pm 0.0$
&94.2 & 83.3\scriptsize$\pm 0.0$
\\
&NN2-$\chi^2$-DRO    &80.5 & 78.5\scriptsize$\pm 1.5$
&77.6 & 73.5\scriptsize$\pm 3.3$
&84.1 & \textbf{74.8}\scriptsize$\pm 1.2$
&81.9 & 75.2\scriptsize$\pm 5.4$
&82.0 & 83.8\scriptsize$\pm 7.2$
&76.7 & 70.2\scriptsize$\pm 2.3$
&92.1 & 80.2\scriptsize$\pm 0.0$
&64.4 & 61.0\scriptsize$\pm 0.0$
&88.4 & 65.4\scriptsize$\pm 0.0$
&93.3 & 83.4\scriptsize$\pm 0.0$
\\
&NN2-CVaR-DORO    &78.6 & 78.2\scriptsize$\pm 1.4$
&75.9 & 72.7\scriptsize$\pm 3.7$
&81.4 & 74.7\scriptsize$\pm 0.9$
&83.0 & \textbf{75.8}\scriptsize$\pm 5.6$
&81.1 & 78.9\scriptsize$\pm 7.9$
&76.9 & 68.1\scriptsize$\pm 2.4$
&91.5 & 79.6\scriptsize$\pm 0.0$
&64.8 & \textbf{61.5}\scriptsize$\pm 0.0$
&88.1 & 66.1\scriptsize$\pm 0.0$
&93.4 & 83.3\scriptsize$\pm 0.0$
\\
&NN2-$\chi^2$-DORO    &78.9 & 78.4\scriptsize$\pm 1.3$
&75.8 & 72.7\scriptsize$\pm 3.7$
&82.4 & 73.8\scriptsize$\pm 1.0$
&78.7 & 71.1\scriptsize$\pm 5.9$
&76.1 & 78.9\scriptsize$\pm 6.7$
&67.0 & 62.2\scriptsize$\pm 1.8$
&91.3 & 79.2\scriptsize$\pm 0.0$
&63.9 & 60.6\scriptsize$\pm 0.0$
&88.6 & 68.3\scriptsize$\pm 0.0$
&93.2 & 83.0\scriptsize$\pm 0.0$
\\
&NN3-CVaR-DRO    &79.8 & 78.5\scriptsize$\pm 1.5$
&77.4 & 73.4\scriptsize$\pm 3.4$
&84.4 & 74.5\scriptsize$\pm 1.4$
&82.1 & 74.3\scriptsize$\pm 6.5$
&72.2 & \textbf{86.4}\scriptsize$\pm 9.0$
&76.1 & 69.8\scriptsize$\pm 2.3$
&91.5 & 80.4\scriptsize$\pm 0.0$
&64.9 & 60.8\scriptsize$\pm 0.0$
&88.6 & 63.8\scriptsize$\pm 0.0$
&94.0 & 83.1\scriptsize$\pm 0.0$
\\
&NN3-$\chi^2$-DRO    &80.3 & 78.6\scriptsize$\pm 1.6$
&76.1 & 73.4\scriptsize$\pm 3.5$
&83.2 & 73.9\scriptsize$\pm 0.7$
&80.5 & 74.2\scriptsize$\pm 5.2$
&75.4 & 85.4\scriptsize$\pm 8.3$
&77.7 & \textbf{70.5}\scriptsize$\pm 2.2$
&91.7 & 80.1\scriptsize$\pm 0.0$
&65.3 & \textbf{61.5}\scriptsize$\pm 0.0$
&88.7 & 64.1\scriptsize$\pm 0.0$
&93.8 & \textbf{83.5}\scriptsize$\pm 0.0$
\\
&NN4-CVaR-DRO    &80.8 & 78.3\scriptsize$\pm 1.5$
&75.8 & 73.4\scriptsize$\pm 3.6$
&84.5 & 73.6\scriptsize$\pm 1.3$
&81.6 & 73.0\scriptsize$\pm 6.5$
&82.2 & 84.1\scriptsize$\pm 7.1$
&76.8 & 70.0\scriptsize$\pm 2.4$
&91.5 & 80.2\scriptsize$\pm 0.0$
&65.0 & 61.1\scriptsize$\pm 0.0$
&88.7 & \textbf{68.6}\scriptsize$\pm 0.0$
&93.8 & 83.1\scriptsize$\pm 0.0$
\\
&NN4-$\chi^2$-DRO    &80.4 & \textbf{78.9}\scriptsize$\pm 1.3$
&75.8 & 73.4\scriptsize$\pm 3.6$
&84.0 & 71.8\scriptsize$\pm 0.9$
&81.5 & 72.8\scriptsize$\pm 6.4$
&84.4 & 83.6\scriptsize$\pm 7.0$
&76.9 & 69.9\scriptsize$\pm 2.0$
&91.1 & 80.1\scriptsize$\pm 0.0$
&64.6 & 61.4\scriptsize$\pm 0.0$
&88.8 & 67.5\scriptsize$\pm 0.0$
&93.3 & 83.1\scriptsize$\pm 0.0$
\\\midrule
\multirow{4}{*}{\begin{tabular}[c]{@{}l@{}}Tree-DRO\\ Methods\\ (base: XGB/LGBM)\end{tabular}}   &XGB-CVaR-DRO    &81.1 & 79.0\scriptsize$\pm 1.4$
&75.8 & \textbf{73.4}\scriptsize$\pm 3.6$
&82.7 & \textbf{75.2}\scriptsize$\pm 1.0$
&85.3 & \textbf{75.8}\scriptsize$\pm 6.5$
&81.4 & 77.6\scriptsize$\pm 6.4$
&79.1 & \textbf{70.4}\scriptsize$\pm 2.6$
&92.1 & \textbf{80.6}\scriptsize$\pm 0.0$
&64.7 & 59.1\scriptsize$\pm 0.0$
&89.6 & 62.8\scriptsize$\pm 0.0$
&94.7 & 85.2\scriptsize$\pm 0.0$
\\
&XGB-KL-DRO    &80.6 & 79.1\scriptsize$\pm 1.3$
&75.8 & \textbf{73.4}\scriptsize$\pm 3.6$
&81.3 & 74.7\scriptsize$\pm 0.8$
&85.1 & 75.7\scriptsize$\pm 6.9$
&67.6 & \textbf{80.0}\scriptsize$\pm 7.7$
&79.2 & 70.2\scriptsize$\pm 2.5$
&92.0 & 80.0\scriptsize$\pm 0.0$
&62.6 & \textbf{60.4}\scriptsize$\pm 0.0$
&80.1 & 64.1\scriptsize$\pm 0.0$
&94.8 & \textbf{85.9}\scriptsize$\pm 0.0$
\\
&LGBM-CVaR-DRO    &80.7 & \textbf{79.2}\scriptsize$\pm 1.3$
&78.1 & 73.0\scriptsize$\pm 2.7$
&75.8 & 73.7\scriptsize$\pm 1.1$
&85.1 & 75.6\scriptsize$\pm 6.6$
&85.6 & 77.2\scriptsize$\pm 8.1$
&78.9 & 70.3\scriptsize$\pm 2.5$
&92.0 & 79.4\scriptsize$\pm 0.0$
&64.7 & 58.7\scriptsize$\pm 0.0$
&89.7 & \textbf{65.0}\scriptsize$\pm 0.0$
&94.7 & 85.5\scriptsize$\pm 0.0$
\\
&LGBM-KL-DRO    &80.1 & 79.2\scriptsize$\pm 1.4$
&75.8 & \textbf{73.4}\scriptsize$\pm 3.6$
&74.8 & 73.1\scriptsize$\pm 1.0$
&85.0 & 75.5\scriptsize$\pm 6.7$
&68.3 & 78.9\scriptsize$\pm 8.6$
&79.2 & 70.2\scriptsize$\pm 2.5$
&92.0 & 79.6\scriptsize$\pm 0.0$
&64.4 & 58.6\scriptsize$\pm 0.0$
&89.3 & 64.8\scriptsize$\pm 0.0$
&94.7 & 85.5\scriptsize$\pm 0.0$
\\\midrule
\multirow{4}{*}{\begin{tabular}[c]{@{}l@{}}Kernel-DRO\\ Methods\\ (base: Kernel)\end{tabular}}   &Kernel-$\chi2$-DRO    &81.3 & 77.5\scriptsize$\pm 2.1$
&76.4 & 73.0\scriptsize$\pm 3.6$
&85.4 & 74.2\scriptsize$\pm 1.6$
&85.2 & 76.4\scriptsize$\pm 6.7$
&82.3 & 75.8\scriptsize$\pm 7.8$
&78.4 & \textbf{71.3}\scriptsize$\pm 1.9$
&91.7 & 80.0\scriptsize$\pm 0.0$
&65.6 & \textbf{61.0}\scriptsize$\pm 0.0$
&88.6 & 60.3\scriptsize$\pm 0.0$
&93.9 & 83.3\scriptsize$\pm 0.0$
\\
&Kernel-CVaR-DRO    &81.2 & \textbf{77.7}\scriptsize$\pm 2.0$
&78.1 & \textbf{73.0}\scriptsize$\pm 3.3$
&86.3 & 73.0\scriptsize$\pm 1.1$
&84.3 & 76.1\scriptsize$\pm 6.3$
&82.8 & \textbf{76.4}\scriptsize$\pm 7.6$
&78.2 & 70.5\scriptsize$\pm 2.4$
&91.9 & 80.4\scriptsize$\pm 0.0$
&64.9 & 60.8\scriptsize$\pm 0.0$
&88.1 & 63.9\scriptsize$\pm 0.0$
&93.5 & \textbf{83.7}\scriptsize$\pm 0.0$
\\
&Kernel-KL-DRO    &81.5 & 77.5\scriptsize$\pm 2.2$
&76.5 & 73.0\scriptsize$\pm 3.6$
&84.9 & 74.0\scriptsize$\pm 1.3$
&84.9 & \textbf{76.4}\scriptsize$\pm 6.3$
&82.0 & 75.3\scriptsize$\pm 7.7$
&78.2 & 70.8\scriptsize$\pm 2.2$
&91.7 & \textbf{80.5}\scriptsize$\pm 0.0$
&65.3 & 61.0\scriptsize$\pm 0.0$
&88.4 & \textbf{64.9}\scriptsize$\pm 0.0$
&93.9 & 83.1\scriptsize$\pm 0.0$
\\
&Kernel-Wasserstein-DRO    &81.1 & 77.5\scriptsize$\pm 2.0$
&77.2 & 72.8\scriptsize$\pm 3.6$
&84.1 & \textbf{74.4}\scriptsize$\pm 1.6$
&84.5 & 75.5\scriptsize$\pm 7.1$
&82.4 & 75.3\scriptsize$\pm 7.8$
&77.7 & 69.2\scriptsize$\pm 2.6$
&91.5 & 79.7\scriptsize$\pm 0.0$
&65.3 & 59.9\scriptsize$\pm 0.0$
&88.6 & 61.8\scriptsize$\pm 0.0$
&93.7 & 83.5\scriptsize$\pm 0.0$
\\\midrule
\multirow{6}{*}{\begin{tabular}[c]{@{}l@{}}Imbalanced\\ Learning\\ \& Fairness\\ Methods \\(base: XGB)\end{tabular}} &SUBY    &80.7 & 75.3\scriptsize$\pm 2.9$
&75.8 & 72.7\scriptsize$\pm 3.7$
&83.7 & 73.0\scriptsize$\pm 1.5$
&77.4 & 72.5\scriptsize$\pm 3.0$
&85.4 & \textbf{86.5}\scriptsize$\pm 7.6$
&73.8 & 70.8\scriptsize$\pm 1.3$
&92.0 & \textbf{80.7}\scriptsize$\pm 0.0$
&65.1 & \textbf{62.8}\scriptsize$\pm 0.0$
&87.2 & \textbf{63.6}\scriptsize$\pm 0.0$
&88.9 & 75.9\scriptsize$\pm 0.0$
\\
&RWY    &81.0 & 75.4\scriptsize$\pm 3.0$
&75.0 & 72.2\scriptsize$\pm 3.7$
&76.2 & \textbf{75.1}\scriptsize$\pm 0.9$
&81.3 & 74.6\scriptsize$\pm 4.0$
&85.3 & 86.4\scriptsize$\pm 7.4$
&74.9 & 70.6\scriptsize$\pm 1.4$
&92.0 & \textbf{80.7}\scriptsize$\pm 0.0$
&65.1 & \textbf{62.8}\scriptsize$\pm 0.0$
&87.8 & 63.1\scriptsize$\pm 0.0$
&94.2 & 84.1\scriptsize$\pm 0.0$
\\
&SUBG    &81.5 & \textbf{77.7}\scriptsize$\pm 2.0$
&78.9 & 73.5\scriptsize$\pm 3.1$
&-&-
&85.5 & \textbf{76.5}\scriptsize$\pm 6.0$
&-&-
&79.4 & 71.6\scriptsize$\pm 2.3$
&92.0 & \textbf{80.7}\scriptsize$\pm 0.0$
&58.4 & 56.3\scriptsize$\pm 0.0$
&-&-
&93.9 & 85.0\scriptsize$\pm 0.0$
\\
&RWG    &81.7 & 77.6\scriptsize$\pm 2.1$
&80.2 & 73.6\scriptsize$\pm 3.1$
&-&-
&85.9 & 76.4\scriptsize$\pm 6.5$
&-&-
&79.7 & 71.6\scriptsize$\pm 2.5$
&92.0 & \textbf{80.7}\scriptsize$\pm 0.0$
&65.7 & 62.2\scriptsize$\pm 0.0$
&-&-
&94.7 & 84.7\scriptsize$\pm 0.0$
\\
&In-processing    &81.7 & 77.6\scriptsize$\pm 2.1$
&80.4 & \textbf{73.8}\scriptsize$\pm 2.9$
&-&-
&86.1 & 76.5\scriptsize$\pm 6.5$
&-&-
&79.4 & \textbf{71.8}\scriptsize$\pm 2.3$
&92.3 & 80.5\scriptsize$\pm 0.0$
&66.1 & 62.7\scriptsize$\pm 0.0$
&-&-
&94.7 & \textbf{85.9}\scriptsize$\pm 0.0$
\\
&Post-processing    &81.3 & 76.8\scriptsize$\pm 2.2$
&79.7 & 73.6\scriptsize$\pm 3.0$
&-&-
&85.6 & 76.5\scriptsize$\pm 6.4$
&-&-
&79.4 & 71.7\scriptsize$\pm 2.3$
&91.7 & 80.1\scriptsize$\pm 0.0$
&65.5 & \textbf{62.8}\scriptsize$\pm 0.0$
&-&-
&87.9 & 85.5\scriptsize$\pm 0.0$
\\
 \bottomrule \bottomrule

\end{tabular}}
\end{table}
\end{landscape}
\begin{landscape}
\begin{table}[!htb]
\caption{Results (Macro-F1 Score) of 10 settings in \Cref{subsec:benchmark-empirical-result} with all 172 target domains, where we report the source macro F1-score, the mean target macro F1-score and standard deviation across multiple target domains in each setting. We run each method with its best configuration (according to the target F1-score), and boldface the best target performance within each class of methods in each setting.}
\vspace{5pt}
\label{table:selected_results_to_all_f1}
\resizebox{21cm}{!}{
\begin{tabular}{@{}llcccccccccccccccccccc@{}}
\toprule\toprule
\multicolumn{2}{l}{\large \textbf{Dataset}}                                                                   & \multicolumn{2}{c}{\large \texttt{ACS Income}}                                                                    & \multicolumn{2}{c}{\large \texttt{ACS Mobility}}                                                                  & \multicolumn{2}{c}{\large \texttt{US Taxi}}                 & \multicolumn{2}{c}{\large \texttt{ACS Pub.Cov}}                                                                   & \multicolumn{2}{c}{\large \texttt{US Accident}}                                                                   & \multicolumn{2}{c}{\large \texttt{ACS Time}}                                                                      & \multicolumn{2}{c}{\large \texttt{Sub-Sampling}}   & \multicolumn{2}{c}{\large \texttt{diabetes}} &  \multicolumn{2}{c}{\large \texttt{assistments}} & \multicolumn{2}{c}{\large \texttt{college}}             \\ 
\multicolumn{2}{l}{\large \textbf{Shift Pattern}}                                                            & \multicolumn{2}{c}{$Y|X$ dominates}                                                                              & \multicolumn{2}{c}{ $Y|X$ dominates}                                                                              & \multicolumn{2}{c}{ $Y|X$ dominates}                       & \multicolumn{2}{c}{ $Y|X$ more}                                                                            & \multicolumn{2}{c}{$Y|X$ more}                                                                              & \multicolumn{2}{c}{$X$ more}                                                                            & \multicolumn{2}{c}{$X$ dominates}  & \multicolumn{2}{c}{$Y|X$ dominates} & \multicolumn{2}{c}{$Y|X$ more} & \multicolumn{2}{c}{$Y|X$ more} \\
\multicolumn{2}{l}{\large Source $\rightarrow$ Target Pair}                                                       & \multicolumn{2}{c}{CA$\rightarrow$ 50 Domains}                                                                        & \multicolumn{2}{c}{MS$\rightarrow$50 Domains}                                                                        & \multicolumn{2}{c}{NYC$\rightarrow$3 Domains}                & \multicolumn{2}{c}{NE$\rightarrow$50 Domains}                                                                        & \multicolumn{2}{c}{CA$\rightarrow$13 Domains}                                                                        & \multicolumn{2}{c}{2010$\rightarrow$3 Domains}                                                                    & \multicolumn{2}{c}{Young$\rightarrow$Old}   & \multicolumn{2}{c}{White$\rightarrow$Others} & \multicolumn{2}{c}{700$\rightarrow$1} & \multicolumn{2}{c}{Normal $\rightarrow$ Special}  \\
\multicolumn{2}{l}{}                                                                          & $i.d.$                                         & $o.o.d$                                         & $i.d.$                                         & $o.o.d$                                         & $i.d.$              & $o.o.d$              & $i.d.$                                         & $o.o.d$                                         & $i.d.$                                         & $o.o.d$                                         & $i.d.$                                         & $o.o.d$                                         & $i.d.$                                         & $o.o.d$                           & $i.d.$                                         & $o.o.d$                                         & $i.d.$                                         & $o.o.d$                                         & $i.d.$                                         & $o.o.d$                                         \\\midrule
\multirow{6}{*}{\begin{tabular}[c]{@{}l@{}}Basic \\ Methods\end{tabular}} &LR    &79.8 & 75.6\scriptsize$\pm 3.1$
&55.8 & 50.2\scriptsize$\pm 2.3$
&81.1 & 73.0\scriptsize$\pm 2.1$
&66.4 & 62.7\scriptsize$\pm 4.7$
&73.3 & 65.4\scriptsize$\pm 8.6$
&46.8 & 43.0\scriptsize$\pm 0.9$
&79.5 & 77.7\scriptsize$\pm 0.0$
&61.1 & 58.1\scriptsize$\pm 0.0$
&83.6 & 42.4\scriptsize$\pm 0.0$
&84.1 & 77.0\scriptsize$\pm 0.0$
\\
&SVM    &80.0 & 75.4\scriptsize$\pm 3.2$
&60.2 & 53.6\scriptsize$\pm 2.2$
&79.6 & 73.3\scriptsize$\pm 2.2$
&65.4 & 61.6\scriptsize$\pm 4.2$
&73.5 & 65.8\scriptsize$\pm 8.5$
&65.5 & 60.3\scriptsize$\pm 1.1$
&78.7 & 77.8\scriptsize$\pm 0.0$
&60.8 & 57.8\scriptsize$\pm 0.0$
&83.5 & 35.8\scriptsize$\pm 0.0$
&83.9 & 76.6\scriptsize$\pm 0.0$
\\
&Kernel-SVM    &79.5 & 75.3\scriptsize$\pm 2.9$
&58.7 & 56.8\scriptsize$\pm 2.2$
&77.2 & 70.8\scriptsize$\pm 1.7$
&58.7 & 60.1\scriptsize$\pm 1.9$
&77.2 & 81.1\scriptsize$\pm 6.8$
&55.3 & 56.3\scriptsize$\pm 0.6$
&80.1 & 76.9\scriptsize$\pm 0.0$
&63.2 & \textbf{61.1}\scriptsize$\pm 0.0$
&77.0 & 62.8\scriptsize$\pm 0.0$
&72.9 & 74.3\scriptsize$\pm 0.0$
\\
&NN2    &79.7 & 76.2\scriptsize$\pm 2.9$
&62.7 & \textbf{61.0}\scriptsize$\pm 1.9$
&78.6 & 72.8\scriptsize$\pm 1.7$
&68.4 & \textbf{67.3}\scriptsize$\pm 2.3$
&81.0 & 80.6\scriptsize$\pm 7.0$
&70.0 & \textbf{66.8}\scriptsize$\pm 1.1$
&80.2 & 78.0\scriptsize$\pm 0.0$
&60.4 & 59.9\scriptsize$\pm 0.0$
&85.1 & \textbf{65.3}\scriptsize$\pm 0.0$
&84.6 & 79.1\scriptsize$\pm 0.0$
\\
&NN3    &78.3 & 76.1\scriptsize$\pm 2.5$
&53.7 & 50.0\scriptsize$\pm 2.6$
&82.6 & 72.4\scriptsize$\pm 1.6$
&67.6 & 64.7\scriptsize$\pm 3.1$
&81.5 & 80.9\scriptsize$\pm 6.9$
&68.9 & 64.4\scriptsize$\pm 1.5$
&79.6 & \textbf{78.2}\scriptsize$\pm 0.0$
&62.0 & 60.4\scriptsize$\pm 0.0$
&84.7 & 64.6\scriptsize$\pm 0.0$
&85.2 & 79.2\scriptsize$\pm 0.0$
\\
&NN4    &80.0 & \textbf{76.3}\scriptsize$\pm 3.1$
&53.0 & 46.4\scriptsize$\pm 2.2$
&81.5 & \textbf{73.6}\scriptsize$\pm 1.8$
&64.0 & 61.4\scriptsize$\pm 3.8$
&83.2 & \textbf{83.0}\scriptsize$\pm 7.3$
&68.2 & 65.2\scriptsize$\pm 1.3$
&78.7 & 77.6\scriptsize$\pm 0.0$
&63.1 & 60.6\scriptsize$\pm 0.0$
&84.4 & 61.2\scriptsize$\pm 0.0$
&84.7 & \textbf{80.0}\scriptsize$\pm 0.0$
\\\midrule
\multirow{4}{*}{\begin{tabular}[c]{@{}l@{}}Tree-based \\ Ensemble \\ Methods\end{tabular}}   &RF    &80.5 & 76.0\scriptsize$\pm 3.2$
&65.1 & 56.8\scriptsize$\pm 1.8$
&80.6 & \textbf{73.9}\scriptsize$\pm 2.0$
&71.4 & 64.2\scriptsize$\pm 4.6$
&84.0 & 75.5\scriptsize$\pm 7.8$
&69.9 & 65.5\scriptsize$\pm 1.6$
&77.6 & 78.4\scriptsize$\pm 0.0$
&62.3 & 61.7\scriptsize$\pm 0.0$
&84.7 & \textbf{66.2}\scriptsize$\pm 0.0$
&86.0 & 80.3\scriptsize$\pm 0.0$
\\
&XGB    &81.1 & 76.1\scriptsize$\pm 3.5$
&64.9 & 59.0\scriptsize$\pm 2.0$
&80.5 & 72.7\scriptsize$\pm 1.6$
&72.6 & 66.4\scriptsize$\pm 4.1$
&85.1 & \textbf{78.1}\scriptsize$\pm 7.8$
&71.2 & 66.9\scriptsize$\pm 1.8$
&79.7 & 78.5\scriptsize$\pm 0.0$
&63.4 & \textbf{61.9}\scriptsize$\pm 0.0$
&85.0 & 62.8\scriptsize$\pm 0.0$
&87.8 & 82.2\scriptsize$\pm 0.0$
\\
&GBM    &81.2 & 76.2\scriptsize$\pm 3.4$
&66.3 & 59.5\scriptsize$\pm 2.1$
&65.5 & 73.8\scriptsize$\pm 1.3$
&73.7 & \textbf{66.6}\scriptsize$\pm 4.2$
&83.8 & 76.3\scriptsize$\pm 7.6$
&71.0 & \textbf{67.3}\scriptsize$\pm 1.9$
&80.0 & \textbf{79.2}\scriptsize$\pm 0.0$
&61.1 & 60.5\scriptsize$\pm 0.0$
&69.4 & 64.6\scriptsize$\pm 0.0$
&87.3 & 81.8\scriptsize$\pm 0.0$
\\
&LGBM    &81.2 & \textbf{76.2}\scriptsize$\pm 3.3$
&66.1 & 59.0\scriptsize$\pm 2.0$
&78.2 & 74.2\scriptsize$\pm 1.6$
&73.2 & 66.3\scriptsize$\pm 4.1$
&85.1 & 77.7\scriptsize$\pm 7.7$
&71.4 & 66.7\scriptsize$\pm 1.9$
&79.9 & 78.1\scriptsize$\pm 0.0$
&63.5 & 62.0\scriptsize$\pm 0.0$
&84.4 & 65.7\scriptsize$\pm 0.0$
&87.4 & \textbf{82.9}\scriptsize$\pm 0.0$
\\\midrule
\multirow{13}{*}{\begin{tabular}[c]{@{}l@{}}Linear-DRO \\  Methods \\ (base: SVM)\end{tabular}}   &KL-DRO    &79.9 & 75.3\scriptsize$\pm 3.5$
&61.1 & 55.3\scriptsize$\pm 2.2$
&80.2 & 73.0\scriptsize$\pm 2.3$
&66.5 & 62.6\scriptsize$\pm 4.3$
&74.2 & 67.3\scriptsize$\pm 8.0$
&65.4 & 60.7\scriptsize$\pm 1.3$
&78.2 & 77.6\scriptsize$\pm 0.0$
&60.8 & 57.9\scriptsize$\pm 0.0$
&83.6 & 38.5\scriptsize$\pm 0.0$
&84.2 & 76.6\scriptsize$\pm 0.0$
\\
&CVaR-DRO    &80.0 & \textbf{75.8}\scriptsize$\pm 3.2$
&62.6 & 59.3\scriptsize$\pm 2.2$
&81.4 & 74.4\scriptsize$\pm 1.7$
&68.8 & 65.3\scriptsize$\pm 3.4$
&74.8 & 67.5\scriptsize$\pm 8.2$
&66.9 & 62.9\scriptsize$\pm 1.2$
&79.4 & \textbf{78.8}\scriptsize$\pm 0.0$
&62.0 & 59.5\scriptsize$\pm 0.0$
&83.3 & 45.8\scriptsize$\pm 0.0$
&84.4 & 76.9\scriptsize$\pm 0.0$
\\
&$\chi^2$-DRO    &79.9 & 75.7\scriptsize$\pm 3.2$
&60.5 & 56.8\scriptsize$\pm 2.1$
&80.7 & 74.3\scriptsize$\pm 2.0$
&67.6 & 63.8\scriptsize$\pm 4.1$
&74.2 & 67.3\scriptsize$\pm 8.1$
&66.8 & 63.1\scriptsize$\pm 1.2$
&78.7 & 77.1\scriptsize$\pm 0.0$
&61.5 & 58.6\scriptsize$\pm 0.0$
&83.6 & 40.3\scriptsize$\pm 0.0$
&84.3 & 76.6\scriptsize$\pm 0.0$
\\
&TV-DRO    &79.9 & 75.1\scriptsize$\pm 3.3$
&61.4 & 56.8\scriptsize$\pm 2.3$
&79.0 & 72.6\scriptsize$\pm 2.2$
&65.7 & 61.7\scriptsize$\pm 5.9$
&74.3 & 67.3\scriptsize$\pm 8.1$
&67.0 & 64.4\scriptsize$\pm 1.2$
&77.2 & 77.5\scriptsize$\pm 0.0$
&59.7 & 58.0\scriptsize$\pm 0.0$
&82.7 & 44.3\scriptsize$\pm 0.0$
&84.2 & 76.4\scriptsize$\pm 0.0$
\\
&Wasserstein-DRO    &79.8 & 75.2\scriptsize$\pm 3.3$
&57.3 & 48.4\scriptsize$\pm 2.9$
&80.6 & 73.2\scriptsize$\pm 2.1$
&64.3 & 61.2\scriptsize$\pm 4.6$
&74.3 & 67.3\scriptsize$\pm 8.1$
&62.9 & 58.8\scriptsize$\pm 1.5$
&77.6 & 75.4\scriptsize$\pm 0.0$
&57.3 & 56.0\scriptsize$\pm 0.0$
&83.4 & 33.1\scriptsize$\pm 0.0$
&83.6 & 77.1\scriptsize$\pm 0.0$
\\
&Aug. Wass.-DRO    &79.8 & 75.2\scriptsize$\pm 3.3$
&63.0 & 59.1\scriptsize$\pm 1.7$
&78.9 & 73.1\scriptsize$\pm 2.0$
&68.5 & \textbf{66.9}\scriptsize$\pm 2.7$
&69.4 & \textbf{77.7}\scriptsize$\pm 7.9$
&65.2 & 61.7\scriptsize$\pm 1.2$
&78.7 & 78.3\scriptsize$\pm 0.0$
&61.3 & 59.4\scriptsize$\pm 0.0$
&64.1 & \textbf{55.2}\scriptsize$\pm 0.0$
&83.7 & 76.9\scriptsize$\pm 0.0$
\\
&Satis. Wass.-DRO    &78.6 & 74.5\scriptsize$\pm 3.0$
&43.1 & 42.1\scriptsize$\pm 1.2$
&73.1 & 66.9\scriptsize$\pm 2.3$
&44.5 & 41.0\scriptsize$\pm 2.9$
&70.0 & 61.9\scriptsize$\pm 8.2$
&41.7 & 37.9\scriptsize$\pm 1.1$
&75.9 & 75.6\scriptsize$\pm 0.0$
&36.9 & 33.6\scriptsize$\pm 0.0$
&82.6 & 32.9\scriptsize$\pm 0.0$
&82.1 & 76.1\scriptsize$\pm 0.0$
\\
&Sinkhorn-DRO    &78.1 & 75.4\scriptsize$\pm 2.5$
&55.9 & 55.5\scriptsize$\pm 1.5$
&81.0 & 74.5\scriptsize$\pm 1.8$
&69.9 & 67.9\scriptsize$\pm 2.5$
&75.7 & 66.9\scriptsize$\pm 8.3$
&68.3 & \textbf{66.6}\scriptsize$\pm 0.5$
&79.4 & 78.3\scriptsize$\pm 0.0$
&56.9 & 56.3\scriptsize$\pm 0.0$
&78.9 & 42.6\scriptsize$\pm 0.0$
&82.8 & \textbf{78.4}\scriptsize$\pm 0.0$
\\
&Unified-DRO($L_2$)    &79.8 & 75.2\scriptsize$\pm 3.3$
&57.3 & 48.4\scriptsize$\pm 2.9$
&79.9 & 72.9\scriptsize$\pm 2.2$
&64.3 & 61.2\scriptsize$\pm 4.6$
&73.9 & 66.7\scriptsize$\pm 8.2$
&62.9 & 58.8\scriptsize$\pm 1.5$
&78.7 & 77.5\scriptsize$\pm 0.0$
&52.1 & 52.5\scriptsize$\pm 0.0$
&83.4 & 33.1\scriptsize$\pm 0.0$
&84.0 & 76.7\scriptsize$\pm 0.0$
\\
&Unified-DRO($L_\text{inf}$)    &79.9 & 75.2\scriptsize$\pm 3.3$
&57.3 & 48.4\scriptsize$\pm 2.9$
&80.1 & 73.4\scriptsize$\pm 2.1$
&64.4 & 61.2\scriptsize$\pm 4.6$
&72.6 & 64.9\scriptsize$\pm 8.3$
&62.9 & 58.8\scriptsize$\pm 1.5$
&78.9 & 77.3\scriptsize$\pm 0.0$
&57.4 & 56.0\scriptsize$\pm 0.0$
&83.4 & 33.1\scriptsize$\pm 0.0$
&84.1 & 76.6\scriptsize$\pm 0.0$
\\
&Marginal-DRO    &80.1 & 75.7\scriptsize$\pm 3.2$
&62.4 & 59.3\scriptsize$\pm 2.0$
&80.0 & 74.3\scriptsize$\pm 1.8$
&69.2 & 65.9\scriptsize$\pm 3.7$
&74.7 & 67.9\scriptsize$\pm 8.2$
&67.5 & 65.2\scriptsize$\pm 1.0$
&78.0 & 77.1\scriptsize$\pm 0.0$
&63.0 & 59.9\scriptsize$\pm 0.0$
&83.9 & 53.8\scriptsize$\pm 0.0$
&84.2 & 76.7\scriptsize$\pm 0.0$
\\
&Conditional-DRO    &80.0 & 75.7\scriptsize$\pm 3.3$
&60.6 & \textbf{60.1}\scriptsize$\pm 1.4$
&80.5 & \textbf{74.7}\scriptsize$\pm 1.6$
&69.2 & 66.2\scriptsize$\pm 3.7$
&75.0 & 77.3\scriptsize$\pm 8.7$
&67.2 & 65.4\scriptsize$\pm 0.8$
&79.9 & 77.4\scriptsize$\pm 0.0$
&62.0 & \textbf{60.3}\scriptsize$\pm 0.0$
&83.4 & 33.4\scriptsize$\pm 0.0$
&84.1 & 76.5\scriptsize$\pm 0.0$
\\
&Holistic-DRO    &77.4 & 74.2\scriptsize$\pm 2.5$
&59.3 & 52.6\scriptsize$\pm 2.4$
&74.6 & 68.1\scriptsize$\pm 2.5$
&63.2 & 60.1\scriptsize$\pm 4.6$
&72.2 & 63.5\scriptsize$\pm 8.9$
&64.4 & 59.4\scriptsize$\pm 1.5$
&75.3 & 73.9\scriptsize$\pm 0.0$
&59.3 & 56.5\scriptsize$\pm 0.0$
&83.4 & 33.4\scriptsize$\pm 0.0$
&82.1 & 74.2\scriptsize$\pm 0.0$
\\\midrule
\multirow{8}{*}{\begin{tabular}[c]{@{}l@{}}NN-DRO\\ Methods\\ (base: NN)\end{tabular}}   &NN2-CVaR-DRO    &80.1 & 76.2\scriptsize$\pm 3.0$
&61.7 & \textbf{61.0}\scriptsize$\pm 1.2$
&77.9 & 73.9\scriptsize$\pm 1.5$
&66.8 & 65.7\scriptsize$\pm 2.6$
&79.9 & 83.5\scriptsize$\pm 7.3$
&67.8 & \textbf{65.8}\scriptsize$\pm 1.5$
&79.0 & \textbf{78.7}\scriptsize$\pm 0.0$
&61.0 & 60.5\scriptsize$\pm 0.0$
&83.9 & 67.7\scriptsize$\pm 0.0$
&85.9 & \textbf{79.8}\scriptsize$\pm 0.0$
\\
&NN2-$\chi^2$-DRO    &80.2 & \textbf{76.4}\scriptsize$\pm 3.1$
&63.0 & 60.2\scriptsize$\pm 2.0$
&82.1 & \textbf{74.1}\scriptsize$\pm 1.7$
&68.0 & \textbf{66.0}\scriptsize$\pm 3.7$
&80.8 & 82.1\scriptsize$\pm 7.2$
&68.7 & 64.9\scriptsize$\pm 1.6$
&80.6 & 78.6\scriptsize$\pm 0.0$
&62.3 & 60.5\scriptsize$\pm 0.0$
&84.3 & 65.0\scriptsize$\pm 0.0$
&85.4 & 79.7\scriptsize$\pm 0.0$
\\
&NN2-CVaR-DORO    &79.0 & 76.1\scriptsize$\pm 2.8$
&60.5 & 56.3\scriptsize$\pm 2.3$
&78.9 & 74.0\scriptsize$\pm 1.4$
&68.5 & 65.8\scriptsize$\pm 3.9$
&79.4 & 76.4\scriptsize$\pm 8.2$
&63.4 & 61.7\scriptsize$\pm 1.1$
&79.5 & 77.6\scriptsize$\pm 0.0$
&62.5 & 61.0\scriptsize$\pm 0.0$
&84.6 & 65.7\scriptsize$\pm 0.0$
&84.7 & 78.9\scriptsize$\pm 0.0$
\\
&NN2-$\chi^2$-DORO    &78.9 & 75.6\scriptsize$\pm 2.9$
&52.7 & 53.7\scriptsize$\pm 1.5$
&79.7 & 73.0\scriptsize$\pm 1.6$
&58.0 & 58.1\scriptsize$\pm 2.1$
&75.3 & 77.0\scriptsize$\pm 6.5$
&62.5 & 60.2\scriptsize$\pm 1.3$
&79.2 & 77.7\scriptsize$\pm 0.0$
&59.9 & 59.3\scriptsize$\pm 0.0$
&84.6 & \textbf{68.3}\scriptsize$\pm 0.0$
&83.1 & 79.2\scriptsize$\pm 0.0$
\\
&NN3-CVaR-DRO    &80.3 & 76.2\scriptsize$\pm 3.1$
&59.5 & 58.4\scriptsize$\pm 1.9$
&81.7 & 73.4\scriptsize$\pm 2.0$
&64.9 & 62.9\scriptsize$\pm 3.2$
&71.8 & \textbf{85.1}\scriptsize$\pm 8.6$
&69.0 & 65.4\scriptsize$\pm 1.6$
&79.2 & 78.7\scriptsize$\pm 0.0$
&61.6 & 60.0\scriptsize$\pm 0.0$
&84.3 & 63.3\scriptsize$\pm 0.0$
&85.3 & 79.3\scriptsize$\pm 0.0$
\\
&NN3-$\chi^2$-DRO    &79.3 & 76.2\scriptsize$\pm 2.7$
&62.4 & 56.1\scriptsize$\pm 2.0$
&80.6 & 73.0\scriptsize$\pm 1.6$
&66.0 & 64.6\scriptsize$\pm 3.2$
&74.9 & 84.0\scriptsize$\pm 8.0$
&68.3 & 65.7\scriptsize$\pm 1.4$
&79.5 & 78.3\scriptsize$\pm 0.0$
&63.3 & 61.3\scriptsize$\pm 0.0$
&84.4 & 64.1\scriptsize$\pm 0.0$
&85.2 & 79.5\scriptsize$\pm 0.0$
\\
&NN4-CVaR-DRO    &79.6 & 76.1\scriptsize$\pm 2.8$
&55.1 & 53.2\scriptsize$\pm 1.9$
&82.1 & 72.1\scriptsize$\pm 1.8$
&63.0 & 63.4\scriptsize$\pm 2.6$
&81.0 & 82.4\scriptsize$\pm 7.2$
&69.1 & 65.0\scriptsize$\pm 1.7$
&79.1 & 78.4\scriptsize$\pm 0.0$
&63.5 & 60.7\scriptsize$\pm 0.0$
&84.6 & 67.0\scriptsize$\pm 0.0$
&85.0 & 79.5\scriptsize$\pm 0.0$
\\
&NN4-$\chi^2$-DRO    &79.6 & 76.2\scriptsize$\pm 2.8$
&57.8 & 55.0\scriptsize$\pm 2.0$
&80.8 & 70.5\scriptsize$\pm 1.4$
&58.8 & 61.4\scriptsize$\pm 1.8$
&80.1 & 81.9\scriptsize$\pm 7.4$
&67.2 & 64.7\scriptsize$\pm 1.4$
&78.7 & 78.1\scriptsize$\pm 0.0$
&63.7 & \textbf{61.1}\scriptsize$\pm 0.0$
&84.6 & 66.3\scriptsize$\pm 0.0$
&84.9 & 79.4\scriptsize$\pm 0.0$
\\\midrule
\multirow{4}{*}{\begin{tabular}[c]{@{}l@{}}Tree-DRO\\ Methods\\ (base: XGB/LGBM)\end{tabular}}  &XGB-CVaR-DRO    &79.8 & \textbf{76.6}\scriptsize$\pm 2.7$
&67.2 & 61.8\scriptsize$\pm 2.0$
&80.5 & \textbf{74.7}\scriptsize$\pm 1.4$
&70.7 & 64.7\scriptsize$\pm 4.1$
&78.1 & 75.1\scriptsize$\pm 6.0$
&68.3 & 64.2\scriptsize$\pm 1.4$
&78.2 & 77.9\scriptsize$\pm 0.0$
&59.3 & 57.4\scriptsize$\pm 0.0$
&86.8 & 62.6\scriptsize$\pm 0.0$
&87.2 & 81.7\scriptsize$\pm 0.0$
\\
&XGB-KL-DRO    &79.2 & 76.5\scriptsize$\pm 2.6$
&66.4 & 61.8\scriptsize$\pm 1.8$
&79.8 & 74.4\scriptsize$\pm 1.1$
&71.3 & 64.5\scriptsize$\pm 4.1$
&67.3 & \textbf{78.2}\scriptsize$\pm 7.3$
&68.0 & \textbf{64.3}\scriptsize$\pm 1.9$
&78.3 & 77.5\scriptsize$\pm 0.0$
&59.1 & \textbf{59.6}\scriptsize$\pm 0.0$
&77.8 & 63.7\scriptsize$\pm 0.0$
&87.6 & \textbf{82.5}\scriptsize$\pm 0.0$
\\
&LGBM-CVaR-DRO    &79.6 & 76.5\scriptsize$\pm 2.7$
&67.6 & 63.1\scriptsize$\pm 2.0$
&73.4 & 72.6\scriptsize$\pm 1.7$
&69.8 & \textbf{64.8}\scriptsize$\pm 3.7$
&83.2 & 74.8\scriptsize$\pm 7.4$
&67.8 & 64.2\scriptsize$\pm 1.4$
&78.0 & 76.8\scriptsize$\pm 0.0$
&58.8 & 57.1\scriptsize$\pm 0.0$
&86.5 & 64.7\scriptsize$\pm 0.0$
&87.2 & 81.8\scriptsize$\pm 0.0$
\\
&LGBM-KL-DRO    &79.1 & 76.5\scriptsize$\pm 2.5$
&66.7 & \textbf{63.3}\scriptsize$\pm 1.9$
&70.6 & 72.2\scriptsize$\pm 1.6$
&71.6 & 64.8\scriptsize$\pm 4.0$
&68.0 & 77.1\scriptsize$\pm 7.9$
&68.0 & 63.8\scriptsize$\pm 1.5$
&78.4 & 76.8\scriptsize$\pm 0.0$
&59.1 & 57.4\scriptsize$\pm 0.0$
&86.6 & \textbf{64.8}\scriptsize$\pm 0.0$
&87.2 & 81.9\scriptsize$\pm 0.0$
\\\midrule
\multirow{4}{*}{\begin{tabular}[c]{@{}l@{}}Kernel-DRO\\ Methods\\ (base: Kernel)\end{tabular}}   &Kernel-$\chi2$-DRO    &80.6 & 75.9\scriptsize$\pm 3.4$
&61.8 & \textbf{59.6}\scriptsize$\pm 1.6$
&82.4 & 73.3\scriptsize$\pm 2.2$
&68.7 & 64.7\scriptsize$\pm 4.8$
&79.0 & 72.6\scriptsize$\pm 7.6$
&68.6 & \textbf{67.3}\scriptsize$\pm 0.7$
&80.4 & 78.8\scriptsize$\pm 0.0$
&61.0 & \textbf{59.7}\scriptsize$\pm 0.0$
&84.2 & 59.6\scriptsize$\pm 0.0$
&85.0 & 78.9\scriptsize$\pm 0.0$
\\
&Kernel-CVaR-DRO    &80.5 & \textbf{76.0}\scriptsize$\pm 3.4$
&61.6 & 58.9\scriptsize$\pm 1.7$
&84.0 & 71.9\scriptsize$\pm 1.9$
&69.0 & \textbf{65.3}\scriptsize$\pm 3.8$
&79.9 & \textbf{73.3}\scriptsize$\pm 7.5$
&67.4 & 66.8\scriptsize$\pm 0.9$
&78.5 & 78.6\scriptsize$\pm 0.0$
&62.0 & \textbf{59.7}\scriptsize$\pm 0.0$
&83.8 & 63.7\scriptsize$\pm 0.0$
&84.1 & \textbf{79.5}\scriptsize$\pm 0.0$
\\
&Kernel-KL-DRO    &80.8 & 75.9\scriptsize$\pm 3.5$
&60.0 & 57.8\scriptsize$\pm 1.8$
&81.6 & 72.9\scriptsize$\pm 1.9$
&69.5 & 65.0\scriptsize$\pm 4.7$
&78.4 & 71.6\scriptsize$\pm 7.9$
&67.9 & 66.8\scriptsize$\pm 0.4$
&79.0 & \textbf{78.9}\scriptsize$\pm 0.0$
&61.1 & 59.6\scriptsize$\pm 0.0$
&84.0 & \textbf{64.4}\scriptsize$\pm 0.0$
&83.6 & 78.8\scriptsize$\pm 0.0$
\\
&Kernel-Wasserstein-DRO    &80.8 & 75.8\scriptsize$\pm 3.5$
&57.5 & 49.8\scriptsize$\pm 2.6$
&82.7 & \textbf{73.3}\scriptsize$\pm 2.0$
&67.4 & 61.7\scriptsize$\pm 5.0$
&79.6 & 71.6\scriptsize$\pm 8.3$
&65.0 & 60.7\scriptsize$\pm 1.8$
&78.0 & 77.8\scriptsize$\pm 0.0$
&59.4 & 56.8\scriptsize$\pm 0.0$
&84.0 & 60.3\scriptsize$\pm 0.0$
&85.0 & \textbf{79.5}\scriptsize$\pm 0.0$
\\\midrule
\multirow{6}{*}{\begin{tabular}[c]{@{}l@{}}Imbalanced\\ Learning\\ \& Fairness\\ Methods \\(base: XGB)\end{tabular}} &SUBY    &80.4 & 74.3\scriptsize$\pm 4.0$
&64.2 & \textbf{62.2}\scriptsize$\pm 1.9$
&82.3 & 72.4\scriptsize$\pm 2.0$
&69.8 & 69.0\scriptsize$\pm 2.5$
&84.4 & \textbf{85.1}\scriptsize$\pm 7.6$
&70.4 & \textbf{69.5}\scriptsize$\pm 0.9$
&81.1 & \textbf{79.4}\scriptsize$\pm 0.0$
&64.4 & \textbf{62.8}\scriptsize$\pm 0.0$
&84.1 & \textbf{63.3}\scriptsize$\pm 0.0$
&81.2 & 74.5\scriptsize$\pm 0.0$
\\
&RWY    &80.7 & 74.4\scriptsize$\pm 4.0$
&64.6 & 61.8\scriptsize$\pm 1.8$
&74.7 & \textbf{74.4}\scriptsize$\pm 1.3$
&71.6 & \textbf{70.0}\scriptsize$\pm 2.4$
&84.3 & 84.9\scriptsize$\pm 7.5$
&71.0 & 69.4\scriptsize$\pm 0.9$
&81.1 & \textbf{79.4}\scriptsize$\pm 0.0$
&64.1 & \textbf{62.8}\scriptsize$\pm 0.0$
&85.1 & 63.1\scriptsize$\pm 0.0$
&86.8 & 81.6\scriptsize$\pm 0.0$
\\
&SUBG    &80.7 & \textbf{76.0}\scriptsize$\pm 3.3$
&65.2 & 59.8\scriptsize$\pm 1.8$
&-&-
&72.4 & 66.6\scriptsize$\pm 3.9$
&-&-
&70.9 & 67.3\scriptsize$\pm 1.4$
&81.1 & \textbf{79.4}\scriptsize$\pm 0.0$
&52.2 & 56.1\scriptsize$\pm 0.0$
&-&-
&85.8 & 81.4\scriptsize$\pm 0.0$
\\
&RWG    &81.0 & 76.0\scriptsize$\pm 3.4$
&65.2 & 59.0\scriptsize$\pm 1.9$
&-&-
&72.8 & 66.5\scriptsize$\pm 4.1$
&-&-
&70.8 & 66.4\scriptsize$\pm 1.8$
&81.1 & \textbf{79.4}\scriptsize$\pm 0.0$
&62.6 & 61.4\scriptsize$\pm 0.0$
&-&-
&86.9 & 81.2\scriptsize$\pm 0.0$
\\
&In-processing    &81.0 & 76.1\scriptsize$\pm 3.4$
&65.3 & 59.1\scriptsize$\pm 2.1$
&-&-
&72.2 & 66.3\scriptsize$\pm 4.4$
&-&-
&70.9 & 66.9\scriptsize$\pm 1.7$
&80.1 & 78.8\scriptsize$\pm 0.0$
&63.7 & 62.6\scriptsize$\pm 0.0$
&-&-
&87.5 & \textbf{82.5}\scriptsize$\pm 0.0$
\\
&Post-processing    &80.4 & 75.1\scriptsize$\pm 3.6$
&65.4 & 59.0\scriptsize$\pm 2.1$
&-&-
&73.2 & 66.3\scriptsize$\pm 4.2$
&-&-
&70.7 & 66.8\scriptsize$\pm 1.6$
&79.0 & 78.9\scriptsize$\pm 0.0$
&62.5 & 62.7\scriptsize$\pm 0.0$
&-&-
&62.5 & 81.8\scriptsize$\pm 0.0$
\\
 \bottomrule \bottomrule

\end{tabular}}
\end{table}
\end{landscape}

\wty{\subsection{Detailed Results of Feasibility in 10 Settings}\label{app:feasibility}
We evaluate each method's feasibility across all 10 settings by averaging over its performance on all target domains. Specifically, for a setting with $T$ target distributions $\{Q_t\}_{t \in [T]}$,  we define the \emph{feasibility} of model $\hat f$ as:
\[\frac{1}{T}\sum_{t \in [T]}\frac{0.5 - \E_{Q_t}[\ell(\hat f(X), Y)]}{0.5 - \min_{\Fscr}\min_{f \in \Fscr}\E_{Q_t}[\ell(f(X), Y)]}.\]
\Cref{fig:one-to-all-relative} presents these feasibility scores for each method. As before, in Sections~\ref{subsec:finding3} and~\ref{subsec:finding4}, we find that DRO shows limited improvements and the DRO performance is strongly correlated with its base model. For brevity, we show only the relative feasibility analog of Figure \ref{fig:one-to-all}, since the rescaling does not affect the conclusions of Findings 1 and 2 (Sections \ref{subsec:acc-on-the-line} and \ref{subsec:benchmark-empirical-result}).
\begin{figure}[!htb]
        \includegraphics[width = \textwidth]{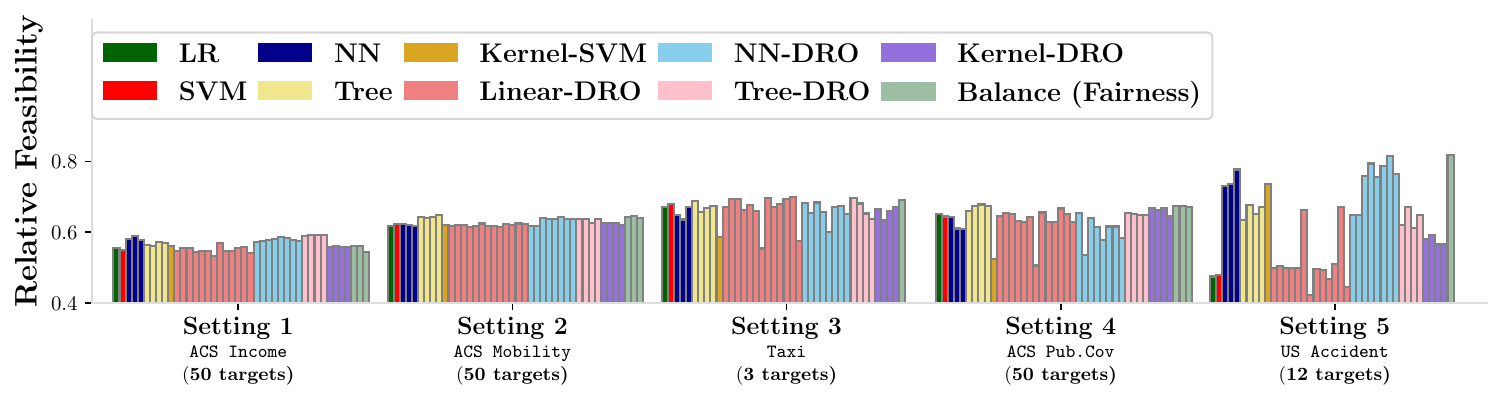}
	\includegraphics[width=\textwidth]{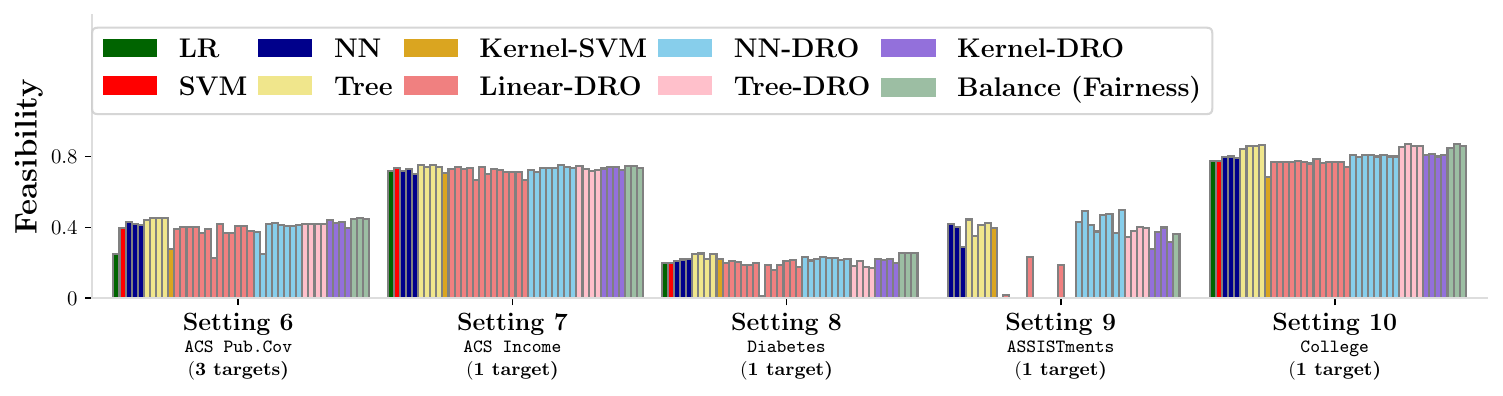}
	\caption{Overall performances (reported as the \emph{feasibility} metric) of all methods on the target data in our
          10 settings in Table \ref{table:overview}. Here we calculate the average feasibility value among multiple target domains in each setting. 
          Note that for Settings 3, 5 and 9, we do not include fairness-enhancing methods since there are no sensitive attributes.}
     \label{fig:one-to-all-relative}
     \vspace{-0.15in}
\end{figure}
}

\wty{
\subsection{Detailed Results of Controlled Shift Settings}\label{app:synthetic}
We construct a distribution shift instance with controlled perturbations, i.e.,  by perturbing features and labels from the original database. We focus on the ACS Income dataset in Setting 1 and set both the source and target domains as data from California while constructing the target domains by perturbing features and (or) labels of the original ACS Income dataset. Denote $x = (x_{\Nscr}, x_{\Cscr})$, where $x_{\Nscr}$ denotes the numerical feature and $x_{\Cscr}$ denotes the categorical feature.  We generate each perturbed row of target domains following \cite{justin2023learning}. More specifically, we allow:
\begin{itemize}
    \item feature shifts: For each numerical feature $x_j, j \in \Nscr$, in the target domain, we perturb $x_j$ with $x_j + \epsilon_j$, where $\epsilon_j \sim N(0, \text{Var}[x_j]/10)$; for each categorical feature $\ell \in \Cscr$, in the target domain, we perturb $x_{\ell}$ with $\tilde x_{\ell} \in \text{domain}(X_{\ell})$ with $P(\tilde x_{\ell} = x_{\ell}) = 0.9$ and $P(\tilde x_{\ell} = u) = \frac{0.1}{|\text{domain}(X_{\ell})| - 1}$ for $u \in \text{domain}(X_{\ell})\backslash \{x_{\ell}\}$. We perturb each sample independently. 
    \item label shifts: We flip the output of each sample with probability 0.2.
\end{itemize}}

\wty{By including at least one shift from the feature or label, we create a ``feature perturbation'' setting with the following four target domains: 
\begin{itemize}[leftmargin=*]
\item [0.] No shifts;
\item [1.] Only label shifts;
\item [2.] Only feature shifts;
\item [3.] Both feature and label shifts.
\end{itemize}
For each target domain, we follow our evaluation pipeline in \Cref{sec:data}, i.e., tune each method based on 128 samples from the target domain, and report results in Table~\ref{tab:synthetic}. Across all four domains, variants of Wasserstein-DRO (Wasserstein-DRO and Unified-DRO) enjoy relatively good performance, often outperforming the base model since such shifts are captured in corresponding ambiguity sets.  When the shifts are severe (e.g., Scenarios 2 or 3), then Unified-DRO achieves better than basic ERM methods across all model classes. This indicates that existing DRO captures such synthetic shifts. 
\begin{table}[!htb]
    \centering
    \caption{Results under the synthetic dataset with 4 target domains, where we run each method with its top-10 configurations (according to the \textit{target} performance and report its mean and standard deviation. We boldface the best target performance within each class of methods in each domain.}
    \label{tab:synthetic}
    \resizebox{\textwidth}{!}{\begin{tabular}{c|c|cccc|cccc}
    \toprule
        & & \multicolumn{4}{c|}{Linear-DRO methods} & \multicolumn{4}{c}{Basic Methods}\\
        Domain & Metric & Wasserstein-DRO & Unified-DRO-$L_2$ & KL-DRO & $\chi^2$-DRO & SVM & LR & XGB & NN2\\
        \midrule
0 & Accuracy & \textbf{80.6}{\scriptsize $\pm$0.1} & \textbf{80.6}{\scriptsize $\pm$0.0} & \textbf{80.6}{\scriptsize $\pm$0.0} & 80.5{\scriptsize $\pm$0.1} & \textbf{80.7}{\scriptsize $\pm$0.0} & \textbf{80.7}{\scriptsize $\pm$0.2} & 80.4{\scriptsize $\pm$0.6} & 80.1{\scriptsize $\pm$0.5} \\
 & F1-score &79.8{\scriptsize $\pm$0.1} & \textbf{79.9}{\scriptsize $\pm$0.0} & 79.8{\scriptsize $\pm$0.0} & 79.8{\scriptsize $\pm$0.2} & \textbf{80.0}{\scriptsize $\pm$0.0} & 79.9{\scriptsize $\pm$0.2} & 79.7{\scriptsize $\pm$0.6} & 79.5{\scriptsize $\pm$0.5} \\
1 & Accuracy & \textbf{79.9}{\scriptsize $\pm$0.0} & \textbf{79.9}{\scriptsize $\pm$0.0} & 79.6{\scriptsize $\pm$0.0} & 79.7{\scriptsize $\pm$0.0} & 79.9{\scriptsize $\pm$0.0} & 79.6{\scriptsize $\pm$0.0} & \textbf{80.1}{\scriptsize $\pm$0.7} & 79.2{\scriptsize $\pm$0.1} \\
 & F1-score & 79.0{\scriptsize $\pm$0.0} & \textbf{79.1}{\scriptsize $\pm$0.0} & 78.7{\scriptsize $\pm$0.0} & 78.8{\scriptsize $\pm$0.0} & 79.0{\scriptsize $\pm$0.0} & 78.8{\scriptsize $\pm$0.0} & \textbf{79.2}{\scriptsize $\pm$0.8} & 78.5{\scriptsize $\pm$0.2} \\
2 & Accuracy &78.6{\scriptsize $\pm$0.1} & \textbf{78.9}{\scriptsize $\pm$0.0} & 78.4{\scriptsize $\pm$0.0} & 78.4{\scriptsize $\pm$0.0} & 78.3{\scriptsize $\pm$0.2} & 78.2{\scriptsize $\pm$0.0} & \textbf{78.7}{\scriptsize $\pm$1.0} & 77.0{\scriptsize $\pm$0.8} \\
 & F1-score &77.6{\scriptsize $\pm$0.1} & \textbf{77.9}{\scriptsize $\pm$0.1} & 77.3{\scriptsize $\pm$0.0} & 77.4{\scriptsize $\pm$0.0} & 77.2{\scriptsize $\pm$0.2} & 77.1{\scriptsize $\pm$0.0} & \textbf{77.5}{\scriptsize $\pm$1.2} & 76.2{\scriptsize $\pm$0.8} \\
3 & Accuracy & 77.0{\scriptsize $\pm$0.0} & \textbf{77.7}{\scriptsize $\pm$0.1} & 76.2{\scriptsize $\pm$0.0} & 76.4{\scriptsize $\pm$0.0} & 76.4{\scriptsize $\pm$0.2} & 76.4{\scriptsize $\pm$0.2} & \textbf{77.0}{\scriptsize $\pm$0.8} & 74.6{\scriptsize $\pm$0.3} \\
 & F1-score &75.5{\scriptsize $\pm$0.1} & \textbf{76.4}{\scriptsize $\pm$0.1} & 74.8{\scriptsize $\pm$0.1} & 74.9{\scriptsize $\pm$0.0} & 75.0{\scriptsize $\pm$0.2} & 74.9{\scriptsize $\pm$0.2} & \textbf{75.5}{\scriptsize $\pm$1.0} & 73.6{\scriptsize $\pm$0.5} \\
      \bottomrule
    \end{tabular}}
\end{table}
}

\wty{\subsection{Difference between Shift Settings in~\ref{app:synthetic} and Settings in~\Cref{table:overview}}\label{app:sythetic-real}
Besides the difference in the selection criterion, we emphasize that the shifts constructed in \Cref{app:synthetic} are inherently easier to address. Specifically, for each setting (including both the setting in \Cref{app:synthetic} and our spatiotemporal shifts studied in \Cref{table:overview}, where we only choose Settings 1,2,4,5 since there are more than 10 domains in each setting), and for each domain -- whether target domains and source domains,  we fit a logistic regression model on the data in that domain and obtain the corresponding coefficient vector. This yields a collection $\{\theta^{(j)}\}_{j \in \Sscr}$ over all domains $\Sscr$ for each setting. We then perform principal component analysis (PCA) on these coefficients and compute the proportion of variance explained by the first $k$ components, for $k = 1,2,\ldots, 5$. }

\wty{
We present the results in \Cref{fig:pca}, where we also compare another ``bootstrap'' setting over the ACS Income dataset while bootstrapping the training data. Unsurprisingly, the coefficients in the ``bootstrap'' setting retain substantial variance even when reduced to one or two components. Interestingly, the coefficients in the setting from \Cref{app:synthetic} exhibit similar variance explained, approximately 10\% higher than our constructed shifts, namely, the ``income'' setting and related settings such as ``accident''. For the same ``income'' setting (Setting 1 and the setting in \Cref{app:synthetic}), we also compute the covariance matrix of $\{\theta^{(j)}\}_{j \in \Sscr}$ for each setting and find that the largest eigenvalue in Setting 1 is significantly larger than that in the synthetic setting ($0.6 > 0.1$). This highlights the greater difficulty of learning under our constructed shifts compared to the setting in \Cref{app:synthetic} with controlled perturbations.
\begin{figure}[!htb]
    \centering
    \includegraphics[width=0.6\linewidth]{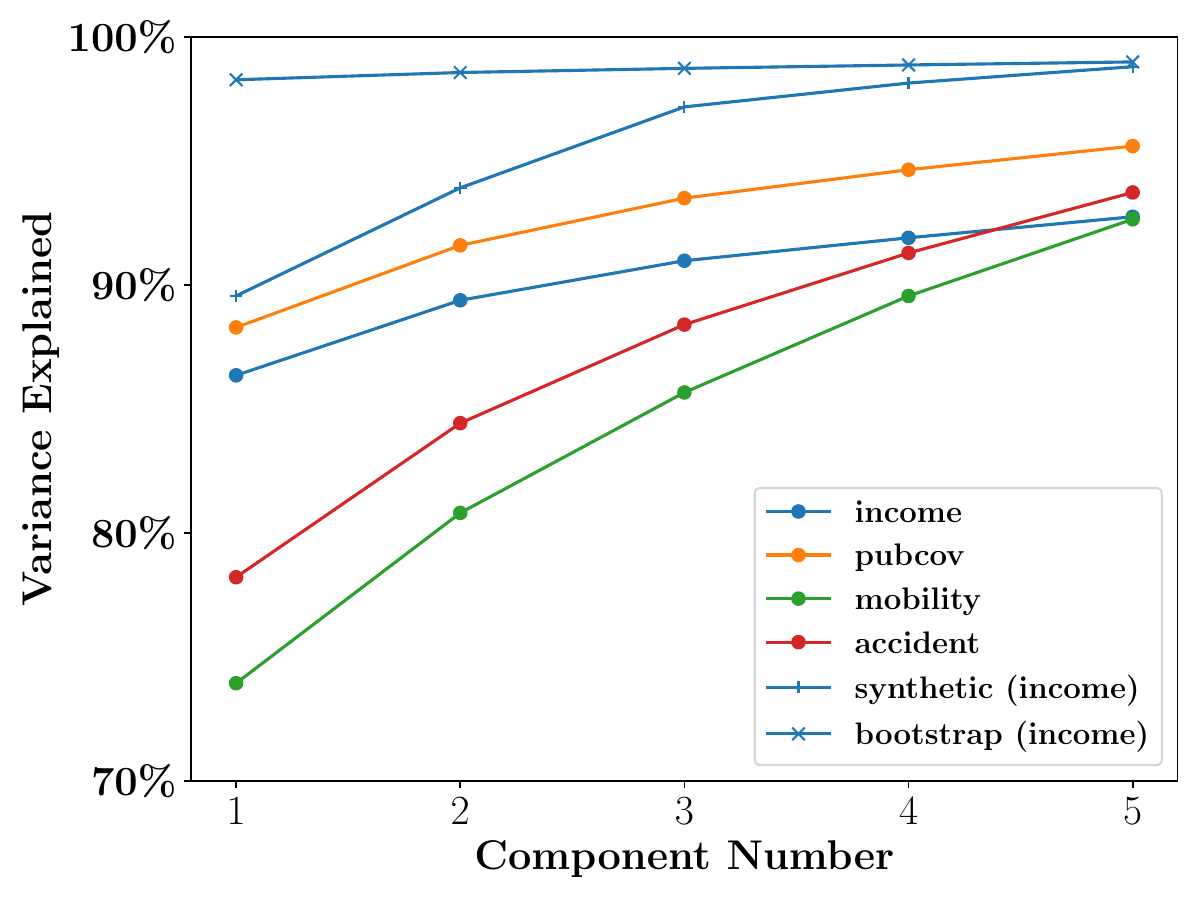}
    \caption{Variance explained by components across different settings, where ``income'', ``pubcov'', ``mobility'', ``accident'' correspond to Settings 1, 4, 2, 5 in \Cref{table:overview} and ``synthetic (income)'' correspond to the setting in \Cref{app:synthetic}}
    \label{fig:pca}
\end{figure}
}
\section{Details in Section~\ref{sec:method-lr}}
\subsection{Definitions of Variables}\label{app:var-def}
Here in \Cref{tab:var_def}, we provide the exact definitions of each variable in the linear regression used in \Cref{subsec:lr-empirical} and further augmented variables that will be used in Appendix~\ref{app:empirical-all}.
\begin{table}[!htb]
    \centering
    \caption{Definition of independent variables used in~\eqref{eq:lr-model} of \Cref{subsec:lr-empirical}}
    \label{tab:var_def}
    \begin{tabular}{llp{10cm}}
    \toprule
    Type & Name& Definition\\
    \midrule
    \multirow{2}{*}{\parbox[l]{1.5cm}{Model\\Class $X_{i,j,s}$}}& LGBM& A dummy variable that takes value one if the underlying model class of the method is LightGBM\\
    & NN-2 & A dummy variable that takes value one if the underlying model class of the method is the neural network with 2 layers\\
    & NN-3 & A dummy variable that takes value one if the underlying model class of the method is the neural network with 3 layers\\
    & NN-4 & A dummy variable that takes value one if the underlying model class of the method is the neural network with 4 layers\\
    & Kernel & A dummy variable that takes value one if the underlying model class of the method is the Kernel method\\
    \midrule
    \multirow{2}{*}{\parbox[l]{1.5cm}{Ambiguity \\Set $D_{i,j,s,t}$}}& Wasserstein & A dummy variable that takes value one if the method belongs to DRO and uses Wasserstein-type metric\\
    & Chi-squared & A dummy variable that takes value one if the method belongs to DRO and uses $\chi^2$-divergence metric\\
    & Kullback-Leibler & A dummy variable that takes value one if the method belongs to DRO and uses KL-divergence metric\\
    & Total Variation & A dummy variable that takes value one if the method belongs to DRO and uses TV-distance metric\\
    & OT-Discrepancy & A dummy variable that takes value one if the method belongs to DRO and uses the optimal transport-discrepancy with conditional moment constraints\\
    & Radius & The rescaled ambiguity size if the method belongs to DRO and equal to zero if the method does not belong to DRO\\ \midrule
    \multirow{2}{*}{\parbox[l]{2cm}{Shift\\Pattern $Z_{i,j}$}} & $Y|X$-ratio & The $Y|X$-shift percentage calculated by DISDE from the source domain to the target domain\\\midrule
    \multirow{2}{*}{\parbox[l]{1.5cm}{Validation\\Type $V_{i,j}$}}& Worst-case & A dummy variable that takes value one if the best configuration is obtained by maximizing the accuracy on target data from the worst-performing target domain\\
    & Average-case & A dummy variable that takes value one if the best configuration is obtained by maximizing the accuracy on target data across all target domains, averaged on each target domain\\
    \bottomrule
    \end{tabular}
\end{table}

\wty{\begin{table}[!htb]
    \centering
    \caption{Additional (Interaction) Variables in Appendix~\ref{app:empirical-all}}
    \label{tab:var_def2}
    \begin{tabular}{lp{10cm}}
    \toprule
    Name& Definition\\
    \midrule
    $\text{Radius}^2$ & The square of the rescaled ambiguity size if the method belongs to DRO methods and zero if the method does not belong to DRO\\
    \hline 
    LGBM-Radius & An interaction variable that takes value to be the robustness radius if the underlying model class of the method is LightGBM\\
    NN2-Radius & An interaction variable that takes value to be the robustness radius if the underlying model class of the method is the neural network with 2 layers\\
    NN3-Radius & An interaction variable that takes value to be the robustness radius if the underlying model class of the method is the neural network with 3 layers\\
    NN4-Radius & An interaction variable that takes value to be the robustness radius if the underlying model class of the method is the neural network with 4 layers\\
    Kernel-Radius & An interaction variable that takes value to be the robustness radius if the underlying model class of the method is kernel-SVM\\
    \hline
    LGBM-$Y|X$-ratio &An interaction variable that takes value to be the Y|X-shift percentage if the underlying model class of the method is LightGBM\\
    NN2-$Y|X$-ratio & An interaction variable that takes value to be the Y|X-shift percentage if the underlying model class of the method is the neural network with 2 layers\\
    NN3-$Y|X$-ratio & An interaction variable that takes value to be the Y|X-shift percentage if the underlying model class of the method is the neural network with 3 layers\\
    NN4-$Y|X$-ratio & An interaction variable that takes value to be the Y|X-shift percentage if the underlying model class of the method is the neural network with 4 layers\\
    Kernel-$Y|X$-ratio & An interaction variable that takes value to be the Y|X-shift percentage if the underlying model class of the method is Kernel SVM\\
    \bottomrule
    \end{tabular}
\end{table}}

\subsection{Scaling of Ambiguity Size}\label{app:ambiguity-scale}
We discuss the choice of rescaled ambiguity size across different DRO methods in \Cref{subsec:lr-empirical}. Note that we consider (squared) Wasserstein Distance and OT-Discrepancy metric (i.e. the distance proposed in Unified-DRO \citep{blanchet2023unifying}), TV, CVaR, KL-divergence, $\chi^2$-divergence in the linear regression. Denote their original ambiguity size hyperparameter in the method training as $\epsilon_{was},\epsilon_{ot}, \epsilon_{tv}, \alpha_{cvar}, \epsilon_{kl}, \epsilon_{\chi^2}$ and the rescaled standard one as $\epsilon_{was'}, \epsilon_{ot'}, \epsilon_{tv'}, \epsilon_{cvar'}, \epsilon_{kl'}, \epsilon_{\chi^{2'}}$

The unsquared Wasserstein Distance, OT Discrepancy metric, and TV distance already form a distance notion, so we do not distinguish the difference between them. Since we use the squared Wasserstein Distance in the DRO formulation, we set $\epsilon_{was'} = \sqrt{\epsilon_{was}}$ in the variable ``Radius'' the linear regression model. And $\epsilon_{ot'} = \epsilon_{ot}, \epsilon_{tv'} = \epsilon_{tv}$.

For the KL and $\chi^2$-divergence, although $\chi^2, KL$ does not form a distance, we can symmetrize these two divergences such that the augmented version becomes a distance. For example, $\tilde d_{KL}(P, Q) = KL(P, Q)+KL(Q,P)$. We do not distinguish the important sizes between KL-divergence and the reversed version. Therefore, we set $\epsilon_{kl'} = 2 \epsilon_{kl}, \epsilon_{\chi^{2'}} = 2 \epsilon_{\chi^2}$.

For the CVaR metric, denote $D_{\infty}(P||Q) = \text{ess}\sup \frac{dP}{dQ}$. Then the uncertainty region becomes the distribution set that CVaR operates on. That is,
\[\Pscr:= \paran{P | D_{\infty}(P || \widehat P) \leq \log \frac{1}{\alpha}} = \paran{P| \text{there exists}~Q, \beta \in [\alpha, 1], \text{s.t.}~\widehat P = \beta P + (1-\beta)Q}.\] 
Therefore, we take $\epsilon_{cvar'} = \log \frac{1}{\alpha_{cvar}}$.

\subsection{Linear Regression Results for Other Settings}
\label{app:empirical-all}
We include linear regression results with respect to \texttt{Best Config} and \texttt{Worst Domain} in other settings in Table~\ref{tab:linear-analysis-all}, where similar patterns hold as in \Cref{subsec:lr-empirical}.

\begin{table}[htbp]
    \centering
    \caption{Regression results on algorithmic design components on the method performance in all other settings besides in \Cref{tab:linear-analysis}}
    \label{tab:linear-analysis-all}
    \sisetup{
    table-format=-1.4, 
    add-integer-zero=false 
    }
    	\resizebox{\textwidth}{!}{\begin{tabular}{llSSSSSSSS}
        \toprule
        & & \multicolumn{8}{c}{Dependent variable: Target accuracy}\\
        & & \multicolumn{4}{c}{\texttt{Best Config}} & \multicolumn{4}{c}{\texttt{Worst Domain}}\\\cmidrule(lr){3-10}
    \multicolumn{2}{c}{Variable Name} & \multicolumn{1}{c}{\ \ Setting 3}& \multicolumn{1}{c}{\ \ Setting 4} & \multicolumn{1}{c}{\ \ Setting 2} & \multicolumn{1}{c}{\ \ Setting 6} & \multicolumn{1}{c}{\ \ Setting 3} & \multicolumn{1}{c}{\ \ Setting 5} & \multicolumn{1}{c}{\ \ Setting 1} & \multicolumn{1}{c}{\ \ Setting 6}\\
         \midrule
\multirow{3}{*}{\parbox[c]{1.5cm}{Model\\Class}}&LGBM&-.0234$^{***}$&-.0189$^{***}$& .0025$^{***}$& .0224$^{***}$&-.0435$^{***}$& .0142$^{***}$& .0054$^{*}$& .0061$^{***}$\\
&&\multicolumn{1}{c}{ \scriptsize\hspace{0.35cm}(.0039)}&\multicolumn{1}{c}{ \scriptsize\hspace{0.35cm}(.0017)}&\multicolumn{1}{c}{ \scriptsize\hspace{0.35cm}(.0006)}&\multicolumn{1}{c}{ \scriptsize\hspace{0.35cm}(.0034)}&\multicolumn{1}{c}{ \scriptsize\hspace{0.35cm}(.0049)}&\multicolumn{1}{c}{ \scriptsize\hspace{0.35cm}(.0030)}&\multicolumn{1}{c}{ \scriptsize\hspace{0.35cm}(.0029)}&\multicolumn{1}{c}{ \scriptsize\hspace{0.35cm}(.0014)}\\
&NN2&-.0108$^{***}$&-.0105$^{***}$& .0039$^{***}$& .0122$^{***}$&-.0934$^{***}$& .0540$^{***}$&-.0221$^{***}$&-.0288$^{***}$\\
&&\multicolumn{1}{c}{ \scriptsize\hspace{0.35cm}(.0040)}&\multicolumn{1}{c}{ \scriptsize\hspace{0.35cm}(.0018)}&\multicolumn{1}{c}{ \scriptsize\hspace{0.35cm}(.0006)}&\multicolumn{1}{c}{ \scriptsize\hspace{0.35cm}(.0034)}&\multicolumn{1}{c}{ \scriptsize\hspace{0.35cm}(.0057)}&\multicolumn{1}{c}{ \scriptsize\hspace{0.35cm}(.0038)}&\multicolumn{1}{c}{ \scriptsize\hspace{0.35cm}(.0036)}&\multicolumn{1}{c}{ \scriptsize\hspace{0.35cm}(.0023)}\\
&NN3&-.0097$^{**}$&-.0249$^{***}$& .0022$^{***}$& .0112$^{***}$&-.1868$^{***}$& .0516$^{***}$&-.0175$^{***}$&-.0201$^{***}$\\
&&\multicolumn{1}{c}{ \scriptsize\hspace{0.35cm}(.0042)}&\multicolumn{1}{c}{ \scriptsize\hspace{0.35cm}(.0018)}&\multicolumn{1}{c}{ \scriptsize\hspace{0.35cm}(.0006)}&\multicolumn{1}{c}{ \scriptsize\hspace{0.35cm}(.0034)}&\multicolumn{1}{c}{ \scriptsize\hspace{0.35cm}(.0057)}&\multicolumn{1}{c}{ \scriptsize\hspace{0.35cm}(.0038)}&\multicolumn{1}{c}{ \scriptsize\hspace{0.35cm}(.0037)}&\multicolumn{1}{c}{ \scriptsize\hspace{0.35cm}(.0025)}\\
&NN4&-.0215$^{***}$&-.0367$^{***}$&-.0005& .0070$^{**}$&-.1839$^{***}$& .0573$^{***}$&-.0158$^{***}$&-.0248$^{***}$\\
&&\multicolumn{1}{c}{ \scriptsize\hspace{0.35cm}(.0044)}&\multicolumn{1}{c}{ \scriptsize\hspace{0.35cm}(.0019)}&\multicolumn{1}{c}{ \scriptsize\hspace{0.35cm}(.0006)}&\multicolumn{1}{c}{ \scriptsize\hspace{0.35cm}(.0035)}&\multicolumn{1}{c}{ \scriptsize\hspace{0.35cm}(.0059)}&\multicolumn{1}{c}{ \scriptsize\hspace{0.35cm}(.0037)}&\multicolumn{1}{c}{ \scriptsize\hspace{0.35cm}(.0036)}&\multicolumn{1}{c}{ \scriptsize\hspace{0.35cm}(.0030)}\\
&Kernel&-.0165$^{***}$&-.0101$^{***}$&-.0015$^{***}$&-.0063$^{**}$&-.0789$^{***}$& .0098$^{***}$&-.0058$^{**}$&-.0010\\
&&\multicolumn{1}{c}{ \scriptsize\hspace{0.35cm}(.0033)}&\multicolumn{1}{c}{ \scriptsize\hspace{0.35cm}(.0016)}&\multicolumn{1}{c}{ \scriptsize\hspace{0.35cm}(.0005)}&\multicolumn{1}{c}{ \scriptsize\hspace{0.35cm}(.0030)}&\multicolumn{1}{c}{ \scriptsize\hspace{0.35cm}(.0049)}&\multicolumn{1}{c}{ \scriptsize\hspace{0.35cm}(.0030)}&\multicolumn{1}{c}{ \scriptsize\hspace{0.35cm}(.0029)}&\multicolumn{1}{c}{ \scriptsize\hspace{0.35cm}(.0015)}\\
\midrule
\multirow{3}{*}{\parbox[c]{1.7cm}{Ambiguity\\Set}}&Wasserstein& .0016& .0046$^{**}$&-.0005& .0051&-.0382$^{***}$&-.0140$^{**}$& .0000&-.0070$^{**}$\\
&&\multicolumn{1}{c}{ \scriptsize\hspace{0.35cm}(.0043)}&\multicolumn{1}{c}{ \scriptsize\hspace{0.35cm}(.0020)}&\multicolumn{1}{c}{ \scriptsize\hspace{0.35cm}(.0006)}&\multicolumn{1}{c}{ \scriptsize\hspace{0.35cm}(.0037)}&\multicolumn{1}{c}{ \scriptsize\hspace{0.35cm}(.0064)}&\multicolumn{1}{c}{ \scriptsize\hspace{0.35cm}(.0064)}&\multicolumn{1}{c}{ \scriptsize\hspace{0.35cm}(.0050)}&\multicolumn{1}{c}{ \scriptsize\hspace{0.35cm}(.0030)}\\
&Chi-squared&-.0006& .0005& .0015$^{***}$& .0075$^{***}$& .0053& .0040$^{*}$& .0159$^{***}$& .0103$^{***}$\\
&&\multicolumn{1}{c}{ \scriptsize\hspace{0.35cm}(.0029)}&\multicolumn{1}{c}{ \scriptsize\hspace{0.35cm}(.0014)}&\multicolumn{1}{c}{ \scriptsize\hspace{0.35cm}(.0004)}&\multicolumn{1}{c}{ \scriptsize\hspace{0.35cm}(.0024)}&\multicolumn{1}{c}{ \scriptsize\hspace{0.35cm}(.0037)}&\multicolumn{1}{c}{ \scriptsize\hspace{0.35cm}(.0023)}&\multicolumn{1}{c}{ \scriptsize\hspace{0.35cm}(.0023)}&\multicolumn{1}{c}{ \scriptsize\hspace{0.35cm}(.0014)}\\
&Kullback-Leibler&-.0015& .0047$^{***}$&-.0031$^{***}$&-.0002& .0188$^{***}$&-.0255$^{***}$&-.0006& .0011\\
&&\multicolumn{1}{c}{ \scriptsize\hspace{0.35cm}(.0037)}&\multicolumn{1}{c}{ \scriptsize\hspace{0.35cm}(.0015)}&\multicolumn{1}{c}{ \scriptsize\hspace{0.35cm}(.0005)}&\multicolumn{1}{c}{ \scriptsize\hspace{0.35cm}(.0035)}&\multicolumn{1}{c}{ \scriptsize\hspace{0.35cm}(.0043)}&\multicolumn{1}{c}{ \scriptsize\hspace{0.35cm}(.0027)}&\multicolumn{1}{c}{ \scriptsize\hspace{0.35cm}(.0025)}&\multicolumn{1}{c}{ \scriptsize\hspace{0.35cm}(.0013)}\\
&Total Variation&-.0067&-.0098$^{***}$&-.0030$^{***}$& .0032&-.0727$^{***}$&-.0410$^{***}$&-.0291$^{***}$&-.0026\\
&&\multicolumn{1}{c}{ \scriptsize\hspace{0.35cm}(.0059)}&\multicolumn{1}{c}{ \scriptsize\hspace{0.35cm}(.0026)}&\multicolumn{1}{c}{ \scriptsize\hspace{0.35cm}(.0008)}&\multicolumn{1}{c}{ \scriptsize\hspace{0.35cm}(.0051)}&\multicolumn{1}{c}{ \scriptsize\hspace{0.35cm}(.0091)}&\multicolumn{1}{c}{ \scriptsize\hspace{0.35cm}(.0055)}&\multicolumn{1}{c}{ \scriptsize\hspace{0.35cm}(.0053)}&\multicolumn{1}{c}{ \scriptsize\hspace{0.35cm}(.0025)}\\
&OT-Discrepancy&-.0039&-.0046$^{*}$&-.0015$^{*}$&-.0055&-.0040&-.0126$^{**}$& .0220$^{***}$&-.0073$^{***}$\\
&&\multicolumn{1}{c}{ \scriptsize\hspace{0.35cm}(.0059)}&\multicolumn{1}{c}{ \scriptsize\hspace{0.35cm}(.0024)}&\multicolumn{1}{c}{ \scriptsize\hspace{0.35cm}(.0009)}&\multicolumn{1}{c}{ \scriptsize\hspace{0.35cm}(.0050)}&\multicolumn{1}{c}{ \scriptsize\hspace{0.35cm}(.0084)}&\multicolumn{1}{c}{ \scriptsize\hspace{0.35cm}(.0052)}&\multicolumn{1}{c}{ \scriptsize\hspace{0.35cm}(.0048)}&\multicolumn{1}{c}{ \scriptsize\hspace{0.35cm}(.0024)}\\
&Radius& .0001& .0005& .0000& .0051$^{***}$&-.0213$^{***}$&-.0101$^{***}$&-.0048$^{***}$&-.0030$^{***}$\\
&&\multicolumn{1}{c}{ \scriptsize\hspace{0.35cm}(.0028)}&\multicolumn{1}{c}{ \scriptsize\hspace{0.35cm}(.0016)}&\multicolumn{1}{c}{ \scriptsize\hspace{0.35cm}(.0004)}&\multicolumn{1}{c}{ \scriptsize\hspace{0.35cm}(.0011)}&\multicolumn{1}{c}{ \scriptsize\hspace{0.35cm}(.0013)}&\multicolumn{1}{c}{ \scriptsize\hspace{0.35cm}(.0009)}&\multicolumn{1}{c}{ \scriptsize\hspace{0.35cm}(.0009)}&\multicolumn{1}{c}{ \scriptsize\hspace{0.35cm}(.0004)}\\
\midrule
\multirow{2}{*}{\parbox[c]{1.5cm}{Shift\\Pattern}}&$Y|X$-ratio& .1976$^{***}$& -0.1218$^{***}$&-.0052$^{***}$& .1967$^{***}$&{-}&{-}&{-}&{-}\\
&&\multicolumn{1}{c}{ \scriptsize\hspace{0.35cm}(.0019)}&\multicolumn{1}{c}{ \scriptsize\hspace{0.35cm}(0.0007)}&\multicolumn{1}{c}{ \scriptsize\hspace{0.35cm}(.0001)}&\multicolumn{1}{c}{ \scriptsize\hspace{0.35cm}(.0010)}&&&&\\
\midrule
\multirow{3}{*}{\parbox[c]{1.5cm}{Validation\\Type}}&Worst-case & .0114$^{***}$&-.0229$^{***}$&-.0002&-.0018&{-}&{-}&{-}&{-}\\
&&\multicolumn{1}{c}{ \scriptsize\hspace{0.35cm}(.0026)}&\multicolumn{1}{c}{ \scriptsize\hspace{0.35cm}(.0011)}&\multicolumn{1}{c}{ \scriptsize\hspace{0.35cm}(.0004)}&\multicolumn{1}{c}{ \scriptsize\hspace{0.35cm}(.0022)}&&&&\\
&Average-case& .0134$^{***}$& .0028$^{***}$& .0015$^{***}$& .0012&{-}&{-}&{-}&{-}\\
&&\multicolumn{1}{c}{ \scriptsize\hspace{0.35cm}(.0026)}&\multicolumn{1}{c}{ \scriptsize\hspace{0.35cm}(.0011)}&\multicolumn{1}{c}{ \scriptsize\hspace{0.35cm}(.0004)}&\multicolumn{1}{c}{ \scriptsize\hspace{0.35cm}(.0022)}&&&&\\
\midrule
\multirow{2}{*}{\parbox[c]{1.5cm}{Fixed\\Effect}}&Setting&{\ \ No}&{\ \ No}&{\ \ No}&{\ \ No}&{\ \ No}&{\ \ No}&{\ \ No}&{\ \ No}\\
&Domain&{\ \ Yes}&{\ \ Yes}&{\ \ Yes}&{\ \ Yes}&{\ \ No}&{\ \ No}&{\ \ No}&{\ \ No}\\
\midrule
\multirow{2}{*}{\parbox[c]{1.5cm}{Overall}}&$N$&{\ \ 287}&{\ \ 3671}&{\ \ 3671}&{\ \ 287}&{\ \ 5056}&{\ \ 4500}&{\ \ 4321}&{\ \ 3352}\\
&Adjusted $R^2$& .878& .841& .929& .868& .356& .251& .053& .135\\
    \bottomrule
    \end{tabular}}
    
    \smallskip
    \footnotesize{\emph{Notes.} $^{***}, ^{**}$ and $^{*}$ show statistical significance at the 1\%, 5\%, and 10\% levels using two-tailed tests, respectively. 
    }
\end{table}

\twy{\subsection{Fine-grained Comparison between the Model Class and the Choice of the Ambiguity Set}\label{app:fine-grain-model-class}
In this subsection, we conduct a fine-grained comparison of accuracy improvements resulting from incremental changes in model class versus algorithmic design of the ambiguity set.}

\twy{We examine accuracy variations across different methods that involve two types of interventions:
\begin{itemize}
    \item Model class changes: Increasing incremental model complexity from linear models to neural networks with the number of hidden layers in $\{2, 3, 4\}$, and to LightGBM with the number of base estimators in $\{64, 128, 256\}$.
    \item DRO ambiguity set designs: Incorporating distributional robustness beyond standard ERM training, through $\chi^2$-DRO and CVaR-DRO in NNs, and KL-DRO and CVaR-DRO in LGBM, using the optimal ambiguity size that maximizes the accuracy in each corresponding target domain. We limit our comparison to these DRO variants with optimal radii, as they are the formulations supported for the respective model classes in our implementation and yield the best accuracies among the considered methods.
\end{itemize}}

\twy{\paragraph{Neural Network Analysis.} In~\Cref{fig:acc-diff-comp-nn}, we report the accuracy gains comparing these two interventions across all distribution shift pairs. Panels $(a)(d)$ in \Cref{fig:acc-diff-comp-nn} show that when the baseline is a linear model (e.g., SVM), changing the model class to NN2 yields larger accuracy improvements than imposing DRO on the linear model. However, when the base model class is already a neural network, the DRO effect becomes relatively more significant, as shown in panels $(b)(c)(e)(f)$. }

\twy{We further corroborate these patterns in \Cref{fig:acc-diff-nn}, which aggregates accuracy differences across all settings as well as representative configurations (Settings 1 and 5), and shows that model class upgrades dominate accuracy gains for linear baselines, while DRO effects remain secondary and model-dependent for neural networks.}

\twy{
\begin{figure}[!htb]
 \centering\captionsetup[subfloat]{labelfont=scriptsize,textfont=scriptsize} 
 \stackunder[3pt]{\includegraphics[width=0.32\textwidth]{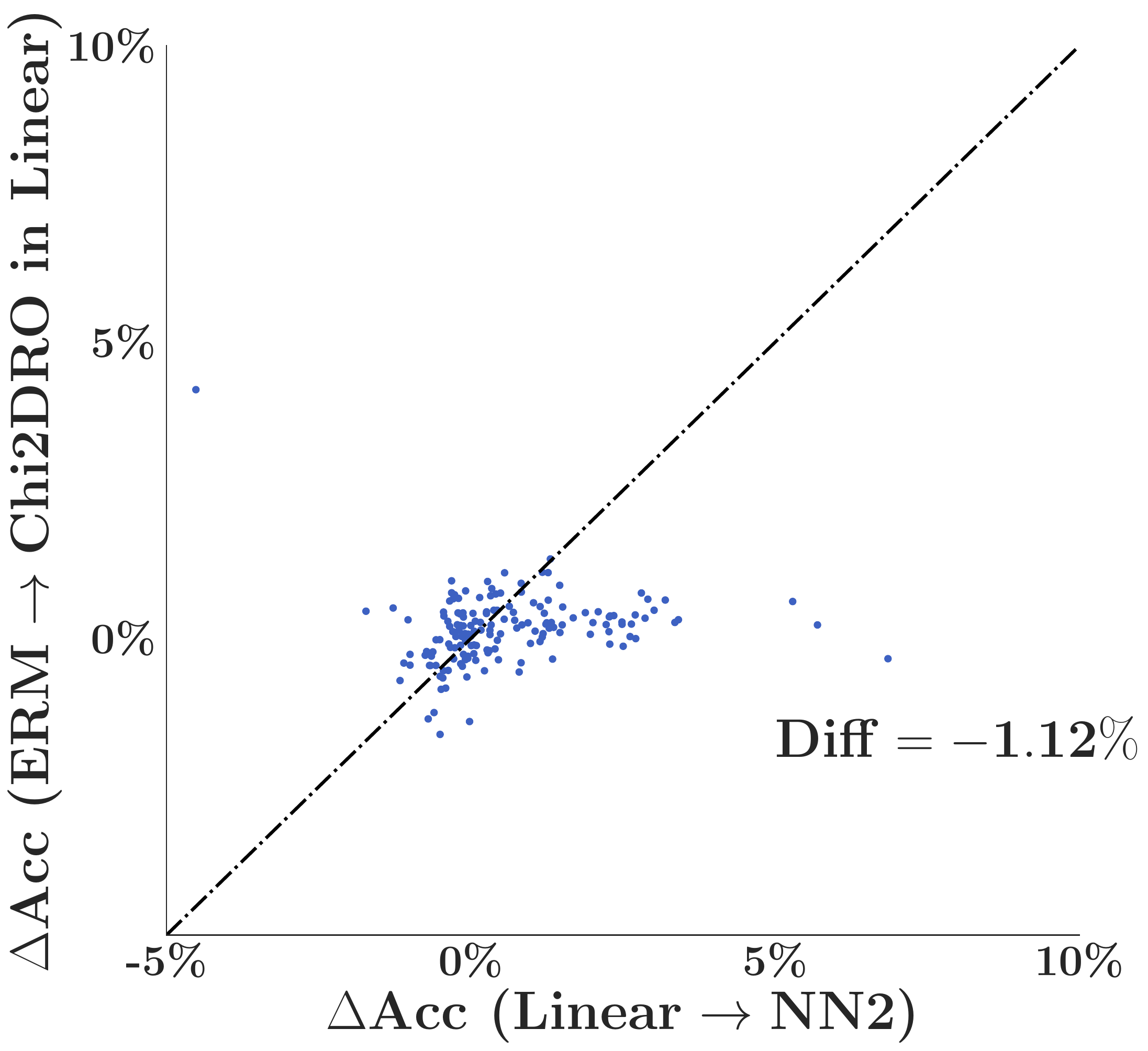}}{(a)}
 \stackunder[3pt]{\includegraphics[width=0.32\textwidth]{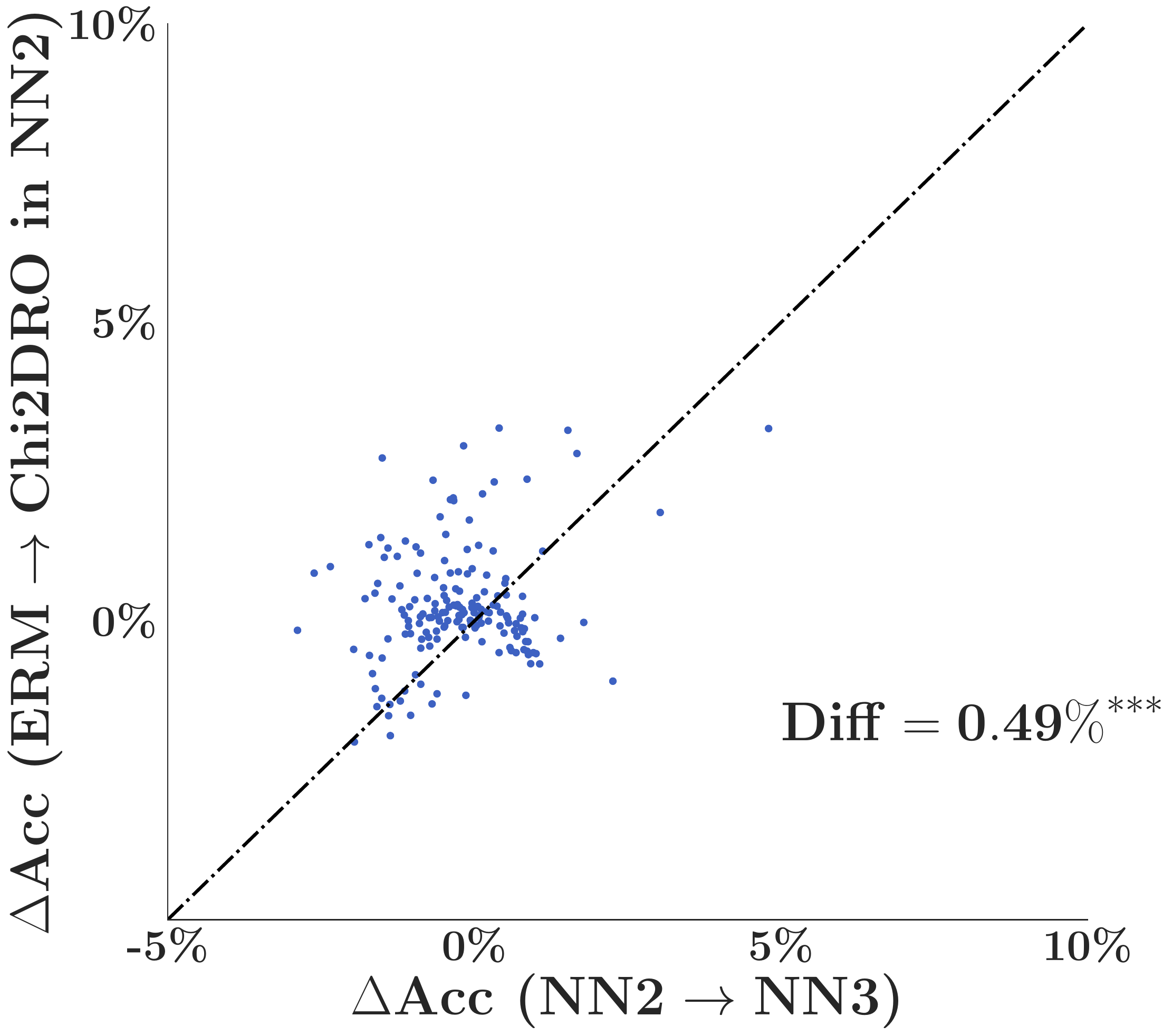}}{(b)}
 \stackunder[3pt]{\includegraphics[width=0.32\textwidth]{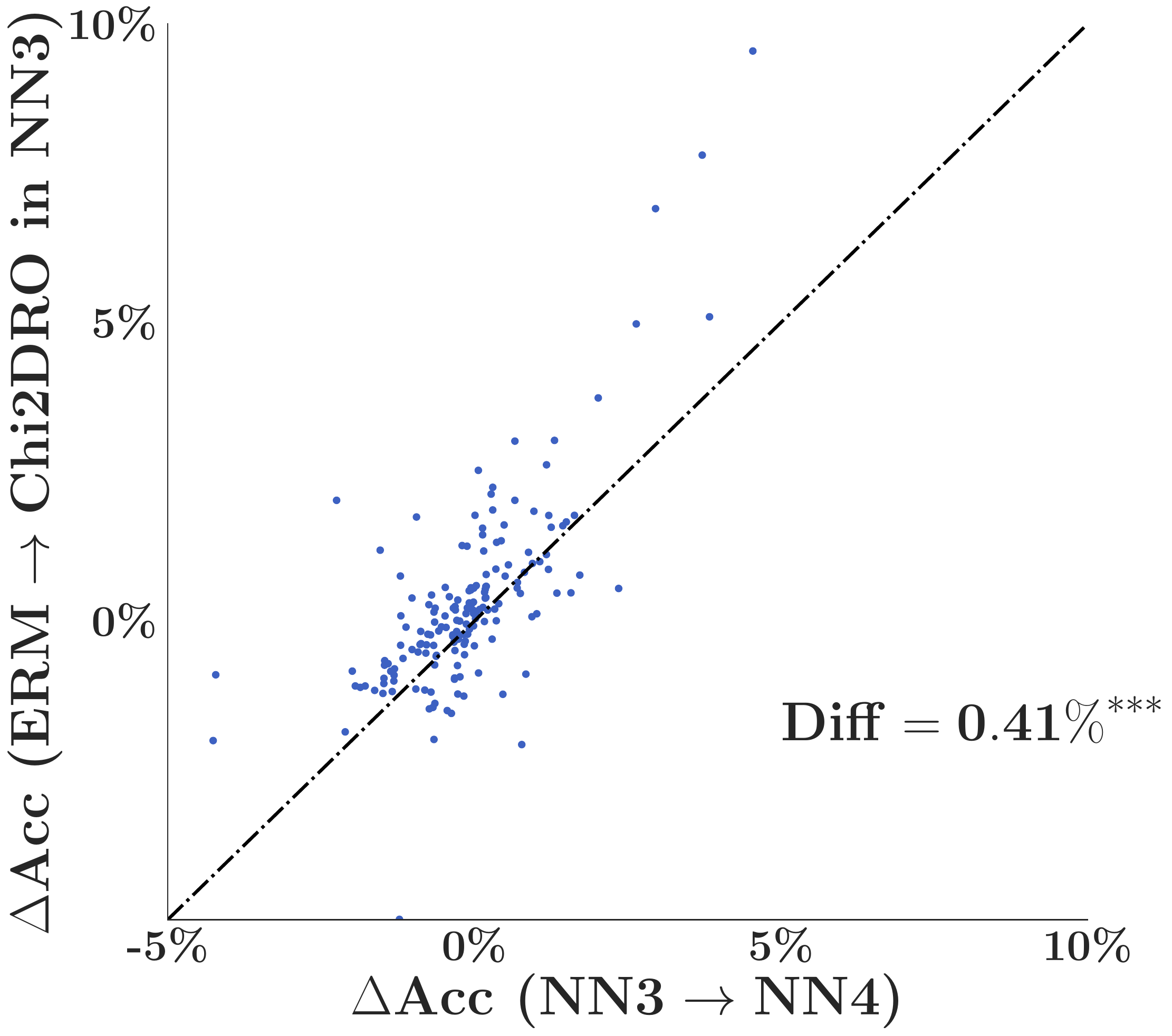}}{(c)}
 \stackunder[3pt]{\includegraphics[width=0.32\textwidth]{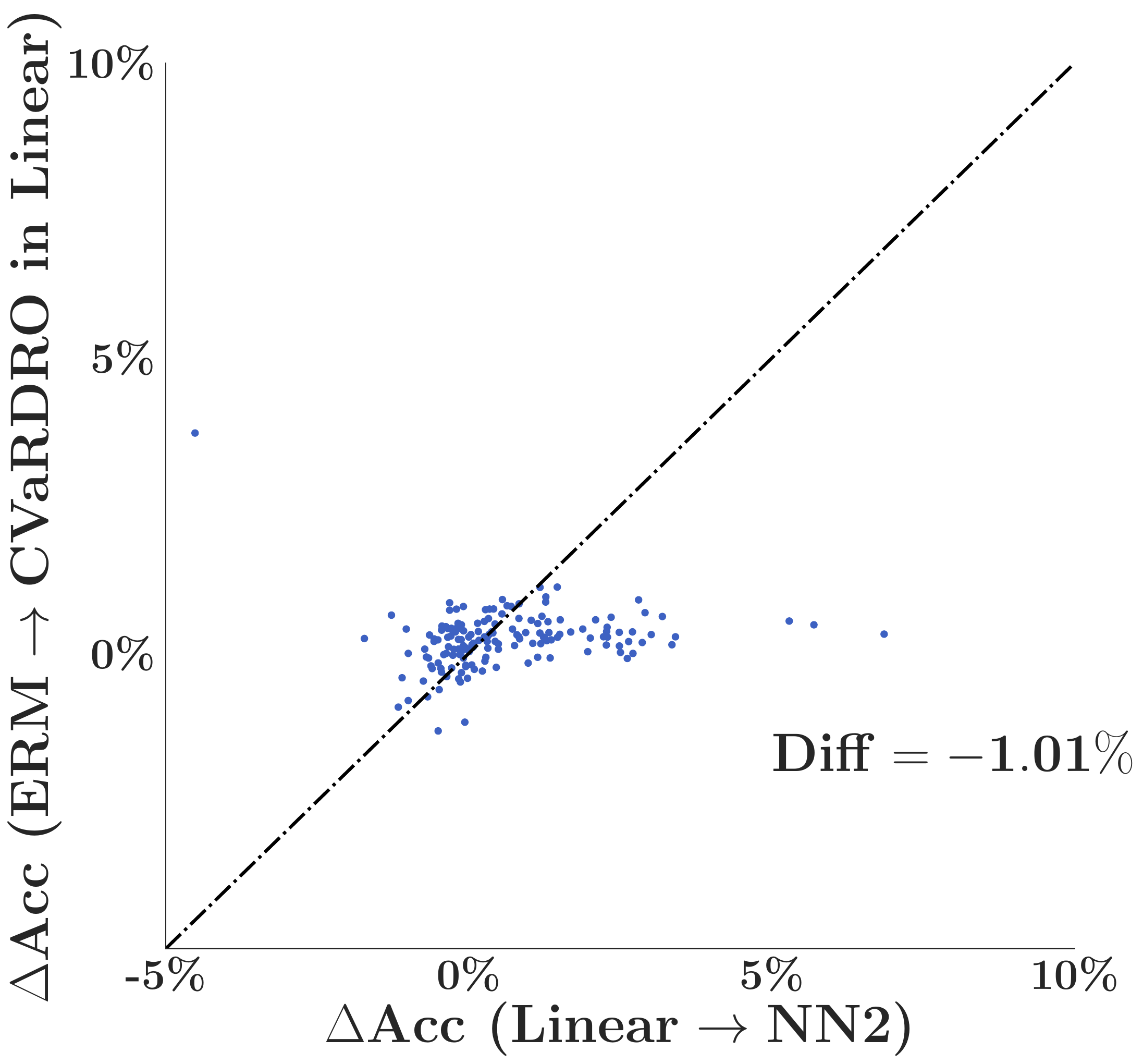}}{(d)}
 \stackunder[3pt]{\includegraphics[width=0.32\textwidth]{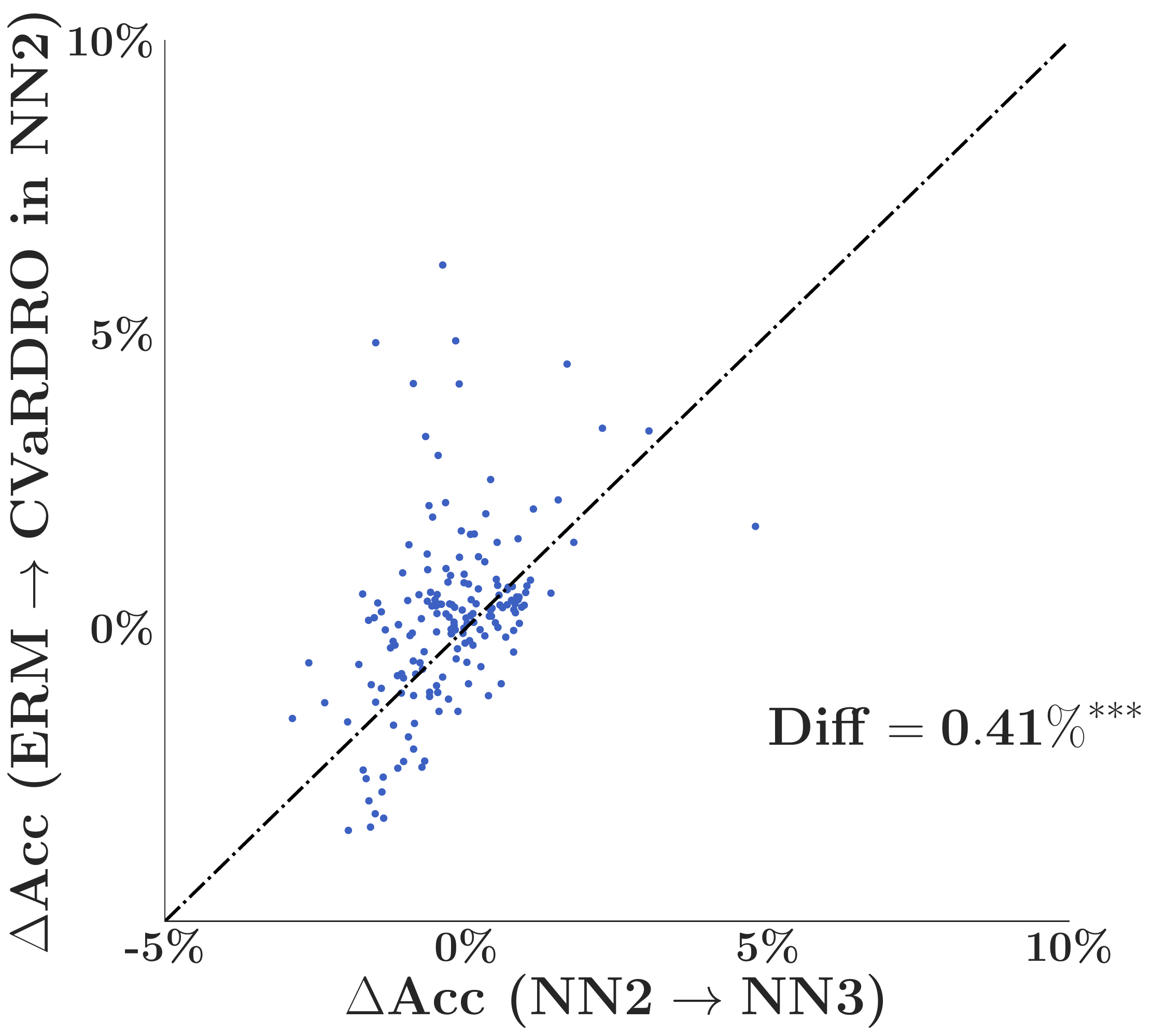}}{(e)}
 \stackunder[3pt]{\includegraphics[width=0.32\textwidth]{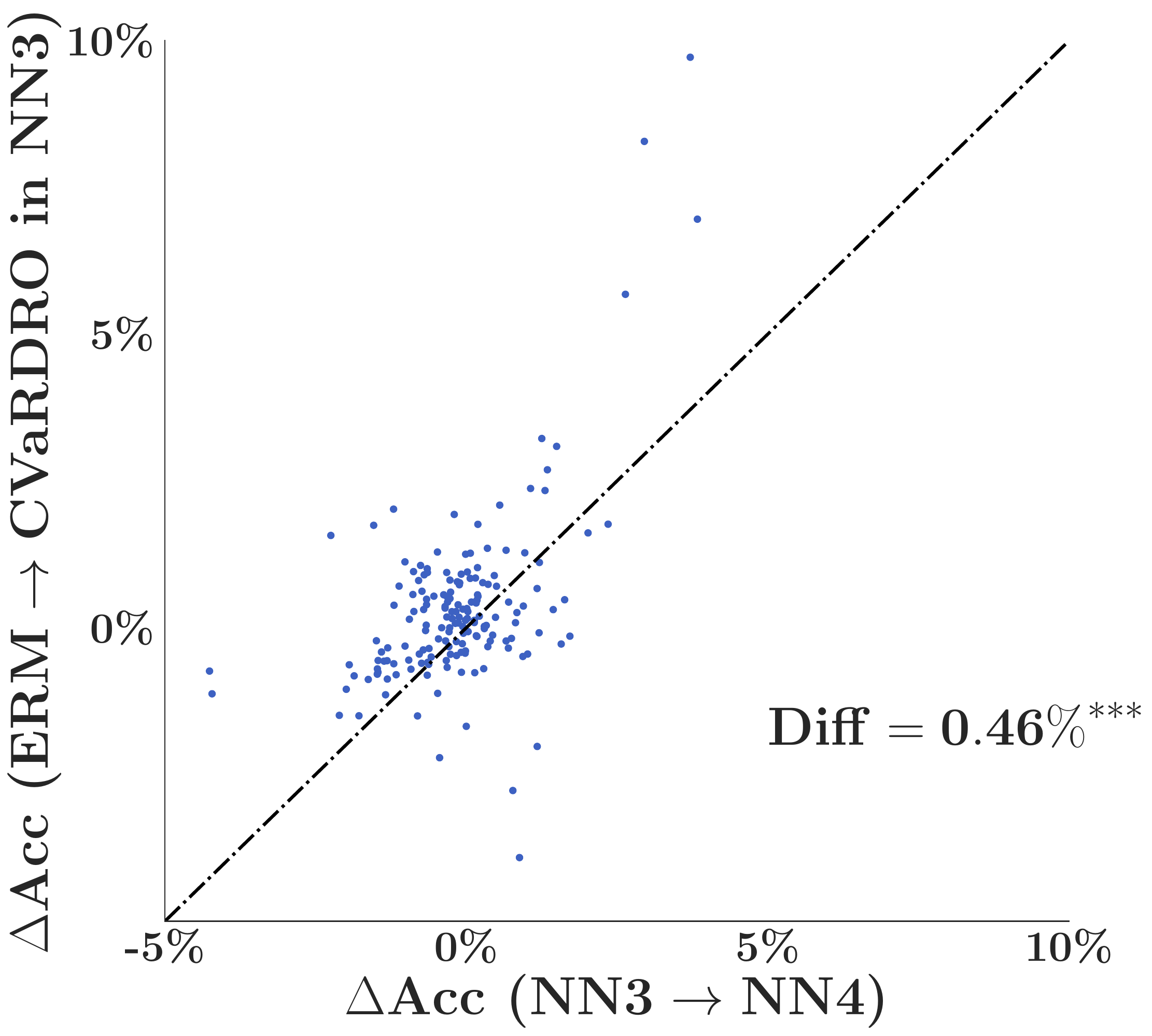}}{(f)}
  \stackunder[3pt]{\includegraphics[width=0.32\textwidth]{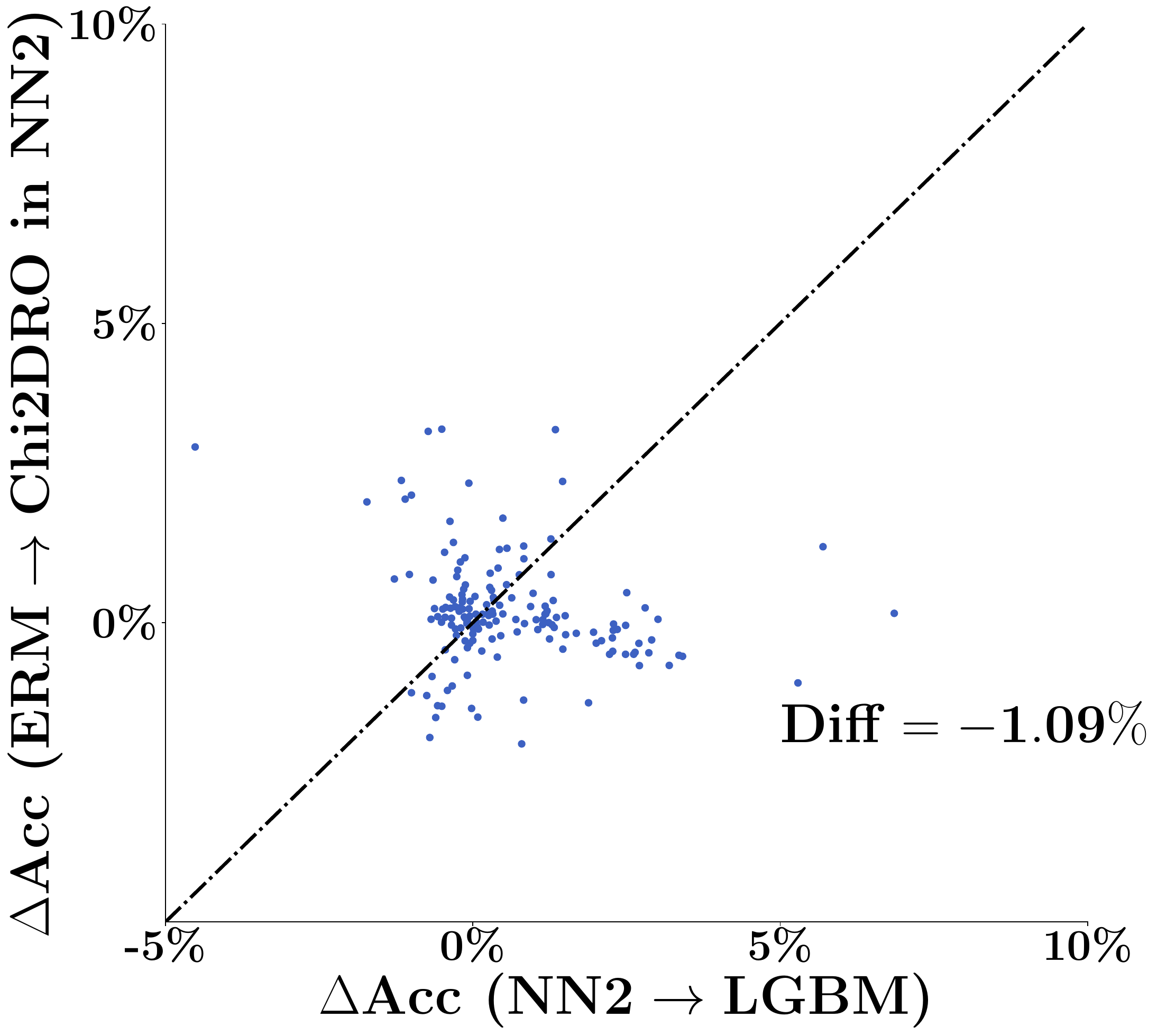}}{(g)}
 \stackunder[3pt]{\includegraphics[width=0.32\textwidth]{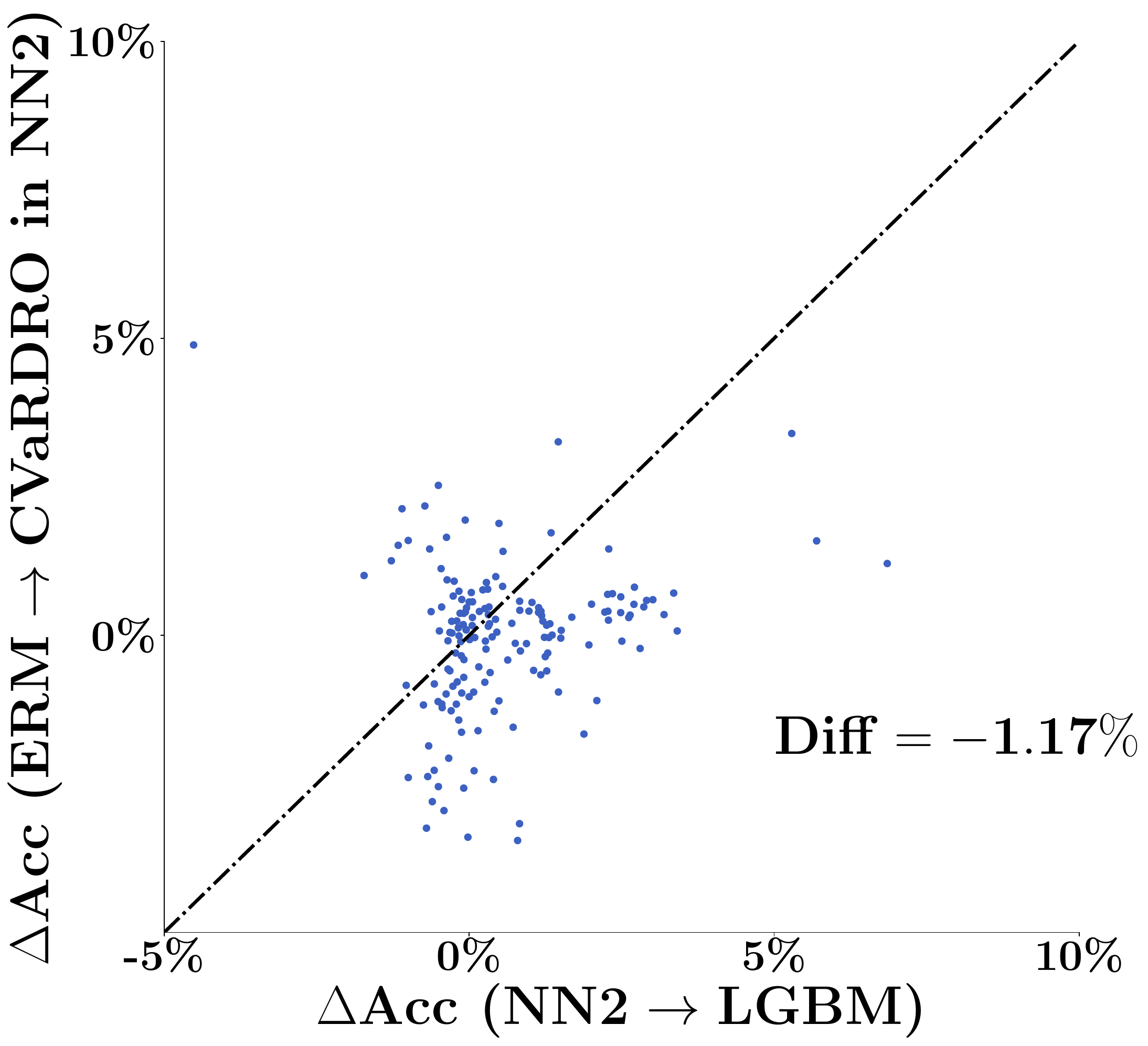}}{(h)}
\vspace{0.1in}
\caption{Scatter plot comparing accuracy gains from interventions across distribution pairs and settings. The horizontal axis shows the accuracy improvement from model-class changes, while the vertical axis shows the accuracy improvement from DRO ambiguity set designs. Each point corresponds to one source–target pair. The difference between the two interventions (vertical minus horizontal) is reported within each panel. Superscripts $^{*}, ^{**}$, and $^{***}$ indicate statistical significance at the 1\%, 5\%, and 10\% levels, respectively, based on one-sided tests of the null hypothesis that the accuracy gain from model-class changes exceeds that from DRO ambiguity design.}
 \label{fig:acc-diff-comp-nn}  
\end{figure}
\begin{figure}[!htb]
 \centering\captionsetup[subfloat]{labelfont=scriptsize,textfont=scriptsize} 
 \stackunder[3pt]{\includegraphics[width=0.8\textwidth]{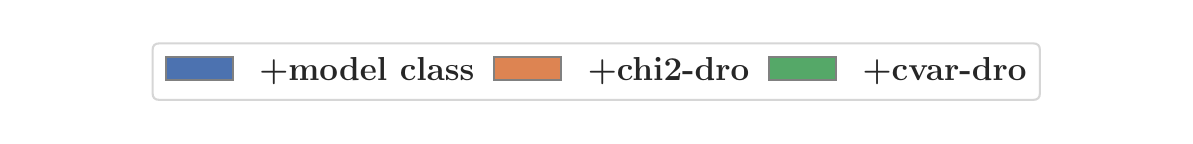}}{}
 \stackunder[3pt]{\includegraphics[width=0.32\textwidth]{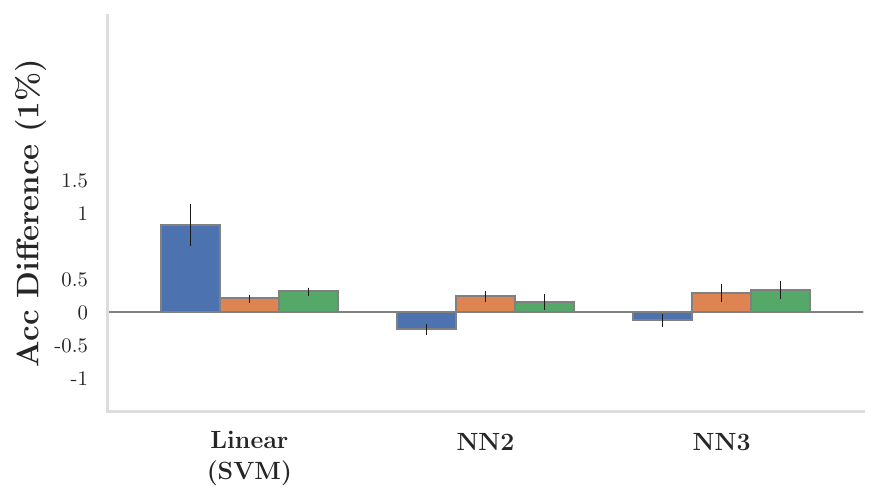}}{(a) All settings}
 \stackunder[3pt]{\includegraphics[width=0.32\textwidth]{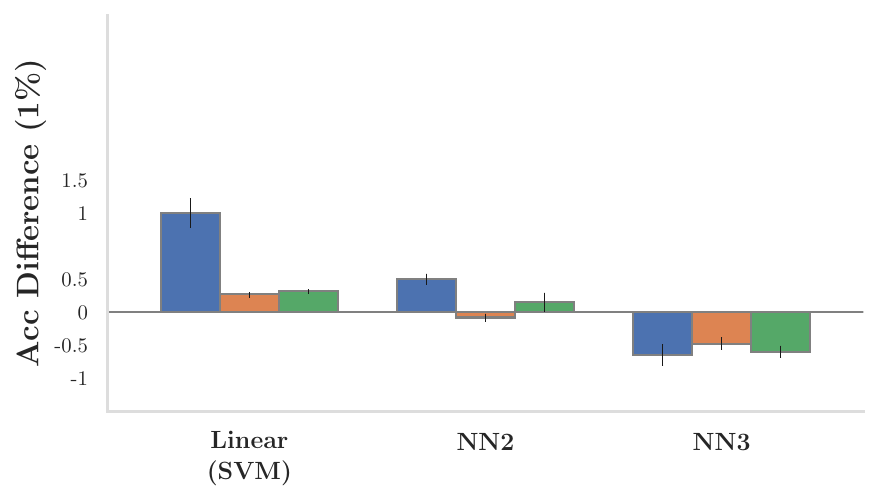}}{(b) Setting 1}
 \stackunder[3pt]{\includegraphics[width=0.32\textwidth]{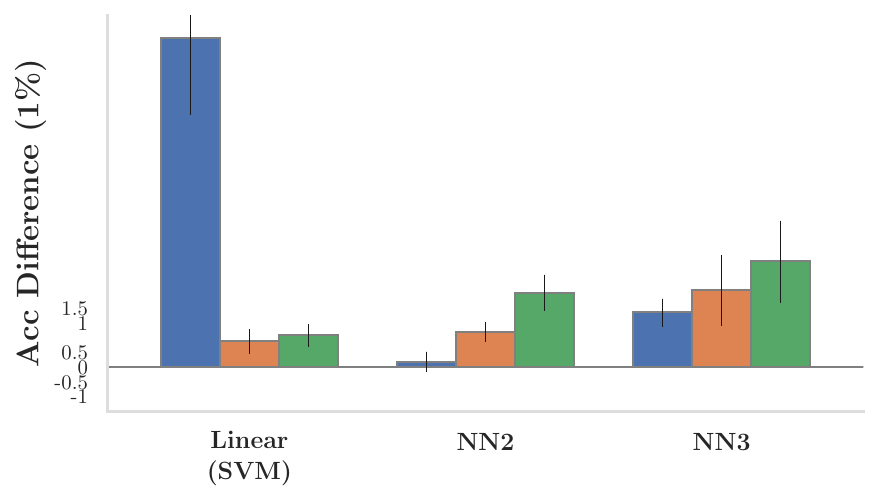}}{(c) Setting 5}
\vspace{0.1in}
\caption{Average accuracy gains (and the standard deviations) from increasing model class (linear to neural networks varying the number of layers versus DRO ambiguity set design among all domains in each setting. Here, ``+model class'' denotes incremental increases in model complexity (e.g., linear to NN2, NN2 to NN3, NN3 to NN4). Each bar group corresponds to a fixed base model class.}
 \label{fig:acc-diff-nn}  
\end{figure}
}

\twy{
This pattern indicates that DRO interventions yield significant improvements when applied across incremental changes within the neural network model class. However, this finding must be interpreted carefully: while DRO applied to NN2 shows substantial relative gains, neural network-based methods overall achieve lower absolute accuracy in our problem setup, indicating that neural networks are not well-suited to our semi-synthetic shifts. Consequently, despite the apparent effectiveness of DRO within the neural network family, these methods do not achieve the best overall performance due to the inherent limitations of the neural network model class for our problems. This interpretation is further supported by panels $(g)$ and $(h)$ in Figure~\ref{fig:acc-diff-comp-nn}, which demonstrate that transitioning from NN2 to LightGBM produces substantially larger accuracy gains than any DRO intervention, highlighting the primacy of model class selection over robustness design when models are mismatched to the problem structure.}

\twy{\paragraph{LightGBM Analysis.} To better isolate the effects of fine-grained model class changes from ambiguity set design, we conduct additional experiments focusing on interventions of LightGBM, which proves to be a better model class with high accuracy in Section~\ref{sec:method-lr}. We represent fine-grained model class changes by controlling the number of base estimators and tuning over other hyperparameters.}

\twy{As shown in all panels of Figure~\ref{fig:acc-diff-comp-tree}, when using a well-suited model class like LightGBM, regardless of the specific base model class, fine-grained changes in model complexity consistently outperform the effects of DRO ambiguity set design. The aggregated results in Figure~\ref{fig:acc-diff-lgbm} further support this finding. Across all settings, incorporating distributional robustness does not even provide improvements when the base model class is LightGBM.}

\twy{\begin{figure}[!htb]
 \centering\captionsetup[subfloat]{labelfont=scriptsize,textfont=scriptsize} 
 \stackunder[3pt]{\includegraphics[width=0.32\textwidth]{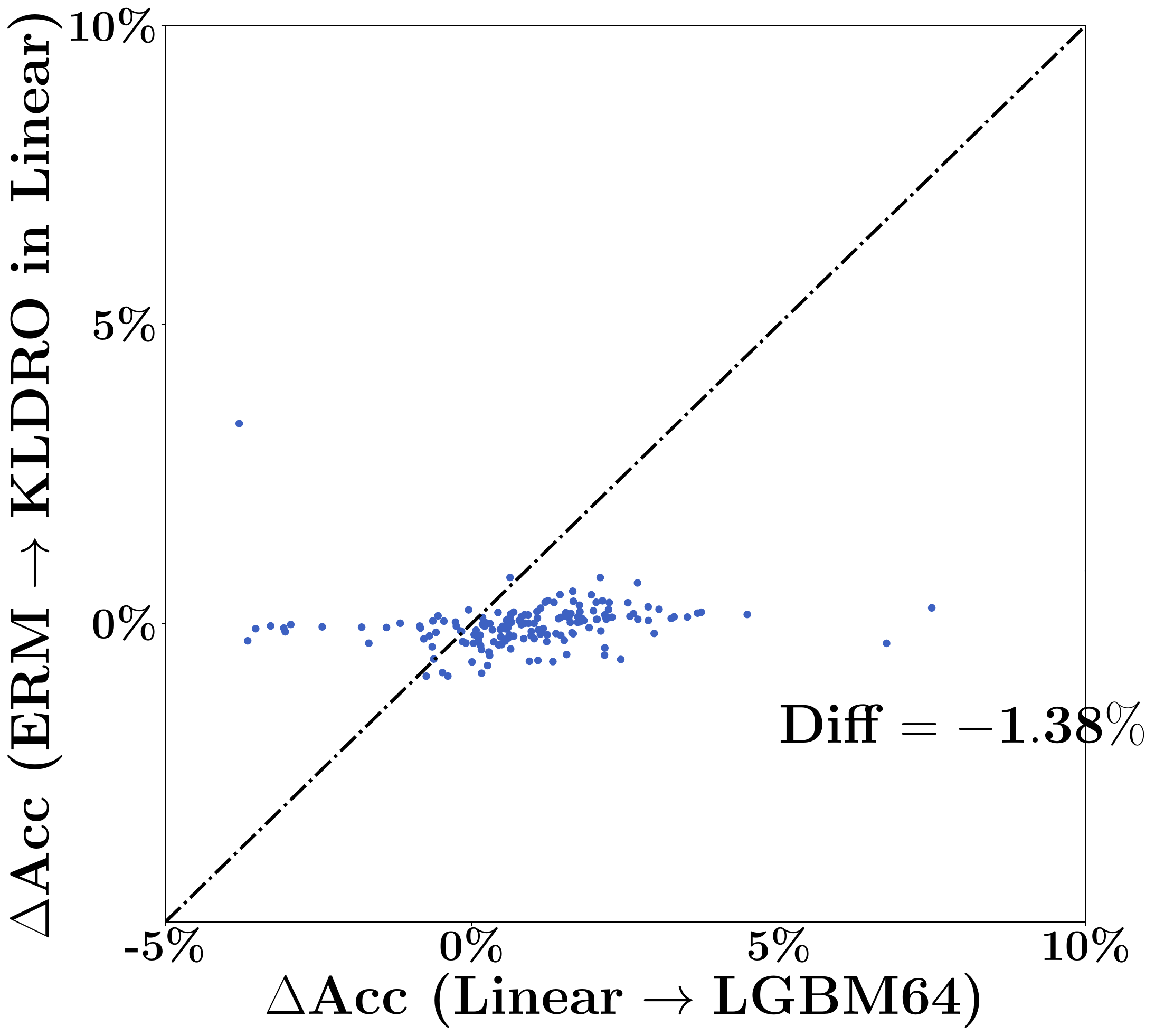}}{(a)}
 \stackunder[3pt]{\includegraphics[width=0.32\textwidth]{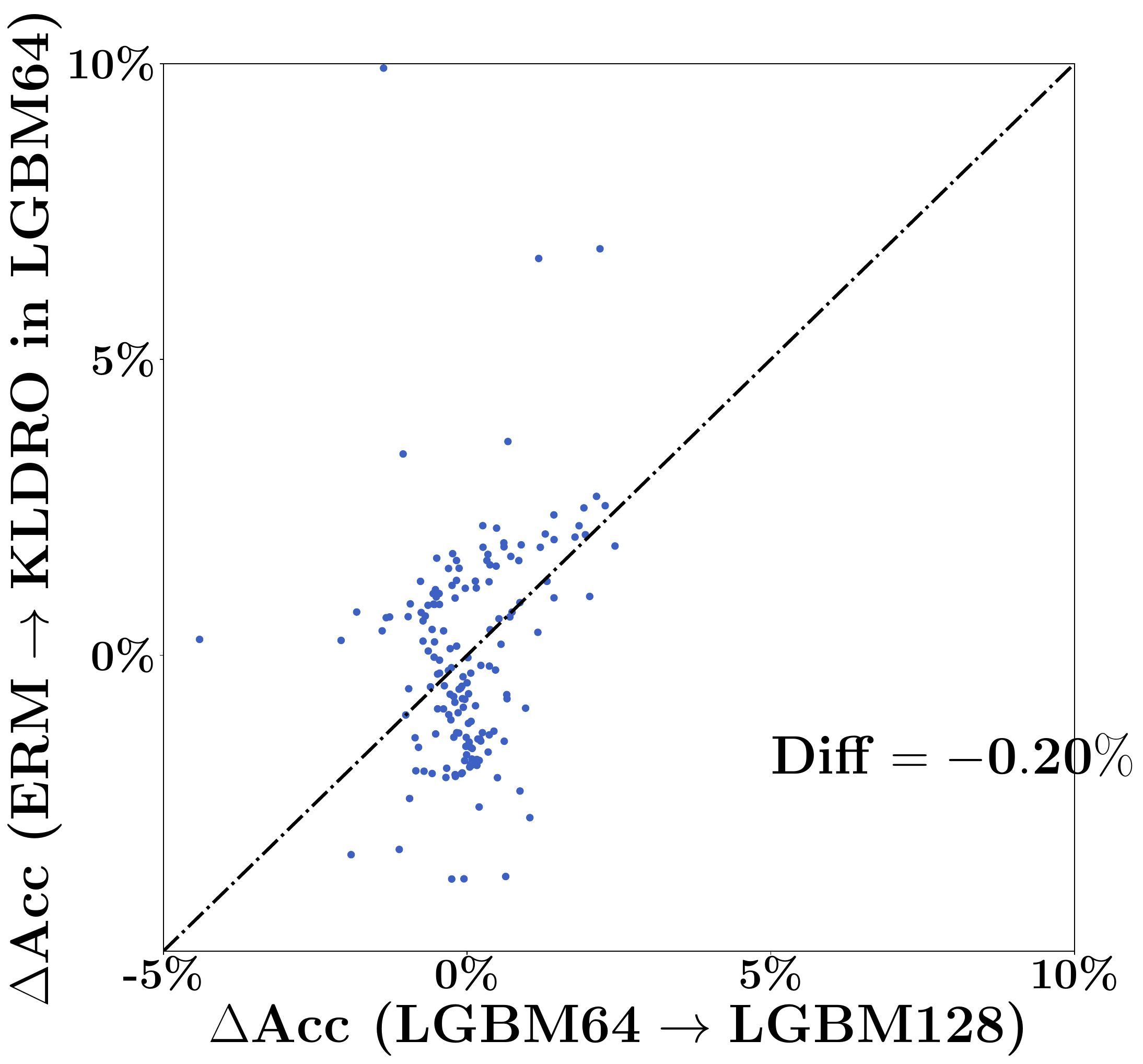}}{(b)}
 \stackunder[3pt]{\includegraphics[width=0.32\textwidth]{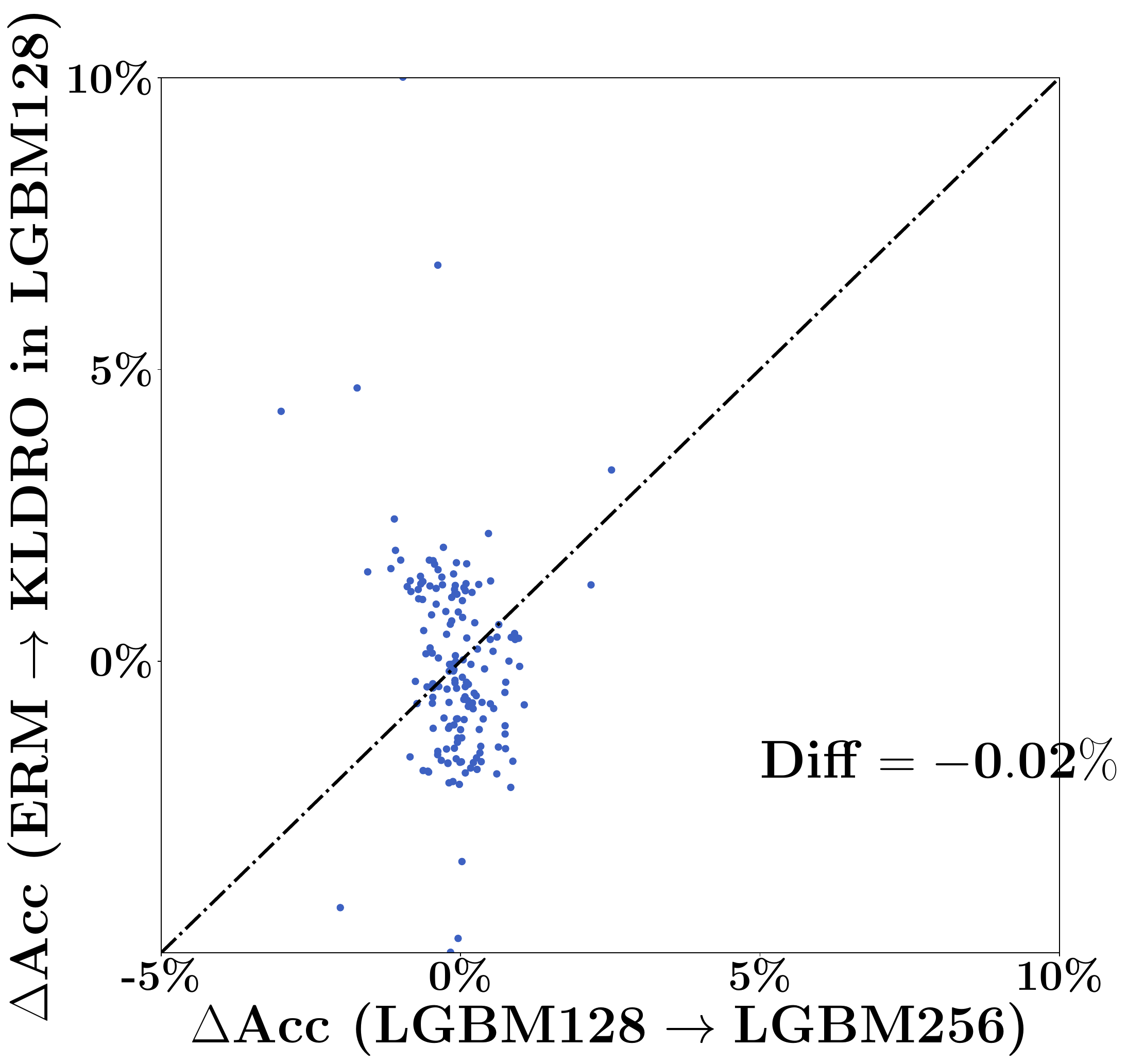}}{(c)}
 \stackunder[3pt]{\includegraphics[width=0.32\textwidth]{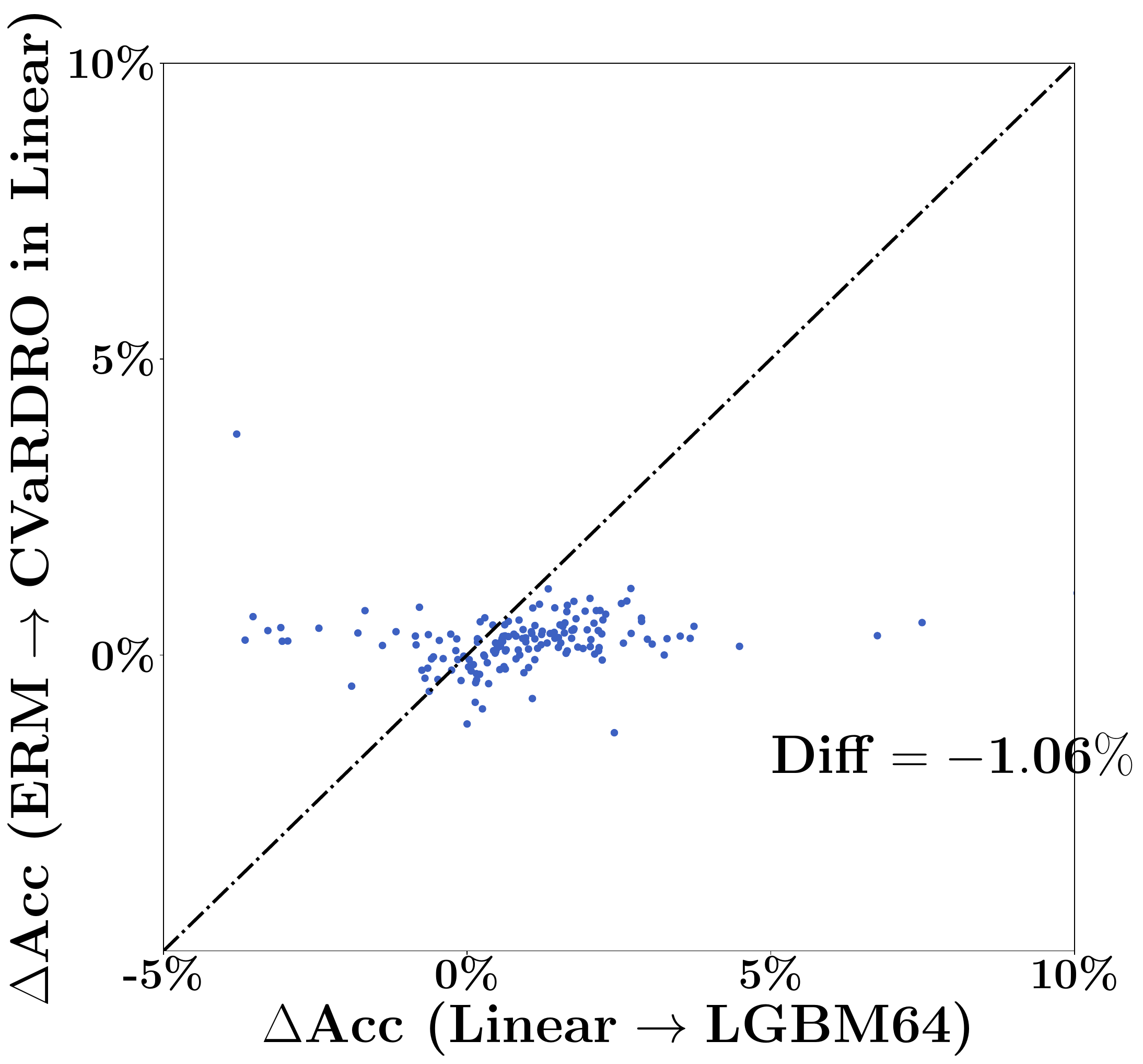}}{(d)}
 \stackunder[3pt]{\includegraphics[width=0.32\textwidth]{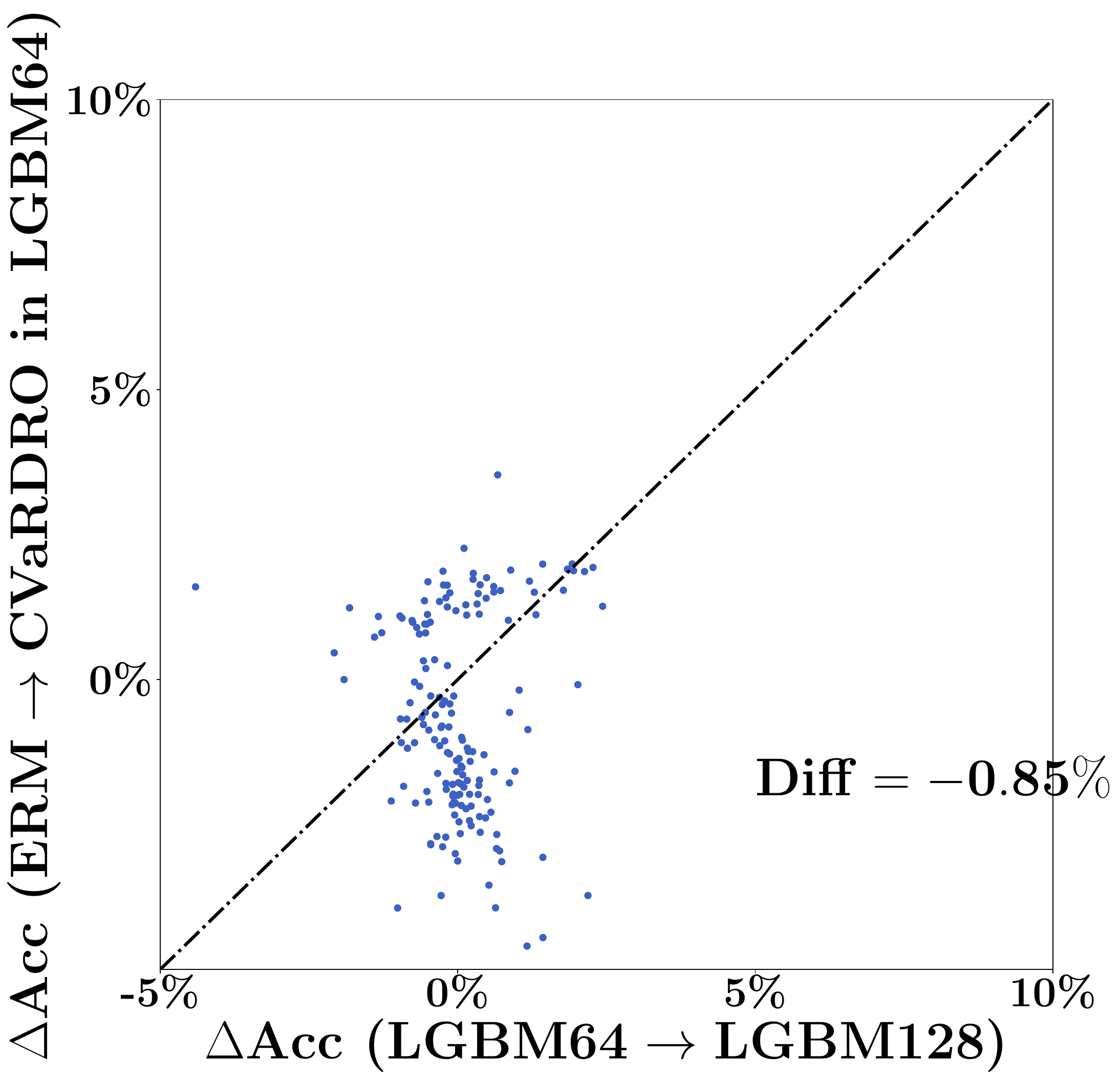}}{(e)}
 \stackunder[3pt]{\includegraphics[width=0.32\textwidth]{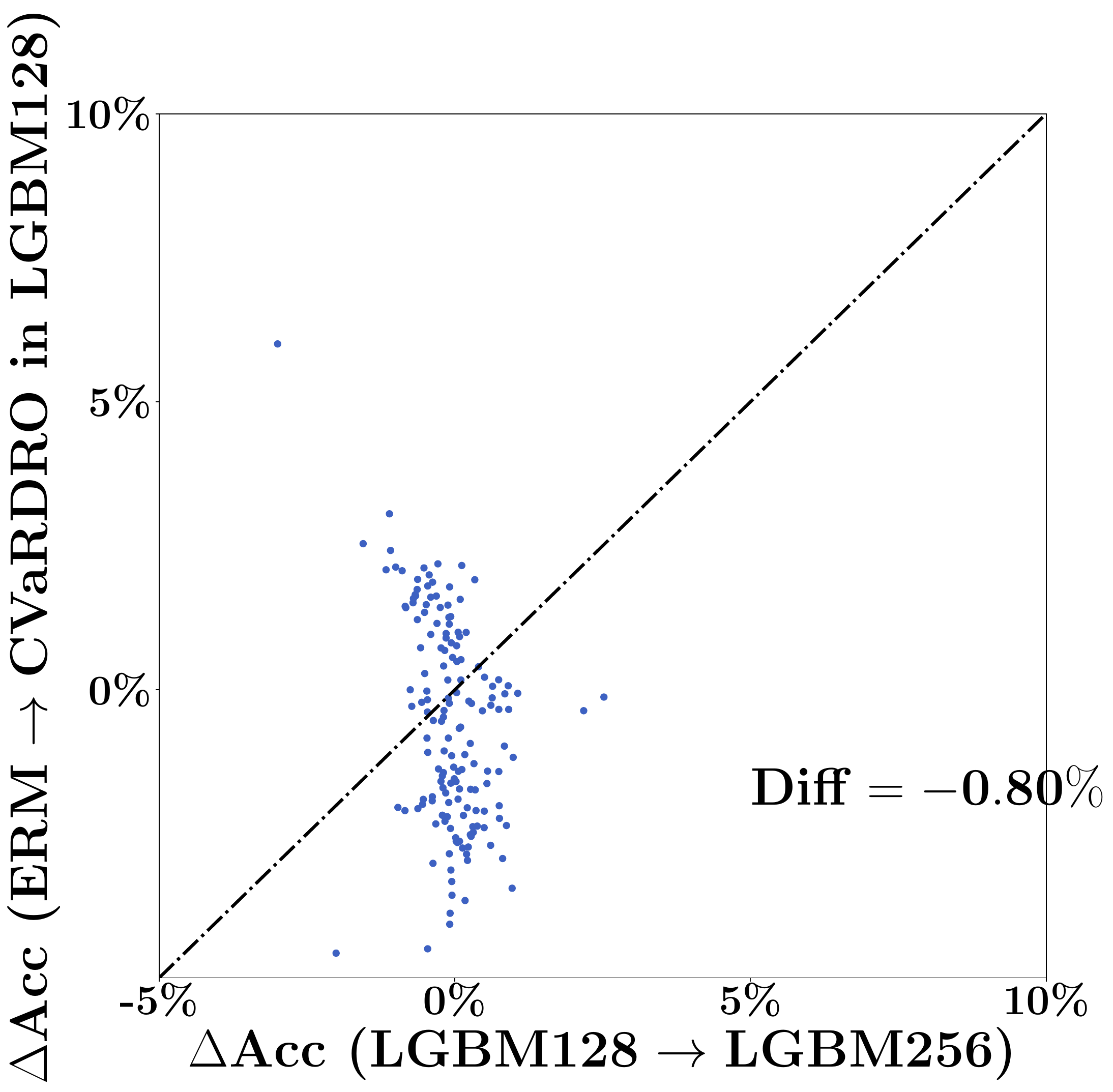}}{(f)}
\vspace{0.1in}
\caption{Scatter plot comparing accuracy gains from interventions across distribution pairs and settings. The horizontal axis shows the accuracy improvement from model-class changes, while the vertical axis shows the accuracy improvement from DRO ambiguity set designs. Each point corresponds to one source-target pair. The difference between the two interventions (vertical minus horizontal) is reported within each panel. Superscripts $^{*}, ^{**}$, and $^{***}$ indicate statistical significance at the 1\%, 5\%, and 10\% levels, respectively, based on one-sided tests of the null hypothesis that the accuracy gain from model-class changes exceeds that from DRO ambiguity design.}
 \label{fig:acc-diff-comp-tree}  
\end{figure}
\begin{figure}[!htb]
 \centering\captionsetup[subfloat]{labelfont=scriptsize,textfont=scriptsize} 
 \stackunder[3pt]{\includegraphics[width=0.8\textwidth]{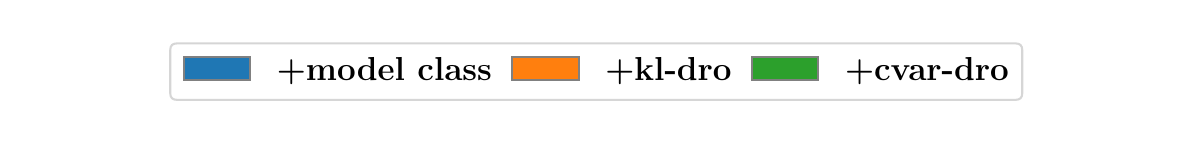}}{}
 \stackunder[3pt]{\includegraphics[width=0.32\textwidth]{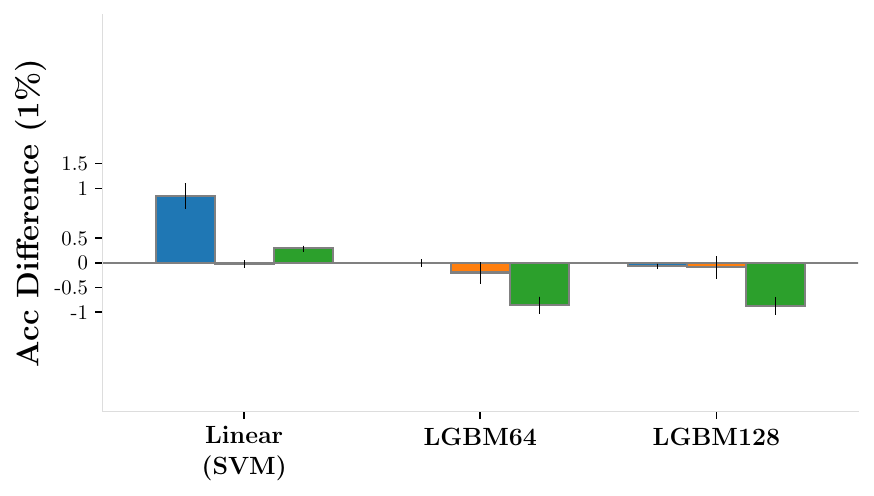}}{(a) All settings}
 \stackunder[3pt]{\includegraphics[width=0.32\textwidth]{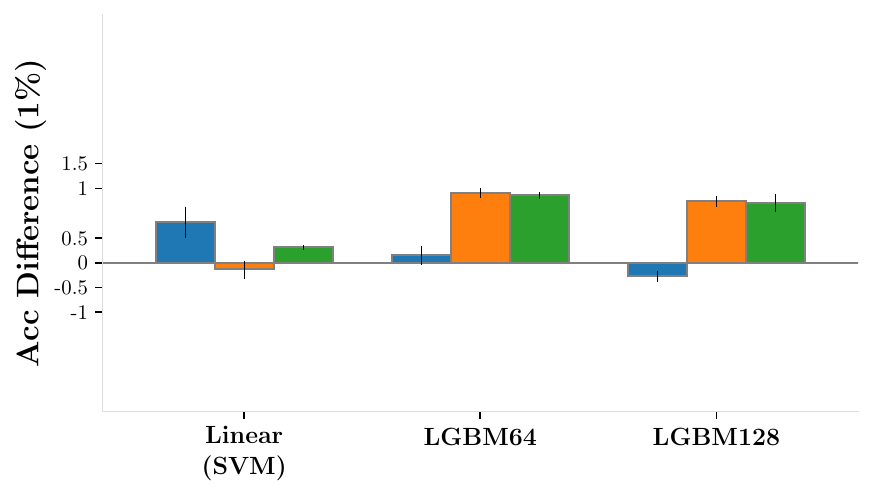}}{(b) Setting 1}
 \stackunder[3pt]{\includegraphics[width=0.32\textwidth]{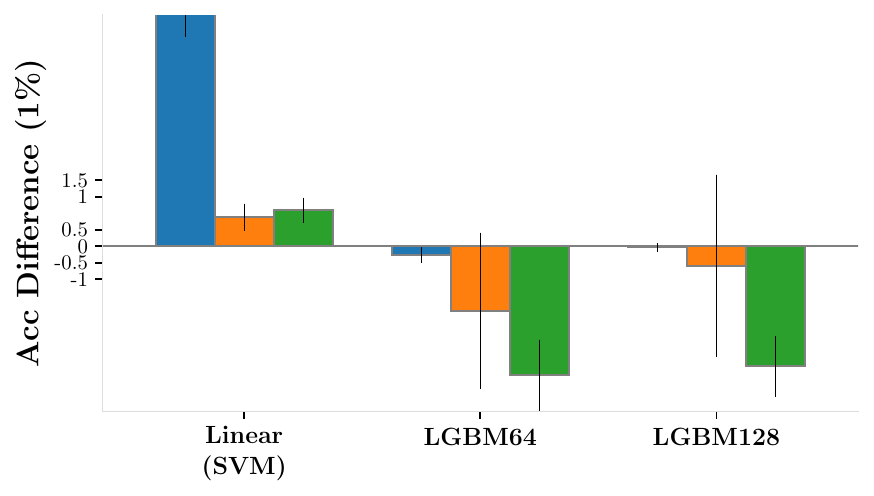}}{(c) Setting 5}
\vspace{0.1in}
\caption{Average accuracy gains (and the standard deviations) from increasing model class (linear to LightGBM varying the number of base estimators versus DRO ambiguity set design among all domains in each setting. Here, ``+model class'' denotes incremental increases in model complexity (e.g., linear to LightGBM with 64 estimators, LightGBM with 64 estimators to 128 estimators). Each bar group corresponds to a fixed base model class.}
 \label{fig:acc-diff-lgbm}  
\end{figure}
\paragraph{Summary.} Our analysis suggests that when the model class is appropriately suitable to the problem (e.g., LightGBM in our case), incremental improvements in model complexity tend to provide larger accuracy gains than distributional robustness interventions. We acknowledge the limitations of this empirical analysis and leave further theoretical quantification of the trade-offs between fine-grained model class selection and robustness design as future work.
}

\subsection{(Non-)Linear Regression Results for Ablation Studies}\label{app:ablation}
We consider the interaction effects of model class and shift patterns, and the nonlinear effects of the radius in the ambiguity set in the linear regression \eqref{eq:lr-model} and provide results in Tables~\ref{tab:linear-analysis-aug} and~\ref{tab:linear-analysis-aug2}. We provide full definitions of variables in \Cref{tab:var_def} and include three categories of variables: $\text{Radius}^2$ as the nonlinear effect of the radius in the ambiguity set, interaction terms between the model class and shift patterns, and interaction terms between the model class and robustness radius in \Cref{tab:var_def2}. We also change the dependent variable to the performance gap~\eqref{eq:src-tgt-acc-def} in~\eqref{eq:lr-model} while keeping all these independent variables in \Cref{tab:var_def} to understand the effects of robust models against distribution shifts and report results in Table~\ref{tab:linear-analysis-aug-gap}.

\begin{table}[htbp]
    \centering
    \caption{Regression results on algorithmic design components (\texttt{Best Config}, incorporating nonlinear effects of the radius and additional interaction effects between the model class and robustness radius, between the model class and shift patterns) on the method performance}
    \label{tab:linear-analysis-aug}
    \sisetup{
    table-format=-1.4, 
    add-integer-zero=false 
    }
    
	\resizebox{\textwidth}{!}{\begin{tabular}{llSSSSSSS}
        \toprule
        & & \multicolumn{7}{c}{Dependent variable: Accuracy}\\
        \cmidrule(lr){3-9}
    \multicolumn{2}{c}{Variable Name} & \multicolumn{1}{c}{\ \ All} & \multicolumn{1}{c}{\ \ Setting 1} & \multicolumn{1}{c}{\ \ Setting 2} & \multicolumn{1}{c}{\ \ Setting 3} & \multicolumn{1}{c}{\ \ Setting 4} &  \multicolumn{1}{c}{\ \ Setting 5} & \multicolumn{1}{c}{\ \ Setting 6}\\
    \midrule
\multirow{3}{*}{\parbox[c]{1.5cm}{Model\\Class}}&LGBM& .0102$^{***}$& .0169$^{***}$& .0101$^{***}$&-.0162$^{**}$&-.0103$^{***}$& .0567$^{***}$& .0313$^{***}$\\
&&\multicolumn{1}{c}{ \scriptsize\hspace{0.35cm}(.0022)}&\multicolumn{1}{c}{ \scriptsize\hspace{0.35cm}(.0016)}&\multicolumn{1}{c}{ \scriptsize\hspace{0.35cm}(.0009)}&\multicolumn{1}{c}{ \scriptsize\hspace{0.35cm}(.0077)}&\multicolumn{1}{c}{ \scriptsize\hspace{0.35cm}(.0027)}&\multicolumn{1}{c}{ \scriptsize\hspace{0.35cm}(.0072)}&\multicolumn{1}{c}{ \scriptsize\hspace{0.35cm}(.0057)}\\
&NN2& .0157$^{***}$& .0140$^{***}$& .0050$^{***}$& .0051& .0055$^{**}$& .0725$^{***}$& .0299$^{***}$\\
&&\multicolumn{1}{c}{ \scriptsize\hspace{0.35cm}(.0020)}&\multicolumn{1}{c}{ \scriptsize\hspace{0.35cm}(.0016)}&\multicolumn{1}{c}{ \scriptsize\hspace{0.35cm}(.0008)}&\multicolumn{1}{c}{ \scriptsize\hspace{0.35cm}(.0074)}&\multicolumn{1}{c}{ \scriptsize\hspace{0.35cm}(.0027)}&\multicolumn{1}{c}{ \scriptsize\hspace{0.35cm}(.0067)}&\multicolumn{1}{c}{ \scriptsize\hspace{0.35cm}(.0057)}\\
&NN3& .0131$^{***}$& .0119$^{***}$&-.0000& .0135$^{*}$&-.0044& .0761$^{***}$& .0273$^{***}$\\
&&\multicolumn{1}{c}{ \scriptsize\hspace{0.35cm}(.0022)}&\multicolumn{1}{c}{ \scriptsize\hspace{0.35cm}(.0016)}&\multicolumn{1}{c}{ \scriptsize\hspace{0.35cm}(.0009)}&\multicolumn{1}{c}{ \scriptsize\hspace{0.35cm}(.0072)}&\multicolumn{1}{c}{ \scriptsize\hspace{0.35cm}(.0028)}&\multicolumn{1}{c}{ \scriptsize\hspace{0.35cm}(.0067)}&\multicolumn{1}{c}{ \scriptsize\hspace{0.35cm}(.0057)}\\
&NN4& .0088$^{***}$& .0149$^{***}$&-.0027$^{***}$& .0227$^{***}$&-.0067$^{**}$& .0803$^{***}$& .0271$^{***}$\\
&&\multicolumn{1}{c}{ \scriptsize\hspace{0.35cm}(.0025)}&\multicolumn{1}{c}{ \scriptsize\hspace{0.35cm}(.0017)}&\multicolumn{1}{c}{ \scriptsize\hspace{0.35cm}(.0009)}&\multicolumn{1}{c}{ \scriptsize\hspace{0.35cm}(.0079)}&\multicolumn{1}{c}{ \scriptsize\hspace{0.35cm}(.0029)}&\multicolumn{1}{c}{ \scriptsize\hspace{0.35cm}(.0081)}&\multicolumn{1}{c}{ \scriptsize\hspace{0.35cm}(.0062)}\\
&Kernel& .0025& .0012&-.0041$^{***}$& .0029&-.0207$^{***}$& .0426$^{***}$&-.0206$^{***}$\\
&&\multicolumn{1}{c}{ \scriptsize\hspace{0.35cm}(.0018)}&\multicolumn{1}{c}{ \scriptsize\hspace{0.35cm}(.0014)}&\multicolumn{1}{c}{ \scriptsize\hspace{0.35cm}(.0006)}&\multicolumn{1}{c}{ \scriptsize\hspace{0.35cm}(.0062)}&\multicolumn{1}{c}{ \scriptsize\hspace{0.35cm}(.0022)}&\multicolumn{1}{c}{ \scriptsize\hspace{0.35cm}(.0055)}&\multicolumn{1}{c}{ \scriptsize\hspace{0.35cm}(.0050)}\\
\midrule
\multirow{3}{*}{\parbox[c]{1.7cm}{Ambiguity\\Set}}&Wasserstein& .0058$^{***}$& .0020$^{**}$& .0024$^{***}$& .0078$^{*}$& .0160$^{***}$&-.0086& .0216$^{***}$\\
&&\multicolumn{1}{c}{ \scriptsize\hspace{0.35cm}(.0019)}&\multicolumn{1}{c}{ \scriptsize\hspace{0.35cm}(.0008)}&\multicolumn{1}{c}{ \scriptsize\hspace{0.35cm}(.0007)}&\multicolumn{1}{c}{ \scriptsize\hspace{0.35cm}(.0041)}&\multicolumn{1}{c}{ \scriptsize\hspace{0.35cm}(.0018)}&\multicolumn{1}{c}{ \scriptsize\hspace{0.35cm}(.0060)}&\multicolumn{1}{c}{ \scriptsize\hspace{0.35cm}(.0033)}\\
&Chi-squared& .0006& .0015$^{**}$& .0032$^{***}$&-.0006&-.0105$^{***}$&-.0027& .0139$^{***}$\\
&&\multicolumn{1}{c}{ \scriptsize\hspace{0.35cm}(.0013)}&\multicolumn{1}{c}{ \scriptsize\hspace{0.35cm}(.0008)}&\multicolumn{1}{c}{ \scriptsize\hspace{0.35cm}(.0006)}&\multicolumn{1}{c}{ \scriptsize\hspace{0.35cm}(.0048)}&\multicolumn{1}{c}{ \scriptsize\hspace{0.35cm}(.0016)}&\multicolumn{1}{c}{ \scriptsize\hspace{0.35cm}(.0075)}&\multicolumn{1}{c}{ \scriptsize\hspace{0.35cm}(.0031)}\\
&Kullback-Leibler& .0003& .0021$^{***}$& .0040$^{***}$& .0013& .0008&-.0095$^{*}$&-.0011\\
&&\multicolumn{1}{c}{ \scriptsize\hspace{0.35cm}(.0015)}&\multicolumn{1}{c}{ \scriptsize\hspace{0.35cm}(.0008)}&\multicolumn{1}{c}{ \scriptsize\hspace{0.35cm}(.0007)}&\multicolumn{1}{c}{ \scriptsize\hspace{0.35cm}(.0046)}&\multicolumn{1}{c}{ \scriptsize\hspace{0.35cm}(.0019)}&\multicolumn{1}{c}{ \scriptsize\hspace{0.35cm}(.0052)}&\multicolumn{1}{c}{ \scriptsize\hspace{0.35cm}(.0030)}\\
&Total Variation&-.0054$^{**}$&-.0006&-.0027$^{***}$&-.0024&-.0153$^{***}$&-.0034& .0043\\
&&\multicolumn{1}{c}{ \scriptsize\hspace{0.35cm}(.0024)}&\multicolumn{1}{c}{ \scriptsize\hspace{0.35cm}(.0010)}&\multicolumn{1}{c}{ \scriptsize\hspace{0.35cm}(.0009)}&\multicolumn{1}{c}{ \scriptsize\hspace{0.35cm}(.0058)}&\multicolumn{1}{c}{ \scriptsize\hspace{0.35cm}(.0027)}&\multicolumn{1}{c}{ \scriptsize\hspace{0.35cm}(.0118)}&\multicolumn{1}{c}{ \scriptsize\hspace{0.35cm}(.0045)}\\
&OT-Discrepancy&-.0008&-.0020$^{**}$&-.0002& .0004&-.0026&-.0058& .0029\\
&&\multicolumn{1}{c}{ \scriptsize\hspace{0.35cm}(.0024)}&\multicolumn{1}{c}{ \scriptsize\hspace{0.35cm}(.0010)}&\multicolumn{1}{c}{ \scriptsize\hspace{0.35cm}(.0009)}&\multicolumn{1}{c}{ \scriptsize\hspace{0.35cm}(.0058)}&\multicolumn{1}{c}{ \scriptsize\hspace{0.35cm}(.0030)}&\multicolumn{1}{c}{ \scriptsize\hspace{0.35cm}(.0081)}&\multicolumn{1}{c}{ \scriptsize\hspace{0.35cm}(.0045)}\\
&Radius& .0124$^{***}$& .0033&-.0147$^{***}$&-.0005& .0153$^{**}$&-.0221& .0254$^{***}$\\
&&\multicolumn{1}{c}{ \scriptsize\hspace{0.35cm}(.0021)}&\multicolumn{1}{c}{ \scriptsize\hspace{0.35cm}(.0024)}&\multicolumn{1}{c}{ \scriptsize\hspace{0.35cm}(.0014)}&\multicolumn{1}{c}{ \scriptsize\hspace{0.35cm}(.0109)}&\multicolumn{1}{c}{ \scriptsize\hspace{0.35cm}(.0077)}&\multicolumn{1}{c}{ \scriptsize\hspace{0.35cm}(.0473)}&\multicolumn{1}{c}{ \scriptsize\hspace{0.35cm}(.0053)}\\
&Radius2&-.0011&-.0001& .0175$^{***}$& .0048& .0004& .0224&-.0137$^{***}$\\
&&\multicolumn{1}{c}{ \scriptsize\hspace{0.35cm}(.0007)}&\multicolumn{1}{c}{ \scriptsize\hspace{0.35cm}(.0020)}&\multicolumn{1}{c}{ \scriptsize\hspace{0.35cm}(.0011)}&\multicolumn{1}{c}{ \scriptsize\hspace{0.35cm}(.0081)}&\multicolumn{1}{c}{ \scriptsize\hspace{0.35cm}(.0059)}&\multicolumn{1}{c}{ \scriptsize\hspace{0.35cm}(.0493)}&\multicolumn{1}{c}{ \scriptsize\hspace{0.35cm}(.0012)}\\
&LGBM-Radius&-.0111$^{***}$& .0014&-.0292$^{***}$&-.0163$^{**}$&-.0238$^{***}$&-.0186&-.0077\\
&&\multicolumn{1}{c}{ \scriptsize\hspace{0.35cm}(.0026)}&\multicolumn{1}{c}{ \scriptsize\hspace{0.35cm}(.0015)}&\multicolumn{1}{c}{ \scriptsize\hspace{0.35cm}(.0021)}&\multicolumn{1}{c}{ \scriptsize\hspace{0.35cm}(.0079)}&\multicolumn{1}{c}{ \scriptsize\hspace{0.35cm}(.0040)}&\multicolumn{1}{c}{ \scriptsize\hspace{0.35cm}(.0128)}&\multicolumn{1}{c}{ \scriptsize\hspace{0.35cm}(.0057)}\\
&NN2-Radius&-.0355$^{***}$&-.1920&-.0049&-.0164&-.0478$^{***}$& .9804&-.0753$^{***}$\\
&&\multicolumn{1}{c}{ \scriptsize\hspace{0.35cm}(.0061)}&\multicolumn{1}{c}{ \scriptsize\hspace{0.35cm}(.1233)}&\multicolumn{1}{c}{ \scriptsize\hspace{0.35cm}(.0100)}&\multicolumn{1}{c}{ \scriptsize\hspace{0.35cm}(.0107)}&\multicolumn{1}{c}{ \scriptsize\hspace{0.35cm}(.0052)}&\multicolumn{1}{c}{ \scriptsize\hspace{0.35cm}(.1164)}&\multicolumn{1}{c}{ \scriptsize\hspace{0.35cm}(.0173)}\\
&NN3-Radius&-.0309$^{***}$& .1406& .0109$^{***}$&-.6364&-.0499$^{***}$& .1362&-.0513$^{***}$\\
&&\multicolumn{1}{c}{ \scriptsize\hspace{0.35cm}(.0048)}&\multicolumn{1}{c}{ \scriptsize\hspace{0.35cm}(.1233)}&\multicolumn{1}{c}{ \scriptsize\hspace{0.35cm}(.0017)}&\multicolumn{1}{c}{ \scriptsize\hspace{0.35cm}(.7382)}&\multicolumn{1}{c}{ \scriptsize\hspace{0.35cm}(.0053)}&\multicolumn{1}{c}{ \scriptsize\hspace{0.35cm}(.1255)}&\multicolumn{1}{c}{ \scriptsize\hspace{0.35cm}(.0173)}\\
&NN4-Radius&-.0314$^{***}$&-.0041& .0196$^{***}$&-.6235$^{**}$&-.1044$^{***}$& .0109&-.0906$^{***}$\\
&&\multicolumn{1}{c}{ \scriptsize\hspace{0.35cm}(.0091)}&\multicolumn{1}{c}{ \scriptsize\hspace{0.35cm}(.0063)}&\multicolumn{1}{c}{ \scriptsize\hspace{0.35cm}(.0034)}&\multicolumn{1}{c}{ \scriptsize\hspace{0.35cm}(.7382)}&\multicolumn{1}{c}{ \scriptsize\hspace{0.35cm}(.0113)}&\multicolumn{1}{c}{ \scriptsize\hspace{0.35cm}(.0354)}&\multicolumn{1}{c}{ \scriptsize\hspace{0.35cm}(.0257)}\\
&Kernel-Radius&-.0028&-.0017& .0079$^{***}$& .0182& .0818$^{***}$&-.0043& .0587$^{***}$\\
&&\multicolumn{1}{c}{ \scriptsize\hspace{0.35cm}(.0027)}&\multicolumn{1}{c}{ \scriptsize\hspace{0.35cm}(.0011)}&\multicolumn{1}{c}{ \scriptsize\hspace{0.35cm}(.0013)}&\multicolumn{1}{c}{ \scriptsize\hspace{0.35cm}(.0112)}&\multicolumn{1}{c}{ \scriptsize\hspace{0.35cm}(.0059)}&\multicolumn{1}{c}{ \scriptsize\hspace{0.35cm}(.0253)}&\multicolumn{1}{c}{ \scriptsize\hspace{0.35cm}(.0069)}\\
\midrule
\multirow{2}{*}{\parbox[c]{1.5cm}{Shift\\Pattern}}&$Y|X$-ratio&-.0038$^{***}$&0.0689$^{***}$ &-.0056$^{***}$& .2029$^{***}$& 0.2029$^{***}$&-.0056$^{***}$& .1958$^{***}$\\
&&\multicolumn{1}{c}{ \scriptsize\hspace{0.35cm}(.0004)}&\multicolumn{1}{c}{ \scriptsize\hspace{0.35cm}(.0014)}&\multicolumn{1}{c}{ \scriptsize\hspace{0.35cm}(.0002)}&\multicolumn{1}{c}{ \scriptsize\hspace{0.35cm}(.0024)}&\multicolumn{1}{c}{ \scriptsize\hspace{0.35cm}(.0012)}&\multicolumn{1}{c}{ \scriptsize\hspace{0.35cm}(.0004)}&\multicolumn{1}{c}{ \scriptsize\hspace{0.35cm}(.0013)}\\
&LGBM-$Y|X$-ratio& .0001&-.0030& .0016$^{***}$& .0050& .0123$^{***}$& .0008& .0085\\
&&\multicolumn{1}{c}{ \scriptsize\hspace{0.35cm}(.0007)}&\multicolumn{1}{c}{ \scriptsize\hspace{0.35cm}(.0019)}&\multicolumn{1}{c}{ \scriptsize\hspace{0.35cm}(.0003)}&\multicolumn{1}{c}{ \scriptsize\hspace{0.35cm}(.0075)}&\multicolumn{1}{c}{ \scriptsize\hspace{0.35cm}(.0030)}&\multicolumn{1}{c}{ \scriptsize\hspace{0.35cm}(.0007)}&\multicolumn{1}{c}{ \scriptsize\hspace{0.35cm}(.0085)}\\
&NN2-$Y|X$-ratio&-.0010&-.0030& .0006$^{**}$&-.0118& .0100$^{***}$& .0008&-.0106\\
&&\multicolumn{1}{c}{ \scriptsize\hspace{0.35cm}(.0007)}&\multicolumn{1}{c}{ \scriptsize\hspace{0.35cm}(.0019)}&\multicolumn{1}{c}{ \scriptsize\hspace{0.35cm}(.0003)}&\multicolumn{1}{c}{ \scriptsize\hspace{0.35cm}(.0075)}&\multicolumn{1}{c}{ \scriptsize\hspace{0.35cm}(.0030)}&\multicolumn{1}{c}{ \scriptsize\hspace{0.35cm}(.0007)}&\multicolumn{1}{c}{ \scriptsize\hspace{0.35cm}(.0085)}\\
&NN3-$Y|X$-ratio&-.0012$^{*}$&-.0026& .0004&-.0234$^{***}$& .0063$^{**}$& .0009&-.0121\\
&&\multicolumn{1}{c}{ \scriptsize\hspace{0.35cm}(.0007)}&\multicolumn{1}{c}{ \scriptsize\hspace{0.35cm}(.0019)}&\multicolumn{1}{c}{ \scriptsize\hspace{0.35cm}(.0003)}&\multicolumn{1}{c}{ \scriptsize\hspace{0.35cm}(.0075)}&\multicolumn{1}{c}{ \scriptsize\hspace{0.35cm}(.0030)}&\multicolumn{1}{c}{ \scriptsize\hspace{0.35cm}(.0007)}&\multicolumn{1}{c}{ \scriptsize\hspace{0.35cm}(.0085)}\\
&NN4-$Y|X$-ratio&-.0013$^{*}$&-.0033$^{*}$& .0009$^{***}$&-.0406$^{***}$& .0097$^{***}$& .0009&-.0120\\
&&\multicolumn{1}{c}{ \scriptsize\hspace{0.35cm}(.0007)}&\multicolumn{1}{c}{ \scriptsize\hspace{0.35cm}(.0019)}&\multicolumn{1}{c}{ \scriptsize\hspace{0.35cm}(.0003)}&\multicolumn{1}{c}{ \scriptsize\hspace{0.35cm}(.0075)}&\multicolumn{1}{c}{ \scriptsize\hspace{0.35cm}(.0030)}&\multicolumn{1}{c}{ \scriptsize\hspace{0.35cm}(.0007)}&\multicolumn{1}{c}{ \scriptsize\hspace{0.35cm}(.0085)}\\
&Kernel-$Y|X$-ratio&-.0008& .0016& .0002&-.0324$^{***}$&-.0037& .0004&-.0091\\
&&\multicolumn{1}{c}{ \scriptsize\hspace{0.35cm}(.0006)}&\multicolumn{1}{c}{ \scriptsize\hspace{0.35cm}(.0016)}&\multicolumn{1}{c}{ \scriptsize\hspace{0.35cm}(.0002)}&\multicolumn{1}{c}{ \scriptsize\hspace{0.35cm}(.0064)}&\multicolumn{1}{c}{ \scriptsize\hspace{0.35cm}(.0025)}&\multicolumn{1}{c}{ \scriptsize\hspace{0.35cm}(.0006)}&\multicolumn{1}{c}{ \scriptsize\hspace{0.35cm}(.0072)}\\
\midrule
\multirow{3}{*}{\parbox[c]{1.5cm}{Validation\\Type}}&Worst-case &-.0056$^{***}$& .0050$^{***}$&-.0002& .0114$^{***}$&-.0229$^{***}$&-.0067$^{**}$&-.0018\\
&&\multicolumn{1}{c}{ \scriptsize\hspace{0.35cm}(.0010)}&\multicolumn{1}{c}{ \scriptsize\hspace{0.35cm}(.0004)}&\multicolumn{1}{c}{ \scriptsize\hspace{0.35cm}(.0003)}&\multicolumn{1}{c}{ \scriptsize\hspace{0.35cm}(.0023)}&\multicolumn{1}{c}{ \scriptsize\hspace{0.35cm}(.0009)}&\multicolumn{1}{c}{ \scriptsize\hspace{0.35cm}(.0033)}&\multicolumn{1}{c}{ \scriptsize\hspace{0.35cm}(.0017)}\\
&Average-case& .0054$^{***}$& .0078$^{***}$& .0015$^{***}$& .0134$^{***}$& .0028$^{***}$& .0202$^{***}$& .0012\\
&&\multicolumn{1}{c}{ \scriptsize\hspace{0.35cm}(.0010)}&\multicolumn{1}{c}{ \scriptsize\hspace{0.35cm}(.0004)}&\multicolumn{1}{c}{ \scriptsize\hspace{0.35cm}(.0003)}&\multicolumn{1}{c}{ \scriptsize\hspace{0.35cm}(.0023)}&\multicolumn{1}{c}{ \scriptsize\hspace{0.35cm}(.0009)}&\multicolumn{1}{c}{ \scriptsize\hspace{0.35cm}(.0033)}&\multicolumn{1}{c}{ \scriptsize\hspace{0.35cm}(.0017)}\\
\midrule
\multirow{2}{*}{\parbox[c]{1.5cm}{Fixed\\Effect}}&Setting&{\ \ Yes}&{\ \ No}&{\ \ No}&{\ \ No}&{\ \ No}&{\ \ No}&{\ \ No}\\
&Domain&{\ \ No}&{\ \ Yes}&{\ \ Yes}&{\ \ Yes}&{\ \ Yes}&{\ \ Yes}&{\ \ Yes}\\
\midrule
\multirow{2}{*}{\parbox[c]{1.5cm}{Overall}}&$N$&{\ \ 12527}&{\ \ 3671}&{\ \ 3671}&{\ \ 287}&{\ \ 3671}&{\ \ 935}&{\ \ 287}\\
&Adjusted $R^2$& .196& .775& .937& .903& .873& .776& .921\\
    \bottomrule
    \end{tabular}}
    \smallskip
    \footnotesize{\emph{Notes.} $^{***}, ^{**}$ and $^{*}$ show statistical significance at the 1\%, 5\%, and 10\% levels using two-tailed tests, respectively. }
\end{table}

\begin{table}[htbp]
    \centering
    \caption{Regression results on algorithmic design components (\texttt{Worst Domain}, incorporating nonlinear effects of the radius and additional interaction effects between the model class and shift patterns) on the method performance}
    \label{tab:linear-analysis-aug2}
    \sisetup{
    table-format=-1.4, 
    add-integer-zero=false 
    }
    
	\resizebox{\textwidth}{!}{\begin{tabular}{llSSSSSSS}
        \toprule
        & & \multicolumn{6}{c}{Dependent variable: Accuracy}\\
        \cmidrule(lr){3-9}
    \multicolumn{2}{c}{Variable Name} & \multicolumn{1}{c}{\ \ All} & \multicolumn{1}{c}{\ \ Setting 1} & \multicolumn{1}{c}{\ \ Setting 2} & \multicolumn{1}{c}{\ \ Setting 3} & \multicolumn{1}{c}{\ \ Setting 4} &  \multicolumn{1}{c}{\ \ Setting 5} & \multicolumn{1}{c}{\ \ Setting 6} \\
    \midrule
\multirow{3}{*}{\parbox[c]{1.5cm}{Model\\Class}}&LGBM& .0004& .0034& .0059$^{**}$&-.0311$^{***}$&-.0120$^{***}$& .0241$^{***}$& .0151$^{***}$\\
&&\multicolumn{1}{c}{ \scriptsize\hspace{0.35cm}(.0017)}&\multicolumn{1}{c}{ \scriptsize\hspace{0.35cm}(.0034)}&\multicolumn{1}{c}{ \scriptsize\hspace{0.35cm}(.0025)}&\multicolumn{1}{c}{ \scriptsize\hspace{0.35cm}(.0057)}&\multicolumn{1}{c}{ \scriptsize\hspace{0.35cm}(.0021)}&\multicolumn{1}{c}{ \scriptsize\hspace{0.35cm}(.0036)}&\multicolumn{1}{c}{ \scriptsize\hspace{0.35cm}(.0016)}\\
&NN2& .0022&-.0068& .0077$^{*}$&-.0509$^{***}$&-.0117$^{***}$& .0634$^{***}$&-.0130$^{***}$\\
&&\multicolumn{1}{c}{ \scriptsize\hspace{0.35cm}(.0023)}&\multicolumn{1}{c}{ \scriptsize\hspace{0.35cm}(.0042)}&\multicolumn{1}{c}{ \scriptsize\hspace{0.35cm}(.0041)}&\multicolumn{1}{c}{ \scriptsize\hspace{0.35cm}(.0066)}&\multicolumn{1}{c}{ \scriptsize\hspace{0.35cm}(.0035)}&\multicolumn{1}{c}{ \scriptsize\hspace{0.35cm}(.0044)}&\multicolumn{1}{c}{ \scriptsize\hspace{0.35cm}(.0026)}\\
&NN3&-.0253$^{***}$& .0004& .0064&-.1512$^{***}$&-.0299$^{***}$& .0586$^{***}$&-.0003\\
&&\multicolumn{1}{c}{ \scriptsize\hspace{0.35cm}(.0023)}&\multicolumn{1}{c}{ \scriptsize\hspace{0.35cm}(.0043)}&\multicolumn{1}{c}{ \scriptsize\hspace{0.35cm}(.0050)}&\multicolumn{1}{c}{ \scriptsize\hspace{0.35cm}(.0066)}&\multicolumn{1}{c}{ \scriptsize\hspace{0.35cm}(.0038)}&\multicolumn{1}{c}{ \scriptsize\hspace{0.35cm}(.0045)}&\multicolumn{1}{c}{ \scriptsize\hspace{0.35cm}(.0029)}\\
&NN4&-.0242$^{***}$&-.0090$^{**}$&-.0304$^{***}$&-.1480$^{***}$&-.0600$^{***}$& .0591$^{***}$&-.0097$^{***}$\\
&&\multicolumn{1}{c}{ \scriptsize\hspace{0.35cm}(.0024)}&\multicolumn{1}{c}{ \scriptsize\hspace{0.35cm}(.0042)}&\multicolumn{1}{c}{ \scriptsize\hspace{0.35cm}(.0054)}&\multicolumn{1}{c}{ \scriptsize\hspace{0.35cm}(.0071)}&\multicolumn{1}{c}{ \scriptsize\hspace{0.35cm}(.0046)}&\multicolumn{1}{c}{ \scriptsize\hspace{0.35cm}(.0044)}&\multicolumn{1}{c}{ \scriptsize\hspace{0.35cm}(.0037)}\\
&Kernel&-.0244$^{***}$&-.0113$^{***}$&-.0065$^{**}$&-.0856$^{***}$&-.0346$^{***}$& .0018& .0003\\
&&\multicolumn{1}{c}{ \scriptsize\hspace{0.35cm}(.0018)}&\multicolumn{1}{c}{ \scriptsize\hspace{0.35cm}(.0034)}&\multicolumn{1}{c}{ \scriptsize\hspace{0.35cm}(.0026)}&\multicolumn{1}{c}{ \scriptsize\hspace{0.35cm}(.0057)}&\multicolumn{1}{c}{ \scriptsize\hspace{0.35cm}(.0020)}&\multicolumn{1}{c}{ \scriptsize\hspace{0.35cm}(.0035)}&\multicolumn{1}{c}{ \scriptsize\hspace{0.35cm}(.0017)}\\
\midrule
\multirow{3}{*}{\parbox[c]{1.7cm}{Ambiguity\\Set}}&Wasserstein&-.0161$^{***}$& .0029& .0019&-.0342$^{***}$&-.0196$^{***}$&-.0035&-.0025\\
&&\multicolumn{1}{c}{ \scriptsize\hspace{0.35cm}(.0027)}&\multicolumn{1}{c}{ \scriptsize\hspace{0.35cm}(.0049)}&\multicolumn{1}{c}{ \scriptsize\hspace{0.35cm}(.0064)}&\multicolumn{1}{c}{ \scriptsize\hspace{0.35cm}(.0063)}&\multicolumn{1}{c}{ \scriptsize\hspace{0.35cm}(.0038)}&\multicolumn{1}{c}{ \scriptsize\hspace{0.35cm}(.0063)}&\multicolumn{1}{c}{ \scriptsize\hspace{0.35cm}(.0029)}\\
&Chi-squared& .0099$^{***}$& .0174$^{***}$& .0133$^{***}$& .0034& .0001& .0074$^{***}$& .0108$^{***}$\\
&&\multicolumn{1}{c}{ \scriptsize\hspace{0.35cm}(.0013)}&\multicolumn{1}{c}{ \scriptsize\hspace{0.35cm}(.0023)}&\multicolumn{1}{c}{ \scriptsize\hspace{0.35cm}(.0022)}&\multicolumn{1}{c}{ \scriptsize\hspace{0.35cm}(.0037)}&\multicolumn{1}{c}{ \scriptsize\hspace{0.35cm}(.0017)}&\multicolumn{1}{c}{ \scriptsize\hspace{0.35cm}(.0023)}&\multicolumn{1}{c}{ \scriptsize\hspace{0.35cm}(.0014)}\\
&Kullback-Leibler& .0016&-.0056$^{**}$& .0012& .0182$^{***}$& .0026$^{*}$&-.0134$^{***}$& .0014\\
&&\multicolumn{1}{c}{ \scriptsize\hspace{0.35cm}(.0014)}&\multicolumn{1}{c}{ \scriptsize\hspace{0.35cm}(.0027)}&\multicolumn{1}{c}{ \scriptsize\hspace{0.35cm}(.0020)}&\multicolumn{1}{c}{ \scriptsize\hspace{0.35cm}(.0046)}&\multicolumn{1}{c}{ \scriptsize\hspace{0.35cm}(.0016)}&\multicolumn{1}{c}{ \scriptsize\hspace{0.35cm}(.0028)}&\multicolumn{1}{c}{ \scriptsize\hspace{0.35cm}(.0014)}\\
&Total Variation&-.0305$^{***}$&-.0304$^{***}$&-.0087$^{**}$&-.0650$^{***}$&-.0400$^{***}$&-.0365$^{***}$& .0011\\
&&\multicolumn{1}{c}{ \scriptsize\hspace{0.35cm}(.0027)}&\multicolumn{1}{c}{ \scriptsize\hspace{0.35cm}(.0052)}&\multicolumn{1}{c}{ \scriptsize\hspace{0.35cm}(.0036)}&\multicolumn{1}{c}{ \scriptsize\hspace{0.35cm}(.0089)}&\multicolumn{1}{c}{ \scriptsize\hspace{0.35cm}(.0032)}&\multicolumn{1}{c}{ \scriptsize\hspace{0.35cm}(.0054)}&\multicolumn{1}{c}{ \scriptsize\hspace{0.35cm}(.0025)}\\
&OT-Discrepancy&-.0021& .0212$^{***}$&-.0043& .0053&-.0304$^{***}$&-.0083&-.0014\\
&&\multicolumn{1}{c}{ \scriptsize\hspace{0.35cm}(.0025)}&\multicolumn{1}{c}{ \scriptsize\hspace{0.35cm}(.0048)}&\multicolumn{1}{c}{ \scriptsize\hspace{0.35cm}(.0034)}&\multicolumn{1}{c}{ \scriptsize\hspace{0.35cm}(.0084)}&\multicolumn{1}{c}{ \scriptsize\hspace{0.35cm}(.0029)}&\multicolumn{1}{c}{ \scriptsize\hspace{0.35cm}(.0053)}&\multicolumn{1}{c}{ \scriptsize\hspace{0.35cm}(.0024)}\\
&Radius& .0011&-.0018& .0054$^{***}$&-.0153$^{***}$& .0299$^{***}$&-.0108$^{***}$& .0076$^{***}$\\
&&\multicolumn{1}{c}{ \scriptsize\hspace{0.35cm}(.0013)}&\multicolumn{1}{c}{ \scriptsize\hspace{0.35cm}(.0026)}&\multicolumn{1}{c}{ \scriptsize\hspace{0.35cm}(.0018)}&\multicolumn{1}{c}{ \scriptsize\hspace{0.35cm}(.0041)}&\multicolumn{1}{c}{ \scriptsize\hspace{0.35cm}(.0015)}&\multicolumn{1}{c}{ \scriptsize\hspace{0.35cm}(.0027)}&\multicolumn{1}{c}{ \scriptsize\hspace{0.35cm}(.0013)}\\
&Radius2&-.0006$^{***}$&-.0007$^{**}$&-.0006$^{**}$& .0011$^{**}$&-.0031$^{***}$&-.0000&-.0012$^{***}$\\
&&\multicolumn{1}{c}{ \scriptsize\hspace{0.35cm}(.0002)}&\multicolumn{1}{c}{ \scriptsize\hspace{0.35cm}(.0004)}&\multicolumn{1}{c}{ \scriptsize\hspace{0.35cm}(.0003)}&\multicolumn{1}{c}{ \scriptsize\hspace{0.35cm}(.0005)}&\multicolumn{1}{c}{ \scriptsize\hspace{0.35cm}(.0002)}&\multicolumn{1}{c}{ \scriptsize\hspace{0.35cm}(.0004)}&\multicolumn{1}{c}{ \scriptsize\hspace{0.35cm}(.0002)}\\
&LGBM-Radius&-.0167$^{***}$& .0096$^{***}$&-.0127$^{***}$&-.0161$^{***}$&-.0363$^{***}$&-.0345$^{***}$&-.0154$^{***}$\\
&&\multicolumn{1}{c}{ \scriptsize\hspace{0.35cm}(.0016)}&\multicolumn{1}{c}{ \scriptsize\hspace{0.35cm}(.0032)}&\multicolumn{1}{c}{ \scriptsize\hspace{0.35cm}(.0021)}&\multicolumn{1}{c}{ \scriptsize\hspace{0.35cm}(.0053)}&\multicolumn{1}{c}{ \scriptsize\hspace{0.35cm}(.0018)}&\multicolumn{1}{c}{ \scriptsize\hspace{0.35cm}(.0034)}&\multicolumn{1}{c}{ \scriptsize\hspace{0.35cm}(.0015)}\\
&NN2-Radius&-.0410$^{***}$&-.0721$^{***}$&-.0391$^{***}$&-.0474$^{***}$&-.0193$^{***}$&-.0229$^{***}$&-.0290$^{***}$\\
&&\multicolumn{1}{c}{ \scriptsize\hspace{0.35cm}(.0017)}&\multicolumn{1}{c}{ \scriptsize\hspace{0.35cm}(.0074)}&\multicolumn{1}{c}{ \scriptsize\hspace{0.35cm}(.0022)}&\multicolumn{1}{c}{ \scriptsize\hspace{0.35cm}(.0040)}&\multicolumn{1}{c}{ \scriptsize\hspace{0.35cm}(.0021)}&\multicolumn{1}{c}{ \scriptsize\hspace{0.35cm}(.0064)}&\multicolumn{1}{c}{ \scriptsize\hspace{0.35cm}(.0027)}\\
&NN3-Radius&-.0548$^{***}$&-.0883$^{***}$&-.0428$^{***}$&-.0389$^{***}$&-.0112$^{***}$&-.0179$^{**}$&-.0357$^{***}$\\
&&\multicolumn{1}{c}{ \scriptsize\hspace{0.35cm}(.0019)}&\multicolumn{1}{c}{ \scriptsize\hspace{0.35cm}(.0076)}&\multicolumn{1}{c}{ \scriptsize\hspace{0.35cm}(.0029)}&\multicolumn{1}{c}{ \scriptsize\hspace{0.35cm}(.0040)}&\multicolumn{1}{c}{ \scriptsize\hspace{0.35cm}(.0029)}&\multicolumn{1}{c}{ \scriptsize\hspace{0.35cm}(.0081)}&\multicolumn{1}{c}{ \scriptsize\hspace{0.35cm}(.0031)}\\
&NN4-Radius&-.0995$^{***}$&-.0522$^{***}$&-.0541$^{***}$&-.0893$^{***}$& .0132& .0004&-.0696$^{***}$\\
&&\multicolumn{1}{c}{ \scriptsize\hspace{0.35cm}(.0055)}&\multicolumn{1}{c}{ \scriptsize\hspace{0.35cm}(.0114)}&\multicolumn{1}{c}{ \scriptsize\hspace{0.35cm}(.0154)}&\multicolumn{1}{c}{ \scriptsize\hspace{0.35cm}(.0118)}&\multicolumn{1}{c}{ \scriptsize\hspace{0.35cm}(.0104)}&\multicolumn{1}{c}{ \scriptsize\hspace{0.35cm}(.0100)}&\multicolumn{1}{c}{ \scriptsize\hspace{0.35cm}(.0175)}\\
&Kernel-Radius& .0019$^{**}$& .0055$^{***}$&-.0078$^{***}$& .0099$^{***}$&-.0054$^{***}$& .0103$^{***}$&-.0014\\
&&\multicolumn{1}{c}{ \scriptsize\hspace{0.35cm}(.0009)}&\multicolumn{1}{c}{ \scriptsize\hspace{0.35cm}(.0018)}&\multicolumn{1}{c}{ \scriptsize\hspace{0.35cm}(.0013)}&\multicolumn{1}{c}{ \scriptsize\hspace{0.35cm}(.0031)}&\multicolumn{1}{c}{ \scriptsize\hspace{0.35cm}(.0011)}&\multicolumn{1}{c}{ \scriptsize\hspace{0.35cm}(.0019)}&\multicolumn{1}{c}{ \scriptsize\hspace{0.35cm}(.0009)}\\
\hline
\multirow{2}{*}{\parbox[c]{1.5cm}{Fixed\\Effect}}&Setting&{\ \ No}&{\ \ No}&{\ \ No}&{\ \ No}&{\ \ No}&{\ \ No}&{\ \ No}\\
&Domain&{\ \ No}&{\ \ No}&{\ \ No}&{\ \ No}&{\ \ No}&{\ \ No}&{\ \ No}\\
\midrule
\multirow{2}{*}{\parbox[c]{1.5cm}{Overall}}&$N$&{\ \ 23776}&{\ \ 4321}&{\ \ 3212}&{\ \ 5056}&{\ \ 3330}&{\ \ 4500}&{\ \ 3352}\\
&Adjusted $R^2$& .483& .113& .267& .410& .340& .291& .205\\
    \bottomrule
    \end{tabular}}
    \smallskip
    \footnotesize{\emph{Notes.} $^{***}, ^{**}$ and $^{*}$ show statistical significance at the 1\%, 5\%, and 10\% levels using two-tailed tests, respectively. }
\end{table}

\wty{\paragraph{Nonlinear effects of the radius and interaction effects of the model class and radius.}  In Tables~\ref{tab:linear-analysis-aug} and~\ref{tab:linear-analysis-aug2}, we introduce the nonlinear effect of the radius in the ambiguity set and interaction effects between model class and radius. This enlarges the coefficient size of the nonlinear and linear components in all settings.  When aggregating all settings, the coefficient size of the radius for linear models (``Radius'') is significantly positive in \Cref{tab:linear-analysis-aug} compared with the negative coefficient size of the radius for nonlinear models (``Radius'' + ``XX-Radius''). This has two implications: (i) Adding robustness does not improve performance for nonlinear models; (ii) For linear models, it may provide slight improvements when the optimal ambiguity radius is not large. Taking into account the scaled size (with Radius $<$0.5 in general DRO models and Radius $<$0.1 in Wasserstein DRO observed from the data), the ultimate aggregated performance improvement is approximately 0.005, which is still smaller than the improvement attributable to the model class. Therefore, the relative improvement incorporating DRO for linear models is not large compared to changing the model class.}

For the nonlinear effects of the radius in the ambiguity set, the coefficients of the quadratic and linear components usually have opposite signs. When plugging the usual choices of the model configuration (with Radius < 0.5) and the optimal radius from the coefficient, the effects on the model accuracy are still relatively small across all settings. For specific settings, the approximated largest effect of the radius on the accuracy is $(-0.014\text{Radius}^2 + 0.025\text{Radius})_{\max} = 0.004$ in Setting 6 in \Cref{tab:linear-analysis-aug}, which is still dominated by the effect of the model class in that setting.

\begin{table}[htbp]
    \centering
    \caption{Regression results on algorithmic design components (\texttt{Best Config}) on the method performance for each model class across all six settings}
    \label{tab:linear-analysis-subgroup}
    \sisetup{
    table-format=-1.4, 
    add-integer-zero=false 
    }
\begin{tabular}{llSSSS}
        \toprule
        & & \multicolumn{4}{c}{Dependent variable: Accuracy}\\
        \cmidrule(lr){3-6}
    \multicolumn{2}{c}{Variable Name} & \multicolumn{1}{c}{Linear} & \multicolumn{1}{c}{LGBM} & \multicolumn{1}{c}{Kernel} & \multicolumn{1}{c}{NN2}\\
    \midrule
\multirow{3}{*}{\parbox[c]{1.7cm}{Ambiguity\\Set}}& Wasserstein & -0.0030 & {-} & 0.0086$^{**}$& {-} \\
    & & \multicolumn{1}{c}{ \scriptsize\hspace{0.35cm}(.0028)} & & \multicolumn{1}{c}{ \scriptsize\hspace{0.35cm}(.0037)}\\
    & Chi-squared & -0.0003 & {-} & 0.0080$^{***}$ & 0.0013\\
    & & \multicolumn{1}{c}{ \scriptsize\hspace{0.35cm}(.0025)} & & \multicolumn{1}{c}{ \scriptsize\hspace{0.35cm}(.0028)} & \multicolumn{1}{c}{ \scriptsize\hspace{0.35cm}(.0026)}\\
    & Kullback-Leibler & -0.0013 & -0.0076$^{***}$ & 0.0107$^{**}$ &{-}\\
    & & \multicolumn{1}{c}{ \scriptsize\hspace{0.35cm}(.0026)} & \multicolumn{1}{c}{ \scriptsize\hspace{0.35cm}(.0028)} & \multicolumn{1}{c}{ \scriptsize\hspace{0.35cm}(.0025)}\\
    & Total Variation & -0.0065$^{***}$ & {-} &{-} & {-} \\
    & & \multicolumn{1}{c}{ \scriptsize\hspace{0.35cm}(.0026)} & & & \\
    & OT-Discrepancy & -0.0040 & {-} & {-}& {-} \\
    & & \multicolumn{1}{c}{ \scriptsize\hspace{0.35cm}(.0026)}\\
    & Radius & 0.0015 & -0.0023 & 0.0076$^{***}$ & -0.0229$^{***}$\\
    & & \multicolumn{1}{c}{ \scriptsize\hspace{0.35cm}(.0030)} & \multicolumn{1}{c}{ \scriptsize\hspace{0.35cm}(.0018)} & \multicolumn{1}{c}{ \scriptsize\hspace{0.35cm}(.0022)} & \multicolumn{1}{c}{ \scriptsize\hspace{0.35cm}(.0058)}\\
\midrule
\multirow{2}{*}{\parbox[c]{1.7cm}{Shift\\Pattern}}&$Y|X$-ratio & -0.0049$^{***}$ & -0.0038$^{***}$ & -0.0046$^{***}$ & -0.0041$^{***}$\\
    & & \multicolumn{1}{c}{ \scriptsize\hspace{0.35cm}(.0004)} & \multicolumn{1}{c}{ \scriptsize\hspace{0.35cm}(.0005)} & \multicolumn{1}{c}{ \scriptsize\hspace{0.35cm}(.0004)} & \multicolumn{1}{c}{ \scriptsize\hspace{0.35cm}(.0005)}\\
\midrule
\multirow{3}{*}{\parbox[c]{1.5cm}{Validation\\Type}}&Worst-case & -0.0007 & -0.0198 $^{***}$ & -0.0008 &-0.0016\\
    & & \multicolumn{1}{c}{ \scriptsize\hspace{0.35cm}(.0020)} & \multicolumn{1}{c}{ \scriptsize\hspace{0.35cm}(.0028)} & \multicolumn{1}{c}{ \scriptsize\hspace{0.35cm}(.0024)} & \multicolumn{1}{c}{ \scriptsize\hspace{0.35cm}(.0027)}\\
    &Average-case & 0.0008 & 0.0036 & 0.0058$^{**}$ & 0.0097$^{***}$\\
    & & \multicolumn{1}{c}{ \scriptsize\hspace{0.35cm}(.0019)} & \multicolumn{1}{c}{ \scriptsize\hspace{0.35cm}(.0030)} & \multicolumn{1}{c}{ \scriptsize\hspace{0.35cm}(.0024)} & \multicolumn{1}{c}{ \scriptsize\hspace{0.35cm}(.0028)}\\
    \midrule
    \multirow{2}{*}{\parbox[c]{1.5cm}{Fixed\\Effect}}&Setting & {\ \ Yes}&{\ \ Yes}&{\ \ Yes}&{\ \ Yes}\\
    &Domain & {\ \ No}&{\ \ No}&{\ \ No}&{\ \ No}\\
    \midrule
    \multirow{2}{*}{\parbox[c]{1.5cm}{Overall}}&$N$& {\ \ 3654} & {\ \ 1566} & {\ \ 2610} & {\ \  1566}\\
    &Adjusted $R^2$& 0.174& 0.271 & 0.178 & 0.288 \\
    \bottomrule
    \end{tabular}
\end{table}

\wty{\paragraph{Subgroup regression per model class.} In Table~\ref{tab:linear-analysis-subgroup}, we run one regression for methods under each model class and report the coefficients of the ambiguity set and other controlled variables. We observe that for most competitive model classes, e.g., linear, LightGBM, NN2, introducing the distributionally robust counterpart does not yield substantial performance gains. The only notable exception is the kernel-DRO methods, which demonstrate significant improvements over the ERM counterparts (i.e., Kernel-SVM) when averaged across target domains. However, this improvement is partly attributable to the increased variability introduced by the random projections used in kernel methods, and the relatively poor performance of Kernel-SVM compared to linear and tree-based ensemble models. In fact, the gains from switching model classes are generally much larger than those obtained by adding robustness within a given class.}

\paragraph{Interaction effects of model class and shift patterns.} For interaction effects of the model class and shift patterns, we do not observe significantly consistent coefficients among different settings and the coefficient sizes for them are too small (smaller than $\approx$ 0.01). This means that when the $Y|X$-shift is larger, no particular model class (e.g., LightGBM or NN or linear model class) suffers a larger performance drop. We observe similar coefficient sizes when we consider the interaction effects of the ambiguity set and shift patterns. This aligns with previous empirical findings in \Cref{subsec:benchmark-empirical-result}.

\paragraph{Setting the dependent variable to the performance gap.} In Tables~\ref{tab:linear-analysis-aug-gap} and~\ref{tab:linear-analysis-aug-gap-worst}, we investigate how DRO models can mitigate the performance gap under natural distribution shifts by isolating the effects of the model class on the target model performance. The coefficient of the tree-based ensemble is negative, which is expected, as vanilla tree-based ensembles do not provide reliable improvements under distribution shifts. However, the coefficient sizes across various ambiguity set designs, including different distance types and radii, are not consistently larger than those of the neural network models in both the overall and most specific settings. This suggests that DRO models fail to sufficiently mitigate performance gaps.

\begin{table}[htbp]
    \centering
    \caption{Regression results on algorithmic design components (\texttt{Best Config}, incorporating nonlinear effects of the radius and additional interaction effects between the model class and shift patterns) on the performance gap, i.e, \textbf{target accuracy - source accuracy}.}
    \label{tab:linear-analysis-aug-gap}
    \sisetup{
    table-format=-1.4, 
    add-integer-zero=false 
    }
    
	\resizebox{\textwidth}{!}{\begin{tabular}{llSSSSSSS}
        \toprule
        & & \multicolumn{7}{c}{Dependent variable: Performance gap}\\
        \cmidrule(lr){3-9}
    \multicolumn{2}{c}{Variable Name} & \multicolumn{1}{c}{\ \ All} & \multicolumn{1}{c}{\ \ Setting 1} & \multicolumn{1}{c}{\ \ Setting 2} & \multicolumn{1}{c}{\ \ Setting 3} & \multicolumn{1}{c}{\ \ Setting 4} &  \multicolumn{1}{c}{\ \ Setting 5} & \multicolumn{1}{c}{\ \ Setting 6} \\
    \midrule
\multirow{3}{*}{\parbox[c]{1.5cm}{Model\\Class}}&LGBM&-.0165$^{***}$& .0057$^{***}$&-.0212$^{***}$&-.0425$^{***}$&-.0452$^{***}$&-.0201$^{***}$&-.0027\\
&&\multicolumn{1}{c}{ \scriptsize\hspace{0.35cm}(.0021)}&\multicolumn{1}{c}{ \scriptsize\hspace{0.35cm}(.0015)}&\multicolumn{1}{c}{ \scriptsize\hspace{0.35cm}(.0009)}&\multicolumn{1}{c}{ \scriptsize\hspace{0.35cm}(.0082)}&\multicolumn{1}{c}{ \scriptsize\hspace{0.35cm}(.0025)}&\multicolumn{1}{c}{ \scriptsize\hspace{0.35cm}(.0072)}&\multicolumn{1}{c}{ \scriptsize\hspace{0.35cm}(.0037)}\\
&NN2& .0064$^{***}$& .0050$^{***}$& .0045$^{***}$&-.0109&-.0021& .0213$^{***}$& .0099$^{***}$\\
&&\multicolumn{1}{c}{ \scriptsize\hspace{0.35cm}(.0020)}&\multicolumn{1}{c}{ \scriptsize\hspace{0.35cm}(.0015)}&\multicolumn{1}{c}{ \scriptsize\hspace{0.35cm}(.0008)}&\multicolumn{1}{c}{ \scriptsize\hspace{0.35cm}(.0080)}&\multicolumn{1}{c}{ \scriptsize\hspace{0.35cm}(.0025)}&\multicolumn{1}{c}{ \scriptsize\hspace{0.35cm}(.0067)}&\multicolumn{1}{c}{ \scriptsize\hspace{0.35cm}(.0037)}\\
&NN3& .0025& .0016& .0032$^{***}$&-.0090&-.0087$^{***}$& .0174$^{***}$& .0098$^{***}$\\
&&\multicolumn{1}{c}{ \scriptsize\hspace{0.35cm}(.0021)}&\multicolumn{1}{c}{ \scriptsize\hspace{0.35cm}(.0015)}&\multicolumn{1}{c}{ \scriptsize\hspace{0.35cm}(.0009)}&\multicolumn{1}{c}{ \scriptsize\hspace{0.35cm}(.0077)}&\multicolumn{1}{c}{ \scriptsize\hspace{0.35cm}(.0025)}&\multicolumn{1}{c}{ \scriptsize\hspace{0.35cm}(.0067)}&\multicolumn{1}{c}{ \scriptsize\hspace{0.35cm}(.0037)}\\
&NN4& .0034& .0056$^{***}$& .0018$^{*}$&-.0006&-.0040& .0189$^{**}$& .0105$^{***}$\\
&&\multicolumn{1}{c}{ \scriptsize\hspace{0.35cm}(.0024)}&\multicolumn{1}{c}{ \scriptsize\hspace{0.35cm}(.0016)}&\multicolumn{1}{c}{ \scriptsize\hspace{0.35cm}(.0010)}&\multicolumn{1}{c}{ \scriptsize\hspace{0.35cm}(.0085)}&\multicolumn{1}{c}{ \scriptsize\hspace{0.35cm}(.0027)}&\multicolumn{1}{c}{ \scriptsize\hspace{0.35cm}(.0082)}&\multicolumn{1}{c}{ \scriptsize\hspace{0.35cm}(.0040)}\\
&Kernel&-.0017&-.0003& .0048$^{***}$& .0044&-.0191$^{***}$& .0104$^{*}$&-.0046\\
&&\multicolumn{1}{c}{ \scriptsize\hspace{0.35cm}(.0017)}&\multicolumn{1}{c}{ \scriptsize\hspace{0.35cm}(.0013)}&\multicolumn{1}{c}{ \scriptsize\hspace{0.35cm}(.0007)}&\multicolumn{1}{c}{ \scriptsize\hspace{0.35cm}(.0067)}&\multicolumn{1}{c}{ \scriptsize\hspace{0.35cm}(.0020)}&\multicolumn{1}{c}{ \scriptsize\hspace{0.35cm}(.0055)}&\multicolumn{1}{c}{ \scriptsize\hspace{0.35cm}(.0032)}\\
\midrule
\multirow{3}{*}{\parbox[c]{1.7cm}{Ambiguity\\Set}}&Wasserstein& .0007&-.0030$^{***}$&-.0048$^{***}$&-.0107$^{**}$& .0051$^{***}$&-.0086& .0059$^{***}$\\
&&\multicolumn{1}{c}{ \scriptsize\hspace{0.35cm}(.0018)}&\multicolumn{1}{c}{ \scriptsize\hspace{0.35cm}(.0008)}&\multicolumn{1}{c}{ \scriptsize\hspace{0.35cm}(.0007)}&\multicolumn{1}{c}{ \scriptsize\hspace{0.35cm}(.0044)}&\multicolumn{1}{c}{ \scriptsize\hspace{0.35cm}(.0016)}&\multicolumn{1}{c}{ \scriptsize\hspace{0.35cm}(.0060)}&\multicolumn{1}{c}{ \scriptsize\hspace{0.35cm}(.0021)}\\
&Chi-squared& .0001& .0039$^{***}$&-.0075$^{***}$& .0003&-.0022&-.0011& .0034$^{*}$\\
&&\multicolumn{1}{c}{ \scriptsize\hspace{0.35cm}(.0013)}&\multicolumn{1}{c}{ \scriptsize\hspace{0.35cm}(.0008)}&\multicolumn{1}{c}{ \scriptsize\hspace{0.35cm}(.0006)}&\multicolumn{1}{c}{ \scriptsize\hspace{0.35cm}(.0052)}&\multicolumn{1}{c}{ \scriptsize\hspace{0.35cm}(.0015)}&\multicolumn{1}{c}{ \scriptsize\hspace{0.35cm}(.0076)}&\multicolumn{1}{c}{ \scriptsize\hspace{0.35cm}(.0019)}\\
&Kullback-Leibler& .0015& .0058$^{***}$&-.0071$^{***}$& .0007& .0001&-.0069&-.0027\\
&&\multicolumn{1}{c}{ \scriptsize\hspace{0.35cm}(.0014)}&\multicolumn{1}{c}{ \scriptsize\hspace{0.35cm}(.0007)}&\multicolumn{1}{c}{ \scriptsize\hspace{0.35cm}(.0007)}&\multicolumn{1}{c}{ \scriptsize\hspace{0.35cm}(.0049)}&\multicolumn{1}{c}{ \scriptsize\hspace{0.35cm}(.0017)}&\multicolumn{1}{c}{ \scriptsize\hspace{0.35cm}(.0052)}&\multicolumn{1}{c}{ \scriptsize\hspace{0.35cm}(.0019)}\\
&Total Variation&-.0031&-.0026$^{***}$&-.0000&-.0002&-.0058$^{**}$& .0002& .0004\\
&&\multicolumn{1}{c}{ \scriptsize\hspace{0.35cm}(.0023)}&\multicolumn{1}{c}{ \scriptsize\hspace{0.35cm}(.0010)}&\multicolumn{1}{c}{ \scriptsize\hspace{0.35cm}(.0010)}&\multicolumn{1}{c}{ \scriptsize\hspace{0.35cm}(.0062)}&\multicolumn{1}{c}{ \scriptsize\hspace{0.35cm}(.0025)}&\multicolumn{1}{c}{ \scriptsize\hspace{0.35cm}(.0118)}&\multicolumn{1}{c}{ \scriptsize\hspace{0.35cm}(.0029)}\\
&OT-Discrepancy& .0040$^{*}$& .0018$^{*}$&-.0013&-.0007& .0100$^{***}$&-.0034& .0024\\
&&\multicolumn{1}{c}{ \scriptsize\hspace{0.35cm}(.0023)}&\multicolumn{1}{c}{ \scriptsize\hspace{0.35cm}(.0010)}&\multicolumn{1}{c}{ \scriptsize\hspace{0.35cm}(.0009)}&\multicolumn{1}{c}{ \scriptsize\hspace{0.35cm}(.0062)}&\multicolumn{1}{c}{ \scriptsize\hspace{0.35cm}(.0027)}&\multicolumn{1}{c}{ \scriptsize\hspace{0.35cm}(.0081)}&\multicolumn{1}{c}{ \scriptsize\hspace{0.35cm}(.0028)}\\
&Radius& .0004&-.0063$^{***}$& .0123$^{***}$&-.0091&-.0067&-.0283& .0056$^{*}$\\
&&\multicolumn{1}{c}{ \scriptsize\hspace{0.35cm}(.0021)}&\multicolumn{1}{c}{ \scriptsize\hspace{0.35cm}(.0023)}&\multicolumn{1}{c}{ \scriptsize\hspace{0.35cm}(.0014)}&\multicolumn{1}{c}{ \scriptsize\hspace{0.35cm}(.0117)}&\multicolumn{1}{c}{ \scriptsize\hspace{0.35cm}(.0071)}&\multicolumn{1}{c}{ \scriptsize\hspace{0.35cm}(.0475)}&\multicolumn{1}{c}{ \scriptsize\hspace{0.35cm}(.0034)}\\
&Radius2& .0011& .0037$^{**}$&-.0199$^{***}$& .0086& .0042& .0226&-.0029$^{***}$\\
&&\multicolumn{1}{c}{ \scriptsize\hspace{0.35cm}(.0007)}&\multicolumn{1}{c}{ \scriptsize\hspace{0.35cm}(.0019)}&\multicolumn{1}{c}{ \scriptsize\hspace{0.35cm}(.0012)}&\multicolumn{1}{c}{ \scriptsize\hspace{0.35cm}(.0087)}&\multicolumn{1}{c}{ \scriptsize\hspace{0.35cm}(.0054)}&\multicolumn{1}{c}{ \scriptsize\hspace{0.35cm}(.0495)}&\multicolumn{1}{c}{ \scriptsize\hspace{0.35cm}(.0007)}\\
&LGBM-Radius& .0024& .0088$^{***}$& .0382$^{***}$&-.0096& .0014&-.0086&-.0038\\
&&\multicolumn{1}{c}{ \scriptsize\hspace{0.35cm}(.0025)}&\multicolumn{1}{c}{ \scriptsize\hspace{0.35cm}(.0014)}&\multicolumn{1}{c}{ \scriptsize\hspace{0.35cm}(.0022)}&\multicolumn{1}{c}{ \scriptsize\hspace{0.35cm}(.0084)}&\multicolumn{1}{c}{ \scriptsize\hspace{0.35cm}(.0037)}&\multicolumn{1}{c}{ \scriptsize\hspace{0.35cm}(.0129)}&\multicolumn{1}{c}{ \scriptsize\hspace{0.35cm}(.0036)}\\
&NN2-Radius&-.0118$^{**}$&-.6542$^{***}$& .0652$^{***}$&-.0092&-.0181$^{***}$& .0714&-.0205$^{*}$\\
&&\multicolumn{1}{c}{ \scriptsize\hspace{0.35cm}(.0060)}&\multicolumn{1}{c}{ \scriptsize\hspace{0.35cm}(.1183)}&\multicolumn{1}{c}{ \scriptsize\hspace{0.35cm}(.0103)}&\multicolumn{1}{c}{ \scriptsize\hspace{0.35cm}(.0115)}&\multicolumn{1}{c}{ \scriptsize\hspace{0.35cm}(.0048)}&\multicolumn{1}{c}{ \scriptsize\hspace{0.35cm}(.1214)}&\multicolumn{1}{c}{ \scriptsize\hspace{0.35cm}(.0110)}\\
&NN3-Radius&-.0084$^{*}$&-.0834& .0083$^{***}$& .4106&-.0266$^{***}$& .1396& .0000\\
&&\multicolumn{1}{c}{ \scriptsize\hspace{0.35cm}(.0047)}&\multicolumn{1}{c}{ \scriptsize\hspace{0.35cm}(.1183)}&\multicolumn{1}{c}{ \scriptsize\hspace{0.35cm}(.0018)}&\multicolumn{1}{c}{ \scriptsize\hspace{0.35cm}(.7917)}&\multicolumn{1}{c}{ \scriptsize\hspace{0.35cm}(.0049)}&\multicolumn{1}{c}{ \scriptsize\hspace{0.35cm}(.1260)}&\multicolumn{1}{c}{ \scriptsize\hspace{0.35cm}(.0110)}\\
&NN4-Radius&-.0133&-.0058& .0435$^{***}$&-.5412$^{*}$&-.0908$^{***}$& .0176& .0144\\
&&\multicolumn{1}{c}{ \scriptsize\hspace{0.35cm}(.0089)}&\multicolumn{1}{c}{ \scriptsize\hspace{0.35cm}(.0061)}&\multicolumn{1}{c}{ \scriptsize\hspace{0.35cm}(.0035)}&\multicolumn{1}{c}{ \scriptsize\hspace{0.35cm}(.7917)}&\multicolumn{1}{c}{ \scriptsize\hspace{0.35cm}(.0104)}&\multicolumn{1}{c}{ \scriptsize\hspace{0.35cm}(.0356)}&\multicolumn{1}{c}{ \scriptsize\hspace{0.35cm}(.0163)}\\
&Kernel-Radius&-.0061$^{**}$&-.0072$^{***}$&-.0041$^{***}$&-.0194& .0282$^{***}$&-.0007& .0129$^{***}$\\
&&\multicolumn{1}{c}{ \scriptsize\hspace{0.35cm}(.0027)}&\multicolumn{1}{c}{ \scriptsize\hspace{0.35cm}(.0011)}&\multicolumn{1}{c}{ \scriptsize\hspace{0.35cm}(.0014)}&\multicolumn{1}{c}{ \scriptsize\hspace{0.35cm}(.0120)}&\multicolumn{1}{c}{ \scriptsize\hspace{0.35cm}(.0054)}&\multicolumn{1}{c}{ \scriptsize\hspace{0.35cm}(.0254)}&\multicolumn{1}{c}{ \scriptsize\hspace{0.35cm}(.0044)}\\
\midrule
\multirow{2}{*}{\parbox[c]{1.5cm}{Shift\\Pattern}}&$Y|X$-ratio&-.0046$^{***}$& -0.0290$^{***}$&-.0074$^{***}$&-.0726$^{***}$&-.1681$^{***}$&-.0059$^{***}$& .0145$^{***}$\\
&&\multicolumn{1}{c}{ \scriptsize\hspace{0.35cm}(.0004)}&\multicolumn{1}{c}{ \scriptsize\hspace{0.35cm}(0.0013)}&\multicolumn{1}{c}{ \scriptsize\hspace{0.35cm}(.0002)}&\multicolumn{1}{c}{ \scriptsize\hspace{0.35cm}(.0026)}&\multicolumn{1}{c}{ \scriptsize\hspace{0.35cm}(0.0007)}&\multicolumn{1}{c}{ \scriptsize\hspace{0.35cm}(.0004)}&\multicolumn{1}{c}{ \scriptsize\hspace{0.35cm}(.0008)}\\
&LGBM-$Y|X$-ratio& .0016$^{**}$&-.0030$^{*}$& .0016$^{***}$& .0050& .0124$^{***}$& .0008& .0085\\
&&\multicolumn{1}{c}{ \scriptsize\hspace{0.35cm}(.0007)}&\multicolumn{1}{c}{ \scriptsize\hspace{0.35cm}(.0018)}&\multicolumn{1}{c}{ \scriptsize\hspace{0.35cm}(.0003)}&\multicolumn{1}{c}{ \scriptsize\hspace{0.35cm}(.0081)}&\multicolumn{1}{c}{ \scriptsize\hspace{0.35cm}(.0027)}&\multicolumn{1}{c}{ \scriptsize\hspace{0.35cm}(.0007)}&\multicolumn{1}{c}{ \scriptsize\hspace{0.35cm}(.0054)}\\
&NN2-$Y|X$-ratio& .0001&-.0030$^{*}$& .0006$^{**}$&-.0118& .0100$^{***}$& .0008&-.0106$^{*}$\\
&&\multicolumn{1}{c}{ \scriptsize\hspace{0.35cm}(.0007)}&\multicolumn{1}{c}{ \scriptsize\hspace{0.35cm}(.0018)}&\multicolumn{1}{c}{ \scriptsize\hspace{0.35cm}(.0003)}&\multicolumn{1}{c}{ \scriptsize\hspace{0.35cm}(.0081)}&\multicolumn{1}{c}{ \scriptsize\hspace{0.35cm}(.0027)}&\multicolumn{1}{c}{ \scriptsize\hspace{0.35cm}(.0007)}&\multicolumn{1}{c}{ \scriptsize\hspace{0.35cm}(.0054)}\\
&NN3-$Y|X$-ratio& .0001&-.0026& .0004&-.0234$^{***}$& .0063$^{**}$& .0009&-.0121$^{**}$\\
&&\multicolumn{1}{c}{ \scriptsize\hspace{0.35cm}(.0007)}&\multicolumn{1}{c}{ \scriptsize\hspace{0.35cm}(.0018)}&\multicolumn{1}{c}{ \scriptsize\hspace{0.35cm}(.0003)}&\multicolumn{1}{c}{ \scriptsize\hspace{0.35cm}(.0081)}&\multicolumn{1}{c}{ \scriptsize\hspace{0.35cm}(.0027)}&\multicolumn{1}{c}{ \scriptsize\hspace{0.35cm}(.0007)}&\multicolumn{1}{c}{ \scriptsize\hspace{0.35cm}(.0054)}\\
&NN4-$Y|X$-ratio& .0001&-.0033$^{*}$& .0009$^{***}$&-.0406$^{***}$& .0097$^{***}$& .0009&-.0120$^{**}$\\
&&\multicolumn{1}{c}{ \scriptsize\hspace{0.35cm}(.0007)}&\multicolumn{1}{c}{ \scriptsize\hspace{0.35cm}(.0018)}&\multicolumn{1}{c}{ \scriptsize\hspace{0.35cm}(.0003)}&\multicolumn{1}{c}{ \scriptsize\hspace{0.35cm}(.0081)}&\multicolumn{1}{c}{ \scriptsize\hspace{0.35cm}(.0027)}&\multicolumn{1}{c}{ \scriptsize\hspace{0.35cm}(.0007)}&\multicolumn{1}{c}{ \scriptsize\hspace{0.35cm}(.0054)}\\
&Kernel-$Y|X$-ratio&-.0001& .0016& .0002&-.0324$^{***}$&-.0037& .0004&-.0091$^{**}$\\
&&\multicolumn{1}{c}{ \scriptsize\hspace{0.35cm}(.0006)}&\multicolumn{1}{c}{ \scriptsize\hspace{0.35cm}(.0015)}&\multicolumn{1}{c}{ \scriptsize\hspace{0.35cm}(.0002)}&\multicolumn{1}{c}{ \scriptsize\hspace{0.35cm}(.0069)}&\multicolumn{1}{c}{ \scriptsize\hspace{0.35cm}(.0023)}&\multicolumn{1}{c}{ \scriptsize\hspace{0.35cm}(.0006)}&\multicolumn{1}{c}{ \scriptsize\hspace{0.35cm}(.0046)}\\
\midrule
\multirow{3}{*}{\parbox[c]{1.5cm}{Validation\\Type}}&Worst-case &-.0056$^{***}$& .0050$^{***}$&-.0002& .0114$^{***}$&-.0228$^{***}$&-.0067$^{**}$&-.0018$^{*}$\\
&&\multicolumn{1}{c}{ \scriptsize\hspace{0.35cm}(.0010)}&\multicolumn{1}{c}{ \scriptsize\hspace{0.35cm}(.0004)}&\multicolumn{1}{c}{ \scriptsize\hspace{0.35cm}(.0004)}&\multicolumn{1}{c}{ \scriptsize\hspace{0.35cm}(.0024)}&\multicolumn{1}{c}{ \scriptsize\hspace{0.35cm}(.0009)}&\multicolumn{1}{c}{ \scriptsize\hspace{0.35cm}(.0033)}&\multicolumn{1}{c}{ \scriptsize\hspace{0.35cm}(.0011)}\\
&Average-case& .0054$^{***}$& .0078$^{***}$& .0015$^{***}$& .0134$^{***}$& .0029$^{***}$& .0202$^{***}$& .0012\\
&&\multicolumn{1}{c}{ \scriptsize\hspace{0.35cm}(.0010)}&\multicolumn{1}{c}{ \scriptsize\hspace{0.35cm}(.0004)}&\multicolumn{1}{c}{ \scriptsize\hspace{0.35cm}(.0004)}&\multicolumn{1}{c}{ \scriptsize\hspace{0.35cm}(.0024)}&\multicolumn{1}{c}{ \scriptsize\hspace{0.35cm}(.0009)}&\multicolumn{1}{c}{ \scriptsize\hspace{0.35cm}(.0033)}&\multicolumn{1}{c}{ \scriptsize\hspace{0.35cm}(.0011)}\\
\midrule
\multirow{2}{*}{\parbox[c]{1.5cm}{Fixed\\Effect}}&Setting&{\ \ Yes}&{\ \ No}&{\ \ No}&{\ \ No}&{\ \ No}&{\ \ No}&{\ \ No}\\
&Domain&{\ \ No}&{\ \ Yes}&{\ \ Yes}&{\ \ Yes}&{\ \ Yes}&{\ \ Yes}&{\ \ Yes}\\
\midrule
\multirow{2}{*}{\parbox[c]{1.5cm}{Overall}}&$N$&{\ \ 12527}&{\ \ 3671}&{\ \ 3671}&{\ \ 287}&{\ \ 3671}&{\ \ 935}&{\ \ 287}\\
&Adjusted $R^2$& .241& .792& .935& .896& .890& .748& .963\\
    \bottomrule
    \end{tabular}}
    \smallskip
    \footnotesize{\emph{Notes.} $^{***}, ^{**}$ and $^{*}$ show statistical significance at the 1\%, 5\%, and 10\% levels using two-tailed tests, respectively. }
\end{table}

\begin{table}[htbp]
    \centering
    \caption{Regression results on algorithmic design components (\texttt{Worst Domain}, incorporating nonlinear effects of the radius and additional interaction effects between the model class and shift patterns) on the performance gap, i.e., \textbf{target accuracy - source accuracy}.}
    \label{tab:linear-analysis-aug-gap-worst}
    \sisetup{
    table-format=-1.4, 
    add-integer-zero=false 
    }
	\resizebox{\textwidth}{!}{\begin{tabular}{llSSSSSSS}
        \toprule
        & & \multicolumn{7}{c}{Dependent variable: Performance gap}\\
        \cmidrule(lr){3-9}
    \multicolumn{2}{c}{Variable Name} & \multicolumn{1}{c}{\ \ All} & \multicolumn{1}{c}{\ \ Setting 1} & \multicolumn{1}{c}{\ \ Setting 2} & \multicolumn{1}{c}{\ \ Setting 3} & \multicolumn{1}{c}{\ \ Setting 4} &  \multicolumn{1}{c}{\ \ Setting 5} & \multicolumn{1}{c}{\ \ Setting 6} \\
    \midrule
\multirow{3}{*}{\parbox[c]{1.5cm}{Model\\Class}}&LGBM&-.0134$^{***}$& .0123$^{***}$& .0193$^{***}$&-.0413$^{***}$&-.0302$^{***}$&-.0343$^{***}$& .0017\\
&&\multicolumn{1}{c}{ \scriptsize\hspace{0.35cm}(.0012)}&\multicolumn{1}{c}{ \scriptsize\hspace{0.35cm}(.0021)}&\multicolumn{1}{c}{ \scriptsize\hspace{0.35cm}(.0012)}&\multicolumn{1}{c}{ \scriptsize\hspace{0.35cm}(.0030)}&\multicolumn{1}{c}{ \scriptsize\hspace{0.35cm}(.0034)}&\multicolumn{1}{c}{ \scriptsize\hspace{0.35cm}(.0026)}&\multicolumn{1}{c}{ \scriptsize\hspace{0.35cm}(.0012)}\\
&NN2& .0056$^{***}$&-.0130$^{***}$& .0341$^{***}$&-.0063$^{*}$& .0252$^{***}$& .0169$^{***}$& .0020\\
&&\multicolumn{1}{c}{ \scriptsize\hspace{0.35cm}(.0015)}&\multicolumn{1}{c}{ \scriptsize\hspace{0.35cm}(.0027)}&\multicolumn{1}{c}{ \scriptsize\hspace{0.35cm}(.0020)}&\multicolumn{1}{c}{ \scriptsize\hspace{0.35cm}(.0035)}&\multicolumn{1}{c}{ \scriptsize\hspace{0.35cm}(.0058)}&\multicolumn{1}{c}{ \scriptsize\hspace{0.35cm}(.0033)}&\multicolumn{1}{c}{ \scriptsize\hspace{0.35cm}(.0019)}\\
&NN3&-.0153$^{***}$&-.0051$^{*}$& .0292$^{***}$&-.0647$^{***}$&-.0127$^{**}$& .0073$^{**}$& .0011\\
&&\multicolumn{1}{c}{ \scriptsize\hspace{0.35cm}(.0015)}&\multicolumn{1}{c}{ \scriptsize\hspace{0.35cm}(.0027)}&\multicolumn{1}{c}{ \scriptsize\hspace{0.35cm}(.0024)}&\multicolumn{1}{c}{ \scriptsize\hspace{0.35cm}(.0035)}&\multicolumn{1}{c}{ \scriptsize\hspace{0.35cm}(.0063)}&\multicolumn{1}{c}{ \scriptsize\hspace{0.35cm}(.0034)}&\multicolumn{1}{c}{ \scriptsize\hspace{0.35cm}(.0021)}\\
&NN4&-.0120$^{***}$&-.0066$^{**}$& .0107$^{***}$&-.0668$^{***}$&-.0313$^{***}$& .0097$^{***}$&-.0024\\
&&\multicolumn{1}{c}{ \scriptsize\hspace{0.35cm}(.0016)}&\multicolumn{1}{c}{ \scriptsize\hspace{0.35cm}(.0027)}&\multicolumn{1}{c}{ \scriptsize\hspace{0.35cm}(.0027)}&\multicolumn{1}{c}{ \scriptsize\hspace{0.35cm}(.0038)}&\multicolumn{1}{c}{ \scriptsize\hspace{0.35cm}(.0076)}&\multicolumn{1}{c}{ \scriptsize\hspace{0.35cm}(.0032)}&\multicolumn{1}{c}{ \scriptsize\hspace{0.35cm}(.0027)}\\
&Kernel&-.0126$^{***}$&-.0022& .0086$^{***}$&-.0442$^{***}$&-.0392$^{***}$&-.0034& .0040$^{***}$\\
&&\multicolumn{1}{c}{ \scriptsize\hspace{0.35cm}(.0012)}&\multicolumn{1}{c}{ \scriptsize\hspace{0.35cm}(.0021)}&\multicolumn{1}{c}{ \scriptsize\hspace{0.35cm}(.0012)}&\multicolumn{1}{c}{ \scriptsize\hspace{0.35cm}(.0031)}&\multicolumn{1}{c}{ \scriptsize\hspace{0.35cm}(.0033)}&\multicolumn{1}{c}{ \scriptsize\hspace{0.35cm}(.0026)}&\multicolumn{1}{c}{ \scriptsize\hspace{0.35cm}(.0013)}\\
\midrule
\multirow{3}{*}{\parbox[c]{1.7cm}{Ambiguity\\Set}}&Wasserstein&-.0112$^{***}$& .0162$^{***}$&-.0049&-.0171$^{***}$&-.0391$^{***}$& .0016&-.0124$^{***}$\\
&&\multicolumn{1}{c}{ \scriptsize\hspace{0.35cm}(.0018)}&\multicolumn{1}{c}{ \scriptsize\hspace{0.35cm}(.0031)}&\multicolumn{1}{c}{ \scriptsize\hspace{0.35cm}(.0031)}&\multicolumn{1}{c}{ \scriptsize\hspace{0.35cm}(.0033)}&\multicolumn{1}{c}{ \scriptsize\hspace{0.35cm}(.0063)}&\multicolumn{1}{c}{ \scriptsize\hspace{0.35cm}(.0047)}&\multicolumn{1}{c}{ \scriptsize\hspace{0.35cm}(.0021)}\\
&Chi-squared&-.0036$^{***}$& .0054$^{***}$& .0046$^{***}$&-.0143$^{***}$&-.0189$^{***}$& .0025&-.0020$^{*}$\\
&&\multicolumn{1}{c}{ \scriptsize\hspace{0.35cm}(.0009)}&\multicolumn{1}{c}{ \scriptsize\hspace{0.35cm}(.0014)}&\multicolumn{1}{c}{ \scriptsize\hspace{0.35cm}(.0011)}&\multicolumn{1}{c}{ \scriptsize\hspace{0.35cm}(.0020)}&\multicolumn{1}{c}{ \scriptsize\hspace{0.35cm}(.0028)}&\multicolumn{1}{c}{ \scriptsize\hspace{0.35cm}(.0017)}&\multicolumn{1}{c}{ \scriptsize\hspace{0.35cm}(.0010)}\\
&Kullback-Leibler&-.0030$^{***}$&-.0101$^{***}$& .0068$^{***}$& .0003&-.0074$^{***}$&-.0099$^{***}$&-.0000\\
&&\multicolumn{1}{c}{ \scriptsize\hspace{0.35cm}(.0009)}&\multicolumn{1}{c}{ \scriptsize\hspace{0.35cm}(.0017)}&\multicolumn{1}{c}{ \scriptsize\hspace{0.35cm}(.0010)}&\multicolumn{1}{c}{ \scriptsize\hspace{0.35cm}(.0025)}&\multicolumn{1}{c}{ \scriptsize\hspace{0.35cm}(.0026)}&\multicolumn{1}{c}{ \scriptsize\hspace{0.35cm}(.0021)}&\multicolumn{1}{c}{ \scriptsize\hspace{0.35cm}(.0010)}\\
&Total Variation&-.0129$^{***}$& .0006&-.0055$^{***}$&-.0099$^{**}$&-.0369$^{***}$&-.0212$^{***}$&-.0005\\
&&\multicolumn{1}{c}{ \scriptsize\hspace{0.35cm}(.0018)}&\multicolumn{1}{c}{ \scriptsize\hspace{0.35cm}(.0033)}&\multicolumn{1}{c}{ \scriptsize\hspace{0.35cm}(.0017)}&\multicolumn{1}{c}{ \scriptsize\hspace{0.35cm}(.0048)}&\multicolumn{1}{c}{ \scriptsize\hspace{0.35cm}(.0053)}&\multicolumn{1}{c}{ \scriptsize\hspace{0.35cm}(.0041)}&\multicolumn{1}{c}{ \scriptsize\hspace{0.35cm}(.0018)}\\
&OT-Discrepancy&-.0057$^{***}$& .0173$^{***}$&-.0042$^{**}$&-.0043&-.0328$^{***}$&-.0039&-.0017\\
&&\multicolumn{1}{c}{ \scriptsize\hspace{0.35cm}(.0017)}&\multicolumn{1}{c}{ \scriptsize\hspace{0.35cm}(.0030)}&\multicolumn{1}{c}{ \scriptsize\hspace{0.35cm}(.0017)}&\multicolumn{1}{c}{ \scriptsize\hspace{0.35cm}(.0045)}&\multicolumn{1}{c}{ \scriptsize\hspace{0.35cm}(.0048)}&\multicolumn{1}{c}{ \scriptsize\hspace{0.35cm}(.0039)}&\multicolumn{1}{c}{ \scriptsize\hspace{0.35cm}(.0017)}\\
&Radius& .0002&-.0059$^{***}$& .0073$^{***}$&-.0060$^{***}$& .0240$^{***}$&-.0085$^{***}$& .0019$^{**}$\\
&&\multicolumn{1}{c}{ \scriptsize\hspace{0.35cm}(.0009)}&\multicolumn{1}{c}{ \scriptsize\hspace{0.35cm}(.0017)}&\multicolumn{1}{c}{ \scriptsize\hspace{0.35cm}(.0009)}&\multicolumn{1}{c}{ \scriptsize\hspace{0.35cm}(.0022)}&\multicolumn{1}{c}{ \scriptsize\hspace{0.35cm}(.0025)}&\multicolumn{1}{c}{ \scriptsize\hspace{0.35cm}(.0020)}&\multicolumn{1}{c}{ \scriptsize\hspace{0.35cm}(.0009)}\\
&Radius2&-.0000& .0005$^{**}$&-.0009$^{***}$& .0005&-.0021$^{***}$& .0002&-.0003$^{***}$\\
&&\multicolumn{1}{c}{ \scriptsize\hspace{0.35cm}(.0001)}&\multicolumn{1}{c}{ \scriptsize\hspace{0.35cm}(.0002)}&\multicolumn{1}{c}{ \scriptsize\hspace{0.35cm}(.0001)}&\multicolumn{1}{c}{ \scriptsize\hspace{0.35cm}(.0003)}&\multicolumn{1}{c}{ \scriptsize\hspace{0.35cm}(.0003)}&\multicolumn{1}{c}{ \scriptsize\hspace{0.35cm}(.0003)}&\multicolumn{1}{c}{ \scriptsize\hspace{0.35cm}(.0001)}\\
&LGBM-Radius&-.0008& .0288$^{***}$& .0063$^{***}$& .0076$^{***}$&-.0241$^{***}$&-.0268$^{***}$&-.0055$^{***}$\\
&&\multicolumn{1}{c}{ \scriptsize\hspace{0.35cm}(.0011)}&\multicolumn{1}{c}{ \scriptsize\hspace{0.35cm}(.0020)}&\multicolumn{1}{c}{ \scriptsize\hspace{0.35cm}(.0010)}&\multicolumn{1}{c}{ \scriptsize\hspace{0.35cm}(.0028)}&\multicolumn{1}{c}{ \scriptsize\hspace{0.35cm}(.0029)}&\multicolumn{1}{c}{ \scriptsize\hspace{0.35cm}(.0025)}&\multicolumn{1}{c}{ \scriptsize\hspace{0.35cm}(.0011)}\\
&NN2-Radius& .0125$^{***}$&-.0246$^{***}$&-.0074$^{***}$& .0111$^{***}$& .0165$^{***}$& .0060& .0147$^{***}$\\
&&\multicolumn{1}{c}{ \scriptsize\hspace{0.35cm}(.0011)}&\multicolumn{1}{c}{ \scriptsize\hspace{0.35cm}(.0047)}&\multicolumn{1}{c}{ \scriptsize\hspace{0.35cm}(.0011)}&\multicolumn{1}{c}{ \scriptsize\hspace{0.35cm}(.0022)}&\multicolumn{1}{c}{ \scriptsize\hspace{0.35cm}(.0034)}&\multicolumn{1}{c}{ \scriptsize\hspace{0.35cm}(.0047)}&\multicolumn{1}{c}{ \scriptsize\hspace{0.35cm}(.0020)}\\
&NN3-Radius& .0177$^{***}$&-.0313$^{***}$&-.0096$^{***}$& .0308$^{***}$& .0295$^{***}$& .0055& .0116$^{***}$\\
&&\multicolumn{1}{c}{ \scriptsize\hspace{0.35cm}(.0012)}&\multicolumn{1}{c}{ \scriptsize\hspace{0.35cm}(.0048)}&\multicolumn{1}{c}{ \scriptsize\hspace{0.35cm}(.0014)}&\multicolumn{1}{c}{ \scriptsize\hspace{0.35cm}(.0022)}&\multicolumn{1}{c}{ \scriptsize\hspace{0.35cm}(.0048)}&\multicolumn{1}{c}{ \scriptsize\hspace{0.35cm}(.0060)}&\multicolumn{1}{c}{ \scriptsize\hspace{0.35cm}(.0023)}\\
&NN4-Radius&-.0086$^{**}$&-.0131$^{*}$& .0207$^{***}$& .0074& .0656$^{***}$& .0115& .0799$^{***}$\\
&&\multicolumn{1}{c}{ \scriptsize\hspace{0.35cm}(.0037)}&\multicolumn{1}{c}{ \scriptsize\hspace{0.35cm}(.0072)}&\multicolumn{1}{c}{ \scriptsize\hspace{0.35cm}(.0075)}&\multicolumn{1}{c}{ \scriptsize\hspace{0.35cm}(.0063)}&\multicolumn{1}{c}{ \scriptsize\hspace{0.35cm}(.0171)}&\multicolumn{1}{c}{ \scriptsize\hspace{0.35cm}(.0074)}&\multicolumn{1}{c}{ \scriptsize\hspace{0.35cm}(.0128)}\\
&Kernel-Radius& .0018$^{***}$& .0020$^{*}$& .0024$^{***}$&-.0013&-.0002& .0077$^{***}$& .0041$^{***}$\\
&&\multicolumn{1}{c}{ \scriptsize\hspace{0.35cm}(.0006)}&\multicolumn{1}{c}{ \scriptsize\hspace{0.35cm}(.0012)}&\multicolumn{1}{c}{ \scriptsize\hspace{0.35cm}(.0006)}&\multicolumn{1}{c}{ \scriptsize\hspace{0.35cm}(.0017)}&\multicolumn{1}{c}{ \scriptsize\hspace{0.35cm}(.0017)}&\multicolumn{1}{c}{ \scriptsize\hspace{0.35cm}(.0014)}&\multicolumn{1}{c}{ \scriptsize\hspace{0.35cm}(.0006)}\\

\multirow{2}{*}{\parbox[c]{1.5cm}{Fixed\\Effect}}&Setting&{\ \ Yes}&{\ \ No}&{\ \ No}&{\ \ No}&{\ \ No}&{\ \ No}&{\ \ No}\\
&Domain&{\ \ No}&{\ \ No}&{\ \ No}&{\ \ No}&{\ \ No}&{\ \ No}&{\ \ No}\\
\midrule
\multirow{2}{*}{\parbox[c]{1.5cm}{Overall}}&$N$&{\ \ 23776}&{\ \ 4321}&{\ \ 3212}&{\ \ 5056}&{\ \ 3330}&{\ \ 4500}&{\ \ 3352}\\
&Adjusted $R^2$& .755& .208& .338& .239& .304& .328& .111\\
    \bottomrule
    \end{tabular}}
    \smallskip
    \footnotesize{\emph{Notes.} $^{***}, ^{**}$ and $^{*}$ show statistical significance at the 1\%, 5\%, and 10\% levels using two-tailed tests, respectively. }
\end{table}

\wty{\paragraph{Nonlinear regression to identify critical components} Due to the limited fitting power of the linear equation, we generalize the linear equation~\eqref{eq:lr-model} in each setting to a nonlinear model for a better explanation and subsequent attribution results. More specifically, following the same notation as~\eqref{eq:lr-model}, we estimate the following equation via fitting a random forest regressor (default setup as \texttt{scikit-learn} with 10 estimators using 80\% of all data points to avoid overfitting):
\begin{equation}\label{eq:nlr-model}
    \text{Accuracy}_{i,j,s,t} = f(X_{i,j,s}, D_{i,j,s,t}, Z_{i,j}, V_{i,j}; \mu_i, \tau_j) + \tilde\eta_{i,j,s,t}
\end{equation}
We do not incorporate interaction terms between algorithmic design since $f(\cdot)$ is a nonlinear function that implicitly takes them into account in its specification. }

\wty{In particular, to quantify the contribution of each algorithmic design component, i.e., model class, validation procedure, and ambiguity‐set configuration, we compute the average marginal performance gain over the baseline linear SVM where all design flags are set to zero across the default configuration set $\Cscr$ of that baseline linear SVM. In particular, for each variable $v \in \Vscr:=\{X_{i,j,s}, D_{i,j,s,t}, Z_{i,j}, V_{i,j}\}$, for the regression fitted in a given setting, we compute its sensitivity (attribution) result $\Ascr(v)$ averaged over $c \in \Cscr$ as follows:
\begin{equation}\label{eq:nlr-attribution}
\Ascr(v;\Delta) = \frac{1}{\Delta |\Cscr|}\sum_{c \in \Cscr}\Para{f(X_{i,j,s}, D_{i,j,s,t}, Z_{i,j}, V_{i,j}; \mu_i, \tau_j)|_{c + \Delta \cdot e_v} - f(X_{i,j,s}, D_{i,j,s,t}, Z_{i,j}, V_{i,j}; \mu_i, \tau_j)|_{c}},    
\end{equation}
where $e_v$ is the unit vector with 1 in the $v$-th coordinate. This metric corresponds to a form of local sensitivity analysis in~\citep{baehrens2010explain}. We report the attribution results in \Cref{tab:nonlinear-attribution} with $\Delta = 1$ (Specifically, we compute the sensitivity value of the ``Radius'' parameter as $\max_{\Delta \in \{0.2, 0.4, \ldots, 1\}} \Ascr(v;\Delta)$ since it is a continuous variable). Note that the random forest regressor produces an $R^2$ exceeding 0.8 across all settings, implying its adequacy to explain algorithmic behavior. Despite that, our main results in fitting~\eqref{eq:lr-model} remain valid, where the effect of the model class still dominates that of the ambiguity set design.
}
\begin{table}[!htb]
    \centering
    \caption{Attribution results of random forest regression on algorithmic design components on the method performance compared with Linear ERM method configurations following~\eqref{eq:nlr-attribution} in the \texttt{Best Config} design, with boldfaced values denoting the largest value in each category.}
    \label{tab:nonlinear-attribution}
    \resizebox{\textwidth}{!}{\begin{tabular}{cc|ccccccc}
\toprule
   \multicolumn{2}{c}{Variable Name}& All & Setting 1 & Setting 2 & Setting 3 & Setting 4 & Setting 5 & Setting 6 \\
    \midrule
    \multirow{3}{*}{\parbox[c]{1.5cm}{Model\\Class}} & LGBM  & 0.0061 & 0.0082 & \textbf{0.0073} & -0.0153 & -0.0099 & 0.0490 & \textbf{0.0244} \\
    & NN2  & \textbf{0.0085} & 0.0073 & 0.0045 & -0.0065 & -0.0044 & 0.0688 & 0.0078 \\
    & NN3         & 0.0032 & 0.0062 & -0.0002 & \textbf{-0.0039} & -0.0154 & \textbf{0.0735} & 0.0048 \\
    & NN4         & 0.0037 & \textbf{0.0085} & -0.0023 & -0.0133 & -0.0150 & 0.0702 & 0.0018 \\
    & Kernel       & -0.0140 & -0.0012 & -0.0072 & -0.0209 & -0.0527 & 0.0515 & -0.0450 \\
    \hline
    \multirow{3}{*}{\parbox[c]{1.7cm}{Ambiguity\\Set}} & Wasserstein  & -0.0003 & -0.0006 & -0.0006 & 0.0006 & -0.0010 & 0.0006 & -0.0052 \\
    & Chi-square        & 0.0006 & \textbf{0.0024} & 0.0004 & -0.0004 & -0.0005 & \textbf{0.0018} & 0.0013 \\
    & Kullback-Leibler          & 0.0000 & -0.0006 & -0.0006 & -0.0002 & 0.0006 & -0.0008 & -0.0003 \\
    & Total Variation   & -0.0010 & -0.0019 & -0.0009 & -0.0019 & -0.0019 & -0.0013 & -0.0001 \\
    & OT-Discrepancy         & -0.0005 & -0.0001 & -0.0006 & -0.0002 & -0.0003 & -0.0000 & -0.0083 \\
    & Radius      & \textbf{0.0030} & 0.0014 & \textbf{0.0012} & \textbf{0.0060} & \textbf{0.0038} & 0.0003 & \textbf{0.0042} \\
    \hline
    Shift Pattern & $Y|X$-ratio        & \textbf{0.0155} & \textbf{0.0339} & \textbf{0.0294} & \textbf{-0.0199} & \textbf{-0.0155} & \textbf{0.0530} & \textbf{0.0141} \\
    \hline
    \multirow{2}{*}{\parbox[c]{1.2cm}{Validation\\Type}} & Worst-case  & -0.0004 & 0.0007 & -0.0001 & 0.0024 & -0.0020 & -0.0030 & 0.0004 \\
    & Average-case    & \textbf{0.0012} & \textbf{0.0035} & \textbf{0.0006} & \textbf{0.0020} & \textbf{-0.0002} & \textbf{0.0024} & \textbf{0.0003} \\
    \hline
    Overall & Out-of-sample $R^2$ & 0.929 & 0.864 & 0.957 & 0.936 & 0.934 & 0.922 & 0.939 \\
    \bottomrule
    \end{tabular}}
\end{table}

\subsection{Training Details in \Cref{subsec:worst-analysis}}\label{subsec:appendix-training-detail-4.2}
\wty{We detail how we compute the worst-case distribution of DRO methods under the Linear-SVM based on $f$-divergence and Wasserstein distances as follows.
\begin{itemize}
    \item For $f$-divergence-DRO, recall the training samples $\{(x_i, y_i)\}_{i \in [n]}$, we denote the loss for each sample by $\ell_{tr}(i) = \ell_{tr}(\what{f}(x_i), y_i)$ and solve the optimization problem as follows:
\[\max_{p \in \Delta_n} \paran{\sum_{i \in [n]} p_i \ell_{tr}(i),~\text{s.t.}~\frac{1}{n}\sum_{i \in [n]} f(n p_i) \leq \epsilon}\]
where $\Delta_n = \{p \in \R_+^n | \sum_{i \in [n]} p_i = 1\}$. This is a linear optimization problem over a convex feasible set.
    \item For Wasserstein-DRO, we solve the optimization problem from Theorem 20 (ii) in \cite{shafieezadeh2019regularization} with $\gamma = \frac{1}{n}$. This is a linear optimization problem.
\end{itemize}
For each optimization problem, we solve it via the interior-point algorithm by calling the MOSEK solver to obtain the worst-case distributions.}

Then we demonstrate the experimental settings in~\Cref{subsec:worst-analysis}, where we focus on Setting 1 with \texttt{ACS Income} dataset and choose CA as the source domain.
We sample 20,000 data points from \texttt{ACS Income} (domain: CA), and obtain the worst-case empirical distribution of DRO, denoted by $\what{P}^\star$.
For the other 50 target domains, we sample 20,000 points from each domain (denoted by $\{\what{Q}_t\}_{t\in [50]}$).
Note that if the number of samples in one domain is less than 40,000, we randomly sample half of the samples from that domain.

For the optimal in-distribution accuracy for each model under the worst-case distribution $\what{P}^\star$ in \eqref{eq:optimal-f}, we calculate its performance via 4-fold validation.
For the 50 target domains $\{\what{Q}_t\}_{t\in [50]}$, we calculate the optimal in-distribution accuracy for each model in the same way (via $4$-fold validation).
For the transfer accuracy in~\eqref{eq:transfer-acc}, we fit models (LR, RF, LGBM, XGB) on $\what{P}^\star$ and test them on $\{\what{Q}_t\}_{t\in [50]}$, respectively.

\subsection{More Results in \Cref{subsec:worst-analysis}}
\label{subsec:more_results_worst}
\begin{figure}[!htb]
 \centering\captionsetup[subfloat]{labelfont=scriptsize,textfont=scriptsize}  
 \stackunder[2pt]{\includegraphics[width=0.23\textwidth]{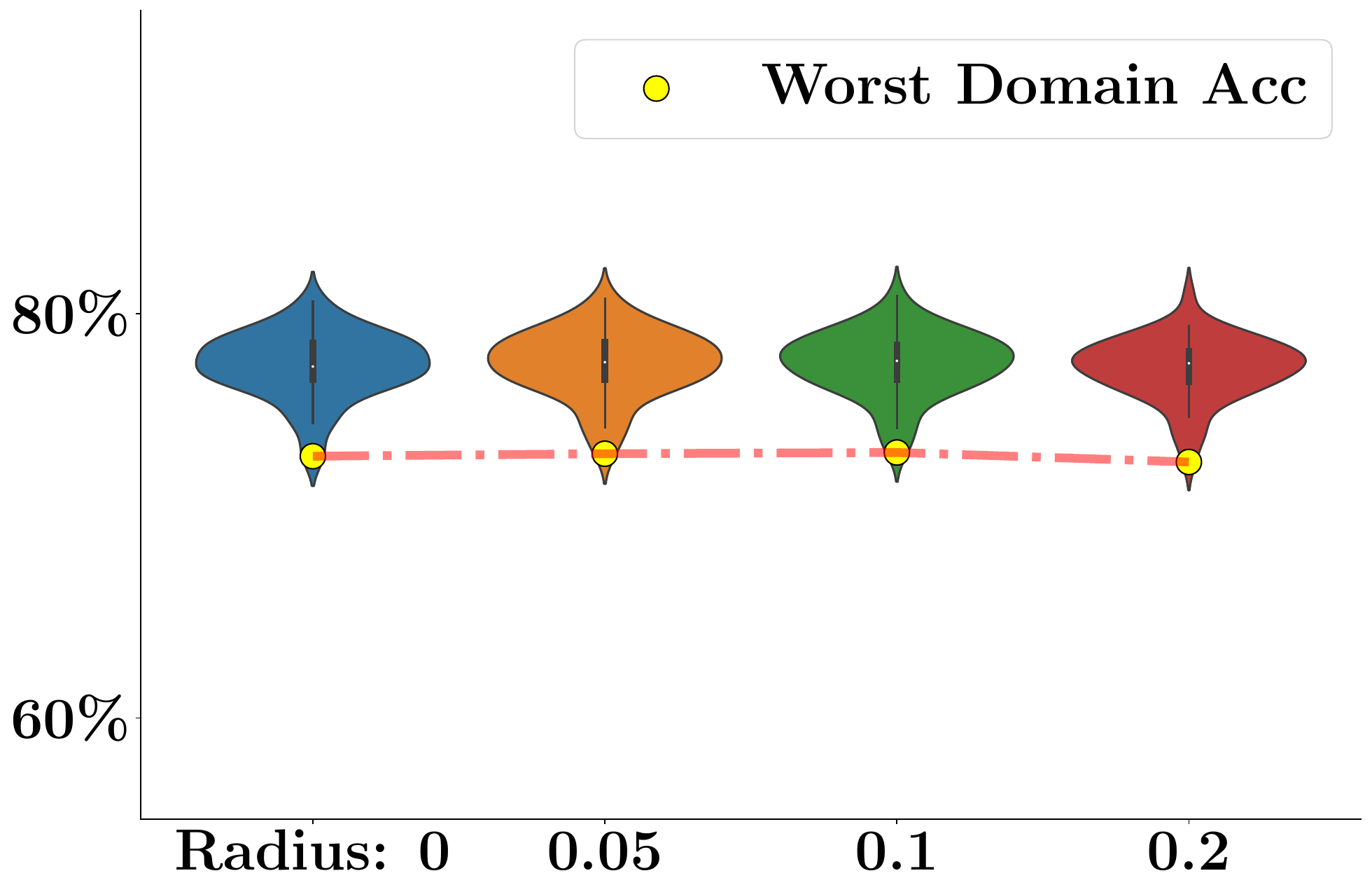}}{\scriptsize  (a) LR}
 \stackunder[2pt]{\includegraphics[width=0.23\textwidth]{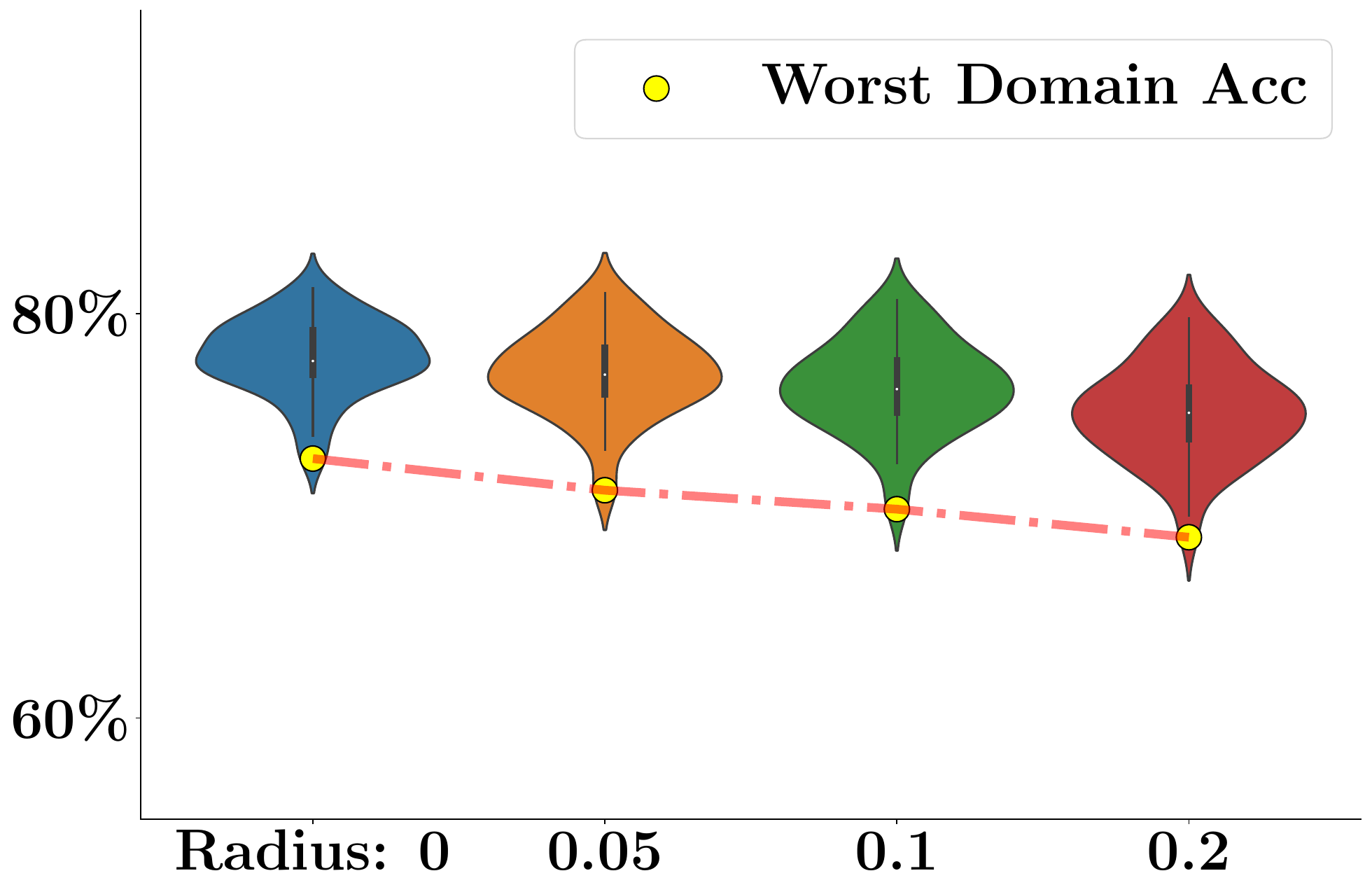}}{\scriptsize (b) RF}
 \stackunder[2pt]{\includegraphics[width=0.23\textwidth]{figures/worst/kl_dro-income-lightgbm.pdf}}{\scriptsize (c) LGBM}
 \stackunder[2pt]{\includegraphics[width=0.23\textwidth]{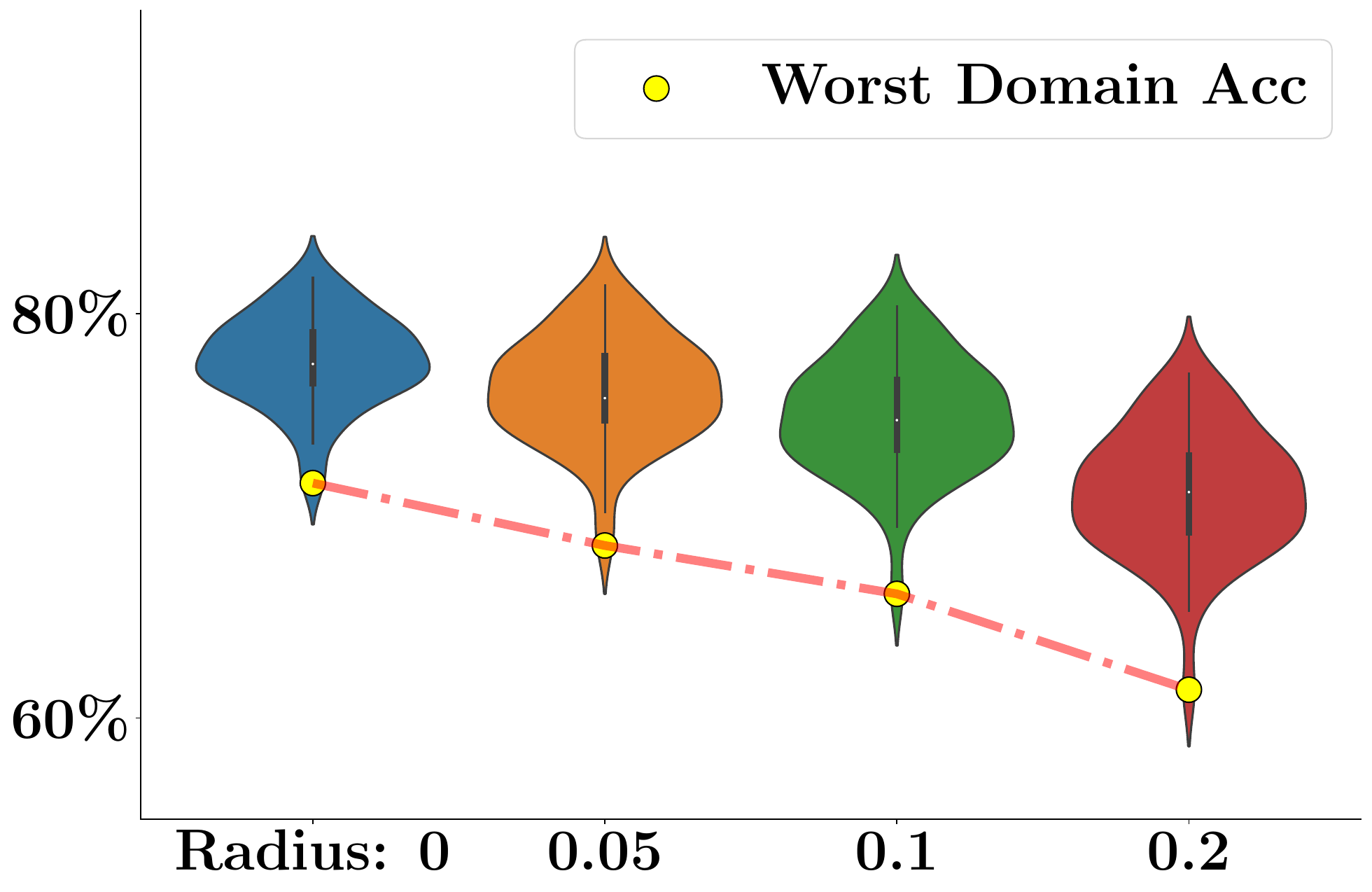}}{\scriptsize (d)  XGB}
\caption{Transfer accuracy~\eqref{eq:transfer-acc} on \texttt{ACS Income} (Setting 1) for different model classes $\Fscr$ with respect to the worst-case distribution of KL-DRO.}
 \label{fig:worst-distribution-more1}  
\end{figure}

\begin{figure}[!htb]
 \centering\captionsetup[subfloat]{labelfont=scriptsize,textfont=scriptsize}  
 \stackunder[2pt]{\includegraphics[width=0.23\textwidth]{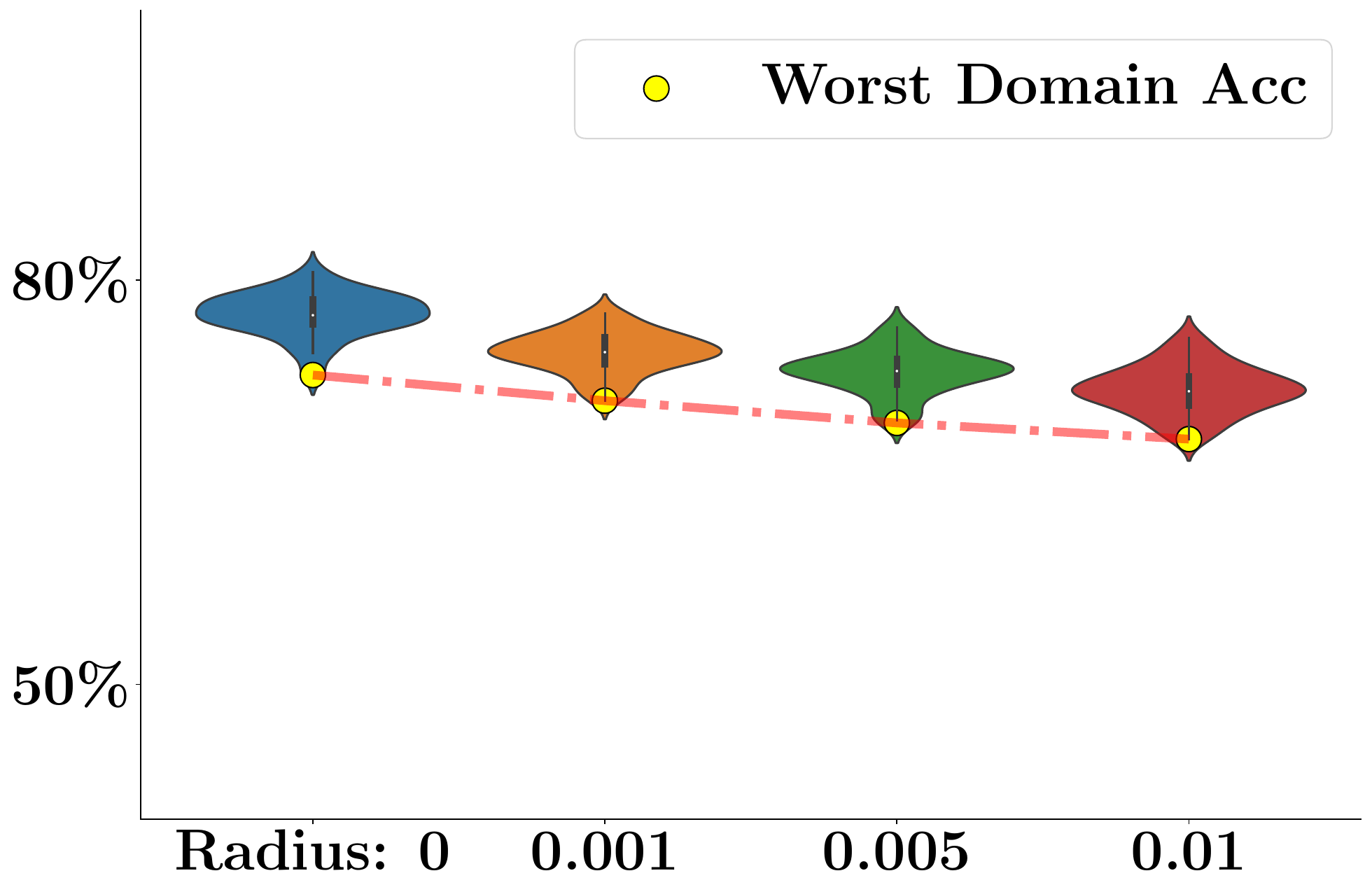}}{\scriptsize  (a) LR}
 \stackunder[2pt]{\includegraphics[width=0.23\textwidth]{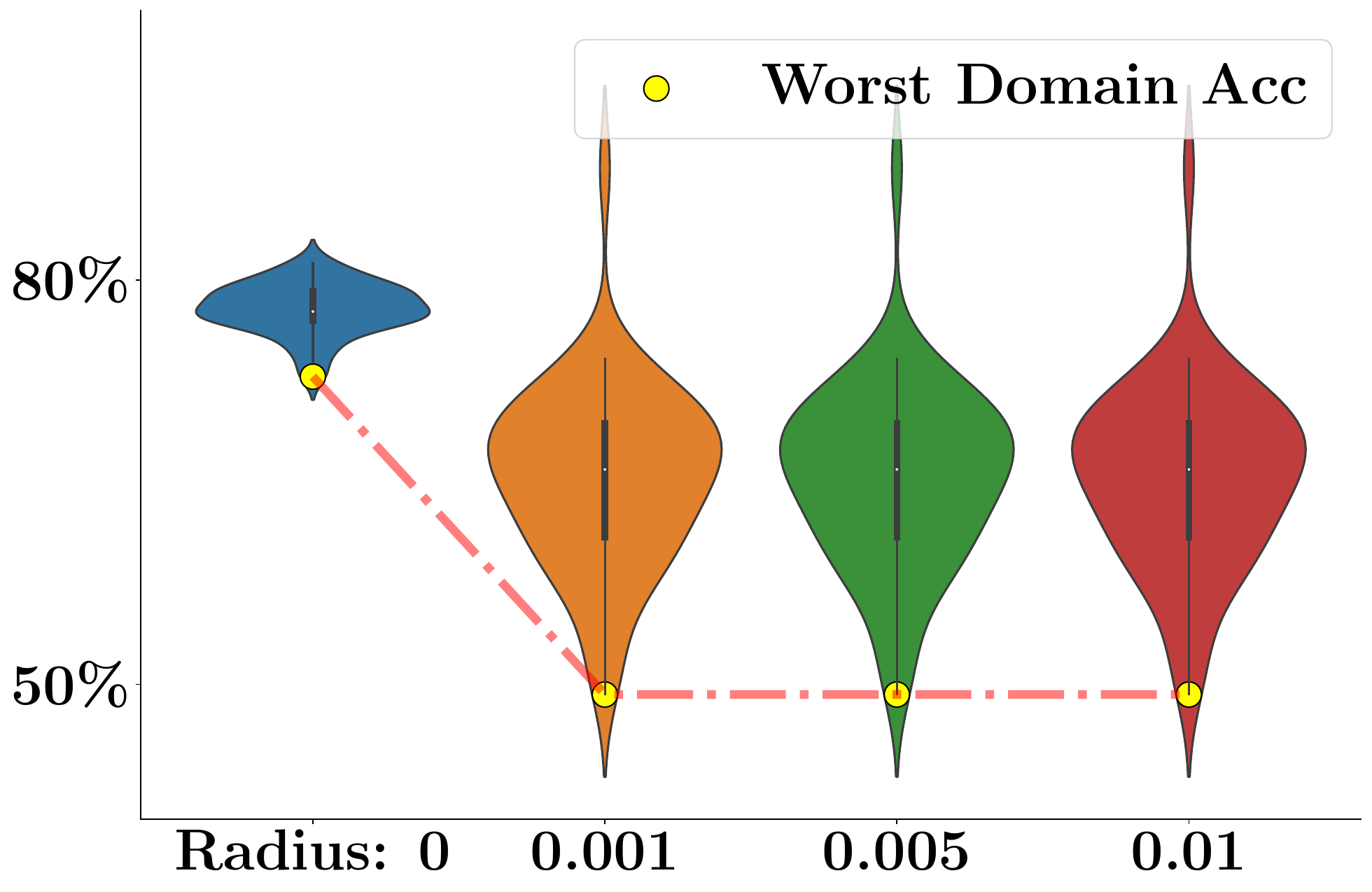}}{\scriptsize (b) RF}
 \stackunder[2pt]{\includegraphics[width=0.23\textwidth]{figures/worst/wasserstein_dro-income-lightgbm.pdf}}{\scriptsize (c) LGBM}
 \stackunder[2pt]{\includegraphics[width=0.23\textwidth]{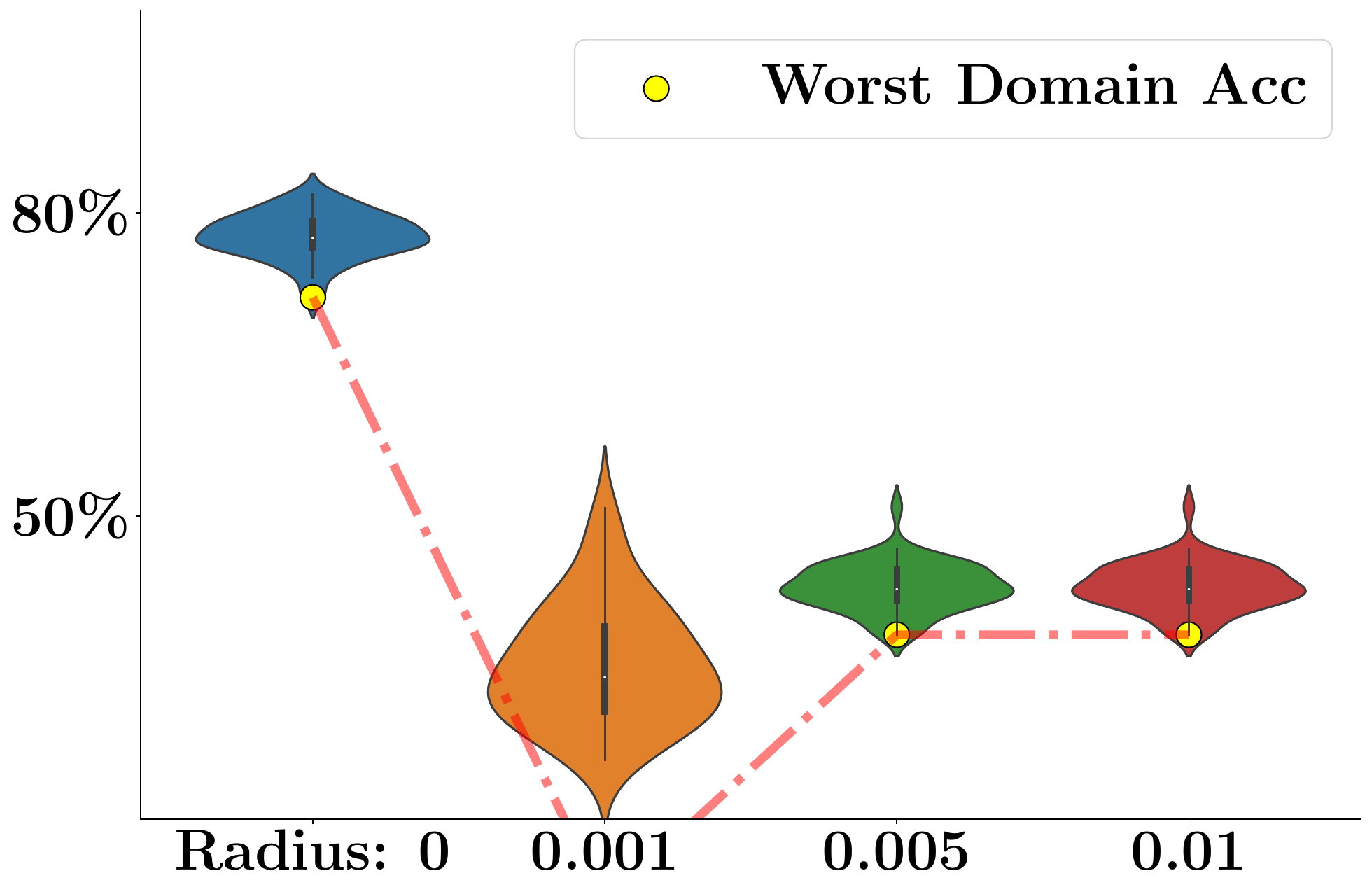}}{\scriptsize (d)  XGB}
\caption{Transfer accuracy~\eqref{eq:transfer-acc} on \texttt{ACS Income} (Setting 1) for different model classes $\Fscr$ with respect to the worst-case distribution of Wasserstein-DRO. }
 \label{fig:worst-distribution-more2}  
\end{figure}

\begin{figure}[!htb]
 \centering\captionsetup[subfloat]{labelfont=scriptsize,textfont=scriptsize}  
  \stackunder[2pt]{\includegraphics[width=0.23\textwidth]{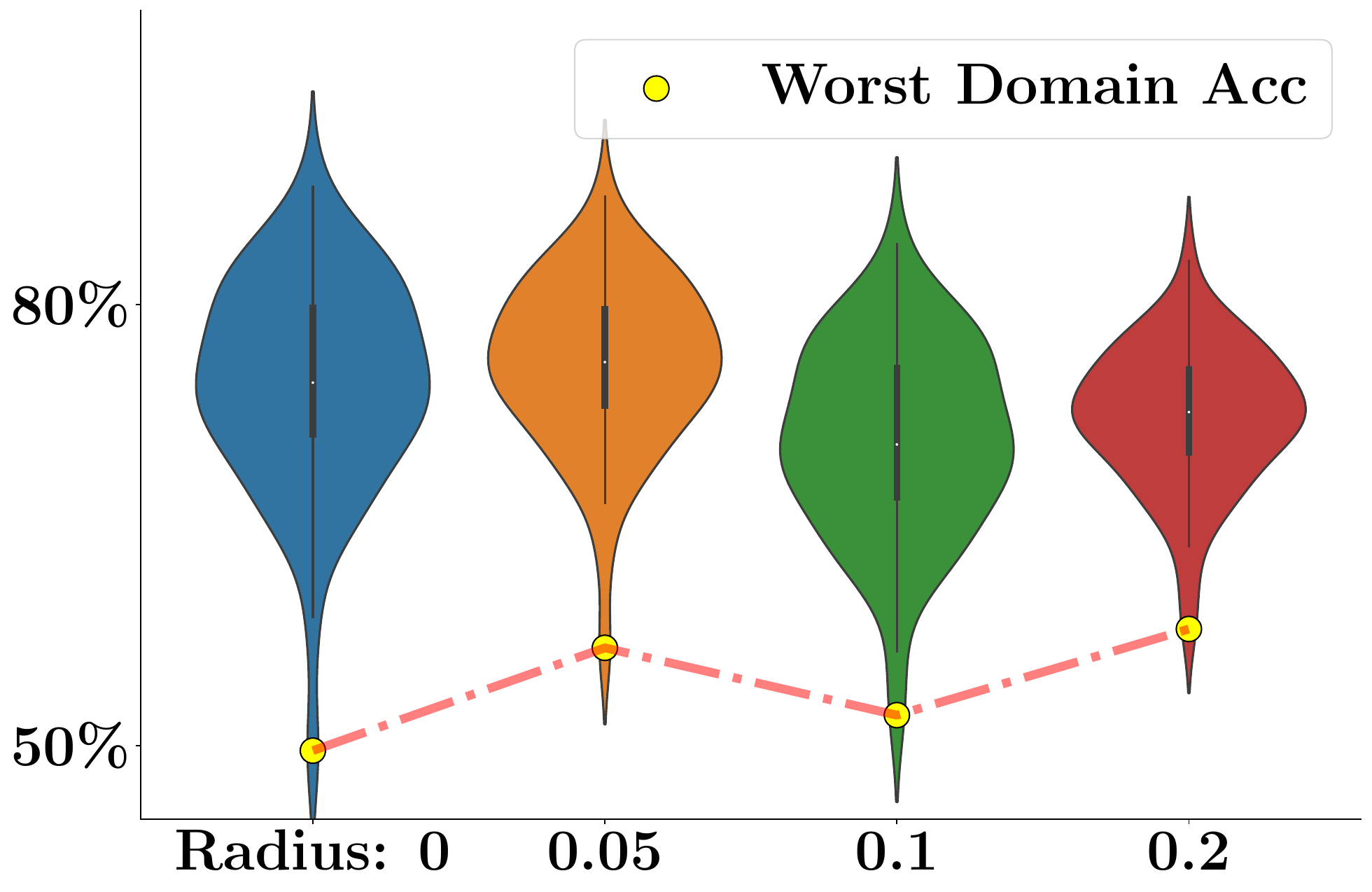}}{\scriptsize (a) LR}
 \stackunder[2pt]{\includegraphics[width=0.23\textwidth]{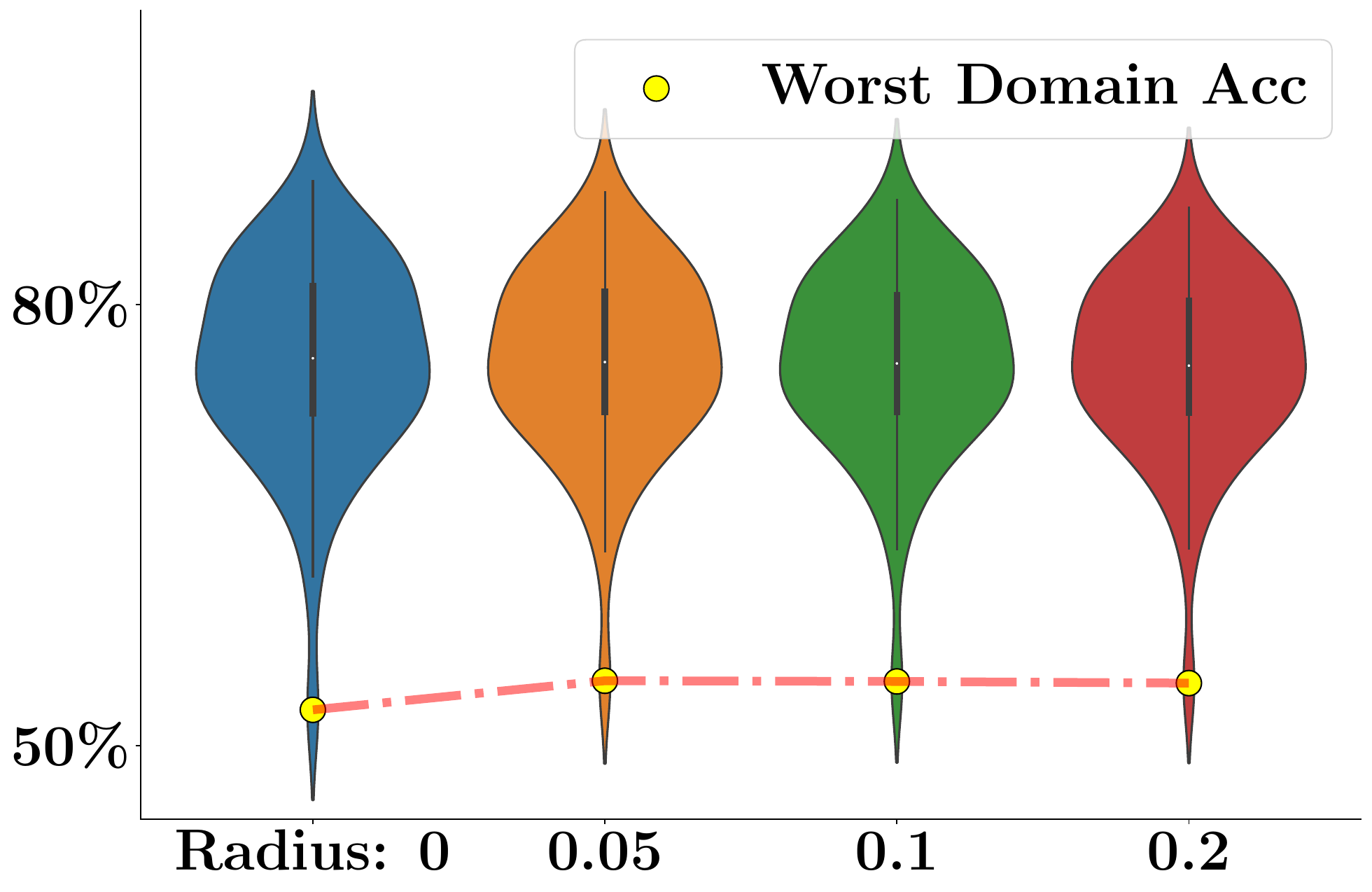}}{\scriptsize (b) RF}
 \stackunder[2pt]{\includegraphics[width=0.23\textwidth]{figures/worst/kl_dro-pubcov-lightgbm.pdf}}{\scriptsize (c) LGBM}
 \stackunder[2pt]{\includegraphics[width=0.23\textwidth]{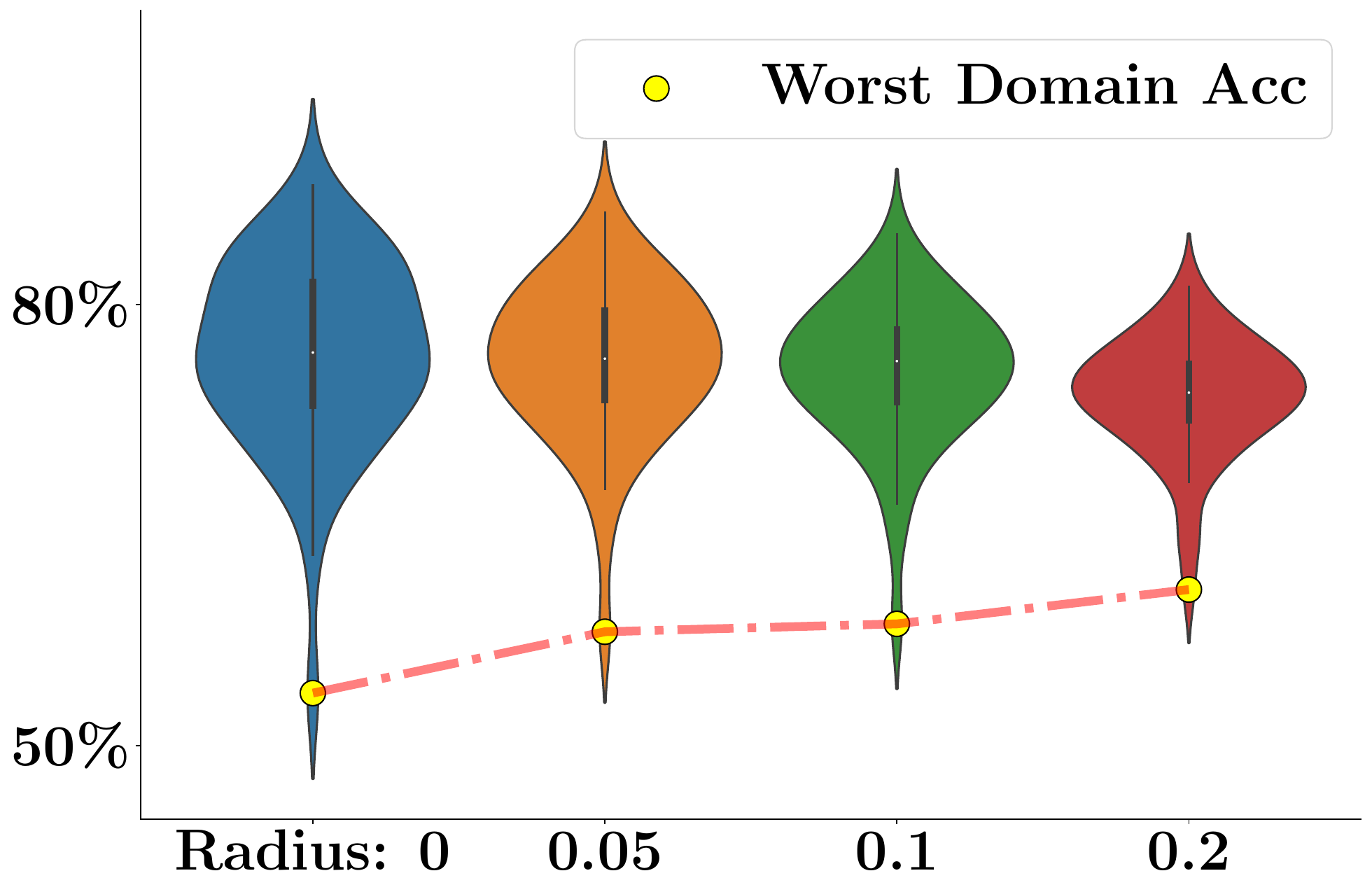}}{\scriptsize (d)  XGB}
\caption{Transfer accuracy~\eqref{eq:transfer-acc} on \texttt{ACS Pub.Cov} (Setting 4) for different model classes $\Fscr$ with respect to the worst-case distribution of KL-DRO. }
 \label{fig:worst-distribution-more3}  
\end{figure}

For the worst-case distribution analysis in~\Cref{subsec:worst-analysis}, we provide results with more basic model classes in~Figures~\ref{fig:worst-distribution-more1},~\ref{fig:worst-distribution-more2} and~\ref{fig:worst-distribution-more3}. Here in each figure, similar to \Cref{fig:worst-distribution}, the first bar represents the transfer accuracies from the training distribution $\what{P}$ to each of the 50 target domains ($\{\text{TAcc}(\what{P},\what{Q}_t)\}_{t\in[50]}$), and the rest three bars represent the transfer accuracy from the worst-case distribution $\what{P}^\star$ to 50 target domains ($\{\text{TAcc}(\what{P}^\star,\what{Q}_t)\}_{t\in[50]}$). The model classes $\Fscr$ used here include LR, RF, LGBM, and XGB respectively.

\wty{\section{Details in~\Cref{sec:alg_intervention}}\label{app:detail-alg-intervention}
\subsection{Feature Attributions in~\Cref{sec:alg_intervention}}\label{app:feature-attribution}
To quantify the shift in feature importance between the source and target domains, we compute Shapley value attributions for each feature using the \texttt{SHAP} framework. Specifically, we train two predictive models (each using a linear SVM) separately on the source data $(X_{\text{train}}, Y_{\text{train}})$ and the target data $(X_{\text{test}}, Y_{\text{test}})$, and use \texttt{shap.Explainer} to obtain SHAP values for each feature of each individual in both domains~\citep{lundberg2017unified}. We then average the SHAP values over all instances within each domain to obtain per-feature mean attributions. Finally, we compute the differences between the mean SHAP values in the source and target domains, resulting in a vector of feature-level attribution shifts. This vector highlights which features contribute differently to predictions across the two domains and thus helps identify critical components affected by distribution shifts.
\subsection{Additional Results in other Settings besides Sections~\ref{sec:casestudy-xshift} and~\ref{sec:casestudy-yshift}}\label{app:result_alg_intervention}
In each setting, we select the top-$K$ features, where $K \in \{1, 2, 3, 4, 5\}$ is chosen adaptively based on the number of features whose importance exceeds that of the remaining features by at least a threshold of 0.01.
\begin{table}[!htb]
    \centering
    \caption{Index of $Z$ selected in each setting}
    \label{tab:attribute-index}
    \begin{tabular}{c|ccccc}
        \toprule
        Setting & 1 & 2 & 3 & 5 & 6 \\
        \midrule
         Shifted $Z$& \{2\} & \{1,2\} & \{3,4\} & \{1,15,19,23,35\} & \{1,2,3,13,27\} \\
        \bottomrule
    \end{tabular}
\end{table}}

\wty{Subsequently, we provide algorithmic intervention results that incorporate shift patterns in \Cref{tab:attribute-index} for other settings, as shown in Tables~\ref{table:intervention-result-acc} and~\ref{table:intervention-result-f1}. Across all five settings, the results remain valid. For settings where $Y|X$-shifts dominate, incorporating such $Y|X$-shift patterns leads to a more refined ambiguity set and (usually) improved model accuracy for the Marginal-DRO and Conditional-DRO methods, especially when there is substantial room for improvement in the base model class, i.e., higher accuracy and macro F1 scores for some tree-based ensemble method compared with SVM. For example, Marginal-DRO in Settings 1 and 5, and Conditional-DRO in Setting 1.
\begin{table}[!htb]
\caption{Results (Accuracy) of 5 other pairs in \Cref{subsec:benchmark-empirical-result}, where we run each method with its top-10 configurations (according to the in-distribution accuracy) and report its mean accuracy and standard deviation. We boldface the best target performance within each class of methods in each setting.}
\vspace{5pt}
\label{table:intervention-result-acc}
\resizebox{\textwidth}{!}{
\begin{tabular}{@{}llccccc@{}}
\toprule\toprule
\multicolumn{2}{l}{\large \textbf{Dataset}}                                                                   & Setting 1 & Setting 2    & Setting 3 & Setting 5 & Setting 6                                                                                 \\ 
\multicolumn{2}{l}{\large \textbf{Shift Pattern}}                                                            & $Y|X$ dominates                                                                              & $Y|X$ dominates                                                                            & $Y|X$ dominates                                                                              & $Y|X$ more                                                                          & $X$ more                                                                            \\
\multicolumn{2}{l}{\large Source $\rightarrow$ Target Pair}                                                       & CA$\rightarrow$PR                                                                       & MS$\rightarrow$HI                                                                        & NYC$\rightarrow$BOG                                                                               & {CA$\rightarrow$OR}                                                                        & {2010$\rightarrow$2017}\\\midrule
\multirow{5}{*}{\begin{tabular}[c]{@{}l@{}}Basic \\ Methods\end{tabular}} &SVM&72.8\scriptsize${\pm 0.0}$&71.2\scriptsize${\pm 0.1}$&\textbf{75.6}\scriptsize${\pm 0.1}$&\textbf{77.2}\scriptsize${\pm 0.1}$&68.7\scriptsize${\pm 0.1}$\\
&LR&73.5\scriptsize${\pm 0.5}$&70.7\scriptsize${\pm 0.5}$&74.9\scriptsize${\pm 0.2}$&77.0\scriptsize${\pm 0.2}$&68.5\scriptsize${\pm 0.2}$\\
&RF&\textbf{76.0}\scriptsize${\pm 0.8}$&\textbf{71.7}\scriptsize${\pm 0.4}$&73.2\scriptsize${\pm 0.2}$&67.9\scriptsize${\pm 0.3}$&\textbf{70.7}\scriptsize${\pm 0.4}$\\
&XGB&69.8\scriptsize${\pm 1.1}$&66.1\scriptsize${\pm 0.5}$&71.4\scriptsize${\pm 0.8}$&66.8\scriptsize${\pm 0.2}$&68.8\scriptsize${\pm 1.5}$\\
&LGBM&75.7\scriptsize${\pm 3.9}$&67.6\scriptsize${\pm 1.1}$&72.8\scriptsize${\pm 2.9}$&67.1\scriptsize${\pm 1.2}$&69.2\scriptsize${\pm 1.1}$\\
\hline
\multirow{6}{*}{\begin{tabular}[c]{@{}l@{}}Linear-DRO \\  Methods \\ (base: SVM)\end{tabular}}  & Wasserstein-DRO&74.2\scriptsize${\pm 2.4}$&\textbf{71.3}\scriptsize${\pm 0.1}$&75.3\scriptsize${\pm 0.1}$&\textbf{76.8}\scriptsize${\pm 0.3}$&67.8\scriptsize${\pm 0.0}$\\
&Wasserstein-DRO-Int&72.8\scriptsize${\pm 1.3}$&71.2\scriptsize${\pm 0.2}$&75.4\scriptsize${\pm 0.1}$&76.0\scriptsize${\pm 0.1}$&67.8\scriptsize${\pm 0.0}$\\
&Marginal-DRO&70.0\scriptsize${\pm 2.5}$&70.8\scriptsize${\pm 0.5}$&76.1\scriptsize${\pm 0.1}$&71.1\scriptsize${\pm 1.8}$&69.0\scriptsize${\pm 0.5}$\\
&Marginal-DRO-Int&75.1\scriptsize${\pm 0.4}$&69.9\scriptsize${\pm 0.6}$&\textbf{76.2}\scriptsize${\pm 0.6}$&\textbf{76.8}\scriptsize${\pm 1.7}$&\textbf{69.2}\scriptsize${\pm 0.4}$\\
&Conditional-DRO&73.2\scriptsize${\pm 0.2}$&70.8\scriptsize${\pm 0.2}$&76.0\scriptsize${\pm 0.1}$&76.0\scriptsize${\pm 0.2}$&69.0\scriptsize${\pm 0.0}$\\
&Conditional-DRO-Int&\textbf{75.7}\scriptsize${\pm 0.0}$&70.5\scriptsize${\pm 0.1}$&76.0\scriptsize${\pm 0.0}$&74.1\scriptsize${\pm 0.0}$&68.9\scriptsize${\pm 0.0}$\\
\bottomrule
\end{tabular}}
\end{table}
\begin{table}[!htb]
\caption{Results (Macro F1 Score) of 5 other pairs in \Cref{subsec:benchmark-empirical-result}, where we run each method with its top-10 configurations (according to the in-distribution accuracy) and report its mean accuracy and standard deviation. We boldface the best target performance within each class of methods in each setting.}
\vspace{5pt}
\label{table:intervention-result-f1}
\resizebox{\textwidth}{!}{
\begin{tabular}{@{}llccccc@{}}
\toprule\toprule
\multicolumn{2}{l}{\large \textbf{Dataset}}                                                                   & Setting 1 & Setting 2    & Setting 3 & Setting 5 & Setting 6                                                                                 \\ 
\multicolumn{2}{l}{\large \textbf{Shift Pattern}}                                                            & $Y|X$ dominates                                                                              & $Y|X$ dominates                                                                            & $Y|X$ dominates                                                                              & $Y|X$ more                                                                          & $X$ more                                                                            \\
\multicolumn{2}{l}{\large Source $\rightarrow$ Target Pair}                                                       & CA$\rightarrow$PR                                                                       & MS$\rightarrow$HI                                                                        & NYC$\rightarrow$BOG                                                                               & {CA$\rightarrow$OR}                                                                        & {2010$\rightarrow$2017}\\\midrule
\multirow{5}{*}{\begin{tabular}[c]{@{}l@{}}Basic \\ Methods\end{tabular}}  &SVM&60.7\scriptsize${\pm 0.0}$&52.9\scriptsize${\pm 0.5}$&\textbf{75.6}\scriptsize${\pm 0.1}$&\textbf{60.9}\scriptsize${\pm 0.1}$&59.8\scriptsize${\pm 0.2}$\\
&LR&61.3\scriptsize${\pm 0.4}$&\textbf{53.8}\scriptsize${\pm 0.3}$&74.8\scriptsize${\pm 0.2}$&59.4\scriptsize${\pm 1.2}$&58.8\scriptsize${\pm 0.3}$\\
&RF&\textbf{62.0}\scriptsize${\pm 0.6}$&51.0\scriptsize${\pm 0.8}$&73.1\scriptsize${\pm 0.2}$&59.8\scriptsize${\pm 0.3}$&64.6\scriptsize${\pm 1.0}$\\
&XGB&57.9\scriptsize${\pm 1.0}$&52.5\scriptsize${\pm 0.5}$&71.3\scriptsize${\pm 0.8}$&60.4\scriptsize${\pm 0.4}$&64.7\scriptsize${\pm 0.9}$\\
&LGBM&62.4\scriptsize${\pm 2.7}$&53.2\scriptsize${\pm 0.4}$&72.6\scriptsize${\pm 3.0}$&59.7\scriptsize${\pm 1.2}$&\textbf{65.1}\scriptsize${\pm 0.7}$\\
\hline
\multirow{6}{*}{\begin{tabular}[c]{@{}l@{}}Linear-DRO \\  Methods \\ (base: SVM)\end{tabular}}  & Wasserstein-DRO&60.6\scriptsize${\pm 0.8}$&51.9\scriptsize${\pm 0.0}$&75.3\scriptsize${\pm 0.1}$&59.6\scriptsize${\pm 1.7}$&58.8\scriptsize${\pm 0.0}$\\
&Wasserstein-DRO-Int&60.7\scriptsize${\pm 0.9}$&51.9\scriptsize${\pm 0.1}$&75.3\scriptsize${\pm 0.1}$&61.7\scriptsize${\pm 0.1}$&58.8\scriptsize${\pm 0.0}$\\
&Marginal-DRO&60.9\scriptsize${\pm 0.5}$&51.7\scriptsize${\pm 0.4}$&71.2\scriptsize${\pm 1.9}$&62.6\scriptsize${\pm 0.5}$&58.6\scriptsize${\pm 1.7}$\\
&Marginal-DRO-Int&61.0\scriptsize${\pm 0.4}$&52.2\scriptsize${\pm 1.1}$&\textbf{76.1}\scriptsize${\pm 0.3}$&\textbf{62.8}\scriptsize${\pm 0.2}$&59.6\scriptsize${\pm 3.1}$\\
&Conditional-DRO&58.6\scriptsize${\pm 1.8}$&\textbf{53.2}\scriptsize${\pm 0.5}$&\textbf{76.1}\scriptsize${\pm 0.1}$&62.4\scriptsize${\pm 0.9}$&\textbf{62.2}\scriptsize${\pm 1.3}$\\
&Conditional-DRO-Int&\textbf{62.2}\scriptsize${\pm 0.3}$&52.7\scriptsize${\pm 1.1}$&73.0\scriptsize${\pm 0.6}$&61.1\scriptsize${\pm 0.3}$&61.6\scriptsize${\pm 1.6}$\\
\bottomrule
\end{tabular}}
\end{table}
}

{\color{black}\subsection{Additional Results in Section~\ref{subsec:llm-shift}}\label{app:llm-shift}
Our prompt to Gemini-2.5-Pro is as follows:



\begin{tcolorbox}[]
\textcolor{black}{Given the ACS Income dataset \url{https://fairlearn.org/main/user_guide/datasets/acs_income.html}, and the following set of one-hot encoded binary features: $\ldots$ (mention feature names), each feature is a binary indicator, encoded via one-hot encoding.
Suppose I train a predictive model using data only from California and aim to generalize its predictions for income to the other 50 states and Puerto Rico (PR).
Which of the above features are most likely to exhibit significant distributional shifts between California and the other states/PR?
Can you provide a list of all features as a dictionary, with keys being the feature and values being the weight?}
\end{tcolorbox}

\vspace{0.1in}

Afterwards, we rescale the vector $\Delta_{\text{LLM}}$ by multiplying its components by a constant so that their sum equals the original feature dimensionality $d$.}

\section{Details in~\Cref{subsec:data-intervention}}
\subsection{Further Discussions for Algorithm \ref{algo}}
\label{appendix:algorithm1}
In this part, we provide a more detailed introduction of our proposed Algorithm \ref{algo}.

\paragraph{Choices of $S_X$.} First, we construct a shared distribution $S_X$ over $X$ whose support is contained in that of both $P_X$ and $Q_X$.
We choose a specific \emph{shared distribution} $S_X$ over $X$ whose support is contained in that of $P_X$ and $Q_X$ (following~\citep{namkoong2023diagnosing}).
\begin{equation}\label{eq:s_X_orginal}
    S_X \propto \frac{p_X(x)q_X(x)}{p_X(x) + q_X(x)}
\end{equation}
\textcolor{black}{Ideally, the chosen shared distribution would exhibit a higher density when both $P_X$ and $Q_X$ densities are high, and a lower density when either of the two possesses a low density.
This strategy effectively allows regions of shared density to be more pronounced.}, which leads to the formulation of~\eqref{eq:s_X_orginal}.
Furthermore, we discuss different choices of $S_X$ to provide more intuitions. As demonstrated in \cite{namkoong2023diagnosing}, we can use many forms of the shared distribution. 
For example, we could choose the following form:
\begin{equation}
\label{appendix:s_X_alternative}
	s_X(x)\propto \text{min}\{p_X(x), q_X(x)\},
\end{equation}
which guarantees that the support of $S_X$ is contained in that of both $P_X$ and $Q_X$.
Another choice is:
\begin{equation*}
	s_X(x)\propto \begin{cases} 
    p_X(x)+q_X(x) & \text{if } \min\paran{\frac{p_X(x)}{q_X(x)}, \frac{q_X(x)}{p_X(x)}}\geq \epsilon, \\
    0 & \text{otherwise,}
\end{cases}
\end{equation*}
for some $\epsilon\geq 0$. 
This form of $S_X$ defines shared samples as those with high likelihood ratios.
Notably, for all the three forms of $S_X$, if $P_X=Q_X$, then $S_X=P_X=Q_X$. And when $p_X(x)\gg q_X(x)$ or $p_X(x)\ll q_X(x)$, \eqref{eq:s_X_orginal} and \eqref{appendix:s_X_alternative} become similar.
In our Algorithm \ref{algo}, we use the form of \eqref{eq:s_X_orginal}, and Cai et al. \cite{namkoong2023diagnosing} observe that in practice the qualitative conclusions are not very sensitive to the specific choice of shared distribution.

\paragraph{Intuitions behind Algorithm \ref{algo}.} 
Since we do not have access to samples from the shared distribution $S_X$, we
reweight samples from $P_X$ and $Q_X$ using the likelihood ratios:
\begin{equation}
	\frac{s_X}{p_X}(x)\propto \frac{q_X(x)}{p_X(x)+q_X(x)} \text{ and }  \frac{s_X}{q_X}(x)\propto \frac{p_X(x)}{p_X(x)+q_X(x)}.
\end{equation}
Then we define $\hat{\alpha}$ as the proportion of the pooled data that comes from  the distribution $Q$:
\begin{equation*}
	\hat{\alpha} = \frac{n_Q}{n_P+n_Q} \quad \text{and}\quad \hat{\pi}(x) = \mathbb{P}(\tilde{X}\text{ from }Q_X|\tilde{X}=x),
\end{equation*}
where $\hat{\pi}(x)$ denotes the probability of a sample to come from $Q_X$.
Using Bayes' rule, we have:
\begin{equation*}
\begin{aligned}
	\hat{\pi}(x)&=\frac{\mathbb{P}(\tilde{X}=x|\tilde{X}\text{ from }Q_X)\mathbb P(\tilde{X}\text{ from }Q_X)}{\mathbb P(x)}=\frac{\hat{\alpha}q(x)}{\hat{\alpha}q(x)+(1-\hat{\alpha})p(x)},\\
	&= \frac{\hat{\alpha}}{\hat{\alpha}+(1-\hat{\alpha})\frac{p(x)}{q(x)}}.
\end{aligned}
\end{equation*}
Noting that the ratio $\hat{\pi}(x)$ can be modeled as the probability that an input $x$ came from $P_X$ vs $Q_X$, we train a binary ``domain'' classifier to estimate the ratios.  (The ``domain'' classifier can be any black-box method, and we use XGBoost throughout).

Then the likelihood ratios that we care about could be reformulated as:
\begin{equation*}
	\frac{s_X}{p_X}(x)\propto \frac{1}{\frac{p_X(x)}{q_X(x)}+1} \text{ and }  \frac{s_X}{q_X}(x)\propto \frac{\frac{p_X(x)}{q_X(x)}}{\frac{p_X(x)}{q_X(x)}+1},
\end{equation*}
which gives that:
\begin{align*}
  \frac{s_X}{p_X}(x) \propto \frac{\hat{\pi}(x)}{(1-\hat{\alpha})\hat{\pi}(x)+\hat{\alpha}(1-\hat{\pi}(x))}\quad \text{and}\quad 	\frac{s_X}{q_X}(x) \propto \frac{1-\hat{\pi}(x)}{(1-\hat{\alpha})\hat{\pi}(x)+\hat{\alpha}(1-\hat{\pi}(x))}.
\end{align*}	
After obtaining the likelihood ratios $\frac{s_X}{p_X}(x)$ and $\frac{s_X}{q_X}(x)$, we could do an apples-to-apples comparison: we estimate $P_{Y|X}$ and $Q_{Y|X}$ over the shared distribution $S_X$ (using XGBoost $\Fscr$):
  \begin{align*}
    f_P := &\arg\min_{f\in \Fscr}\left\{ \mathbb{E}_{S_X}\bigg[\mathbb{E}_P[\ell_{tr}(f(X),Y)|X]\bigg] = \mathbb{E}_P\bigg[\ell_{tr}(f(X),Y)\frac{dS_X}{dP_X}(X)\bigg]\right\},\\
    f_Q := &\arg\min_{f\in \Fscr}\left\{ \mathbb{E}_{S_X}\bigg[\mathbb{E}_Q[\ell_{tr}(f(X),Y)|X]\bigg] = \mathbb{E}_Q\bigg[\ell_{tr}(f(X),Y)\frac{dS_X}{dQ_X}(X)\bigg]\right\}.
\end{align*}	
Then, for any threshold $b\in [0, 1]$,
$\{x \in \Xscr : |f_P(x)-f_Q(x)| \geq b\}$ suggests a region that may suffer
model performance degradation due to $Y|X$-shifts.

\subsection{Alternative Approach of Identifying Covariate Region}\label{app:others}
In Algorithm \ref{algo}, we propose a simple way to identify the covariate region to explain the cause of $Y|X$-shifts.
And the identified region could be used to guide the collecting process of target data, which could help to further reduce the effects of performance degradation.
However, in practice, when the amount of target samples is quite small, it is difficult to train the $f_Q$ only with target samples accurately.

Following this idea, we propose a sample-efficient alternative for identifying covariate regions in Algorithm \ref{algo:alternative}, which does not need $f_Q$ to fit $Q_{Y|X}$ on the target distribution.
Since the training data is enough, it is feasible to fit $f_P$ on the shared distribution $S_X$ as:
\begin{equation}
\label{equ:region-appendix}
	f_P := \argmin_{f\in\mathcal F}\mathbb{E}_P\left[\frac{dS_X}{dP_X}\ell_{tr}(f(X),Y)\right].
\end{equation} 
Then for $n_Q$ samples from target distribution $Q$, we could use a prediction model $h(x)$ to approximate $|Y-f_P(X)|$ on the shared distribution (by reweighting density ratio $\frac{dS_X}{dQ_X}$).
Note that this method does not need to train $f_Q$ on target samples, but the quality of the density ratio $dS_X / dP_X$ and the prediction model $h(x)$ still depend on the target samples. 

\begin{algorithm}[t]
    \KwIn{Source samples $\{(x_i^P, y_i^P)\}_{i \in [n_P]}\overset{\text{i.i.d}}{\sim} P$ and target samples
      $\{(x_j^Q, y_j^Q)\}_{j \in [n_Q]}\overset{\text{i.i.d}}{\sim} Q$. Model
      discrepancy threshold $b$.}
    Estimate $\hat{\pi}(x)\approx\mathbb{P}(\tilde{X}\sim Q_x|\tilde{X}=x)$ by training a classifier on the source and target samples.\\
    Calculate density ratios $w_{\mu}(\hat\pi(x), \hat\alpha)$ according to Equation \eqref{equ:region1} and \eqref{equ:region2} for $\mu = P, Q$.\\
    Fit the prediction model $f_{P}$ according to Equation \eqref{equ:region-appendix}, where we set $\Fscr$ as XGBoost; (\emph{only fit $f_P$ here})\\
    Fit a model $h(x)$ to predict $|f_P(x)-y|$ using \emph{target} samples $\{(x_j^Q, y_j^Q)\}_{j \in [n_Q]}$ with the weight $\lambda_j^Q = \frac{w_{Q}(\hat\pi(x_j^{Q}),\hat\alpha)}{\sum_{k \in [n_{Q}]} w_{Q}(\hat\pi(x_k^{Q}),\hat\alpha)}, \forall j \in [n_{Q}]$.  \BlankLine \KwOut{Region $\Rscr = \{x \in \Xscr: h(x)\geq b\}$.}
    \BlankLine
  \caption{Sample-Efficient Alternative for Identifying Regions}
  \label{algo:alternative}
\end{algorithm}

\subsection{Other Results}\label{app:other-dataintervention}

\subsubsection{Analysis of Decision Tree}\label{app:tree-split}
In Algorithm \ref{algo}, we use a shallow \emph{decision tree} $h(x)$ to approximate $y=|f_P(x)-f_Q(x)|$ on the shared distribution $S_X$ to find the covariate region with highest discrepancy.
In our decision tree, we use the \emph{squared error} as the splitting criterion.
And below we demonstrate that this criterion is equivalent to maximizing the discrepancy between two children nodes.

Suppose there are $N$ samples with outcomes $\{y_i\}_{i \in [N]}$ belonging to tree node $fa$, and these samples are split into two children nodes $s_1,s_2$, where the node $s_1, s_2$ denote the set of sample indices in the two children nodes respectively. The squared error criterion to split $fa$ into $s_1$ and $s_2$ is:
\begin{equation}\label{eq:sq-split}
	\min_{s_1,s_2} \bigg \{\mathcal L(s_1,s_2):=\frac{1}{N}\bigg(\sum_{i\in s_1}(y_i- \mu_{Y, 1})^2 + \sum_{i\in s_2}(y_i-\mu_{Y, 2})^2 \bigg)\bigg\},
\end{equation} 
where 
\begin{align}
	\mu_{Y, 1} := \frac{\sum_{i=1}^{N} y_i \mathbf{1}_{\{i \in s_1\}}}{\sum_{i = 1}^N \mathbf{1}_{\{i \in s_1\}}},\quad \mu_{Y, 2} := \frac{\sum_{i=1}^{N} y_i \mathbf{1}_{\{i \in s_2\}}}{\sum_{i = 1}^N \mathbf{1}_{\{i \in s_2\}}}
\end{align}
denote the mean values of the outcome $Y$ with samples in children nodes $s_1$ and $s_2$. Denote the distribution of the outcome $Y$ follows the empirical distribution over the $N$ samples $\{y_i\}_{i \in [N]}$. Simplifying~\eqref{eq:sq-split}, we have:
\begin{equation}
	\mathcal L(s_1,s_2) = P(Y\in s_1)\text{Var}_{s_1}(Y)+P(Y\in s_2)\text{Var}_{s_2}(Y) = \mathbb{E}_S[\text{Var}(Y|S)],
\end{equation}
where $\text{Var}_{s}(Y)$ denotes the variance of the outcome variable $Y$ in node $s$, $S=\{s_1,s_2\}$ is the variable representing the children nodes.
Therefore, given that $\text{Var}_{fa}(Y)(:=\text{Var}_S(\mathbb E[Y|S]) + \mathbb{E}_S[\text{Var}(Y|S)])$ is constant, the minimal $\mathbb{E}_S[\text{Var}(Y|S)]$ corresponds with the largest $\text{Var}_S(\mathbb E[Y|S])$, which maximizes the discrepancy of the outcome between two children nodes.

\subsubsection{Details of Data-centric Interventions}\label{app:nonalg-intervene}
In Section \ref{sec:model_intervention}, we propose two potential data-centric interventions to mitigate the performance degradation. 
In this section, we introduce in detail the intervention of collecting specific data from the target.

\paragraph{Experiment setup.} We focus on the income prediction task using the \texttt{ACS Income} dataset. 
Consider a practical scenario where the training set consists of 20,000 samples from California (CA) and the trained model was deployed in Puerto Rico (PR) in trial.
After the trial deployment in PR, we got a \emph{small amount} of samples from PR with labels and observed performance degradation.
Under this setting, we investigate the effect of non-algorithmic interventions.

\begin{figure}[htbp]
\vspace{-0.15in}
  \centering
  \subfloat[$N=100$]{\includegraphics[width=0.48\textwidth]{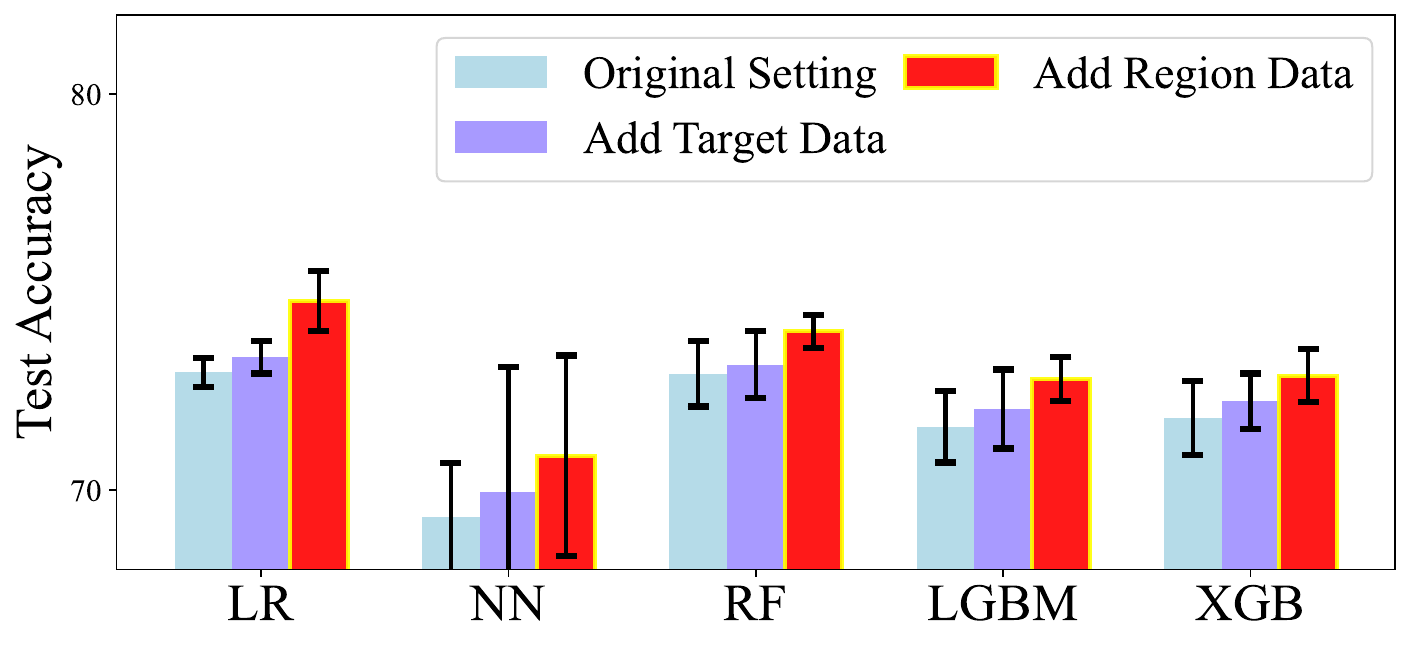}\label{fig:appendix_case1}}
  \hfill
    \subfloat[$N=200$]{\includegraphics[width=0.48\textwidth]{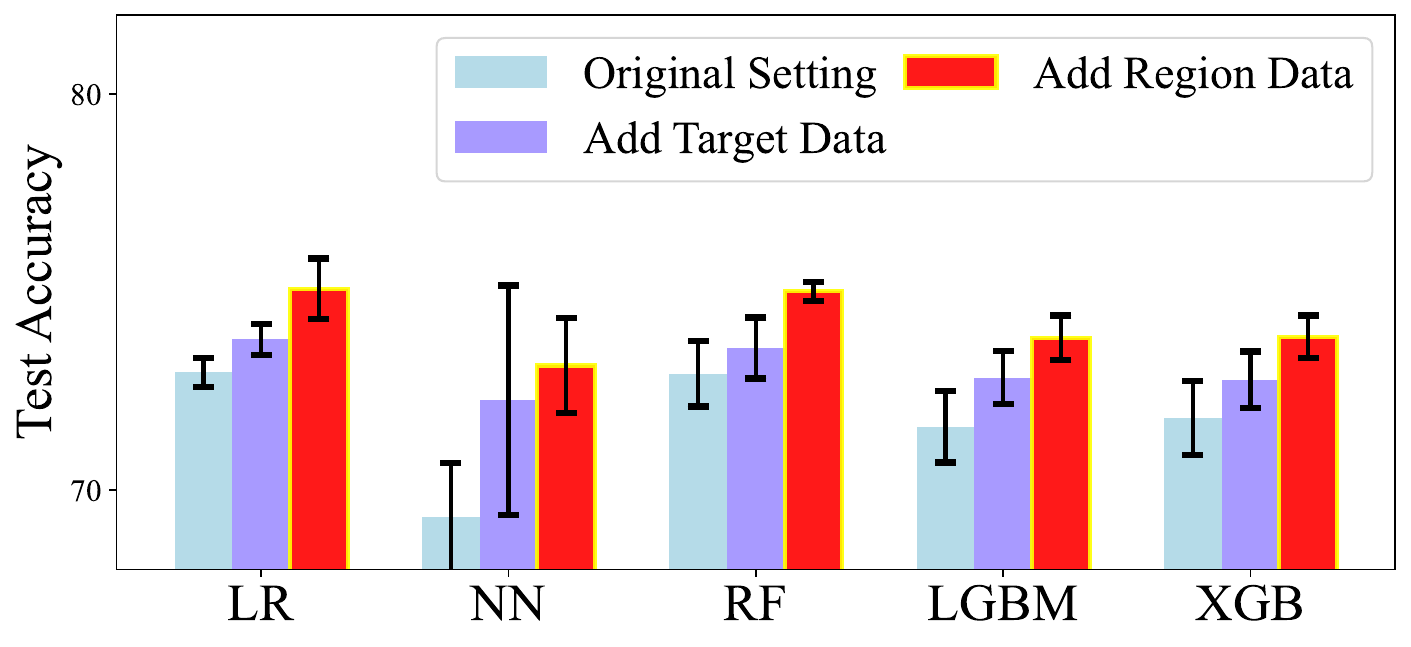}\label{fig:appendix_case2}}
\hfill
    \centering\subfloat[$N=300$]{\includegraphics[width=0.49\textwidth]{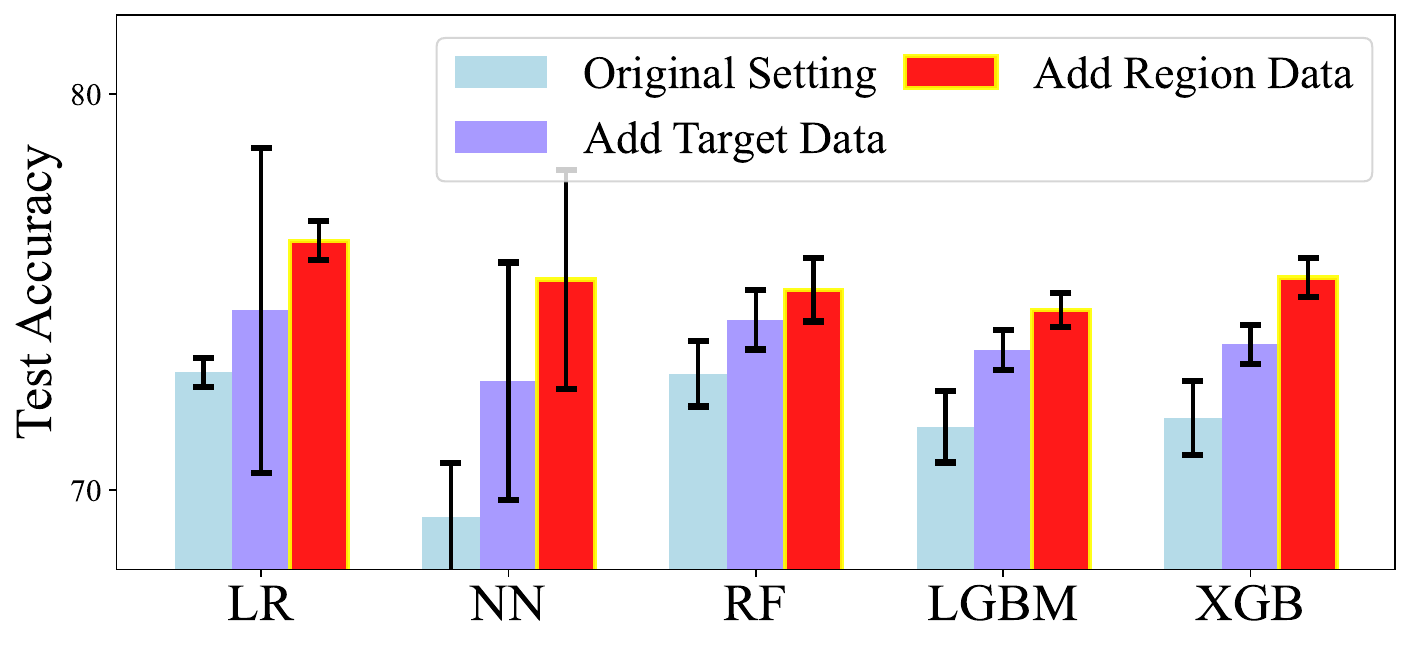}\label{fig:appendix_case3}}
\caption{Test accuracies of different ways to incorporate data.}
 \label{fig:appendix_analysis}  
 \vspace{-10pt}
\end{figure}

\paragraph{Collect specific data from the target.} We first identify the regions with high discrepancy between source and target.
Note that the sample size of the target state is small compared to the training samples.
Then for basic ERM and tree-based ensemble methods like LR, NN, RF, LightGBM and XGB, we compare the performances of:
\begin{itemize}
	\vspace{-5pt}
	\item original setting (only 20,000 samples from the source);
	\vspace{-5pt}
	\item original setting with $N$ additional random samples drawn from the whole target state;
	\vspace{-5pt}
	\item original setting with $N$ additional random samples drawn from the \emph{covariate region} of the target domains.
	\vspace{-5pt}
\end{itemize} 
In this experiment, we first select the best configuration of each method according to the $i.i.d$ validation set in the original setting (only samples from CA), and fix it for the other two interventions.
We vary $N$ as 100, 200, 300 and the results are shown in \Cref{fig:appendix_analysis}.
From \Cref{fig:appendix_analysis}, incorporating data from the identified covariate region leads to a \emph{stable improvement} on typical algorithms even for small target sample sizes.
However, we observe that LightGBM and XGBoost would easily overfit $f_Q$ on the target data, and we use random forest under this setup as an alternative. It is worth investigating approaches to find covariate regions effectively under small/imbalanced sample sizes in the future. The approach mentioned here is a simple way of non-algorithmic and explainable interventions and we hope it could inspire further research in this direction.

\end{document}